\documentclass{article}
\usepackage[final]{corl_2025} 
\usepackage{wrapfig}  
\usepackage{booktabs}
\usepackage{multirow}
\usepackage{pifont}
\usepackage{amssymb}
\usepackage{algorithm}
\usepackage{algpseudocode}
\usepackage{graphicx}
\usepackage{subcaption}
\usepackage{tabularx}
\usepackage{makecell}
\usepackage{adjustbox}
\usepackage{soul}
\usepackage{comment}
\usepackage{bbm}
\usepackage{amsmath}
\usepackage{amssymb}
\usepackage[dvipsnames]{xcolor}

\newcommand{\zhenyang}[1]{\textcolor{orange}{(Zhenyang: #1)}}

\newcommand{\danfei}[1]{\textcolor{blue}{(Danfei: #1)}}
\newcommand{\woochul}[1]{\textcolor{ForestGreen}{(Woochul: #1)}}
\newcommand{\yangcen}[1]{\textcolor{purple}{(Yangcen: #1)}}
\newcommand{\harish}[1]{\textcolor{violet}{(Harish: #1)}}

\title{ImMimic: Cross-Domain Imitation from Human Videos via Mapping and Interpolation}

\author{
  Yangcen Liu\thanks{Equal contribution},\quad
  Woo Chul Shin\footnotemark[1],\quad
  Yunhai Han,\quad
  Zhenyang Chen,\\
  \textbf{Harish Ravichandar},\quad
  \textbf{Danfei Xu} \\
  College of Computing, Georgia Institute of Technology, United States \\
  \texttt{\{yliu3735, wshin49, yhan389, zchen927, harish.ravichandar, danfei\}@gatech.edu}
}

\begin{document}
\maketitle

\newcolumntype{C}{>{\centering\arraybackslash}m{}} 
\vspace{-1em}
\begin{abstract}
Learning robot manipulation from abundant human videos offers a scalable alternative to costly robot-specific data collection.
However, domain gaps across visual, morphological, and physical aspects hinder direct imitation. To effectively bridge the domain gap, we propose \textbf{ImMimic}, an embodiment-agnostic co-training framework that leverages both human videos and a small amount of teleoperated robot demonstrations. 
ImMimic uses Dynamic Time Warping (DTW) with either action- or visual-based mapping to map retargeted human hand poses to robot joints, followed by MixUp interpolation between paired human and robot trajectories. Our key insights are (1) retargeted human hand trajectories provide informative action labels, and (2) interpolation over the mapped data creates intermediate domains that facilitate smooth domain adaptation during co-training. Evaluations on four real-world manipulation tasks (Pick and Place, Push, Hammer, Flip) across four robotic embodiments (Robotiq, Fin Ray, Allegro, Ability) show that ImMimic improves task success rates and execution smoothness, highlighting its efficacy to bridge the domain gap for robust robot manipulation. The project website can be found at \url{https://sites.google.com/view/immimic}.

\end{abstract}


\begin{figure}[H]
  \centering
       \centering
       \vspace{-2em}
       \includegraphics[width=0.92\linewidth]{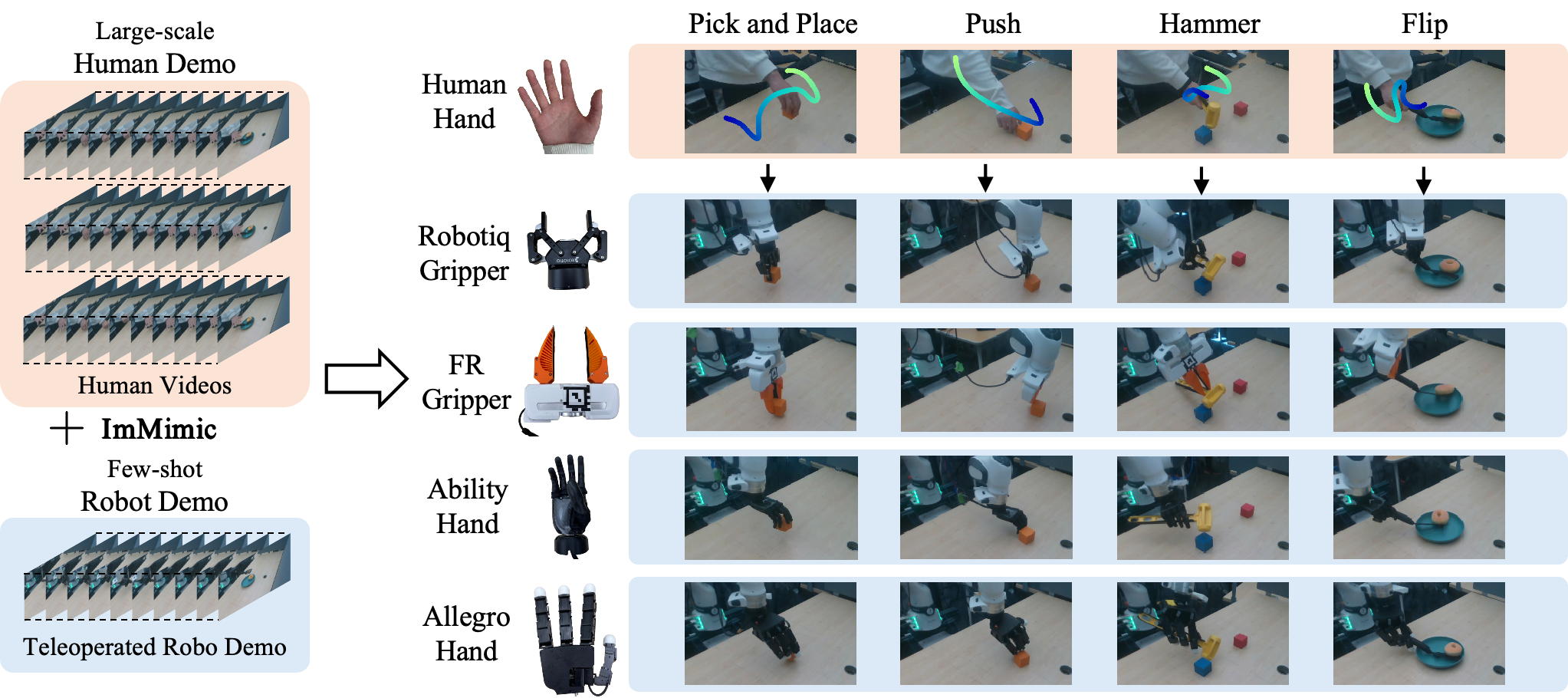}
       \caption{
       \textbf{ImMimic} enables embodiment-agnostic co-training between human and robot demonstrations. 
        It leverages large-scale human videos and a small amount of teleoperated robot data, using a MixUp interpolation to enable smooth domain transfer. 
        We validate ImMimic on four diverse manipulation tasks across four robotic embodiments. 
}
       \label{fig:teaserfig}
       \vspace{-1em}
\end{figure}

\section{Introduction}

Teaching robots to perform diverse manipulation tasks in real-world environments remains a significant challenge because collecting robot-specific demonstration data is expensive.
As an alternative, human videos have emerged as a promising resource, offering abundant examples of people engaging in everyday manipulation activities~\cite{srivastava2022behavior, li2024behavior}. 
Leveraging these human videos for robot training provides a scalable and cost-effective approach to enhance robotic skills without extensive robot demonstration collection or simulation~\cite{mimicplay,egomimic,videodex}. 
However, learning robot skills from human demonstration videos still faces an inherent limitation: a substantial domain gap arising from stark differences in visual appearance, embodiment structure, physical constraints, and other factors.

In general, the challenge of enabling robots to learn human demonstrations can be formulated as a domain adaptation problem: the robot (representing the target domain) aims to emulate the behaviour of the human demonstrator (representing the source domain). 
A relevant application of this concept can be seen in several recent, well-established vision-based teleoperation systems~\cite{anyteleop, bunnyvision, telekinesis} where the human demonstrator often first practices with the teleoperation setup before being able to collect high-quality robot data. This process indeed reflects an instance of \textit{inverse} adaptation, where the human demonstrator adapts to the robot system rather than the way around. However, such inverse adaptation is absent in human demonstration videos, as human demonstrators do not consider the robot's subsequent operation. Therefore, to enable effective adaptation when learning robot skills from such human data, recent works often preprocess the input data of both domains—for example, masking out embodiments in images~\cite{bahl2022human, egomimic} to mitigate visual differences, or restricting the action space to only 3D translations to address the embodiment-specific action gap~\cite{mimicplay, ditto}. 
Additionally, another line of works encourages the adaptation of the latent spaces of visual inputs from both domains using unsupervised learning objectives~\cite{xu2023xskill, xu2024flow, lin2024flowretrieval, strap}, but often overlooks human actions, i.e., the hand trajectories, instead learning the action decoder solely from robot demonstrations.

To develop a more generalizable adaptation method across diverse robot embodiments and manipulation tasks, we introduce \textbf{ImMimic (Interpolation-via-Mapping Mimic)}, an embodiment-agnostic co-training framework that learns jointly from human demonstration videos and robot teleoperations.
Our key insights are: (1) beyond the visual contexts, the retargeted human hand trajectories can serve as action labels for human demonstrations, (2) creating intermediate domains via interpolation leads to robust adaptation, and (3) establishing an effective mapping between human and robot data for interpolation is essential for co-training. Specifically, we begin by retargeting human hand poses into the robot’s action space.
We then perform sequence‑level mapping via Dynamic Time Warping (DTW), using either visual features or action distance to pair each human timestep with its best‑matching robot timestep.
Finally, inspired by MixUp‑based adaptation~\cite{mixup,dlow}, we interpolate both condition and predicted actions along these DTW mapping, enabling adaptation through intermediate (human–robot) domains.

To demonstrate the benefits of the ImMimic, we conduct comprehensive experiments across four different types of embodiments: Robotiq Gripper, FR Gripper~\cite{FRGripper,umi,liu2024fastumi}, Allegro Hand, and Ability Hand. 
These embodiments are evaluated on four manipulation tasks: Pick and Place, Push, Hammer, and Flip.
We show that ImMimic achieves higher success rates and smoother motions across all embodiments and tasks compared to baseline methods. 
We observe that action-based mapping provides greater improvements than visual-based mapping, suggesting that the rich action information of the human hand trajectories is equally, if not more, beneficial for co-training. Furthermore, we find that performance improves when the average action distance between human and robot is smaller, and interestingly, due to the factors such as hand-arm mounting conditions and arm kinematics, solely using a human-mimetic end-effector does not necessarily result in smaller action distances.  Finally, we analyze failure cases in terms of hardware structure and algorithmic factors. 
\section{Related Work}
\label{sec:related}

\textbf{Learning from Human Videos.}
Human videos offer an efficient and scalable source of supervision for robot learning. 
Retrieval‑based approaches~\cite{r+x,strap,mimiclabs, xu2023xskill, xu2024flow} search large video corpora for sequences that resemble the desired behavior for augmented learning, while typically relying on the robot’s own data for action decoding.
To better utilize human videos, two-stage methods~\cite{videodex,theia,r3m,mimicplay,screwmimic,mimictouch} first learn high‑level policies on human data and then adapt to robot demonstrations, but limit low-level action learning.
To address this, co‑training~\cite{egomimic} jointly optimizes on human and robot data, yet typically relies on heavy visual preprocessing or simplified action spaces, leaving the core domain shift largely unaddressed.
Our work adopts a co-training pipeline, and treating human videos as domain-adaptive supervision to smoothly address the domain shifts.

\textbf{Dexterous Retargeting.}
Dexterous retargeting translates human hand poses 
into robot joint poses, ensuring that human trajectories are mapped into the same action space used for robot execution. 
Most recent works adopt this technique to generate robot action commands for teleoperation by observing a human demonstrator moving their own hand~\cite{anyteleop, bunnyvision, openteach, telekinesis, shaw2024bimanual}. 
However, continuous guidance from the human demonstrator is crucial for the robot's success during teleoperation~\citep{li2021survey}.
Other works~\cite{chen2022dextransfer, pmlr-v205-chen23b, videodex, arunachalam2023dexterous, han2024learning, dexmv, she2022learning, chen2024object, dexcap} leverage retargeted robot trajectories to learn robot policies through bespoke refinements tailored to dexterous grasping~\cite{chen2022dextransfer, pmlr-v205-chen23b}, reinforcement learning~\cite{she2022learning, han2024learning, dexmv, chen2024object}, fine-tuning with additional robot data~\cite{videodex, arunachalam2023dexterous}, or human-in-the-loop corrections~\cite{dexcap}. 
This additional process underscores the domain gap between retargeted trajectories and actual robot execution, which arises from various factors such as visual, kinematic, and physical differences.
Our work handles such gap through a novel embodiment-agnostic co-training framework that smoothly adapts human demonstrations to the robot domain.

\textbf{Domain Adaptation.}  
To bridge domain gap, classic methods include adversarial feature adaptation~\cite{ganin2016dann} and cycle-consistent data translation~\cite{hoffman2018cycada}.
Recent work inserts intermediate domains via domain flow~\cite{gong2019dlow} with MixUp~\cite{abnar2021gift, mixup, xu2020domainmixup} to build the source-to-target path.
In robotic imitation learning, methods learn domain-invariant representations—such as viewpoint-agnostic and visual-invariant encoders—to handle sim-to-real and human-to-robot perception gaps~\cite{stadie2017third, choi2023visual}.
Meta-learned and latent-policy adaptation approaches enable rapid embodiment transfer and observation-to-action alignment~\cite{yu2018one, edwards2019imitation}.
Structural adaptation via optimal transport or point-cloud matching, combined with modality-invariant representations and MixUp-augmented offline RL~\cite{rlmixup}, helps bridge domain gap~\cite{fickinger2022cross, zhang2025moma, gao2025learning,daob}.
We extend this line by applying MixUp to visual features and actions in latent space, yielding a continuous bridge from the human domain to the robot domain.

\begin{figure*}[t]
  \centering
  \includegraphics[width=0.99\linewidth]{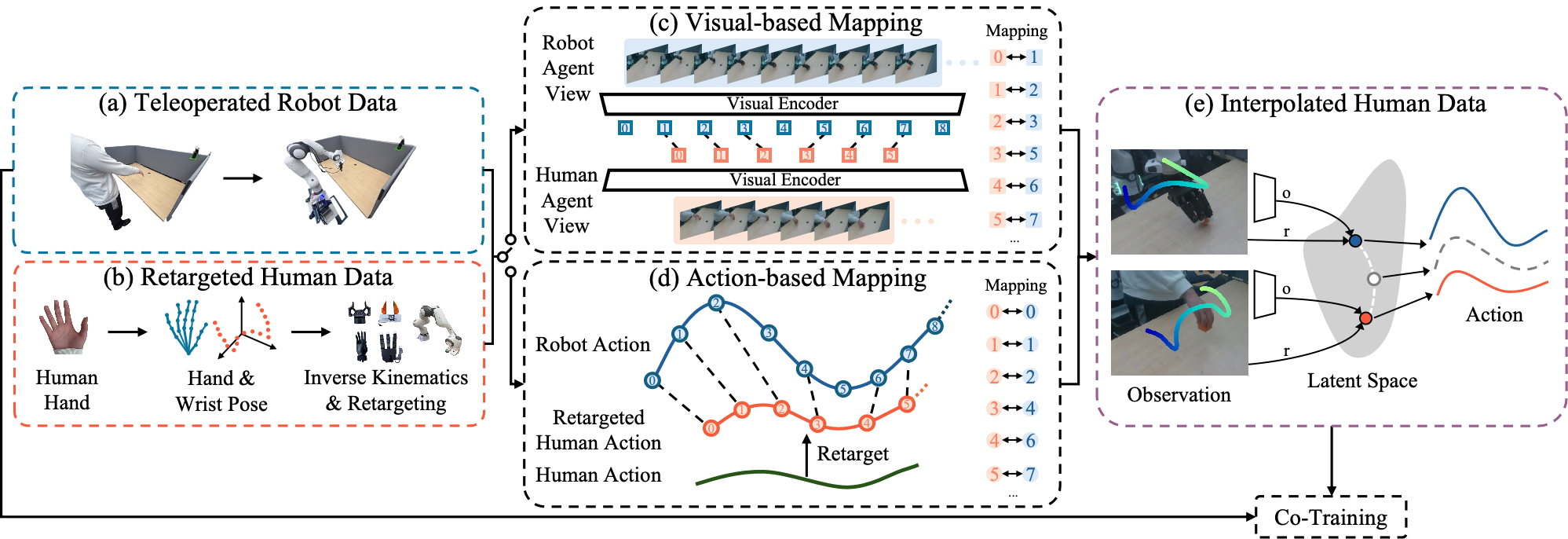}
  \caption{Overview of how we collect, map, and interpolate human and robot data.
  (a) Robot demonstrations are collected via visual teleoperation. 
  (b) Human actions are retargeted from videos. 
  (c, d) Visual or action based DTW maps retargeted human and robot trajectories. 
  (e) MixUp: Mapped human-robot pairs are interpolated in both the latent space and action space to generate interpolated human data. 
  Finally, we co-train the interpolated human data together with the robot data. 
  See Fig.~\ref{fig:pipeline} for the co-training pipeline.}
  \label{fig:overall}
  \vspace{-1em}
\end{figure*}

\section{Embodiment-Agnostic Co-Training Framework}
\label{sec:learning}

We aim to learn robotic manipulation policies from large-scale, easily collected human videos, with only a small number of teleoperated robot demonstrations.  
For each task, the model has access to a large corpus of human demonstrations $\{ \mathbf{I}_t^{\mathrm{a,h}} \}_{t=1}^T$, where each frame $\mathbf{I}_t^{\mathrm{a,h}} \in \mathbb{R}^{H \times W \times 3}$ is an agent-view RGB image.  
In parallel, it also receives a smaller set of robot demonstrations, each containing agent-view video $\{ \mathbf{I}_t^{\mathrm{a,r}} \}_{t=1}^T$, wrist-view video $\{ \mathbf{I}_t^{\mathrm{w,r}} \}_{t=1}^T$, and proprioception $\{ \mathbf{r}_t \}_{t=1}^T$, where $\mathbf{r}_t \in \mathbb{R}^D$ includes end-effector pose and finger joint positions.

This setting presents a domain adaptation challenge in the context of human-to-robot imitation learning. 
Specifically, human (source domain) and robot (target domain) data differ in both visual and action: (1) a visual covariate shift between human and robot observations due to embodiment appearance differences, and (2) an action gap arising from differences in embodiment structure and physical constraints, which can lead to variations in how the same task is performed.
Our goal is to bridge these gaps to better adapt human demonstrations to robot execution, enabling the policy to effectively leverage both large-scale human videos and few-shot robot demonstrations.

\begin{figure*}[t]
  \centering
  \includegraphics[width=1.0\linewidth]{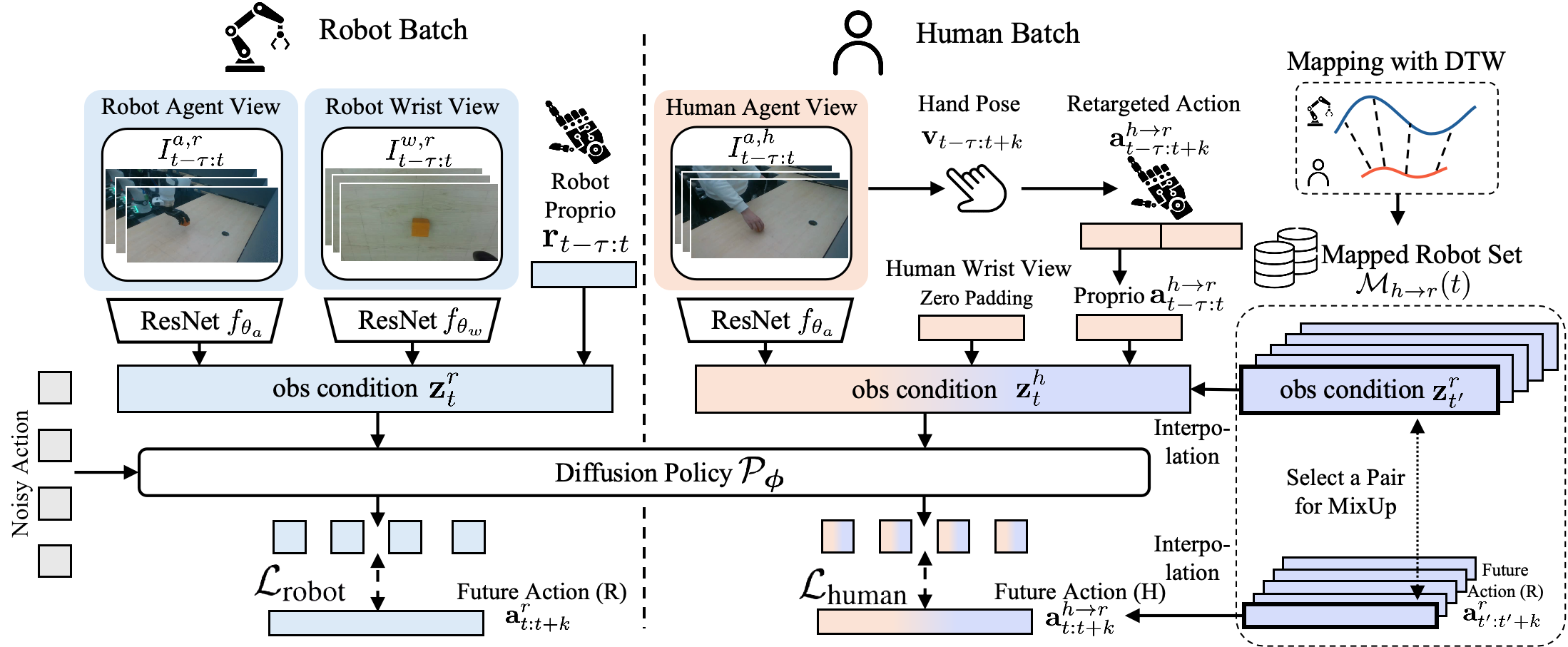}
  \caption{Overview of our embodiment-agnostic co-training framework \textbf{ImMimic}. 
  For \textbf{robot demonstration}, we train the policy using agent- and wrist-view images encoded with ResNet $f_{\theta_a}$, $f_{\theta_w}$, along with proprioception $\mathbf{r}_t$.
  All are combined into the observation condition $\mathbf{z}^r_t$ to predict future actions. 
  For \textbf{human demonstration}, we train the same diffusion policy $\mathcal{P}_{\boldsymbol{\phi}}$ using human videos. 
  A hand pose retargeting module generates retargeted robot actions $\mathbf{a}^{h \rightarrow r}_t$, which serve as both the future action and proprioception for training. 
  Mapping with DTW, we apply MixUp (Fig.~\ref{fig:overall}(e)) for human data with paired robot data.
  The interpolation enables human data to smoothly adapt to the robot data.
  The model is optimized upon the sum of reconstruction losses $\mathcal{L}_{\text{human}}$ and $\mathcal{L}_{\text{robot}}$.
}
  \label{fig:pipeline}
  \vspace{-1em}
\end{figure*}

To achieve this, we first retarget estimated human hand trajectories from videos to the robot trajectories (Sec.~\ref{handdetection}). We then jointly train the policy on both human and robot demonstrations (Sec.~\ref{cotraining}).
During co-training, to achieve a smooth domain adaptation from human to robot, we pair human-robot samples using DTW, and further perform MixUp with interpolation over mapped pairs (Sec.~\ref{mixup}). 
An overview framework is shown in Fig.~\ref{fig:pipeline}.


\subsection{Hand Pose Retargeting System}
\label{handdetection}

To fully leverage human videos, we extract both visual context and human hand trajectories, and then retarget hand trajectories to robot embodiments following recent advanced methods~\cite{videodex, anyteleop}.

\textbf{Hand and Wrist Pose Estimation.}
We use MediaPipe~\cite{mediapipe} to localize and crop the human hand in each frame.
Each patch is fed into FrankMocap~\cite{frankmocap}, whose SMPL‑X regressor produces precise 3D positions for 21 hand joints in the local wrist frame.
By projecting these joints into the depth map and solving a Perspective‑$n$‑Point problem, we recover the wrist 6D pose in camera frame. 

\textbf{Retargeting.}
Following AnyTeleop~\cite{anyteleop}, we map human keypoints \(\mathbf p_t^i\) to robot joint angles \(\mathbf q_t\) via
\begin{equation}
\label{eq:retarget}
\min_{\mathbf{q}_t} \sum_{i=1}^N \left\| \alpha\,\mathbf{p}_t^i - f_i(\mathbf{q}_t) \right\|^2
+ \beta\,\bigl\| \mathbf{q}_t - \mathbf{q}_{t-1} \bigr\|^2,
\quad
\mathbf{q}_l \le \mathbf{q}_t \le \mathbf{q}_u,
\end{equation}
where \(f_i\) is the robot’s forward‑kinematics, and \(\alpha,\beta\) balance scale and temporal smoothness.

\subsection{Co-Training}
\label{cotraining}
While prior work often treats human videos as auxiliary pretraining data~\cite{videodex, mimicplay}, recent studies such as EgoMimic~\cite{egomimic} demonstrate the benefits of co-training on both human and robot data.
Motivated by this, we adopt a co-training strategy that treats human and robot demonstrations equally, allowing the policy to learn from both domains throughout a single training.
Regarding the policy backbone, our framework builds on the Diffusion Policy~\cite{diffusionpolicy}. 

\textbf{Robot Prediction Loss.}  
At each timestep $t$, where images and proprioception are denoted as $\{(\mathbf{I}_t^{\mathrm{a,r}}, \mathbf{I}_t^{\mathrm{w,r}}, \mathbf{r}_t)\}$, we incorporate temporal context using a history of length $\tau$. Thus, the robot condition at each timestep is:
$\mathbf{z}_t^r = \left[\mathbf{z}_{t-\tau:t}^{\mathrm{a,r}} \;\middle\|\; \mathbf{z}_{t-\tau:t}^{\mathrm{w,r}} \;\middle\|\; \mathbf{r}_{t-\tau:t}\right] \in \mathbb{R}^{(d_a + d_w + d_a)\times \tau},$ where $\mathbf{z}_t^{\mathrm{a,r}} = f_{\theta_a}(\mathbf{I}_t^{\mathrm{a,r}})$ and $\mathbf{z}_t^{\mathrm{w,r}} = f_{\theta_w}(\mathbf{I}_t^{\mathrm{w,r}})$ are feature embeddings extracted by separate ResNet18~\cite{resnet} encoders $f_{\theta_a}$ and $f_{\theta_w}$ respectively.

We denote the future action sequence as:
$\mathbf{a} = \left(\mathbf{a}_{t+1}^{h \rightarrow r}, \ldots, \mathbf{a}_{t+k}^{h \rightarrow r}\right)$, where $k$ is the prediction horizon. 
A diffusion policy $\mathcal{P}_{\boldsymbol{\phi}}$ reconstructs $\mathbf{a}$ from a noisy action $\tilde{\mathbf{a}}$ using denoising steps conditioned on $\mathbf{z}_t$. The training objective minimizes an $\ell_2$ loss:
$
\mathcal{L}_{\text{robot}}(\boldsymbol{\phi}) =
\sum_{i=1}^{k} \left\|\,\mathbf{a}_{t+i}^r - \hat{\mathbf{a}}_{t+i}^r\right\|_2^2,
$ where
$\hat{\mathbf{a}}_{t:t+k}^r = \mathcal{P}_{\boldsymbol{\phi}}\left(\mathbf{\tilde{a}}^r_{t:t+k} \,\middle|\, \mathbf{z}_t^r\right).
$

\textbf{Human Prediction Loss.}  
For a human video $\{\mathbf{I}_{t}^{\mathrm{a,h}}\}_{t=1}^{T_{\mathrm{vid}}}$, at each timestep $t$, the condition input includes both image features and retargeted actions:
$\mathbf{z}^h_t = \left[\mathbf{z}_{t-\tau:t}^{\mathrm{a,h}} \;\middle\|\; \mathbf{0}_{t-\tau:t} \;\middle\|\; \mathbf{a}_{t-\tau:t}^{h \rightarrow r} \right] \in \mathbb{R}^{(d_a + d_w + d_a)\times \tau},$
where each $\mathbf{z}_{t}^{\mathrm{a,h}} = f_{\theta_h}(\mathbf{I}_{t}^{\mathrm{a,h}}) \in \mathbb{R}^{d_a}$ is extracted using the same ResNet encoder $f_{\theta_a}$, and $\mathbf{a}_{t}^{h \rightarrow r}$ is the retargeted action (from Sec.~\ref{handdetection}). Similarly, we compute the $\ell_2$ loss as 
$
\mathcal{L}_{\text{human}}(\boldsymbol{\phi}) = 
\sum_{i=1}^{k} \left\|\,\mathbf{a}_{t+i}^{h \rightarrow r} - \hat{\mathbf{a}}_{t+i}^{h \rightarrow r}\right\|_2^2,
$ where 
$
\hat{\mathbf{a}}_{t:t+k}^{h \rightarrow r} = \mathcal{P}_{\boldsymbol{\phi}}\left(\tilde{\mathbf{a}}_{t:t+k}^{h \rightarrow r} \,\middle|\, \mathbf{z}_t^h\right).
$ 

\textbf{Co-training Loss.} During co-training, each batch includes an equal proportion of robot and human data, and the total loss is the sum of both: $\mathcal{L}_{\text{total}}(\boldsymbol{\phi})=\mathcal{L}_{\text{robot}}(\boldsymbol{\phi})+\mathcal{L}_{\text{human}}(\boldsymbol{\phi})$.

\subsection{Mapping-guided MixUp}
\label{mixup}

To create a continuum of intermediate domains in latent space such that the source and target domain on a smooth manifold~\cite{dlow}, we propose a mapping-guided MixUp method. 

\textbf{Mapping.} To construct the affinity between human and robot demonstrations, we compute a mapping $\mathcal{M}_{h \rightarrow r}$ between human demonstration $\mathcal{D}^h$ and
robot demonstration $\mathcal{D}^r$ using \textit{Dynamic Time Warping (DTW)}~\cite{dtw}, based on either visual or action distance. Mapping $\mathcal{M}_{h \rightarrow r}(t)$ denotes the set of given human timestep $t$ mapped with robot timesteps across multiple demonstrations.
This mapping assumes that mapped human and robot segments with similar visual or action patterns correspond to shared states~\cite{dynamo}. 
Similar to retrieval-based methods~\cite{strap}, DTW ensures temporal consistency and avoids implausible supervision. 
We explore two mapping strategies: (1)  \textit{Action-based Mapping.}  
We define the action distance between a retargeted human demonstration and a robot demonstration as a weighted sum of several components:
$
d_{\text{act}} = \|\mathbf{t}^{h \rightarrow r} - \mathbf{t}^{r} \|_1 
+ \lambda_1 \| \mathbf{p}^{h \rightarrow r} - \mathbf{p}^{r} \|_1 
+ \lambda_2\, d_{\text{rot}}\bigl(\mathbf{o}^{h \rightarrow r}, \mathbf{o}^{r}\bigr),
$
where $\mathbf{t}$ denotes the translation, $\mathbf{p}$ the hand pose, $\mathbf{o}$ the orientation, $d_{\text{rot}}$ the angular distance and $\lambda_1, \lambda_2$ are weighting coefficients, and (2) \textit{Visual-based Mapping.}  
Here, we compute the frame-wise distance using extracted visual features and temporal alignment:
$
d_{\text{vis}} = \| \mathbf{f}^{h \rightarrow r} - \mathbf{f}^{r} \|_2,
$
where $\mathbf{f}$ represents visual features extracted from a pretrained encoder.

\textbf{MixUp-based Interpolation.} 
After establishing the mapping, we apply MixUp~\cite{mixup} to interpolate between original human and robot data, creating interpolated human data. 
During co-training, we train jointly on both the interpolated human data and the original robot data, serving as both regularization and augmentation. 

During training, we apply MixUp on both the condition inputs and the predicted actions.
At each training iteration, for each human timestep $t$, we randomly sample a robot timestep $t' \in \mathcal{M}_{h \rightarrow r}(t)$ and construct the mixed condition input and predicted action as:
\begin{equation}
\mathbf{z}_t^{\text{mix}} = \alpha \cdot \mathbf{z}^h_t + (1 - \alpha) \cdot \mathbf{z}_{t'}^r, \quad
\mathbf{a}_{t:t+k}^{\text{mix}} = \alpha \cdot \mathbf{a}_{t:t+k}^{h \rightarrow r} + (1 - \alpha) \cdot \mathbf{a}_{t':t'+k}^r
\end{equation}
where $\mathbf{z}^h_t$ and $\mathbf{a}_{t:t+k}^{h \rightarrow r}$ are the condition input and retargeted action of the raw human data, and $\mathbf{z}_{t'}^r$, $\mathbf{a}_{t':t'+k}^r$ come from the mapped robot demonstration. 
Inspired by DLOW~\cite{dlow}, we adopt a progressive interpolation strategy that gradually decreases the coefficient $\alpha$ during training, enabling smoother domain adaptation.

\subsection{Inference}

At test time, actions are predicted at a fixed inference frequency in the timestep of $k$. 
At each inference step $t$, an upsample rate $\gamma$, which is calculated from the duration of teleoperation and consistent with the rate used in training, is applied to both observations and predicted actions  (details in the Supp. A).
The condition $\mathbf{z}^r_t$ is constructed using an observation history of length $\tau$.
A future action sequence $\mathbf{a} = (\hat{\mathbf{a}}_{t+1}, \ldots, \hat{\mathbf{a}}_{t+k})$ is predicted from random noise.
For stability, temporal ensembling is applied with a decaying weight factor to average overlapping predictions across timesteps.

\section{Experiment Setups}
\label{sec:results}


\textbf{Hardware setup.}
We conduct experiments using a Franka Emika Panda robot arm equipped with four types of end-effectors (see Fig.~\ref{fig:teaserfig}): (1) Robotiq 2F-85 Fripper (2-finger), (2) Fin Ray Fripper (2-finger), (3) Allegro Hand (4-finger), and (4) Ability Hand (5-finger). These devices provide a range of dexterity and serve to evaluate embodiment transfer under different hardware configurations.

\textbf{Tasks.}
We introduce two categories of manipulation tasks, desinged to target increasing levels of control difficulty and embodiment demands:
\emph{(1) Basic Object Manipulation.}
These tasks assess coarse end-effector control and general spatial positioning:
\textbf{Pick and Place:} The robot must pick up a cube from a random initial position and place it precisely at a designated goal location.
\textbf{Push:} The robot must push a cube across a tabletop surface into a specified goal region.
\emph{(2) Tool-based Manipulation.}
These tasks evaluate the robot’s ability to manipulate external tools as a proxy for object interaction:
\textbf{Hammer:} The robot picks up a hammer and strikes a target point.
\textbf{Flip:} The robot uses a spatula to flip a bagel off the surface.

\textbf{Baselines.}
We compare against the following baselines in our experiments: Robot-only (Training diffusion policy using only robot data), Two-stage Fine-Tuning (Pretraining on human videos, followed by fine-tuning with robot data), Vanilla Co-Training (Simultanous training on both human and robot data), Random Mapping (Randomly pairing human and robot data for MixUp), Visual Mapping (ImMimic-V, using DTW with visual feature for mapping), and Action Mapping (ImMimic-A, using DTW with action for mapping).

\textbf{Metrics.}
We evaluate performance using three key metrics:
\emph{(1). Success Rate (SR).}
The proportion of 10 rollouts that successfully complete the task, scored in a binary manner (success or failure).
\emph{(2). Trajectory Smoothness (SPARC).}
Trajectory smoothness is quantified using the Spectral Arc Length (SPARC)~\cite{sparc}, which measures the smoothness in the frequency domain.
\emph{(3). Action Distance (AD).}
The average distance of translation and orientation after DTW for trajectory similarity.

\begin{table}[!tbp]
    \centering
    \resizebox{0.9\linewidth}{!}{
    \begin{tabular}{c|cccc|cccc}
        \toprule
        \multirow{2}{*}{\textbf{Setting}} 
        & \multicolumn{4}{c|}{\textbf{Pick and Place}} 
        & \multicolumn{4}{c}{\textbf{Push}} \\
        & \textbf{Robotiq} & \textbf{FR} & \textbf{Allegro} & \textbf{Ability} 
        & \textbf{Robotiq} & \textbf{FR} & \textbf{Allegro} & \textbf{Ability} \\
        \midrule
        Robot Only & 0.40 & 1.00 & 0.00 & 0.80 & 0.00 & 0.60 & 1.00 & 1.00 \\
        Co-Training & 0.40 & 1.00 & 1.00 & 0.80 & 0.20 & 0.60 & 1.00 & 1.00 \\
        ImMimic-A & \textbf{1.00} & 1.00 & \textbf{1.00} & \textbf{1.00} & \textbf{0.40} & \textbf{0.70} & 1.00 & 1.00 \\
        \midrule
        \multirow{2}{*}{\textbf{Setting}} 
        & \multicolumn{4}{c|}{\textbf{Hammer}} 
        & \multicolumn{4}{c}{\textbf{Flip}} \\
        & \textbf{Robotiq} & \textbf{FR} & \textbf{Allegro} & \textbf{Ability} 
        & \textbf{Robotiq} & \textbf{FR} & \textbf{Allegro} & \textbf{Ability} \\
        \midrule
        Robot Only & 0.20 & 0.90 & 0.00 & 0.00 & 0.60 & 0.60 & 0.00 & 0.60 \\
        Co-Training & 0.40 & 0.80 & 0.00 & 0.00 & 0.60 & 0.80 & 0.00 & 0.90 \\
        ImMimic-A & \textbf{0.50} & \textbf{1.00} & \textbf{0.20} & 0.00 & \textbf{1.00} & \textbf{0.80} & \textbf{0.20} & \textbf{1.00} \\
        \bottomrule
    \end{tabular}
    }
    \caption{Success rates of Robot-Only, Co-Training, and ImMimic-A across four embodiments and four tasks. Policies are trained using 5 robot demonstrations and 100 human demonstrations.}
        \vspace{-1em}
    \label{tab:maintab}
\end{table}

\begin{table}[!tbp]
    \centering
    \begin{minipage}{0.52\textwidth}
        \centering
        \vspace{-3pt}
        \resizebox{\textwidth}{!}{
        \begin{tabular}{l|cc|cc}
            \toprule
            \multirow{2}{*}{\textbf{Setting}} 
            & \multicolumn{2}{c|}{\textbf{Robotiq}} 
            & \multicolumn{2}{c}{\textbf{Ability}} \\
            & \textbf{Pick and Place} & \textbf{Flip} 
            & \textbf{Pick and Place} & \textbf{Flip} \\
            \midrule
            Robot Only  & 0.40 & 0.60 & 0.80 & 0.60 \\
            \midrule
            Fine-Tuning & 0.80 & 0.70 & 0.50 & 0.40 \\
            Co-Training & 0.40 & 0.80 & 0.80 & 0.90 \\
            \midrule
            Random Mapping & 0.40 & 0.50 & 0.80 & 0.50 \\
            ImMimic-V & \textbf{1.00} & 0.50 & 0.90 & 0.70 \\
            ImMimic-A & \textbf{1.00} & \textbf{1.00} & \textbf{1.00} & \textbf{1.00} \\
            \bottomrule
        \end{tabular}
        }
        \caption{Comparison of success rate across two embodiments (Robotiq, Ability) and two tasks (Pick and Place, Flip), with 5 robot demos and 100 human demos.}
        \vspace{-1em}
        \label{tab:ablation_study}
    \end{minipage}
    \hfill
    \begin{minipage}{0.47\textwidth}
        \centering
        \vspace{3pt}
        \resizebox{\textwidth}{!}{
        \begin{tabular}{l|c|c|c}
            \toprule           \makecell{\textbf{Embodi-}\\ \textbf{{ment}}} & \makecell{Rollout \\ (Robot Only)} & \makecell{Rollout \\ (Co-Training)} & \makecell{Rollout \\ (ImMimic-A)} \\
            \midrule
            Robotiq & -12.7694 & -9.6533 & \textbf{-9.4424}  \\
            FR & -24.4935 & \textbf{-14.3644} & -15.6430 \\
            \midrule
            Ability & -13.9228 & -10.9241 & \textbf{-10.8409} \\
            Allegro & N/A & -17.1312 & \textbf{-13.8940} \\
            \bottomrule
        \end{tabular}
        }
        \caption{Spectral Arc Length (SPARC) smoothness scores ($\uparrow$) on Pick and Place. A higher score indicates a smoother trajectory. We evaluate average scores on 5 successful rollouts over 3 methods.}
        \label{tab:sparc_smoothness}
    \end{minipage}
\end{table}

\begin{figure}[!tbp]
    \centering
    \vspace{-0.5em}
    \begin{minipage}{0.24\textwidth}
        \centering
        \includegraphics[width=\linewidth]{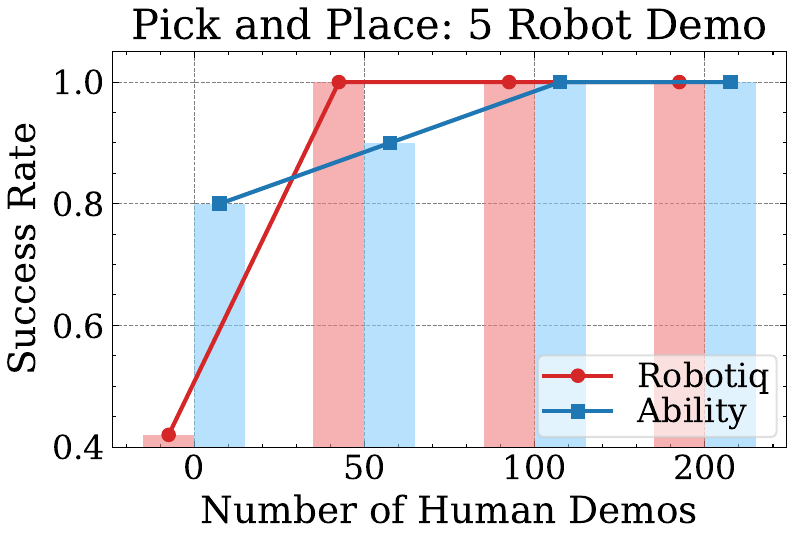}
    \end{minipage}
    \hfill
    \begin{minipage}{0.24\textwidth}
        \centering
        \includegraphics[width=\linewidth]{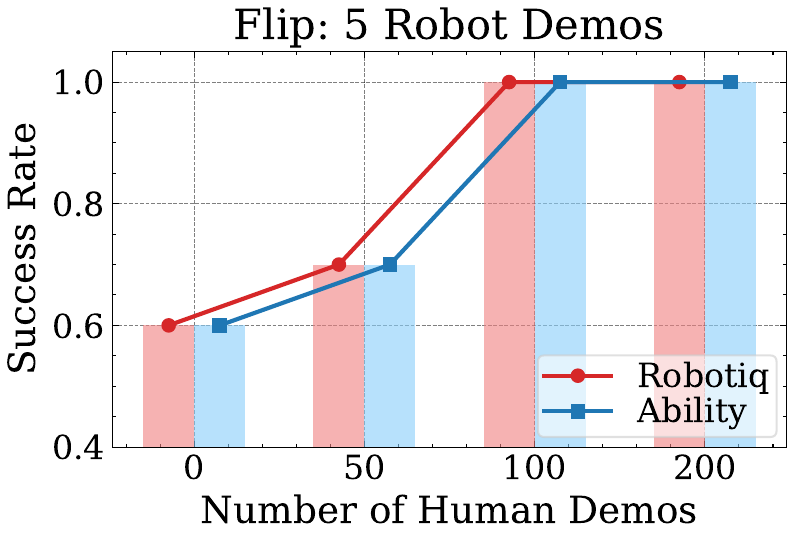}
    \end{minipage}
    \hfill
    \begin{minipage}{0.24\textwidth}
        \centering
        \includegraphics[width=\linewidth]{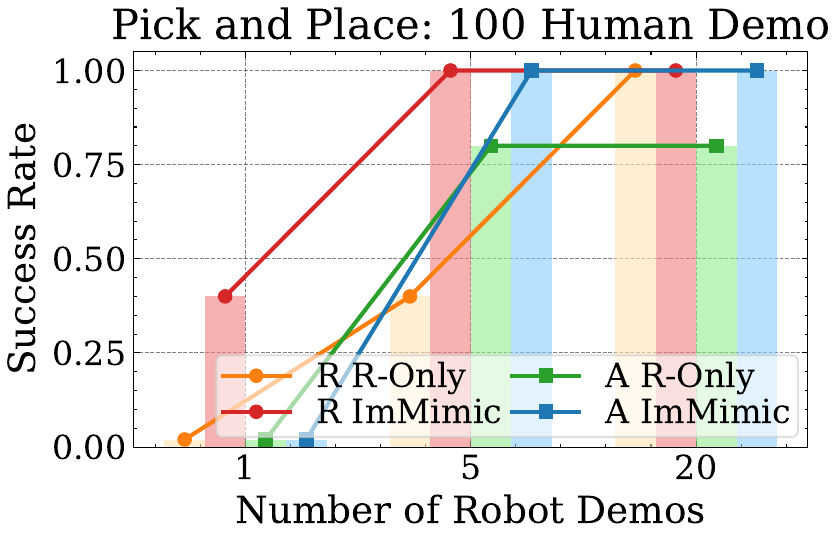}
    \end{minipage}
    \hfill
    \begin{minipage}{0.24\textwidth}
        \centering
        \includegraphics[width=\linewidth]{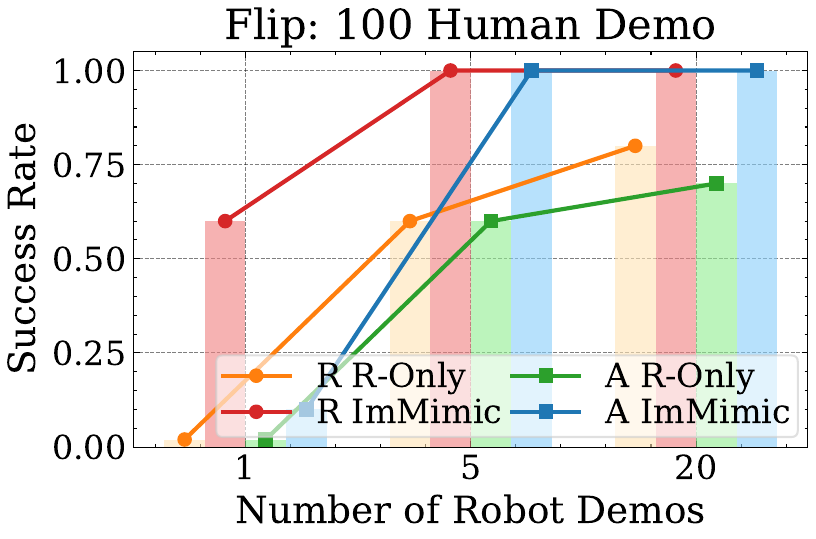}
    \end{minipage}
    
    \begin{minipage}{0.49\textwidth}
        \centering
        \caption{Sample efficiency of ImMimic-A with varying numbers of human demonstrations.}
        \label{fig: Human sample}
    \end{minipage}
    \hfill
    \begin{minipage}{0.49\textwidth}
        \centering
        \caption{Sample efficiency of ImMimic-A and Robot-Only with varying numbers of robot demonstrations.}
        \label{fig: Robot sample}
    \end{minipage}
    \vspace{-2em}
\end{figure}

\section{Core Results}

\textbf{Human videos enhance the robustness and smoothness of learned policies.}  
Leveraging human videos substantially improves policy success rates as shown in Tab.~\ref{tab:maintab}. 
As shown in Tab.~\ref{tab:ablation_study}, policies trained with ImMimic-A consistently achieve higher success rates across all tasks and embodiments compared to robot-only, two-stage fine-tuning, and co-training baselines. 
These results indicate that learning from human videos using our method improves the robustness of robot rollouts, as the interpolated human data effectively serves as data augmentation for the limited robot data. For example, ImMimic-A is more robust to variations in object positions (Fig.~\ref{fig:failures}(c)).
Furthermore, ImMimic improves action smoothness. In Tab.~\ref{tab:sparc_smoothness}, we show that it achieves higher SPARC scores compared to robot-only policies, indicating smoother trajectories. 
It also outperforms vanilla co-training on three out of four embodiments. These results together suggest that our method effectively enhances the robot policy by leveraging prior knowledge from human demonstrations.

\begin{wrapfigure}{r}{0.4\textwidth}
  \vspace{-1em}
  \begin{minipage}{\linewidth}
    \centering
    \includegraphics[width=\linewidth]{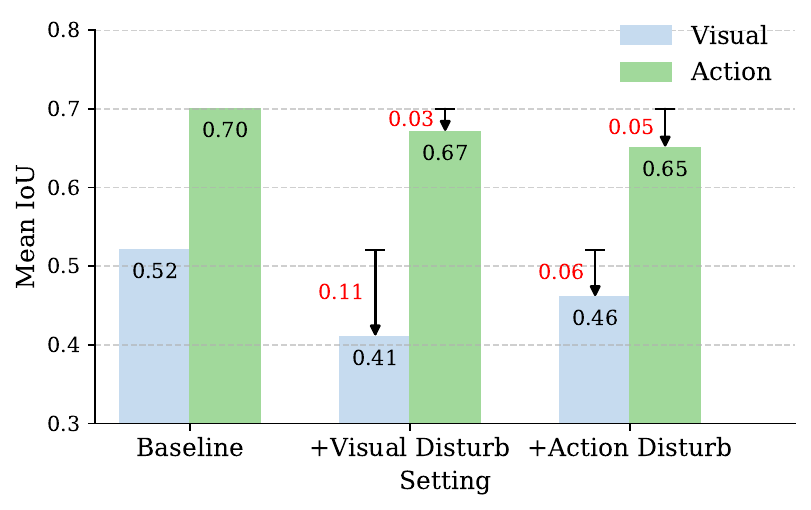}
    \vspace{-2em}
    \caption{Comparison of Mean IoU across different disturbance settings.}
    \label{fig:mean_iou}
    \vspace{-1em}
  \end{minipage}
\end{wrapfigure}
\textbf{Interpolation with Action-based Mapping leads to better performance.}
We compare action-based and visual-based mapping to evaluate their effectiveness in bridging human-robot domain gap. 
As shown in Tab.~\ref{tab:ablation_study}, action-based mapping (ImMimic-A) consistently outperforms visual-based mapping (ImMimic-V) and random mapping. 
This performance gain is attributed to the fact that retargeted human actions, aligned via kinematic constraints, are structurally more similar to robot actions than visual features are to robot observations.
In a separate long-horizon video retrieval task (details in Supp. C), we extend DTW to retrieve robot-relevant subsequences from unsegmented human videos. 
The results in Fig.~\ref{fig:mean_iou} show that action-based mapping can be more accurate and robust with visual and action disturbance. 
Fig.~\ref{fig:failures}(e) shows ImMimic-V failing due to poor mapping, causing the robot to loop in place. Especially in task with subtle action transitions, weak visual mapping degrades performance, highlighting that mapping quality is critical, and training with action-based mapping leads to a more reliable robot policy.

\textbf{ImMimic leads to consistent improvement across embodiments.}
ImMimic consistently enhances policy performance across different end-effectors, regardless of their morphological similarity to the human hand. 
As shown in Tab.~\ref{tab:maintab}, ImMimic-A improves task success across all embodiments compared to the Robot-Only baseline, and outperforms or matches the performance of Co-Training. This demonstrates that ImMimic-A effectively adapts to various tasks and embodiments.

However, for certain embodiment-task, success rates remain low despite using our method.
For Hammer with Ability Hand ($0.0$ SR), Fig.~\ref{fig:failures}(d) shows that the short thumb causes the index finger to make unintended contact with the hammer, leading to misoriented grasp.
For Flip with Allegro Hand ($\le 0.2$ SR), Fig.~\ref{fig:failures}(g) shows a failure case where the hand cannot firmly grasp the spatula due to its large size.
These cases show the essential effects of embodiment structure on task performance.

\begin{figure*}[t]
  \centering
  \includegraphics[width=1.0\linewidth]{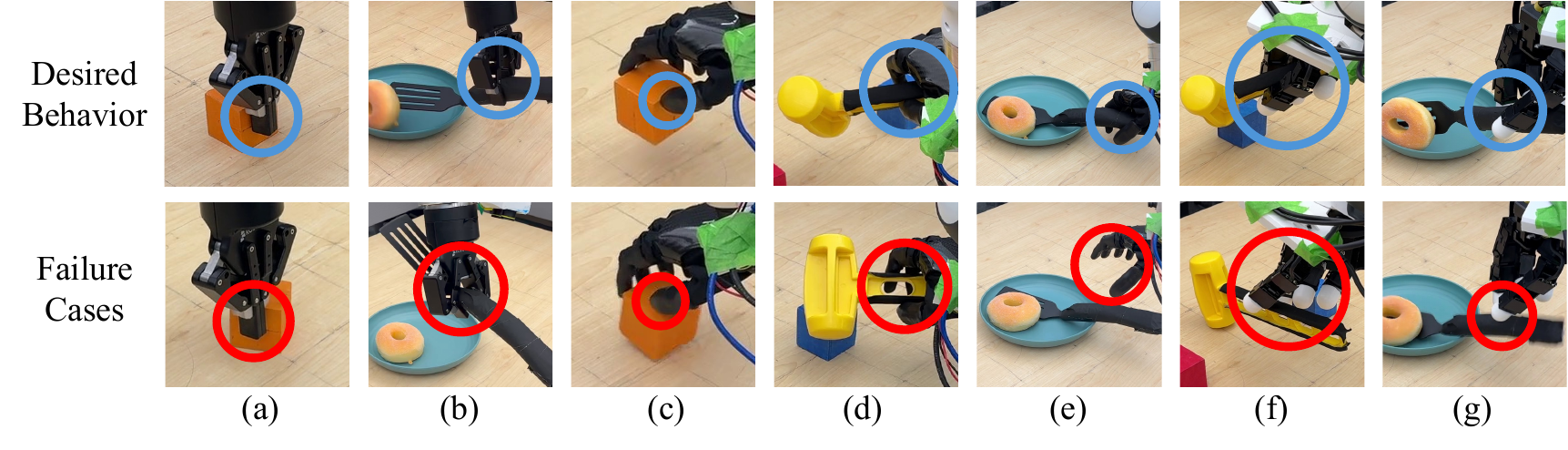}
  \caption{Desired behavior and corresponding failure cases. (a) Unstable push due to thin tip. (b) Unstable grasp from structural gap. (c) Grasp failure due to variations in object positions. (d) Poor hammer grasp from bad initial contact. (e) Motion trap due to weak visual mapping. (f) Insufficient gripping force under heavy hammer weight. (g) Infirm grasp of spatula due to large hand.}
  \label{fig:failures}
  \vspace{-1em}
\end{figure*}

\textbf{More human-mimetic embodiments don't necessarily lead to better transfer.} Intuitively speaking, human-mimetic embodiments should exhibit smaller action distance to human demonstrations, but our results show otherwise. 
Average AD (Action Distance) shows that two dexterous hands demonstrate larger action distances (Allegro: 0.078, Ability: 0.075) compared to the two grippers (Robotiq: 0.066, FR: 0.065).
Moreover, Tab.~\ref{tab:maintab} show that policies benefit more from human videos when the action distance is smaller, regardless of its mechanical structure. This is potentially due to the fact that in addition to hand design, mounting configuration and arm kinematics also influence the action retargeting and the way robot performs the task.

These observations are likely to offer useful insights for end-effector design as well.
In Fig.~\ref{fig:failures}(a,b), Robotiq's thin fingertips and palm gap cause unstable contact and slipping.  
In Fig.~\ref{fig:failures}(d), Ability's shorter thumb may contribute to misaligned grasps when position offsets are present.
In Fig.~\ref{fig:failures}(f,g), Allegro's larger size appears to limit its ability to lift the heavy hammer or grasp the spatula firmly.  
Overall, features such as longer fingertips, extended thumb reach, and higher grasping force may support more robust performance across a range of manipulation tasks.

\noindent\textbf{The scale and diversity of human demonstrations enhance learning performance.} 
Human videos exhibit greater diversity than robot data, as reflected by a higher intra-dataset Action Distance (AD) (0.012 vs.\ 0.005). In Fig.~\ref{fig: Human sample}, for Pick and Place, adding 50 human videos raises success rate (SR) from 0.4 to 1.0 for the Robotiq and from 0.8 to 0.9 for the Ability; both reach 1.0 by 100 videos. 
Conversely, in Fig.~\ref{fig: Robot sample}, with 100 human videos fixed, ImMimic-A achieves 1.0 SR with only 5 robot demonstrations, while the robot-only baseline requires 20 demos but still underperforms. 
These results suggest that incorporating human data can significantly improve sample efficiency when combined with a small amount of robot data.

\section{Conclusion}
We present ImMimic, a novel embodiment-agnostic co-training framework that unites large-scale human videos with few-shot robot demonstrations. To bridge the domain gap between human and robot data, ImMimic leverages DTW-based mapping and MixUp to interpolate between mapped human-robot pairs, creating intermediate domains that enable smooth domain adaptation during co-training. Evaluation on four tasks and four embodiments demonstrates consistent improvements in task success rate and rollout smoothness. Additionally, we find that mapping based on action similarity between retargeted human and robot actions, rather than visual context, leads to improved policy performance, suggesting that human hand trajectories offer rich supervision for robot learning. We also identify several failure cases, attributed to either hardware design or limitations in the learning method, and observe that a more human-like hand does not necessarily yield better performance.
\section{Limitations}

Exisiting limitation of ImMimic includes:
(1) \textbf{Large domain gap leads to performance drop.}  
Although ImMimic outperforms baselines across embodiments in most of the tasks, its performance is still degraded under even larger domain gaps, such as significant differences in average action distances between embodiments and humans, or major visual appearance differences.
Future directions may include improved representation learning to better align the features even across larger domain gaps.
(2) \textbf{Inconsistent gains across embodiments potentially indicate that policy performance is influenced by the robot's structural design.}
While ImMimic consistently improves success rates and smoothness across all four embodiments shown in Tab.~\ref{tab:maintab} and Tab.~\ref{tab:ablation_study}, the magnitude of these gains varies.
In future work, we aim to empirically investigate how embodiment design impacts policy performance when learning from human demonstrations, with the ultimate goal of developing a unified system that enables robots to more effectively acquire and adapt human skills.

\section*{Acknowledgements}
This project is partially supported by the Samsung Research America LEAP-U Program and a gift from Meta Platforms, Inc.

\clearpage
\bibliography{main}  
\section*{Appendix} 
\addcontentsline{toc}{section}{Appendix} 

\renewcommand{\thesection}{\Alph{section}} 

\counterwithin{figure}{section}
\counterwithin{table}{section}
\numberwithin{equation}{section}

\appendix
\renewcommand{\thesection}{\Alph{section}} 

\section{Demonstration Collection System}
\begin{figure*}[h]
  \centering
       \centering
       \includegraphics[width=1.0\linewidth]{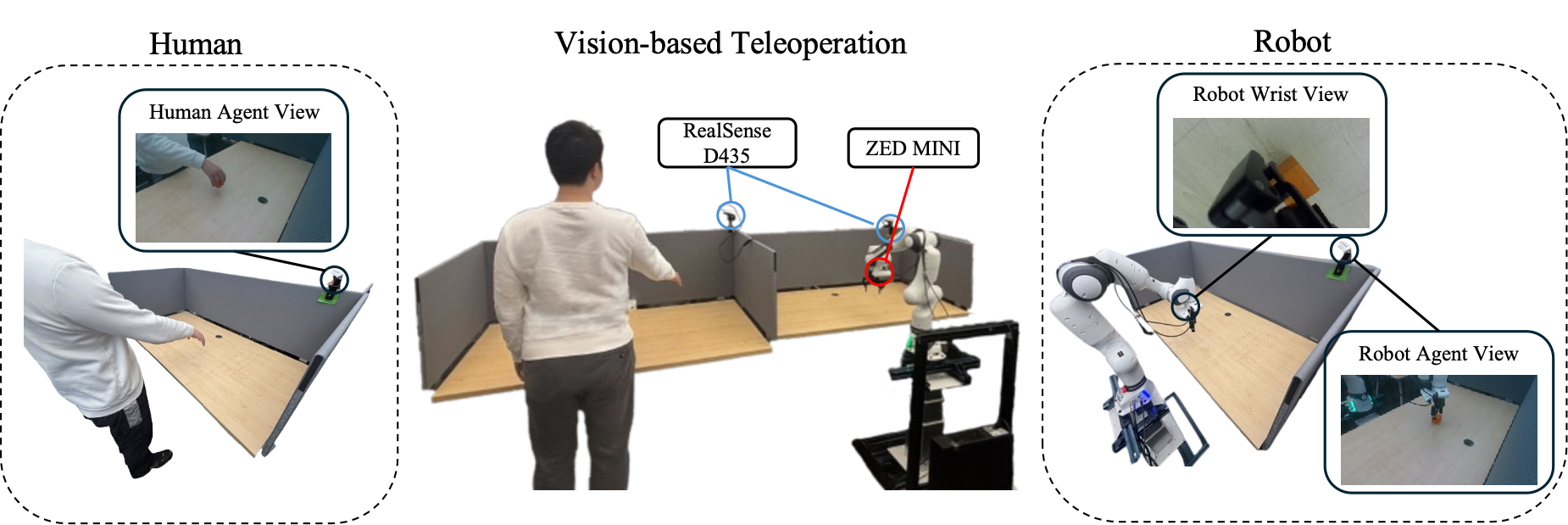}
       \caption{Overall data collection system. (1) For human demonstrations, only the agent-view camera is used. (2) For robot demonstrations, both the agent-view and wrist-view cameras are used to enable precise control. (3) For teleoperation, a separate workspace is placed to the left of the robot, and a camera with identical intrinsics and calibration is used for vision-based control.}
       \label{fig:system}
       \vspace{-1em}
\end{figure*}

The overall data collection system is illustrated in Fig.~\ref{fig:system}. 
We collect both human demonstration videos and robot teleoperation data to establish a comprehensive dataset for our study.
To minimize visual gap between human and robot demonstrations, we use the same RealSense D435 camera for both. 
Demonstrations are recorded from a fixed viewpoint that captures the entire workspace and clearly shows hand-object interactions.

\subsection{Data Collection Throughput}
As shown in Tab.~\ref{tab:teleop}, we report the teleoperation throughput for each embodiment on each task in terms of: (1) \textbf{Frequency} -- the average number of successful demonstrations recorded per minute, (2) \textbf{Success Rate} -- the ratio of successful demonstrations to total attempts, and (3) \textbf{Duration} -- the average length of all successful demonstrations.
Due to structural differences and varying task difficulty, these metrics differ across embodiments and tasks.
These trends also strongly correlate with the final policy performance.
For Hammer, using Allegro Hand and Ability Hand for teleoperation shows low success rates ($\leq 0.3$) and  require longer durations due to the need for precise wrist angle adjustments during teleoperation. 
This aligns with the policy rollout results, where the policies learned with these embodiments also exhibit low rollout success rates ($\leq 0.2$). 
In contrast, for the same tasks, using the Robotiq Gripper and FR Gripper for teleoperation shows better performance, and the policies trained for these embodiments achieve higher performance.

\begin{table}[h]
    \centering
    \resizebox{0.8\textwidth}{!}{%
    \begin{tabular}{l|l|cccccc}
        \toprule
        \textbf{Method} & \textbf{Metric} & \textbf{Pick and Place} & \textbf{Push} & \textbf{Hammer} & \textbf{Flip} \\
        \midrule
        \multirow{3}{*}{Human Demo}  
            & Frequency     & 5.4 & 6.7  & 2.8 & 3.4 \\
            & Success Rate  & 1.00 & 1.00  & 1.00 & 0.98 \\
            & Duration  & 2.66 & 1.59 & 4.66 & 2.52 \\
        \midrule
        \multirow{3}{*}{\makecell[l]{Vision-based Teleop\\(Robotiq)}}  
            & Frequency     & 1.47 & 1.33   & 1.05 & 0.45 \\
            & Success Rate  & 0.82 & 0.88   & 0.48 & 0.28 \\
            & Duration  & 8.33 & 9.17  & 12.73 & 7.54 \\
        \midrule
        \multirow{3}{*}{\makecell[l]{Vision-based Teleop\\(FR)}}  
            & Frequency     & 1.4 & 1.52  & 0.83 & 0.76 \\
            & Success Rate  & 0.83 & 0.88& 0.52 & 0.46 \\
            & Duration  & 12.87 & 17.04 & 16.23  & 11.04 \\
        \midrule
        \multirow{3}{*}{\makecell[l]{Vision-based Teleop\\(Allegro)}}  
            & Frequency     & 1.42 & 1.67  & 0.12 & 0.43 \\
            & Success Rate  & 0.70 & 0.86  & 0.04 & 0.32 \\
            & Duration  & 15.43 & 10.99  & 21.31 & 14.78 \\
        \midrule
        \multirow{3}{*}{\makecell[l]{Vision-based Teleop\\(Ability)}}  
            & Frequency     & 1.21 & 2.05  & 0.38 & 0.59 \\
            & Success Rate  & 0.68 & 0.91  & 0.22 & 0.45 \\
            & Duration  & 16.09 & 10.12 & 18.28 & 13.86 \\
        \bottomrule
    \end{tabular}
    }
    \caption{Frequency (number of successful demonstrations collected per minute), Success Rate (ratio of successful demonstrations) and Duration (average duration of all demonstrations) for human demonstrations and vision-based teleoperation across four tasks using four different end-effectors: Robotiq, Fin Ray, Allegro, Ability.}
    \label{tab:teleop}
    \vspace{-1em}
\end{table}

\begin{table}[h]
    \centering
    \begin{tabular}{l|ccccccc}
        \toprule
        \textbf{Method} & \textbf{Pick and Place} & \textbf{Push} & \textbf{Hammer} & \textbf{Flip} \\
        \midrule
        \multirow{1}{*}{Human Demo}  
            & 32 & 32 & 32 & 32 \\
        \midrule
        \multirow{1}{*}{\makecell[l]{Robotiq}}  
            & 100 & 185 & 87 & 96 \\
        \midrule
        \multirow{1}{*}{\makecell[l]{FR}}  
            & 155 & 343 & 112 & 140 \\
        \midrule
        \multirow{1}{*}{\makecell[l]{Allegro}}  
            & 185 & 221 & 146 & 188 \\
        \midrule
        \multirow{1}{*}{\makecell[l]{Ability}}  
            & 193 & 204 & 126 & 176 \\
        \bottomrule
    \end{tabular}
    \caption{Sample rate $\gamma$ used during training and inference. It is computed as the ratio between the durations of human and robot demonstrations and is used to subsample robot data during training and upsample predicted actions during inference.}
    \label{samplerate}
    \vspace{-1em}
\end{table}

\subsection{Sample Rate Normalization}
To enable consistent training and inference across human and robot demonstrations, we define a sample rate $\gamma$ that compensates for the difference in demonstration durations. 
As shown in Tab.~\ref{tab:teleop}, human demonstrations tend to be faster, while teleoperated robot demonstrations take longer time. 
To align their temporal coverage, we fix the action sequence length $k = 32$ for human demonstrations, then compute $\gamma$ as the ratio of robot to human demonstration durations. Using this value, we uniformly subsample $\gamma$-spaced frames from each robot demonstration to produce a $k$-step sequence that spans a comparable duration.

During training, we use an observation history length $\epsilon = 2$, where the policy predicts $k$ future actions based on $\epsilon$ past observations. For robot data, these observations are offset by $\gamma$, allowing the model to learn over a similar time horizon as in human data. This normalization helps mitigate issues caused by overly short prediction horizons in slower-paced robot trajectories.

At inference, we upsample the predicted $k$-step sequence using $\gamma$ to recover the original robot execution speed. The model performs inference every $k$ steps, and intermediate steps are filled via temporal ensembling of previously predicted actions with a decaying weight. This ensures smooth, continuous motion during rollout while maintaining consistency with the teleoperated control pace.

\subsection{Camera Calibration}
\label{calibration}


Accurate camera calibration is essential for both human and robot demonstrations. Before data collection, we calibrate the agent-view RealSense D435 camera used across our settings. For vision-based teleoperation, we use a separate RealSense D435 camera positioned over a dedicated workspace to the left of the robot for RGBD-based hand pose estimation and retargeting. This camera shares the same intrinsic parameters and calibration with the agent-view camera.

We now describe the camera calibration method used to transform retargeted human trajectories (extracted from human demonstration videos) from the camera coordinate frame to the robot base frame. Specifically, we aim to estimate the rigid transformation that maps 3D points and orientations from the camera frame to the robot base frame, denoted as:
\begin{equation}
    {}^{\text{base}}\mathbf{T}_{\text{cam}} =
    \begin{bmatrix}
        \mathbf{R} & \mathbf{t} \\
        \mathbf{0}^{\top} & 1
    \end{bmatrix},
\end{equation}
where $\mathbf{R} \in \mathrm{SO}(3)$ is a rotation matrix and $\mathbf{t} \in \mathbb{R}^3$ is a translation vector. In homogeneous coordinates, any point $\mathbf{p}_{\text{cam}}$ in the camera frame is transformed to the robot base frame via:
\begin{equation}
    \begin{bmatrix}
        \mathbf{p}_{\text{base}}\\[4pt]
        1
    \end{bmatrix}
    \;=\;
    {}^{\text{base}}\mathbf{T}_{\text{cam}}
    \begin{bmatrix}
        \mathbf{p}_{\text{cam}}\\[4pt]
        1
    \end{bmatrix}.
\end{equation}

\begin{figure}[ht!]
    \centering
    \begin{subfigure}[t]{0.45\linewidth}
        \centering
        \includegraphics[width=0.9\linewidth]{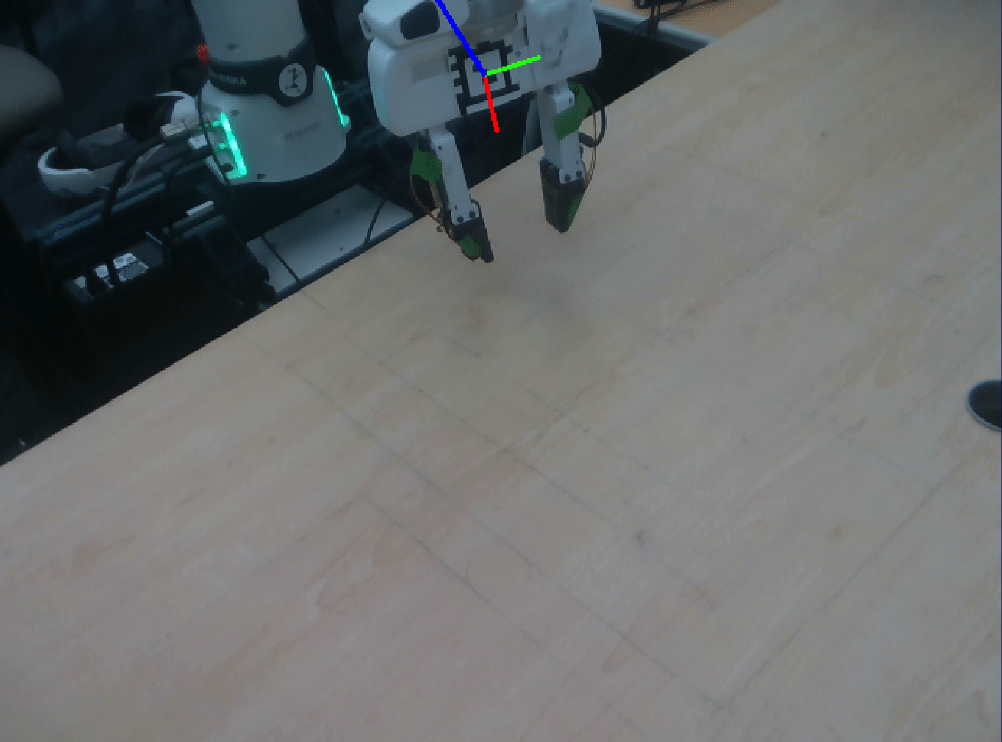}
        \caption{AprilTag used for camera calibration, enabling precise estimation of its 6-DoF pose in the camera frame.}
        \label{fig:apriltag}
    \end{subfigure}
    \hspace*{1em}
    \begin{subfigure}[t]{0.45\linewidth}
        \centering
        \includegraphics[width=0.9\linewidth]{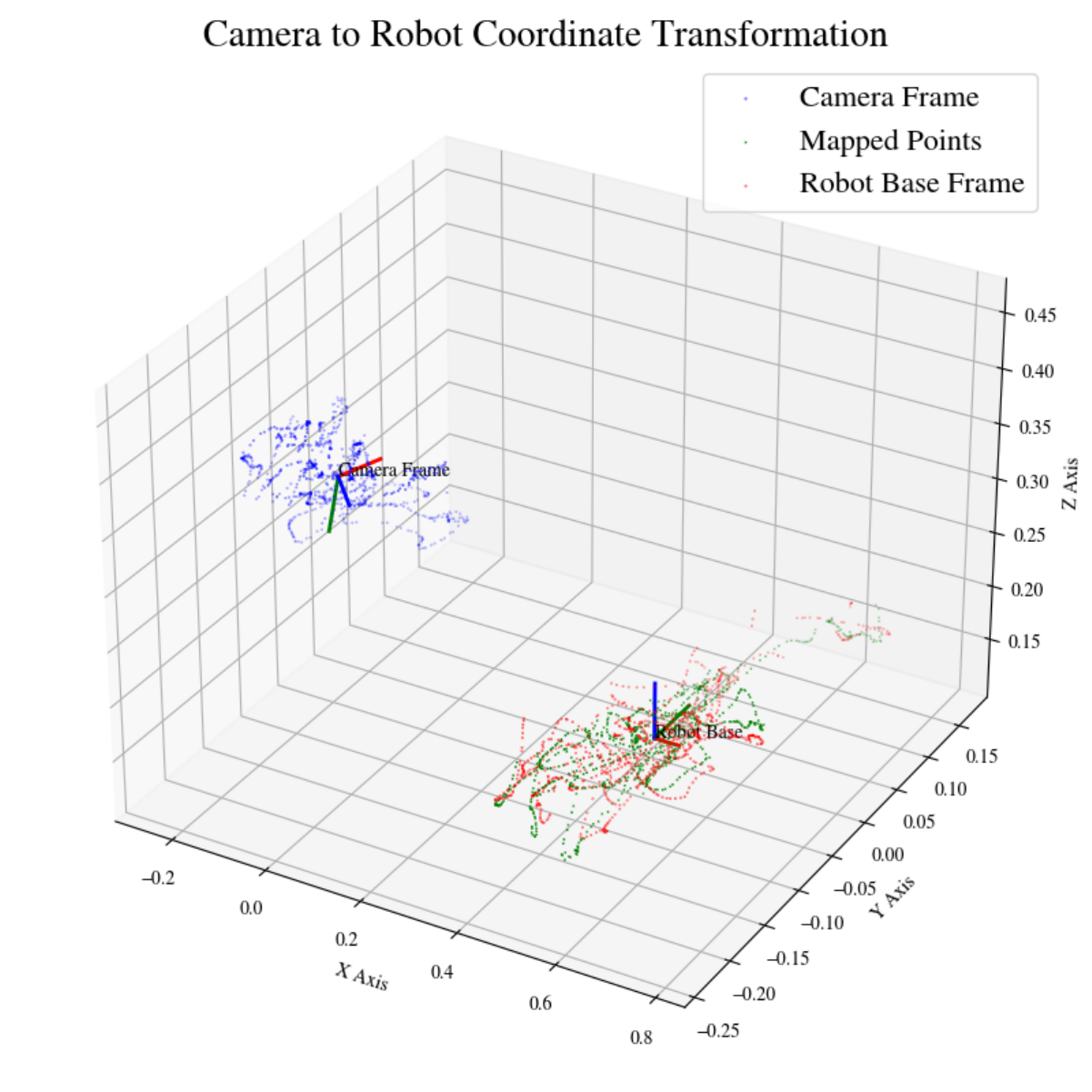}
        \caption{Calibration from the camera coordinate frame to the robot base frame.}
        \label{fig:calibration}
    \end{subfigure}
    \caption{Camera calibration process.}
    \vspace{-1em}
\end{figure}

To perform calibration, we attach an AprilTag to a known location (Fig~\ref{fig:apriltag}) such that its pose relative to the robot base is known, yielding ${}^{\text{base}}\mathbf{T}_{\text{tag}}$. The camera observes the tag, yielding ${}^{\text{tag}}\mathbf{T}_{\text{cam}}$. Combining these yields:
\begin{equation}
    {}^{\text{base}}\mathbf{T}_{\text{cam}} 
    = {}^{\text{base}}\mathbf{T}_{\text{tag}} 
      \,\bigl( {}^{\text{tag}}\mathbf{T}_{\text{cam}} \bigr)^{-1},
\end{equation}
Multiple such measurements enable us to refine $(\mathbf{R},\mathbf{t})$ using a best-fit procedure. Given $N$ pairs of corresponding points $\mathbf{p}_i^{\text{cam}}$ (in the camera frame) and $\mathbf{p}_i^{\text{rob}}$ (in the robot base frame), we estimate $(\mathbf{R},\mathbf{t})$ by minimizing:
\begin{align}
    \mathcal{L}(\mathbf{R}, \mathbf{t}) & = 
    \sum_{i=1}^{N} \|\mathbf{p}_i^{\text{rob}} - (\mathbf{R}\,\mathbf{p}_i^{\text{cam}} + \mathbf{t})\|^2, \\
    & \text{s.t.} \mathbf{R}^T\mathbf{R}=\mathbf{I}.
\end{align}
We use a quaternion-based parameterization of $\mathbf{R}$ to enforce $\mathrm{SO}(3)$ constraint and solve the problem via nonlinear least squares. 
The overall calibration procedure is illustrated in Fig~\ref{fig:calibration}.

\section{Retargeting}

In both human videos processing and vision-based teleoperation for robot data collection, we perform retargeting from human hand motion to robot actions. While the overall retargeting pipeline is shared across both settings, there are key differences. 
For human demonstration videos, we use offline retargeting based on RGB inputs and apply position retargeting, where absolute 3D joint positions are mapped to robot actions. 
For real-time vision-based teleoperation, we apply online retargeting that replaces wrist pose estimation with a more stable depth-based method and adopts vector retargeting~\cite{anyteleop}, which aligns finger segment orientations rather than absolute positions for teleoperation. 
This section provides additional details on the retargeting process.

\textbf{Human Pose Estimation and Wrist Localization.}
To estimate human hand pose, we use MediaPipe~\cite{mediapipe}, a real‑time pipeline that provides robust hand bounding boxes.  
Each cropped hand region is then passed to FrankMocap~\cite{frankmocap}, which outputs shape and pose parameters for an SMPL‑X model~\cite{smpl}, resulting in  accurate 3D coordinates for 21 knuckle joints in the local wrist frame.  

To improve spatial accuracy, particularly important for teleoperation, we replace FrankMocap's estimated wrist translation with a wrist point derived from depth data captured by an RGBD camera.
For wrist orientation, we apply the \textbf{Perspective‑n‑Point (PnP) algorithm}~\cite{handsdet}, solving:
\begin{equation}
R^*,t^* = \arg\min_{R,t}\sum_{i} \left\| \mathbf p_i - \Pi(R\mathbf P_i + t) \right\|^2
\end{equation}
where $\mathbf P_i$ are the 3D keypoints in the local frame, $\mathbf p_i$ are their 2D projections, $R\in SO(3)$ is the orientation matrix, $t$ is the translation vector, and $\Pi$ is the camera projection function.  
This yields a refined 6-DoF wrist pose that is consistent with the observed depth.

\textbf{Online Retargeting for Real-time Teleoperation.}
For real-time teleoperation, we adopt vector retargeting to ensure responsiveness and avoid kinematic singularities. Instead of matching absolute joint positions, we optimize finger orientations to follow the directions of human keypoint vectors.
Given keypoint vectors $\mathbf v_t^i$ from MediaPipe~\cite{mediapipe}, we solve for the robot joint configuration by minimizing:
\begin{equation}
\label{eq:vector_retarget}
\min_{\mathbf{q}_t}\sum_{i=1}^N \left\| \alpha\,\mathbf v_t^i - \mathbf R\,f_i(\mathbf q_t) \right\|^2
+ \beta\,\bigl\| \mathbf q_t - \mathbf q_{t-1} \bigr\|^2,
\quad \text{s.t.}\quad \mathbf q_l \le \mathbf q_t \le \mathbf q_u,
\end{equation}
where $f_i(\cdot)$ maps to the corresponding robot finger vector, $\mathbf R$ aligns coordinate frames, and $\alpha,\beta$ control the scaling and temporal smoothness.  
We solve this constrained optimization problem in under 10 ms per frame. 
To further reduce latency and improve motion continuity, we apply a low-pass filter with a smoothing parameter of 0.2 to suppress sudden keypoint fluctuations. This enables stable control and recording at 30 Hz.

\section{Long Raw Human Video Retrieval}

\begin{figure}[t]
    \centering

    \begin{subfigure}{0.9\textwidth}
        \centering
        \begin{minipage}{0.25\textwidth}
            \includegraphics[width=\linewidth]{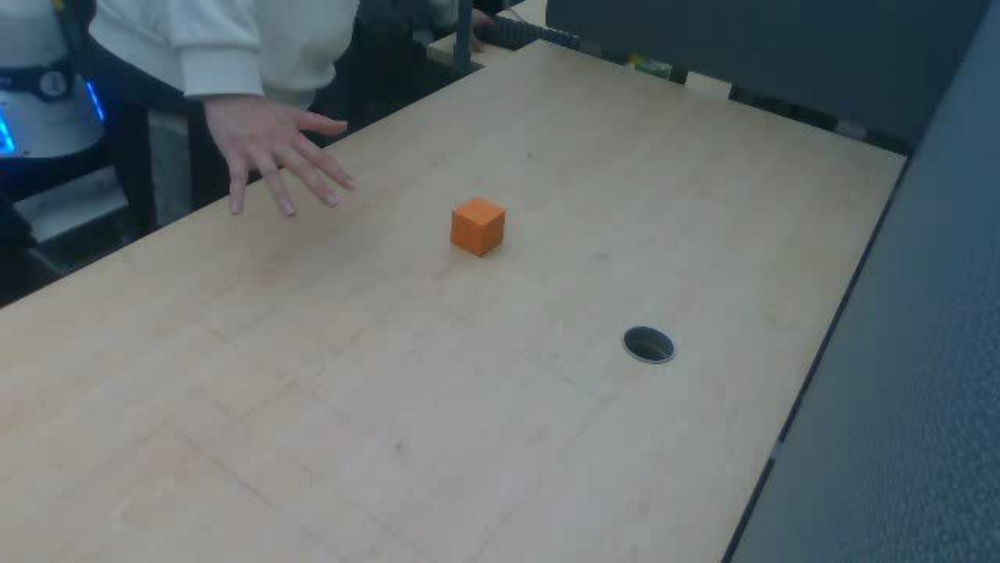}
        \end{minipage}
        \hfill
        \begin{minipage}{0.72\textwidth}
            \includegraphics[width=\linewidth]{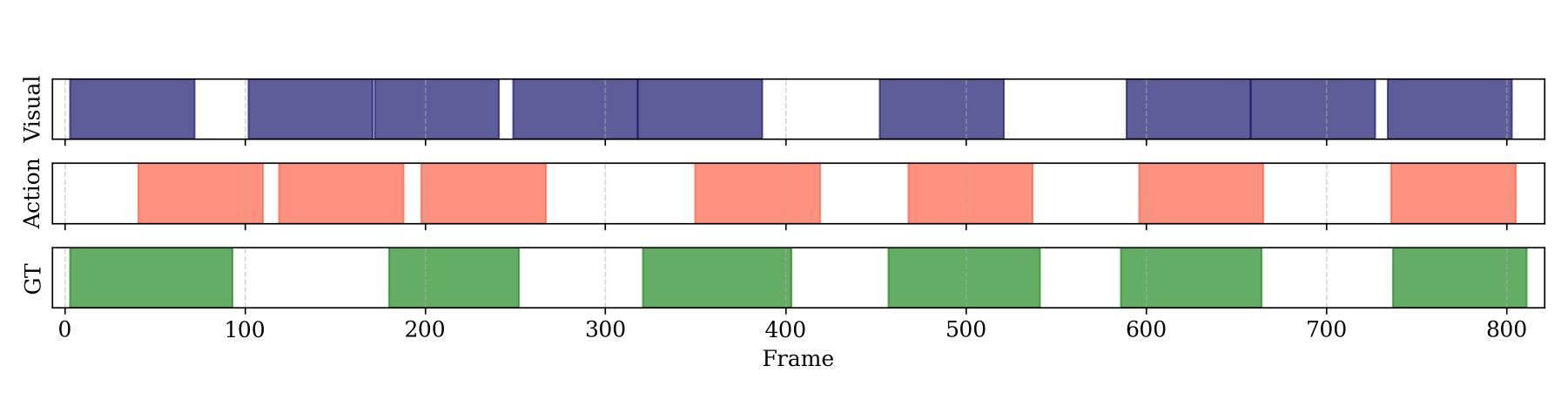}
        \end{minipage}
        \caption{Baseline retrieval for a Pick and Place task, the same setting we used for training.}
        \label{fig:baseline}
    \end{subfigure}

    \vspace{1.5em}

    \begin{subfigure}{0.9\textwidth}
        \centering
        \begin{minipage}{0.25\textwidth}
            \includegraphics[width=\linewidth]{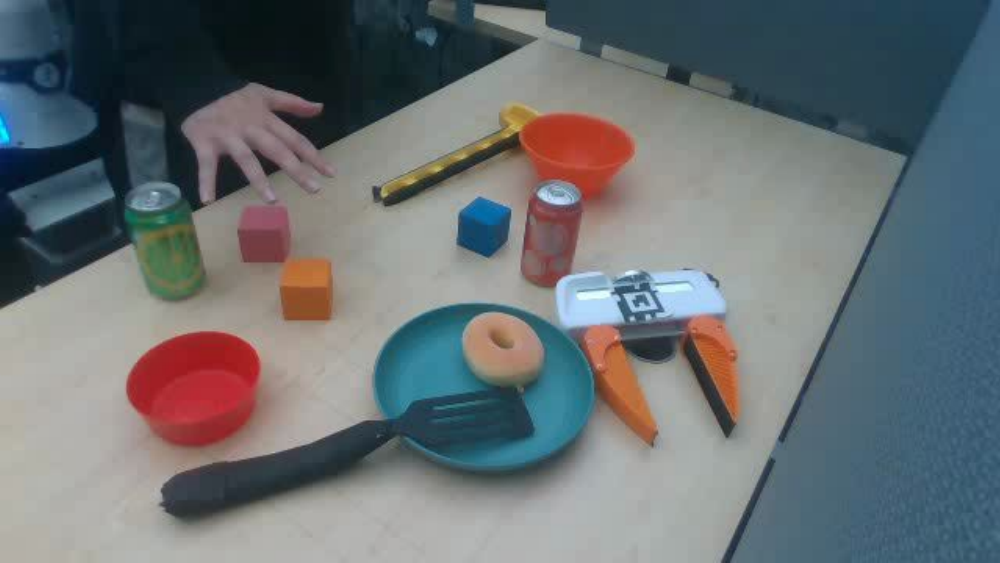}
        \end{minipage}
        \hfill
        \begin{minipage}{0.72\textwidth}
            \includegraphics[width=\linewidth]{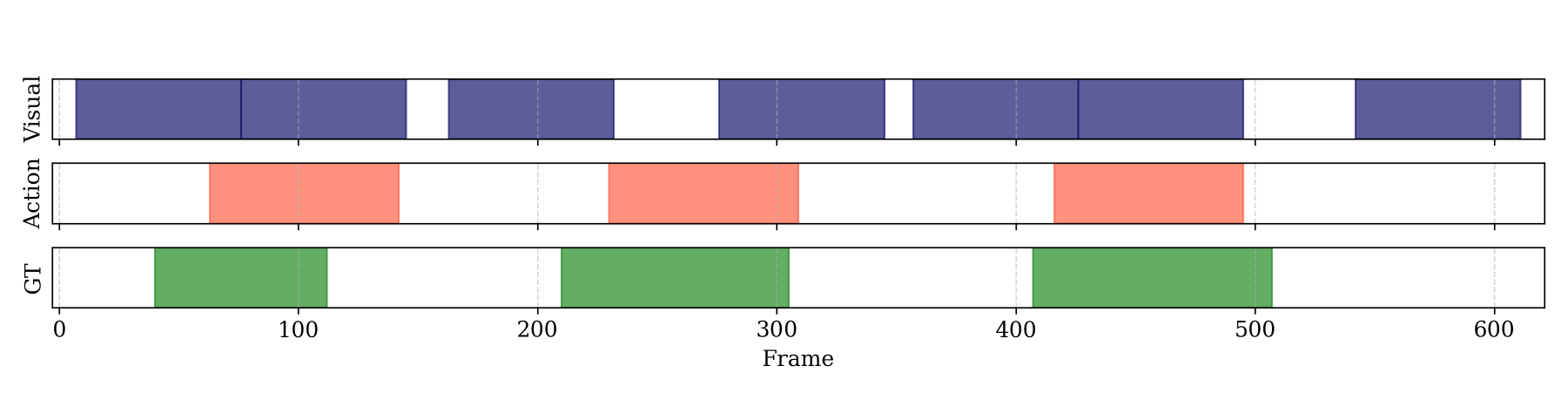}
        \end{minipage}
        \caption{Retrieval with visual disturbance: additional objects and background change.}
        \label{fig:visual_dist}
    \end{subfigure}

    \vspace{1.5em}

    \begin{subfigure}{0.9\textwidth}
        \centering
        \begin{minipage}{0.25\textwidth}
            \includegraphics[width=\linewidth]{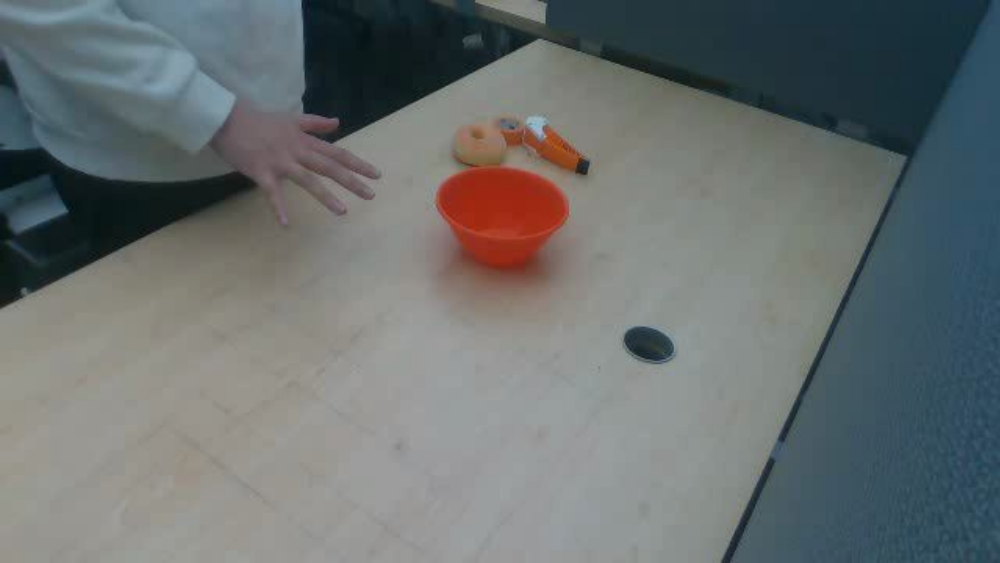}
        \end{minipage}
        \hfill
        \begin{minipage}{0.72\textwidth}
            \includegraphics[width=\linewidth]{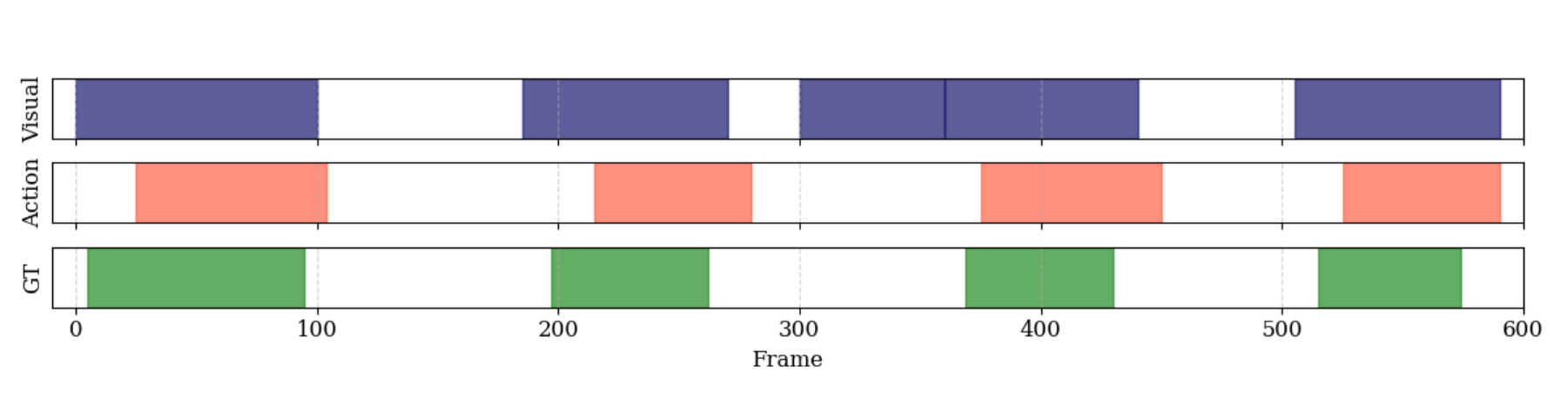}
        \end{minipage}
        \caption{Retrieval with action disturbance: Pick and Place different objects.}
        \label{fig:action_dist}
    \end{subfigure}

    \caption{Comparison of visual- and action-based mapping methods under baseline, visual disturbance, and action disturbance conditions. The results indicate that visual-based mapping suffers a more noticeable performance drop under visual disturbances, while action-based mapping remains comparatively robust.}
    \label{fig:actionvsvisual_all}
    \vspace{-1em}
\end{figure}

\subsection{Greedy Multi‑Segment Subsequence DTW (GMS‑SDTW).}
In our current setup, we use segmented human and robot demonstrations recorded in the same workspace while performing the same task. 
This controlled design minimizes the visual and action gap and simplifies the mapping process.
In contrast, more practical scenario involves long, untrimmed human videos that include disturbances and task-irrelevant actions.
In such cases, identifying an accurate mapping strategy becomes even more critical.
To extract useful segments from these raw videos, recent retrieval-based methods attempts to match human segments with corresponding robot behaviors, most often relying on visual features~\cite{strap}.
We formulate a retrieval task using long human videos, enabling a comparison between visual- and action-based mapping strategies to clarify which modality yields higher accuracy.
To address this, we propose \textbf{Greedy Multi-Segment Subsequence DTW (GMS-SDTW)}, an extension of our current mapping algorithm.

\begin{wrapfigure}{r}{0.4\textwidth}
  \vspace{-1em}
  \begin{minipage}{\linewidth}
    \begin{center}
    \resizebox{\linewidth}{!}{%
      \begin{tabular}{l|l|c|c}
      \toprule
      \textbf{Setting} & \textbf{Method} & \textbf{mIoU} & \textbf{Acc@0.5} \\
      \midrule
      \multirow{2}{*}{Baseline} 
        & Visual & 0.52 & 66.7 \\
        & Action & 0.70 & 71.4 \\
      \midrule
      \multirow{2}{*}{+ Visual Disturb.} 
        & Visual & 0.41\textcolor{red}{\scriptsize$\downarrow$0.11} & 33.3\textcolor{red}{\scriptsize$\downarrow$33.4} \\
        & Action & 0.67\textcolor{red}{\scriptsize$\downarrow$0.03} & 66.7\textcolor{red}{\scriptsize$\downarrow$4.7} \\
      \midrule
      \multirow{2}{*}{+ Action Disturb.} 
        & Visual & 0.46\textcolor{red}{\scriptsize$\downarrow$0.06} & 40.0\textcolor{red}{\scriptsize$\downarrow$26.7} \\
        & Action & 0.65\textcolor{red}{\scriptsize$\downarrow$0.05} & 75.0\textcolor{green!60!black}{\scriptsize$\uparrow$3.6} \\
      \bottomrule
      \end{tabular}
    }
    \end{center}
    \captionof{table}{Comparison of mean IoU and Acc@0.5 for visual- and action-based mappings under different disturbance conditions on long raw videos.}
    \label{iou_accuracy}
    \vspace{-1em}
  \end{minipage}
\end{wrapfigure}

\textbf{Overview of GMS-SDTW.}
Given a long human trajectory $\mathbf{H}\!=\!\{\mathbf{h}_t\}_{t=1}^{T_h}$  
that contains an unknown number of action subsequences, and a single robot trajectory  
$\mathbf{R}\!=\!\{\mathbf{r}_s\}_{s=1}^{T_r}$, our goal is to identify \emph{all} the human subsequences that best match the robot trajectory.
We extend classical Subsequence DTW (S‑DTW) by scanning through $\mathbf{H}$ using a sliding window method, greedily selecting mapped subsequences whose distance to the robot trajectory is below a predefined threshold~$\epsilon$. The sliding window length $L$ is varied within a predefined range $L\!\in\![L_{\min},L_{\max}]$.
The algorithm is presented in Alg.~\ref{long_dtw}.

\textbf{S‑DTW.}  
The cumulative distance matrix $D(i,j)$ is initialized to support open-ended matching in the candidate sequence $\mathbf{R}$:
\begin{equation}
\begin{aligned}
D(0, 0) = &d(0, 0), \quad
D(i, 1) = \sum_{k=1}^{i} d(k, 1), \quad D(1, j) = d(1, j) \\
&\text{for } i = 1,\dots,T_h \text{ and } j = 1,\dots,T_r.
\end{aligned}
\end{equation}
where $d(i, j)$ is the pairwise distance between the $i$-th human frame and $j$-th robot frame.

The recursive update is follows the standard DTW formulation:
\begin{equation}
D(i, j) = d(i, j) + \min \left\{ D(i-1, j-1),\, D(i-1, j),\, D(i, j-1) \right\}.
\end{equation}
The best-matching endpoint is chosen as \( j^\star = \arg\min_j D(T_h, j) \), and the start index is 
recovered via backtracking from \( (i, j^\star) \).

\textbf{Greedy search.}  
Starting at frame $t=1$, we evaluate subsequence  
$\mathbf{H}_{t:t+L}$ for lengths $L \in [L{\min}, L_{\max}]$ via  
S‑DTW, get the subsequence with the minimum distance, and store it if  
$d^\star<\epsilon$. 
Stored subsequences are recorded as segments  
$(t,t+L^\star,k^\star,j^\star)$ and the search resumes from $t\!=\!t+L^\star+1$.  
Otherwise we increment $t\!\leftarrow\!t+1$. The algorithm runs in  
$\mathcal{O}\!\bigl((L_{\max}-L_{\min})\,T_r,T_h\bigr)$ time, and each robot frame is only assessed within the S‑DTW dynamic‑programming table.

\begin{algorithm}[t]
\caption{Greedy Multi-Segment Subsequence DTW (GMS-SDTW)}
\begin{algorithmic}[1]
\Require Human trajectory $\mathbf{H}$ of length $T_h$; robot trajectory $\mathbf{R}$ of length $T_r$;\\
\hspace{1.5em} window bounds $L_{\min}, L_{\max}$; distance threshold $\epsilon$
\Ensure Set $\mathcal{P}$ of matched segments $(h_\text{start}, h_\text{end}, r_\text{start}, r_\text{end}, d)$
\State $\mathcal{P} \gets \emptyset$; \quad $t \gets 1$
\While{$t + L_{\min} - 1 \le T_h$}
    \State $d_\text{best} \gets +\infty$
    \For{$L = L_{\min}$ to $\min(L_{\max}, T_h - t + 1)$}
        \State $\mathbf{q} \gets \mathbf{H}_{t:t+L-1}$
        \State $(d, j_\text{start}, j_\text{end}, \_) \gets \text{S-DTW}(\mathbf{q}, \mathbf{R})$
        \If{$d < d_\text{best}$}
            \State $d_\text{best} \gets d$
            \State $L^\star \gets L$
            \State $j_\text{start}^\star \gets j_\text{start}$
            \State $j_\text{end}^\star \gets j_\text{end}$
        \EndIf
    \EndFor
    \If{$d_\text{best} < \epsilon$}
        \State Add $(t, t+L^\star-1, j_\text{start}^\star, j_\text{end}^\star, d_\text{best})$ to $\mathcal{P}$
        \State $t \gets t + L^\star$ \Comment{Skip matched subsequence}
    \Else
        \State $t \gets t + 1$
    \EndIf
\EndWhile
\State \Return $\mathcal{P}$
\end{algorithmic}
\label{long_dtw}
\end{algorithm}

\textbf{Complexity.}
Each S‑DTW distance has a time complexity of $\mathcal{O}(T_hT_r)$. With a linear scan over
$T$ frames and at most $L_{\max}-L_{\min}+1$ window lengths, the overall complexity is $\mathcal{O}\!\bigl((L_{\max}-L_{\min}+1)\,T_rT_h\bigr)$, which is tractable in practice since $L_{\max}\ll T$.

\subsection{Visual- and Action-based Long Raw Video Retrieval}

As discussed in \textbf{Core Results}, action-based mapping tends to offer more robust performance than visual mapping.
To further compare their performance, we propose a long raw video retrieval task~\cite{pseudoformer} as an intuitive way to assess the robustness of each mapping method under varying conditions. 
In addition to segmented human demos with well-defined start and end boundaries, we also explore extended videos containing multiple irrelevant visual and action segments.

We evaluate the following three scenarios:
(1) \textbf{Baseline}: Videos captured under standard clear conditions.
(2) \textbf{With Visual Disturbance}: Videos that include background clutter or additional distracting objects, simulating more realistic visual environments.
(3) \textbf{With Action Disturbance}: Videos where the demonstrated action is slightly altered (e.g., grasping a different object), introducing minor motion variations.

Our proposed GMS-SDTW method processes each long video to detect and maps subsequences corresponding to Pick and Place robot demonstration trajectory. 
As shown in Fig.~\ref{fig:actionvsvisual_all}, action-based retrieval yields more precise results showing resilience to visual disturbances. Quantitative results, including mean Intersection over Union (mIoU) and accuracy at a threshold of 0.5, are presented in Tab.~\ref{iou_accuracy}.
By focusing on action similarity, our system more accurately localizes the relevant segments while reducing sensitivity to irrelevant visual content. Overall, while visual-based mapping may suffer from real-world visual variations, action-based mapping remains robust and reliable.

\section{Additional Baseline Comparison via Visual Retrieval}
We compare ImMimic-A with the current state-of-the-art  retrieval-based method STRAP~\cite{strap}.
STRAP leverages a strong vision foundation model, DINOv2~\cite{dinov2} to embed each video frame and employs S‑DTW to retrieve relevant subtrajectories.
Following STRAP, each robot demonstration is first segmented into variable‑length sub‑trajectories using the low‑level end‑effector motion heuristic. We then extract DINOv2 features from agent‑view videos for both human and robot data. 
Treating robot subtrajectories as a query, we apply S‑DTW to locate matching subsequences in human videos.
We cap the number of matches per query at $K = 500$ where $K$ denotes the maximum number of matched segments per query.
As shown in Tab.~\ref{tab:strap_study}, STRAP outperforms the Robot Only baseline, while ImMimic-A still achieves even higher performance.
STRAP is designed for robot-to-robot transfer via retrieval-based matching and therefore does not explicitly address the domain distribution gap present in human-to-robot transfer.
Moreover, while STRAP employs a strong visual encoder for feature similarity, action information can offer more robust correspondence in the presence of a human-to-robot visual gap.

\section{Additional Experimental Results and Details}

\subsection{Domain Gap}
Learning from human videos poses two critical gaps that often hinder policy transfer to robots: the \emph{visual gap} and the \emph{action gap}~\cite{mimicplay, egomimic}. 
The visual gap arises due to significant differences in appearance between humans and robots. 
The action gap stems from differences in kinematic constraints, motion dynamics, embodiment size, and task execution strategies.

\textbf{Visual Gap.}
In Tab.~\ref{datavis}, we present sample demonstration clips highlighting how human and robot embodiments differ significantly in their visual observations. While a shared workspace setup can help reduce background-related visual discrepancies, notable appearance differences between human and robot demonstrations remain.

\begin{table}[!tbp]
    \centering
    \resizebox{0.7\textwidth}{!}{
    \begin{tabular}{l|cc|cc}
        \toprule
        \multirow{2}{*}{\textbf{Setting}} 
        & \multicolumn{2}{c|}{\textbf{Robotiq}} 
        & \multicolumn{2}{c}{\textbf{Ability}} \\
        & \textbf{Pick and Place} & \textbf{Flip} 
        & \textbf{Pick and Place} & \textbf{Flip} \\
        \midrule
        Robot Only  & 0.40 & 0.60 & 0.80 & 0.60 \\
        Co-Training & 0.40 & 0.80 & 0.80 & 0.90 \\
        \midrule
        STRAP & 0.50 & 0.60 & 0.90 & 0.90 \\ 
        ImMimic-A & \textbf{1.00} & \textbf{1.00} & \textbf{1.00} & \textbf{1.00} \\
        \bottomrule
    \end{tabular}
    }
    \caption{Comparison of success rates between Robot Only, Co-Training, STRAP, and our ImMimic-A across two embodiments and two tasks, using 5 robot demonstrations and 100 human demonstrations.}
    \label{tab:strap_study}
    \vspace{-1em}
\end{table}

\begin{table}[!tbp]
    \centering
    \resizebox{0.7\textwidth}{!}{
    \begin{tabular}{l|cc|cc}
        \toprule
        \multirow{2}{*}{\textbf{Setting}} 
        & \multicolumn{2}{c|}{\textbf{Robotiq}} 
        & \multicolumn{2}{c}{\textbf{Ability}} \\
        & \textbf{Pick and Place} & \textbf{Flip} 
        & \textbf{Pick and Place} & \textbf{Flip} \\
        \midrule
        ImMimic-A ($\beta$-dist) & 0.90 & 0.90 & 0.90 & \textbf{1.00} \\ 
        ImMimic-A (linear) & \textbf{1.00} & \textbf{1.00} & \textbf{1.00} & \textbf{1.00} \\
        \bottomrule
    \end{tabular}
    }
    \caption{Comparison between ImMimic-A ($\beta$-dist), which samples the MixUp ratio $\alpha$ from a $\beta$-distribution, and ImMimic-A (linear), which uses a linearly decreasing schedule for $\alpha$. Success rates are reported across two embodiments and two tasks, using 5 robot demonstrations and 100 human demonstrations.}
    \label{tab:beta_dist_study}
    \vspace{-1em}
\end{table}

\begin{figure}[tbp]
    \centering

    \begin{subfigure}[t]{0.23\textwidth}
        \centering
        \includegraphics[width=\linewidth]{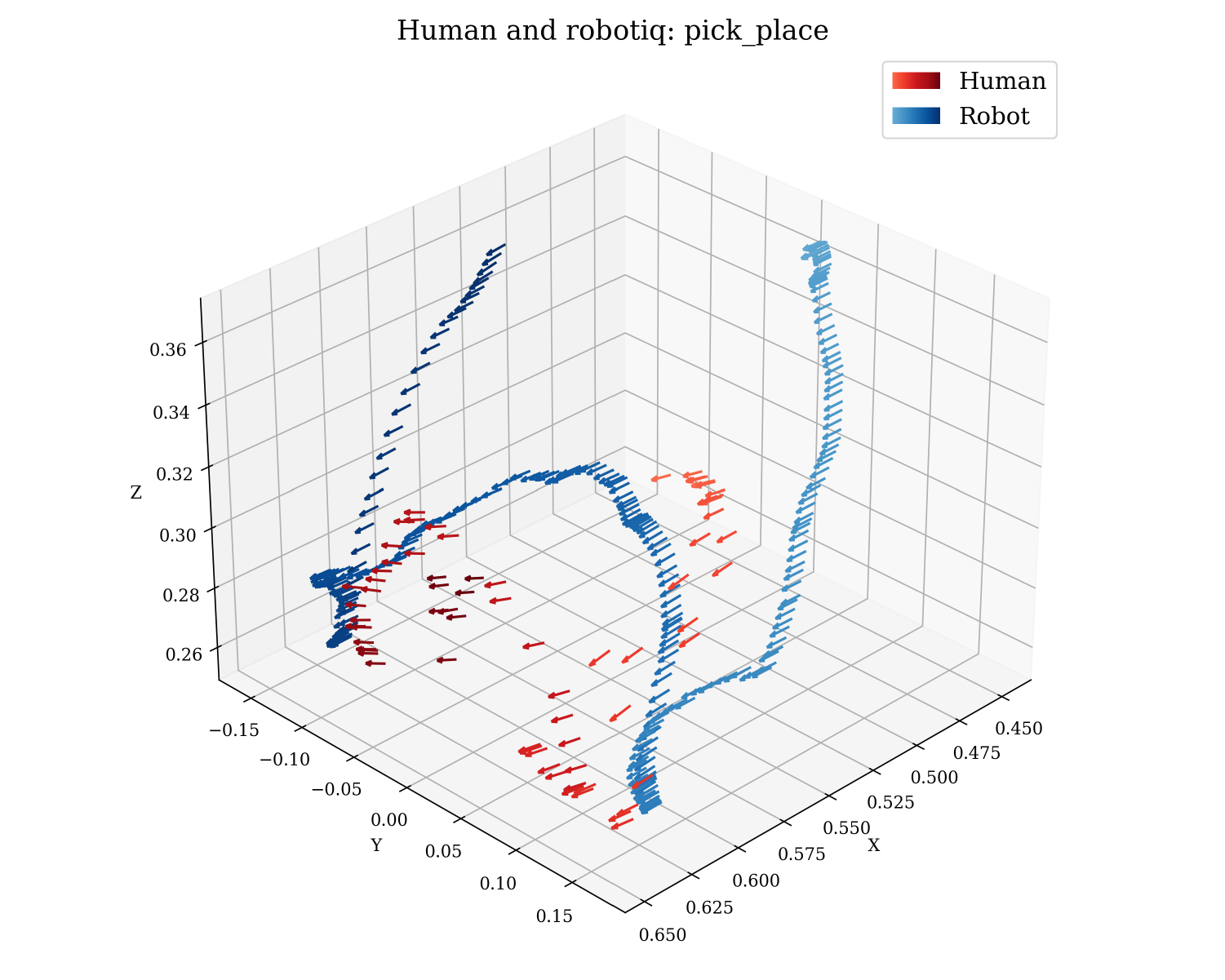}
        \caption{Pick and Place: Robotiq}
    \end{subfigure}
    \begin{subfigure}[t]{0.23\textwidth}
        \centering
        \includegraphics[width=\linewidth]{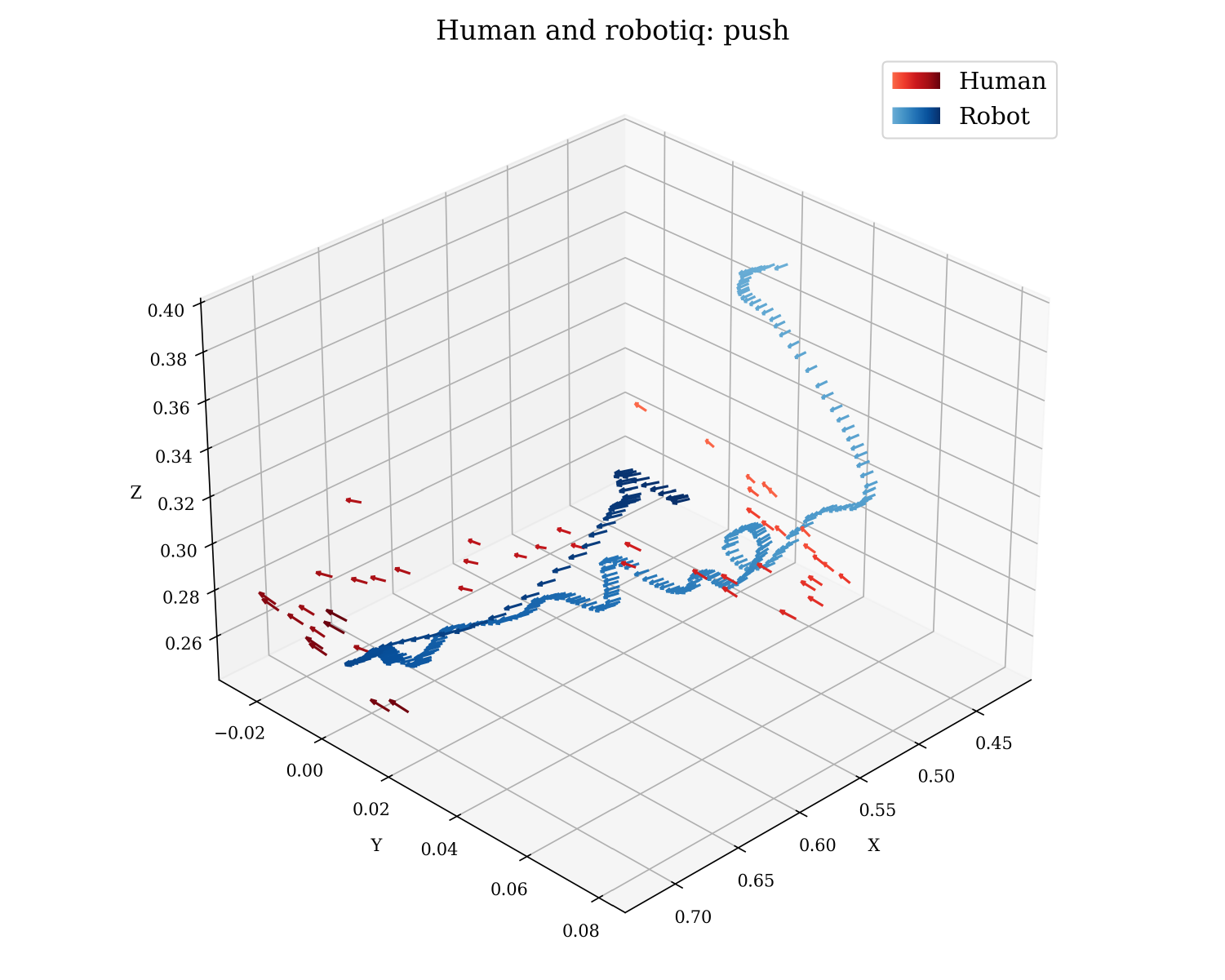}
        \caption{Push: Robotiq}
    \end{subfigure}
    \begin{subfigure}[t]{0.23\textwidth}
        \centering
        \includegraphics[width=\linewidth]{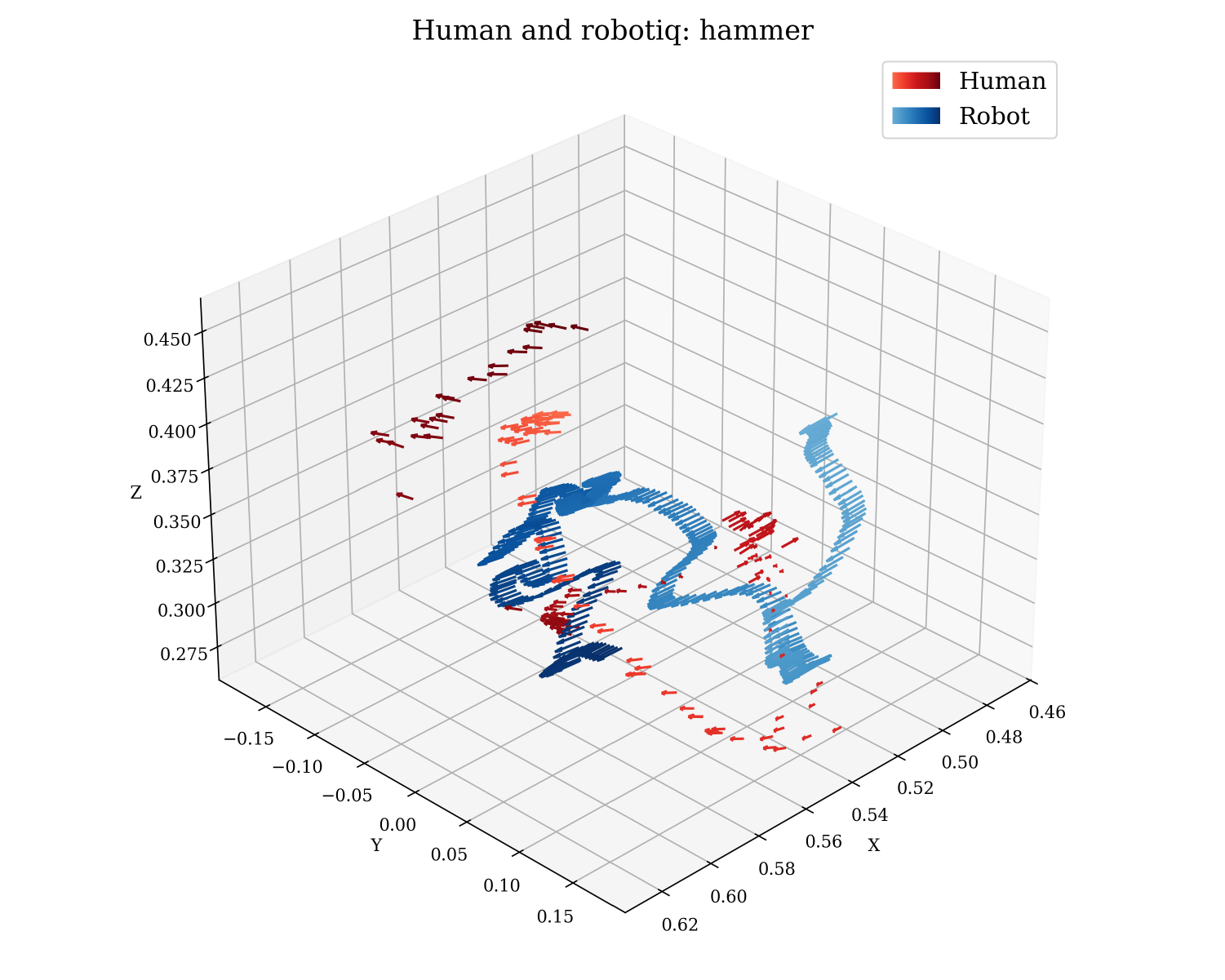}
        \caption{Hammer: Robotiq}
    \end{subfigure}
    \begin{subfigure}[t]{0.23\textwidth}
        \centering
        \includegraphics[width=\linewidth]{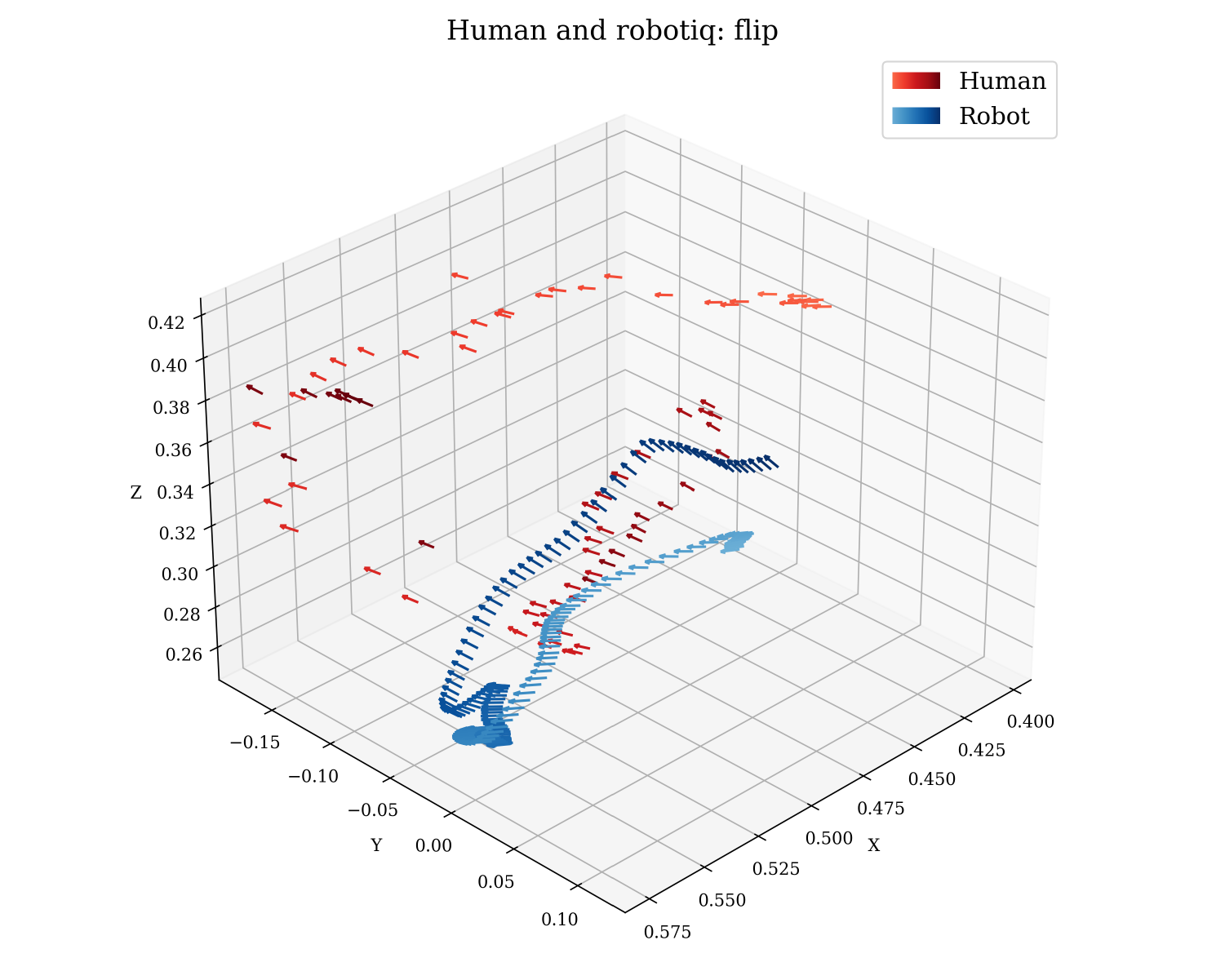}
        \caption{Flip: Robotiq}
    \end{subfigure}

    \begin{subfigure}[t]{0.23\textwidth}
        \centering
        \includegraphics[width=\linewidth]{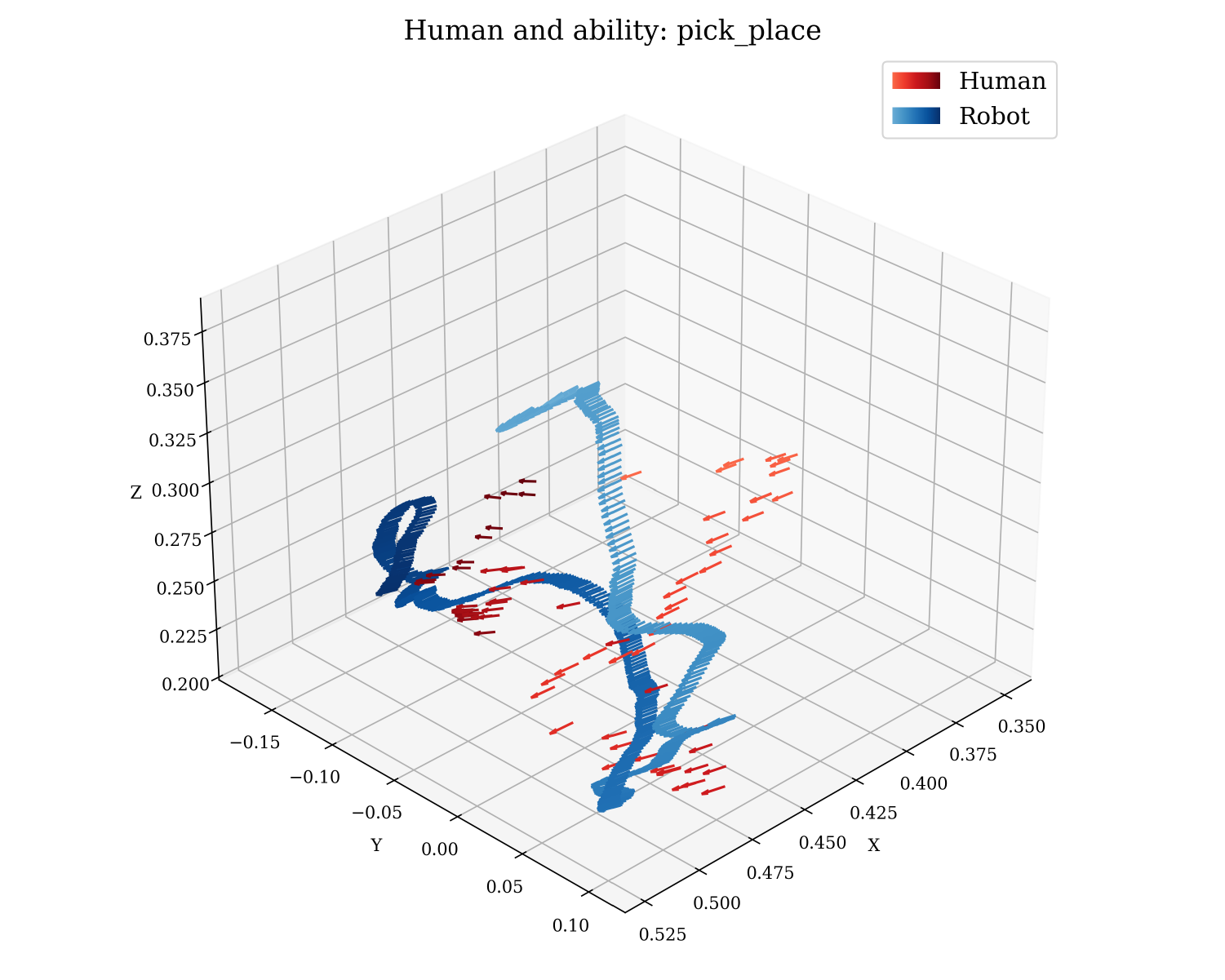}
        \caption{Pick and Place: Ability}
    \end{subfigure}
    \begin{subfigure}[t]{0.23\textwidth}
        \centering
        \includegraphics[width=\linewidth]{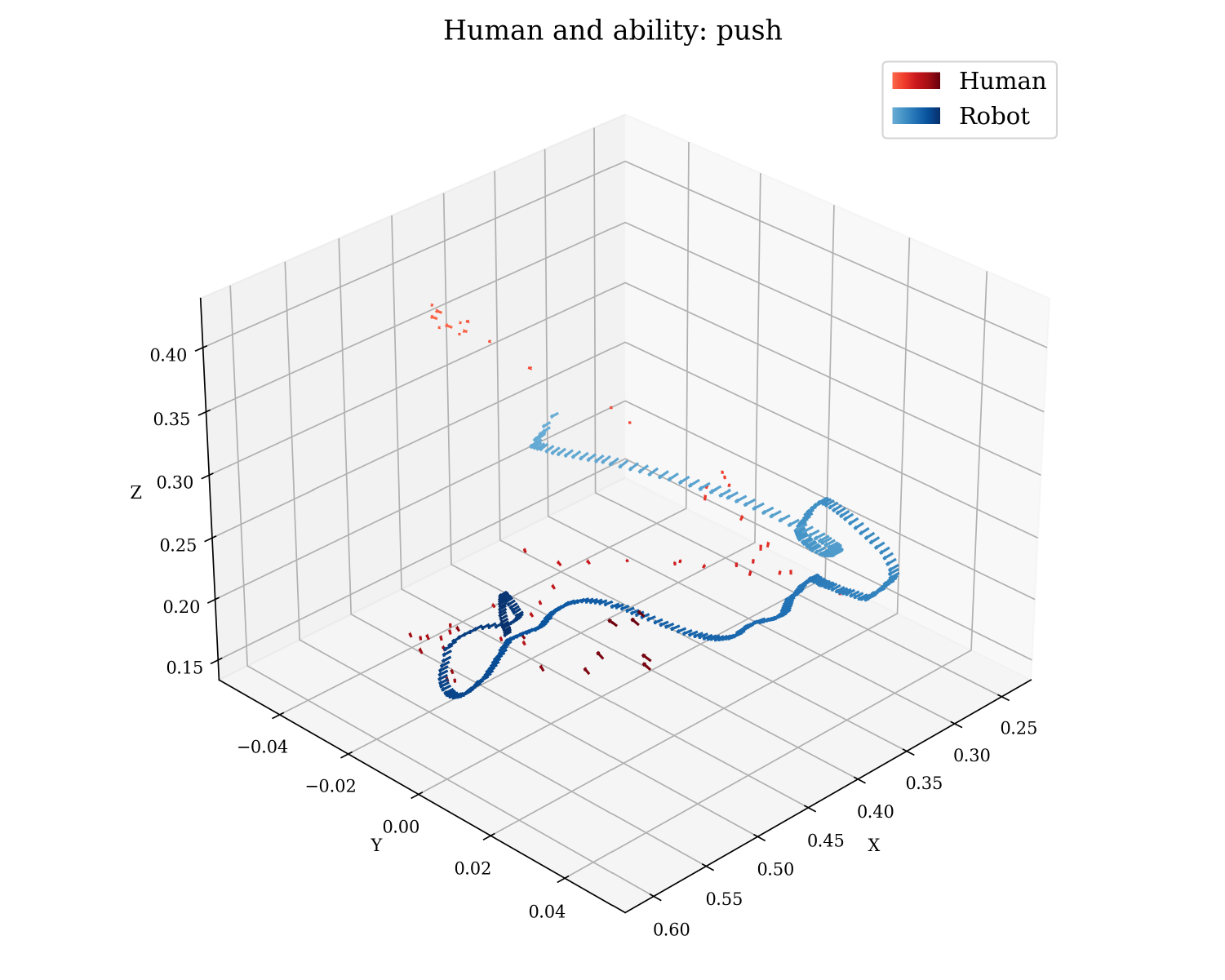}
        \caption{Push: Ability}
    \end{subfigure}
    \begin{subfigure}[t]{0.23\textwidth}
        \centering
        \includegraphics[width=\linewidth]{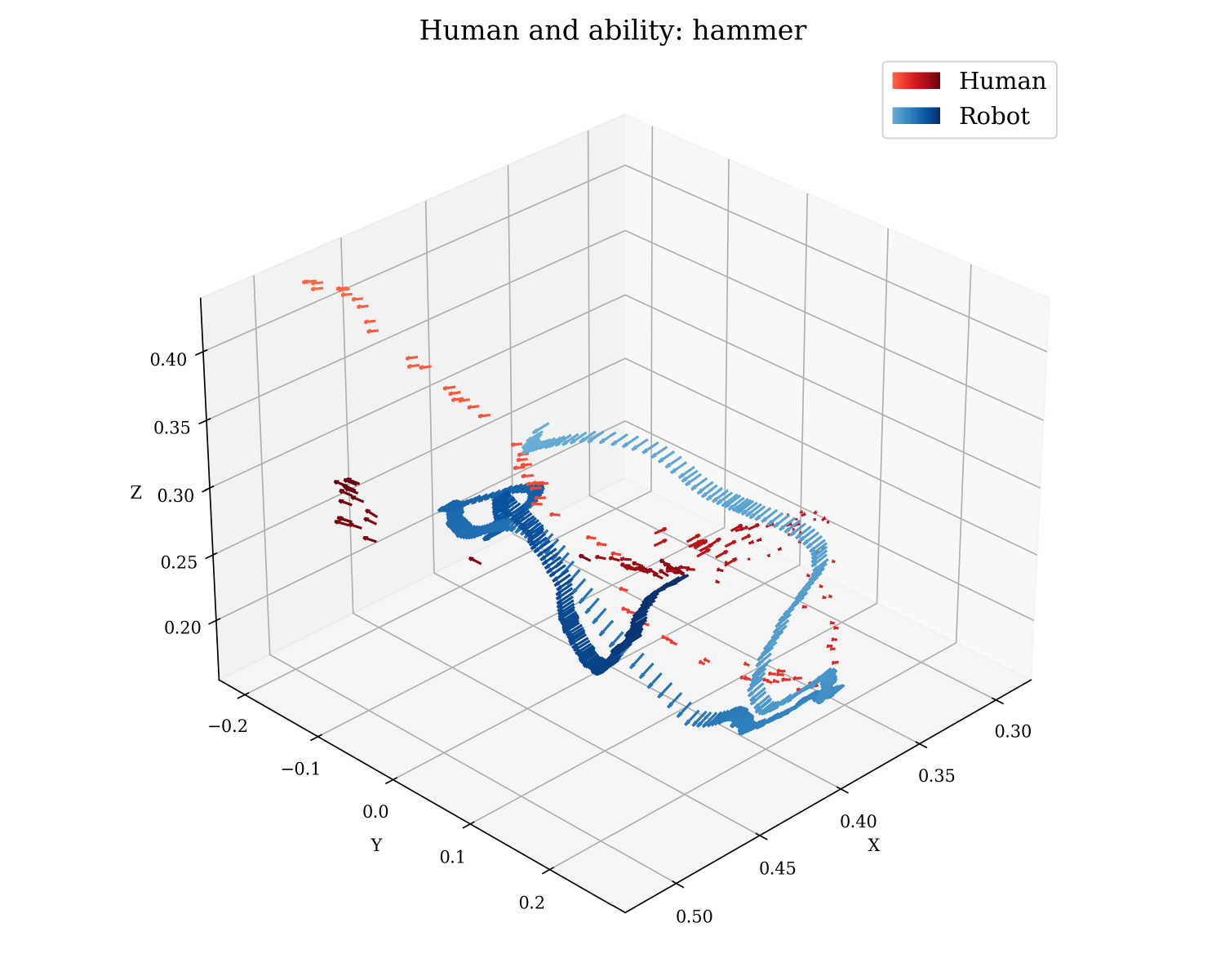}
        \caption{Hammer: Ability}
    \end{subfigure}
    \begin{subfigure}[t]{0.23\textwidth}
        \centering
        \includegraphics[width=\linewidth]{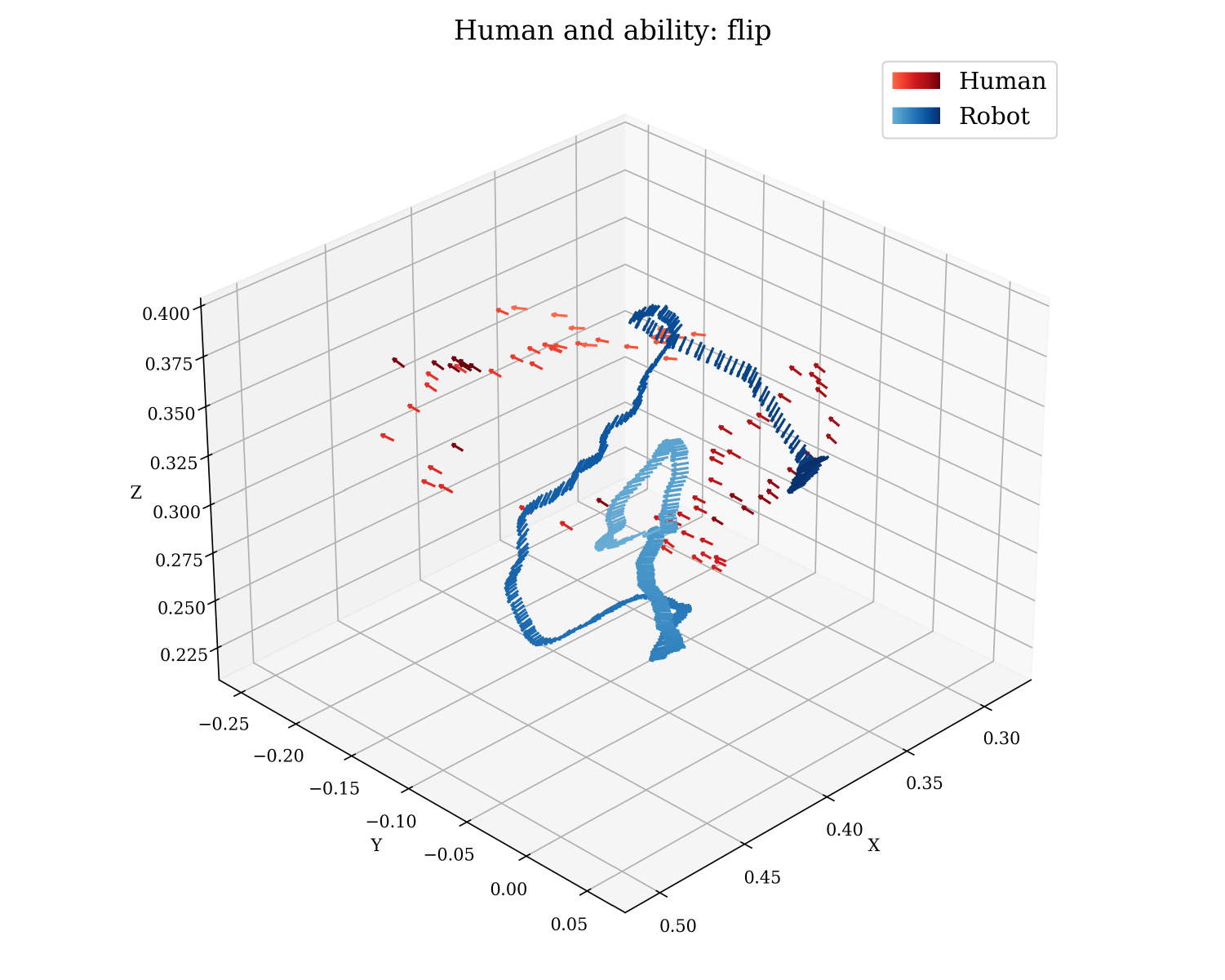}
        \caption{Flip: Ability}
    \end{subfigure}
    \caption{Visualization of sample trajectories pairs: the human retargeted trajectory and the corresponding robot demonstration trajectory. Arrows indicate orientation.}
    \label{fig:traj}
    \vspace{-1em}
\end{figure}

\begin{table}[ht]
\centering
\begin{tabular}{l|cccc|c}
\toprule
\textbf{Embodiment} & \textbf{Pick and Place} & \textbf{Push} & \textbf{Hammer} & \textbf{Flip} & \textbf{AVG} \\
\midrule
Robotiq & 0.031 & 0.067 & 0.085 & 0.083 & 0.066 \\
FR     & 0.028 & 0.077 & 0.068 &  0.089 & 0.065 \\
\midrule
Allegro & 0.063 & 0.065 & 0.089 & 0.094 & 0.078 \\
Ability & 0.047 & 0.056 & 0.091 & 0.106 & 0.075 \\
\bottomrule
\end{tabular}
\caption{Average action similarity across different embodiments and tasks. Grippers generally exhibit a smaller action gap compared to dexterous hands.}
\label{tab:actionsim}
\vspace{-1em}
\end{table}
\textbf{Action Gap.}
Fig.~\ref{fig:traj} shows human demonstration trajectories overlaid with their corresponding teleoperated robot trajectories.
Despite structural differences in design, retargeting aligns human and robot motions by emphasizing underlying physical similarities.
Tab~\ref{tab:actionsim} further quantifies human–robot action similarity.

\subsection{Visualization of the Mapping}

\begin{figure*}[h]
  \centering
       \centering \includegraphics[width=0.82\linewidth]{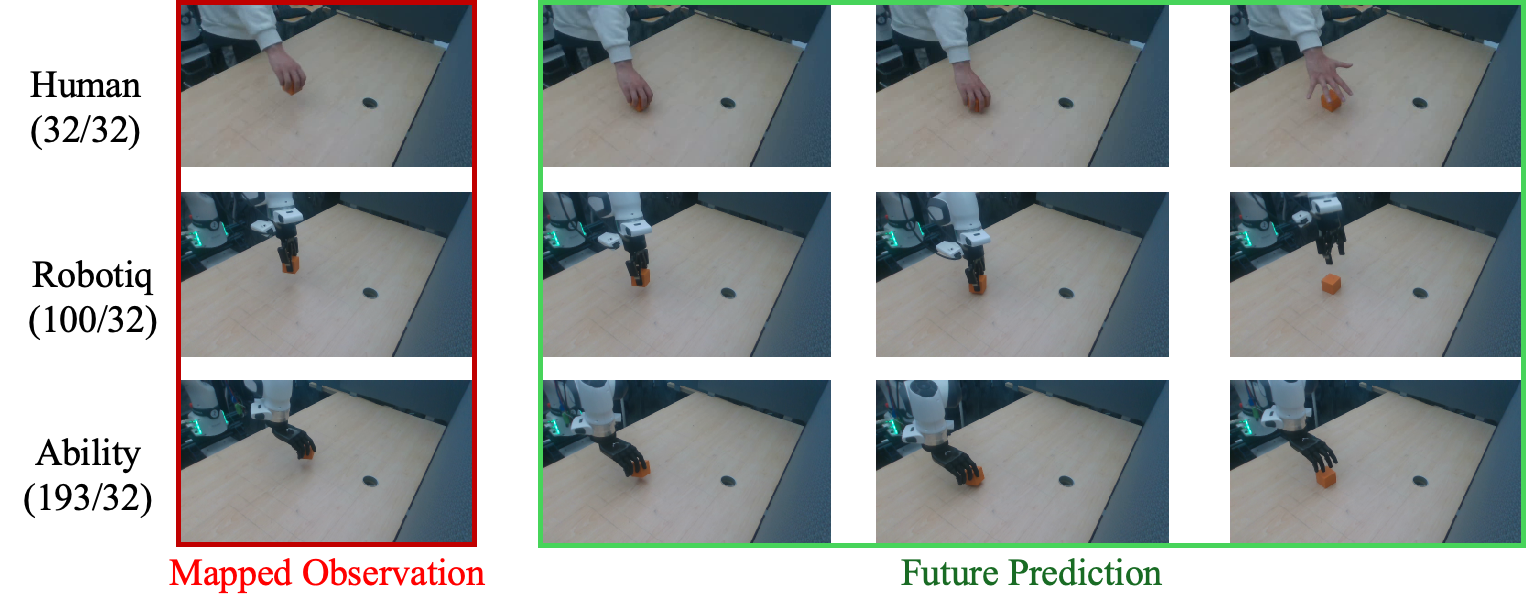}
       \caption{An example of mapped pairs at the same timestep used for MixUp. As shown in Tab.~\ref{samplerate}, we set sample rates $\gamma$ (Human: 32/32, Robotiq: 100/32, Ability: 193/32) based on average durations  to ensure consistent execution speed.}
       \label{fig:mapped}
\end{figure*}

During MixUp, our mapping strategy ensures that interpolated demonstration pairs remain plausible to avoid generating infeasible demonstrations.
Experimental results show that Random Mapping fails to improve performance, and ImMimic‑V with its lower mapping quality, underperforms compared to ImMimic‑A.
We visualize an example of our action mapping at certain timesteps for the Robotiq Gripper and Ability Hand performing Pick and Place (Fig~\ref{fig:mapped}).
By sampling at different rates, we minimize the speed discrepancy between human and robot demonstrations to match their average durations.
As shown in the figure, our mapping strategy effectively mpas observations and future states across embodiments, ensuring task-relevant consistency.

\subsection{Visualization of Domain Flow}

\begin{figure*}[h]
  \centering
       \centering
       \includegraphics[width=0.95\linewidth]{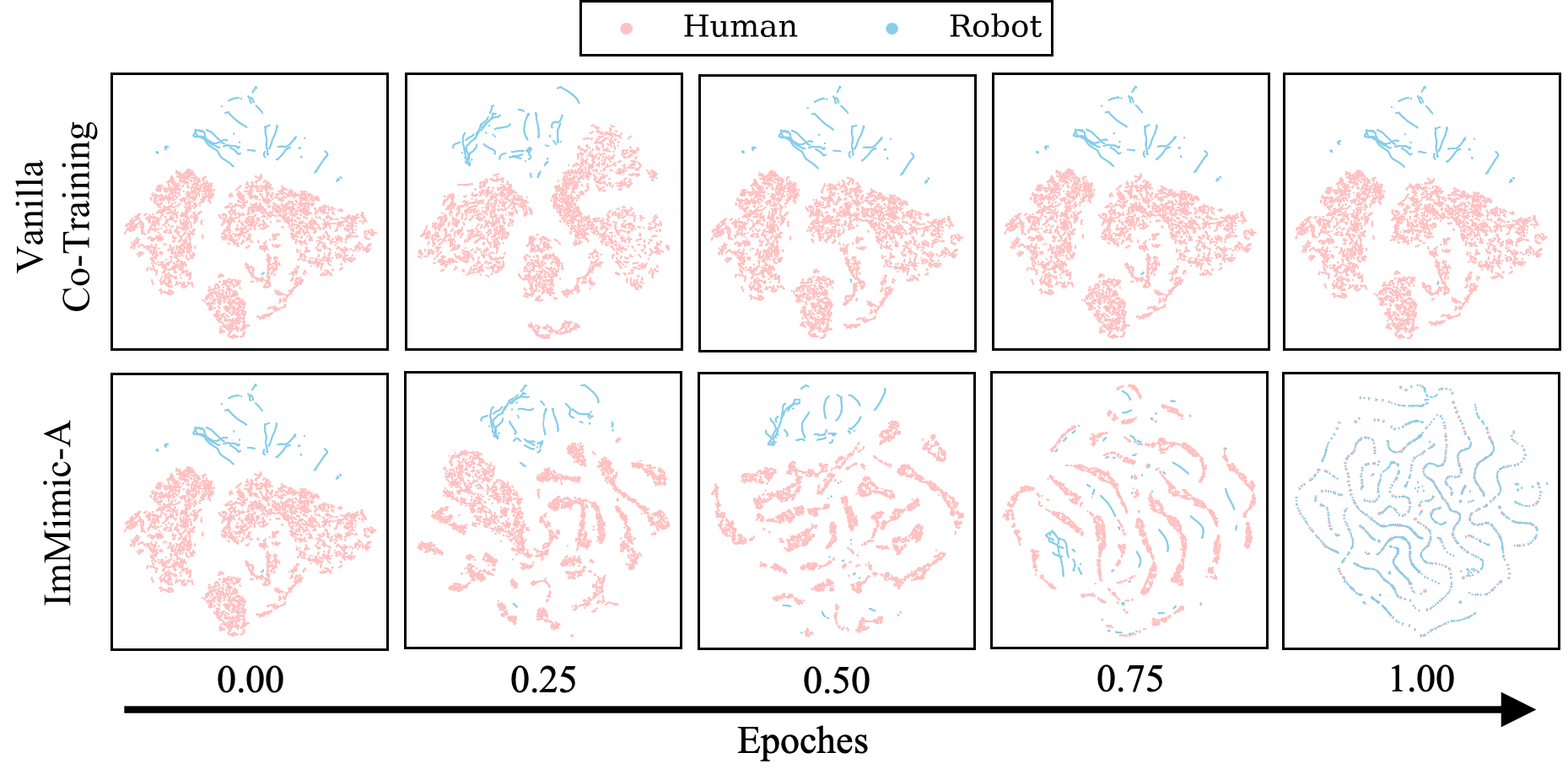}
       \caption{t-SNE visualization of input conditions at each timestep from human and robot datasets during training. We compare ImMimic-A with Co-Training, showing that ImMimic-A generates a smooth domain flow for the human data, enabling effective domain adaptation.}
       \label{fig:domainflow}
\end{figure*}

To illustrate how our methods adapts the across domains, we visualize the t-SNE~\cite{tsne} embeddings of human and robot conditions in Fig.~\ref{fig:domainflow}.  
Each point in the scatter plot represents a condition at a specific timestep from either human or the robot dataset.  
Under Vanilla Co-Training, human and robot data distributions remain clearly separated, highlighting the domain gap.  
This separation between the source (human) and target (robot) data indicates that, without explicit domain adaptation, the model cannot fully leverage human data for robot training.  
Similar to DLOW~\cite{dlow}, which employs a continuous “domainness” variable to transition from source to target domains, ImMimic-A uses the mixing coefficient $\alpha$ to control how far each sample is adapted toward the robot domain.

\begin{table}[tbp]
    \centering
    \label{fig:dataset_viz}
    \vspace{-1pt}
    \setlength{\tabcolsep}{1pt}
    \renewcommand{\arraystretch}{1.}
    \begin{tabularx}{\textwidth}{c|c|X X X X X X}
        \toprule
        \textbf{Task} & \textbf{Embodiment} & \multicolumn{6}{c}{\textbf{Agent view}} \\
        \midrule
        \multirow{5}{*}{\vspace*{-10ex}Pick and Place}
            & Human & 
                \includegraphics[width=0.9\linewidth]{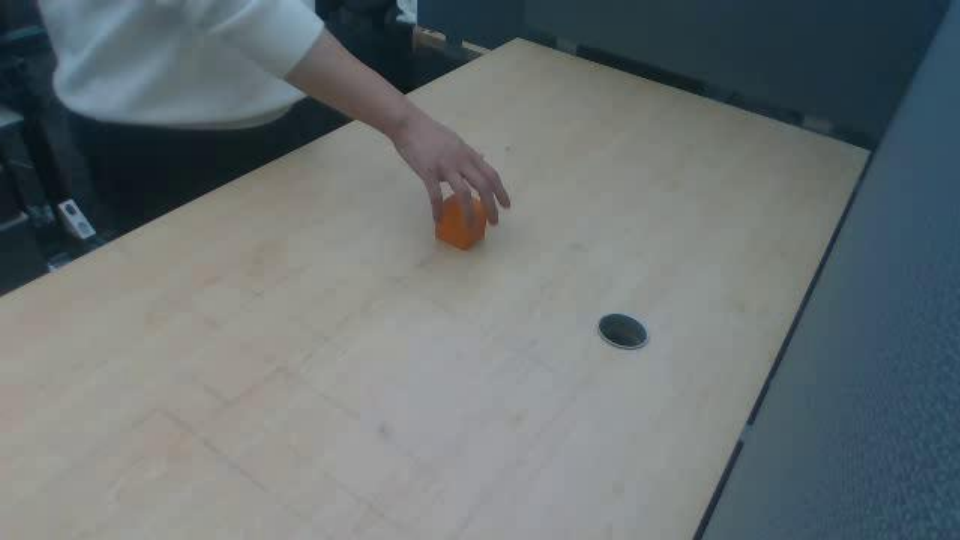} & 
                \includegraphics[width=0.9\linewidth]{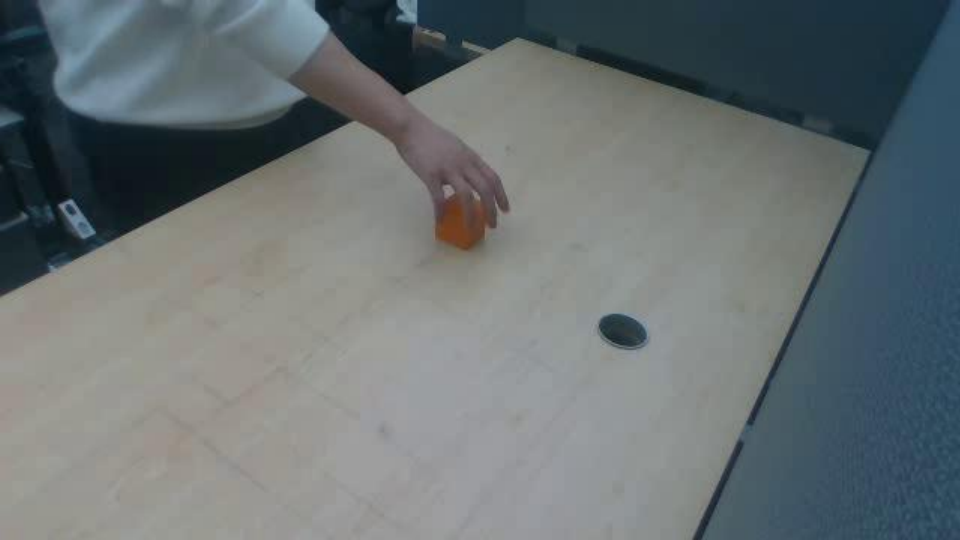} & 
                \includegraphics[width=0.9\linewidth]{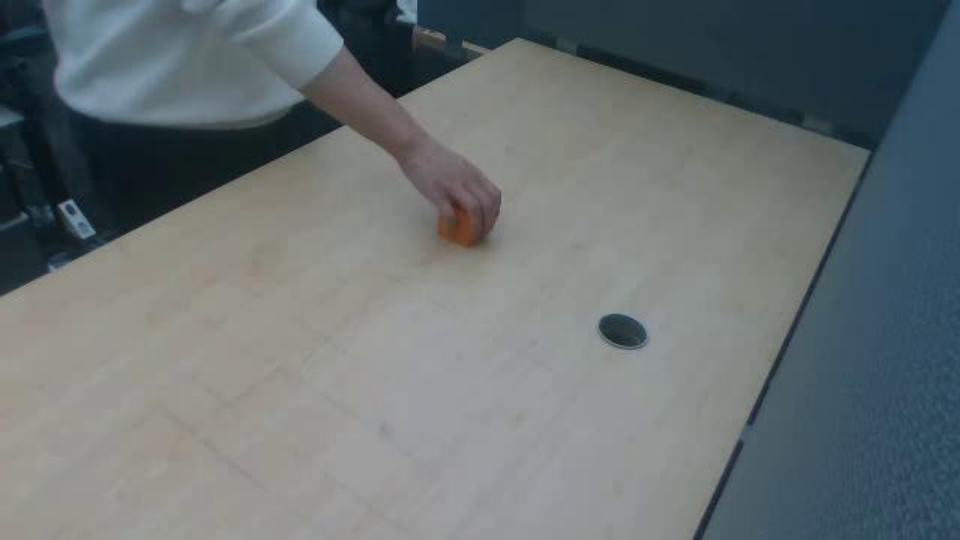} & 
                \includegraphics[width=0.9\linewidth]{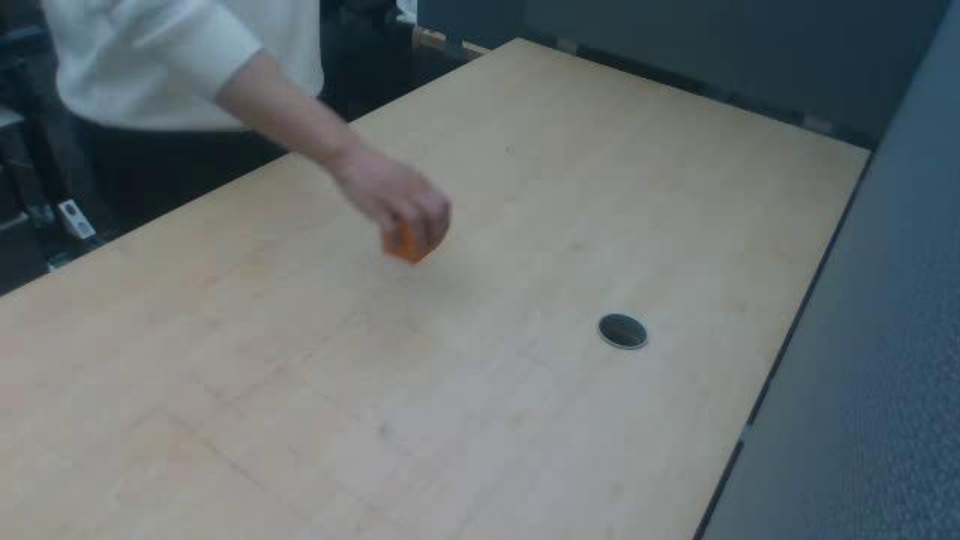} & 
                \includegraphics[width=0.9\linewidth]{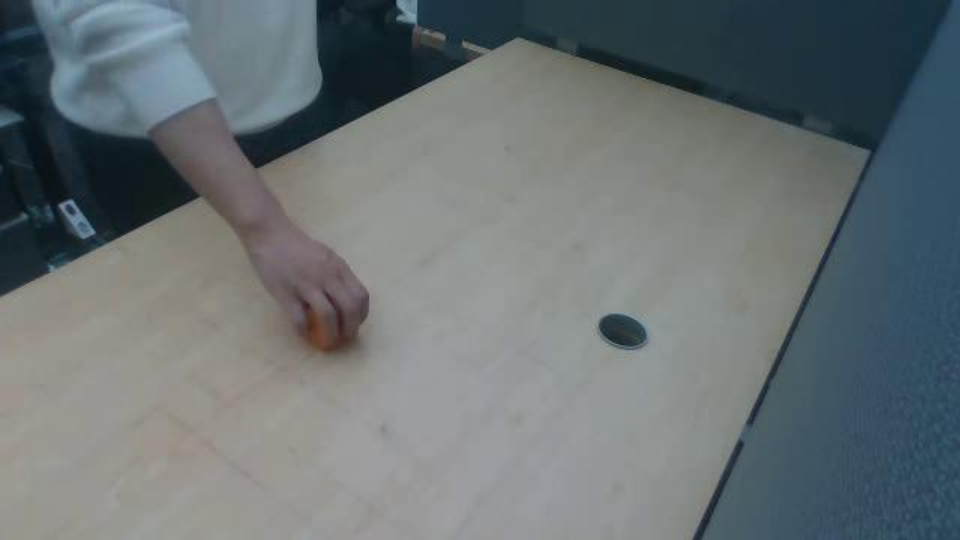} & 
                \includegraphics[width=0.9\linewidth]{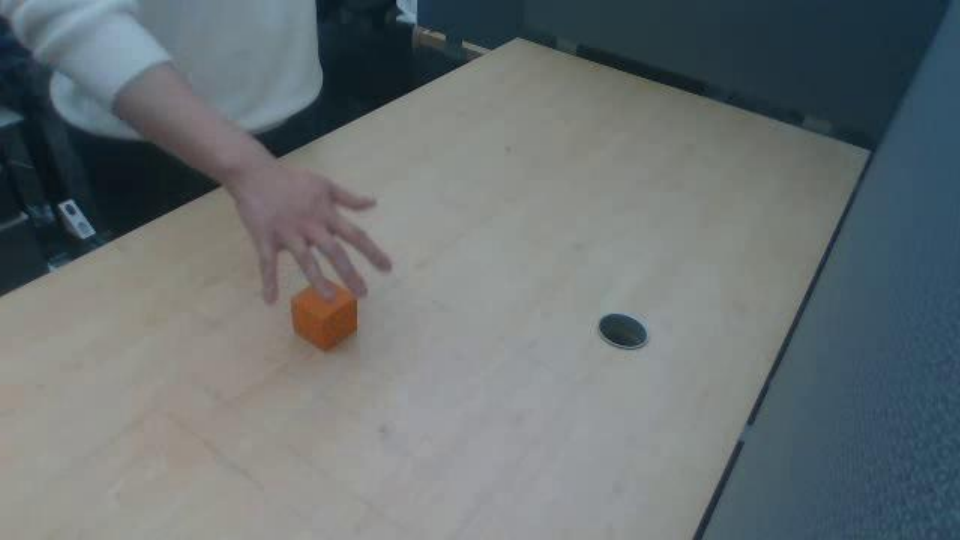} \\
            & FR & 
                \includegraphics[width=0.9\linewidth]{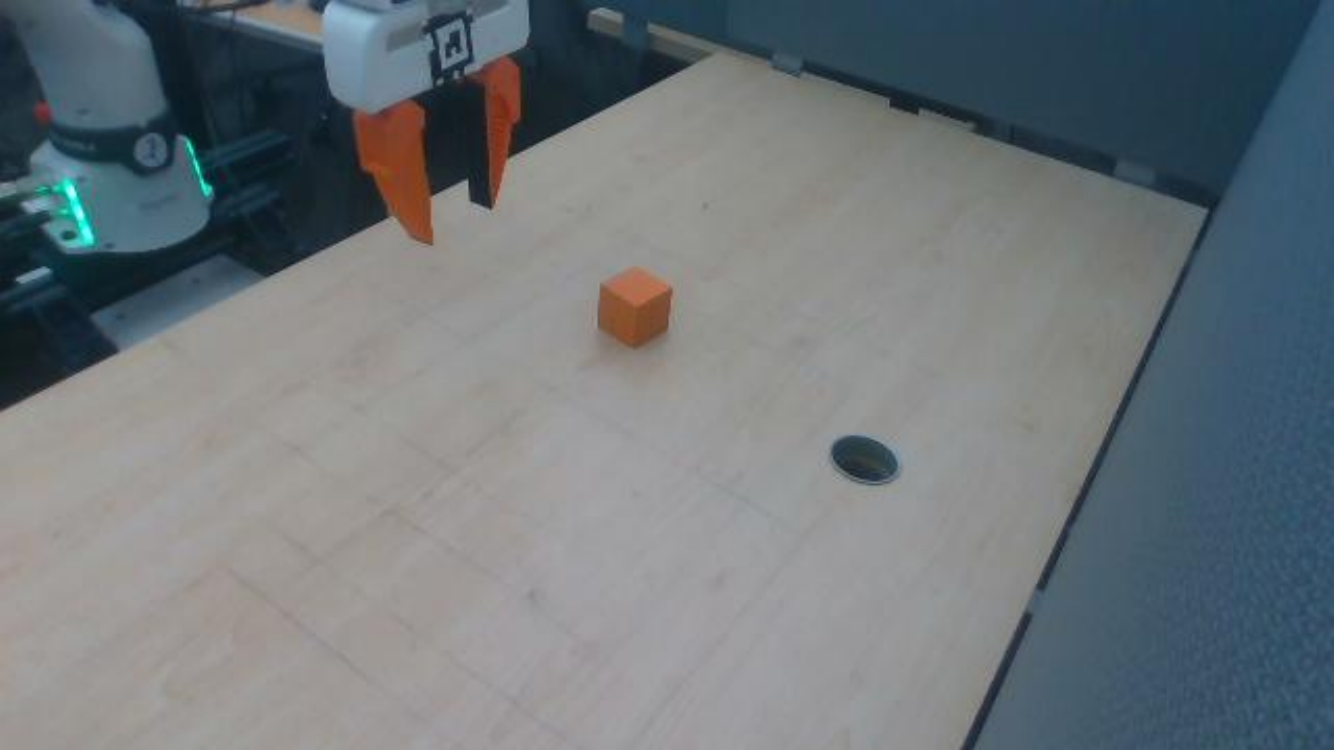} & 
                \includegraphics[width=0.9\linewidth]{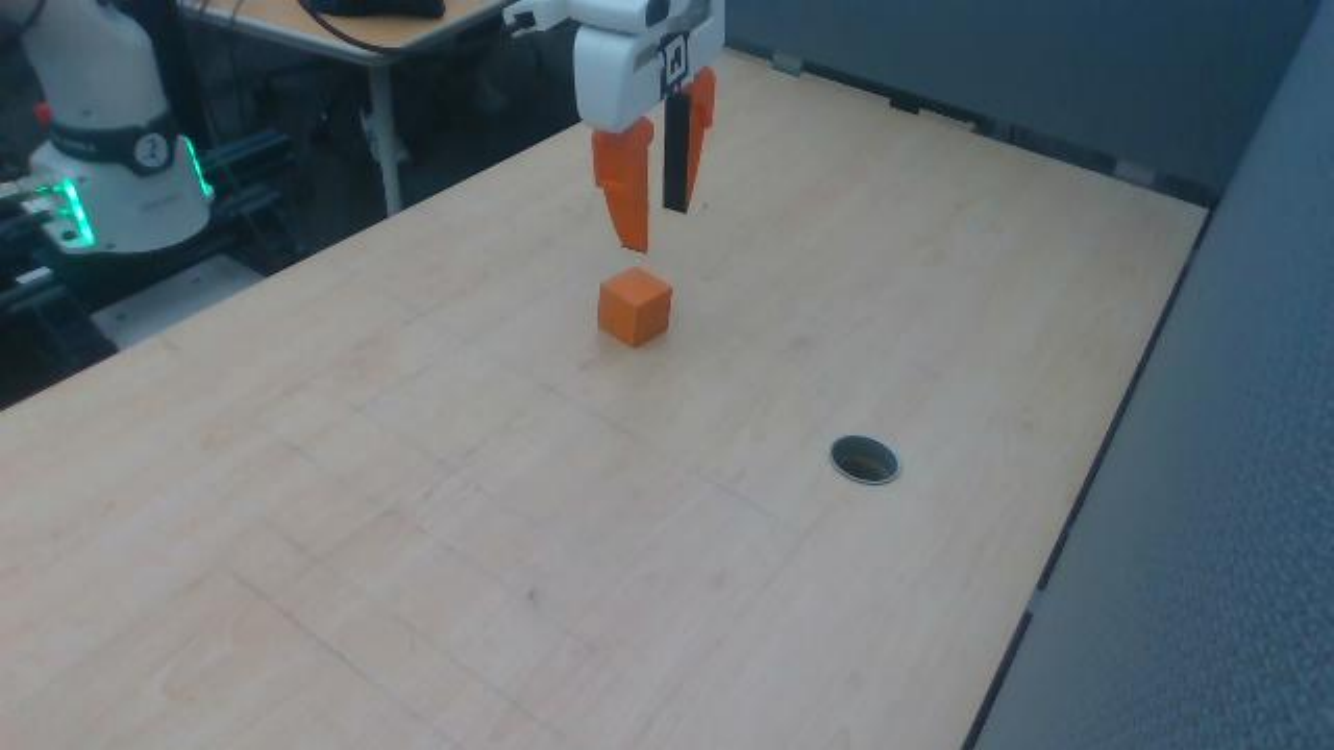} & 
                \includegraphics[width=0.9\linewidth]{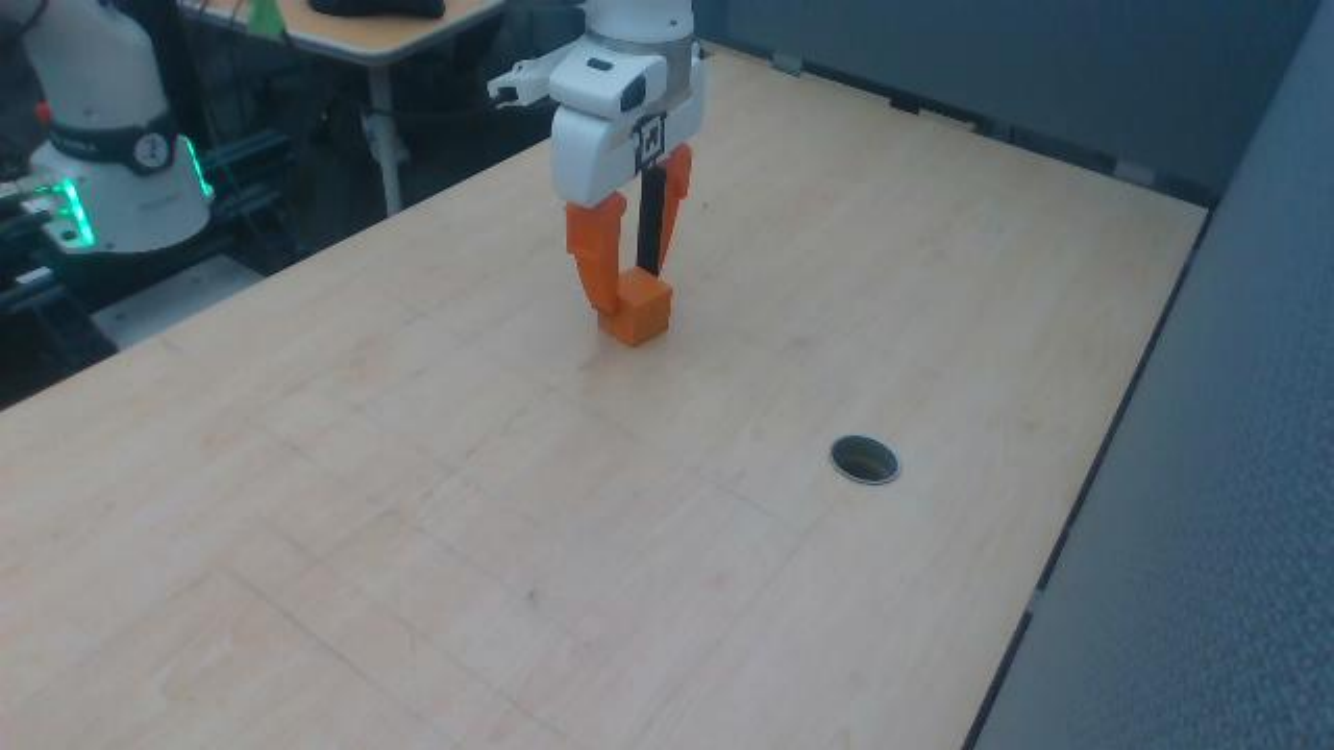} & 
                \includegraphics[width=0.9\linewidth]{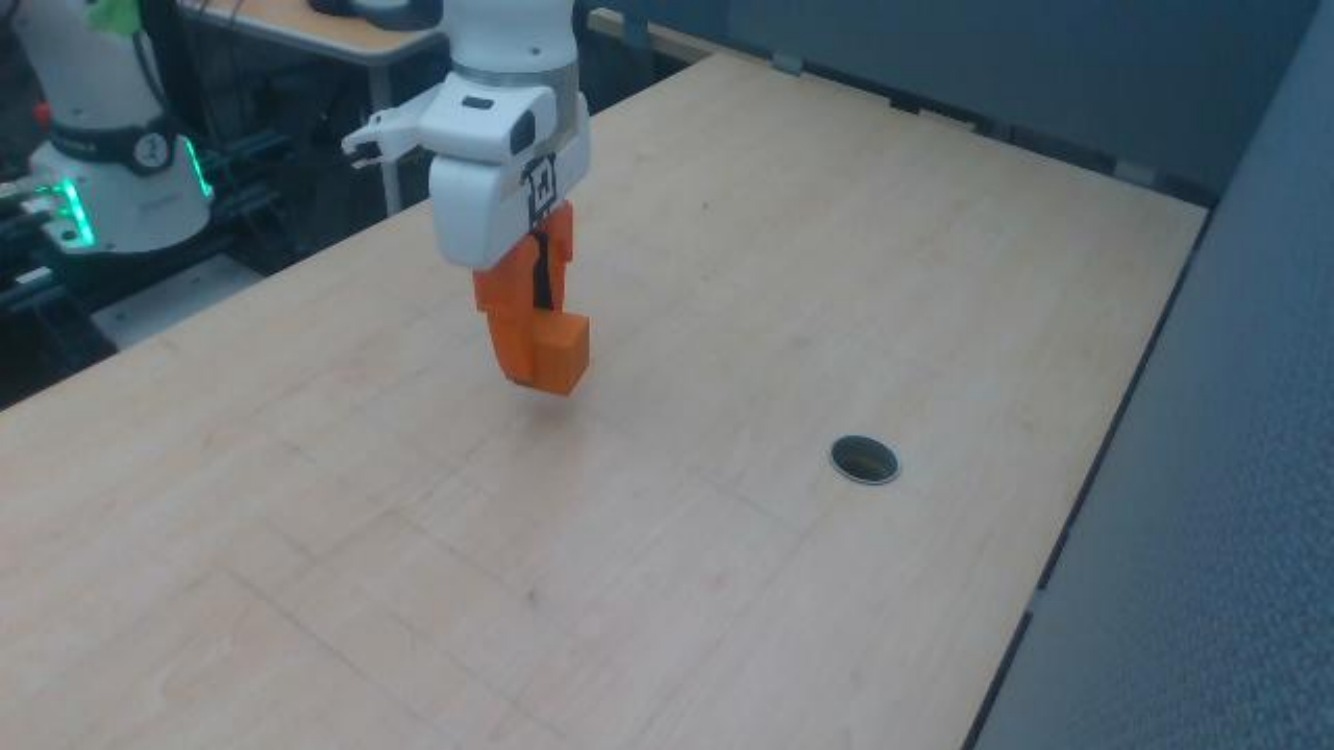} & 
                \includegraphics[width=0.9\linewidth]{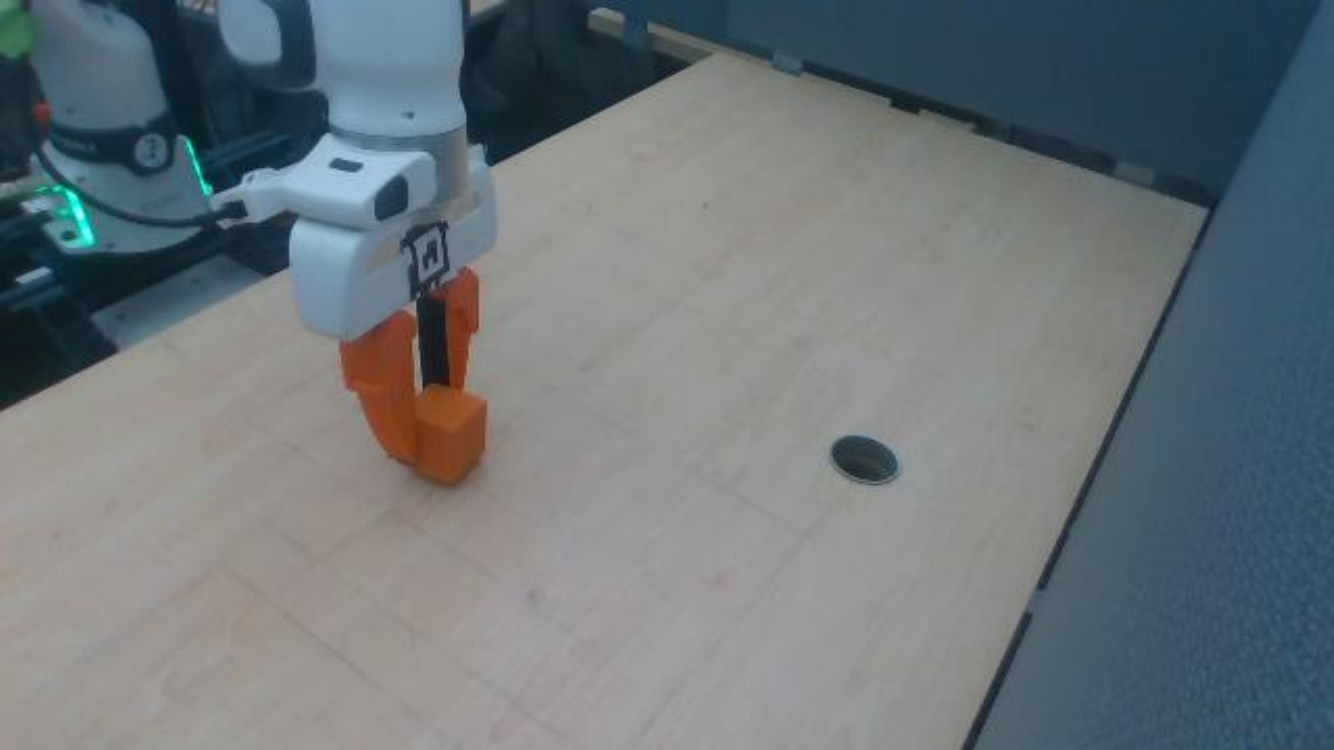} & 
                \includegraphics[width=0.9\linewidth]{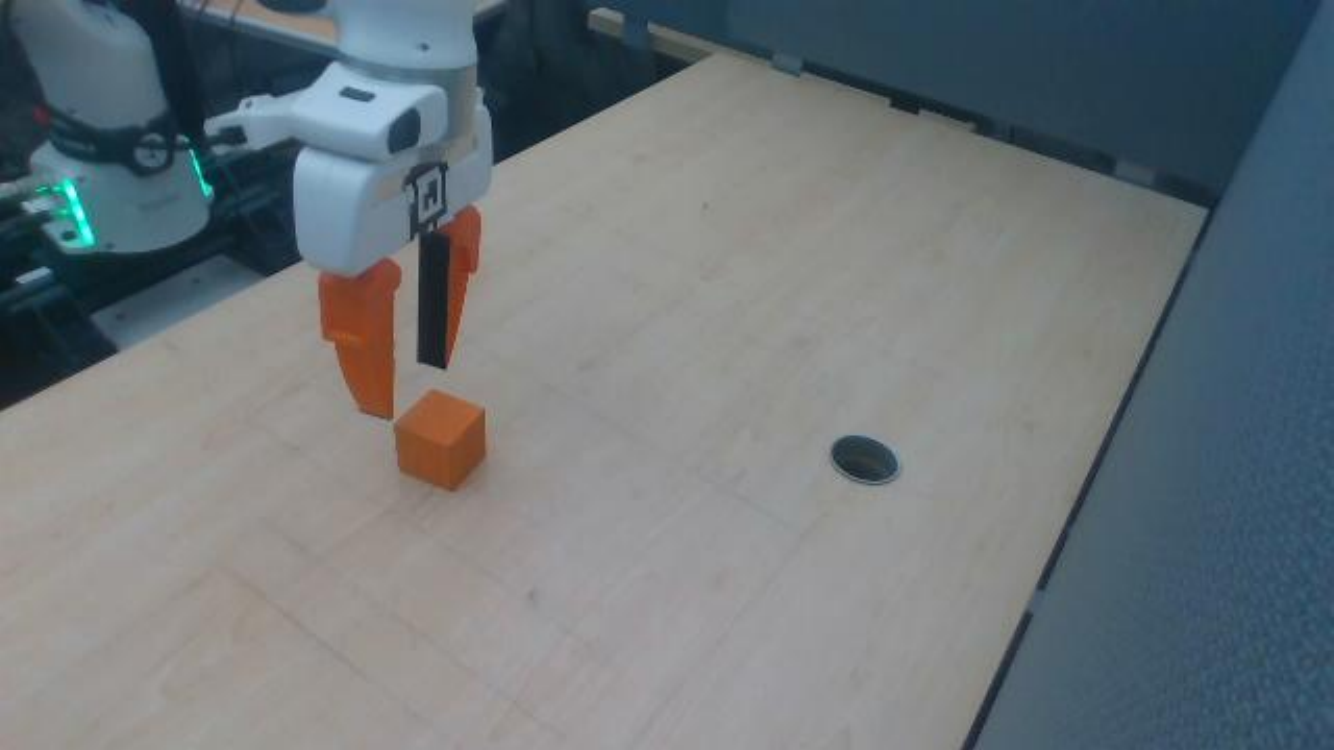} \\
            & Robotiq & 
                \includegraphics[width=0.9\linewidth]{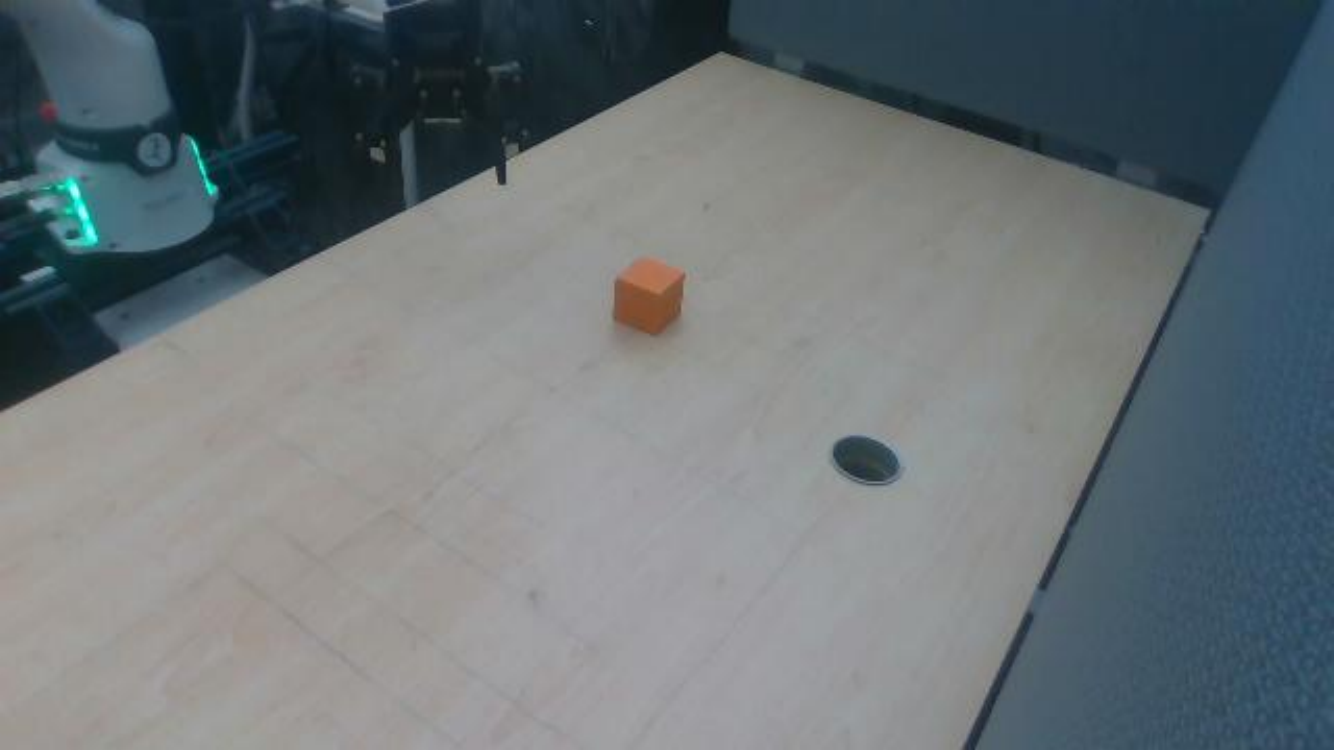} & 
                \includegraphics[width=0.9\linewidth]{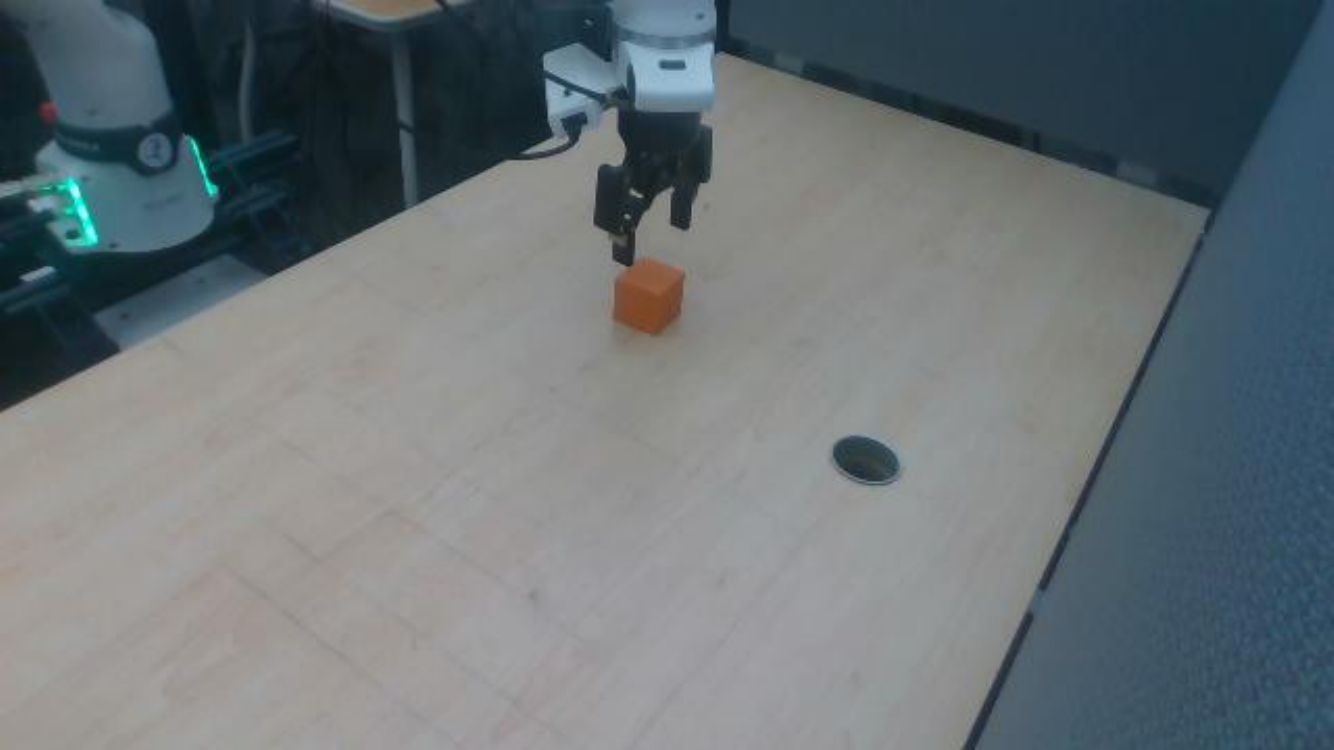} & 
                \includegraphics[width=0.9\linewidth]{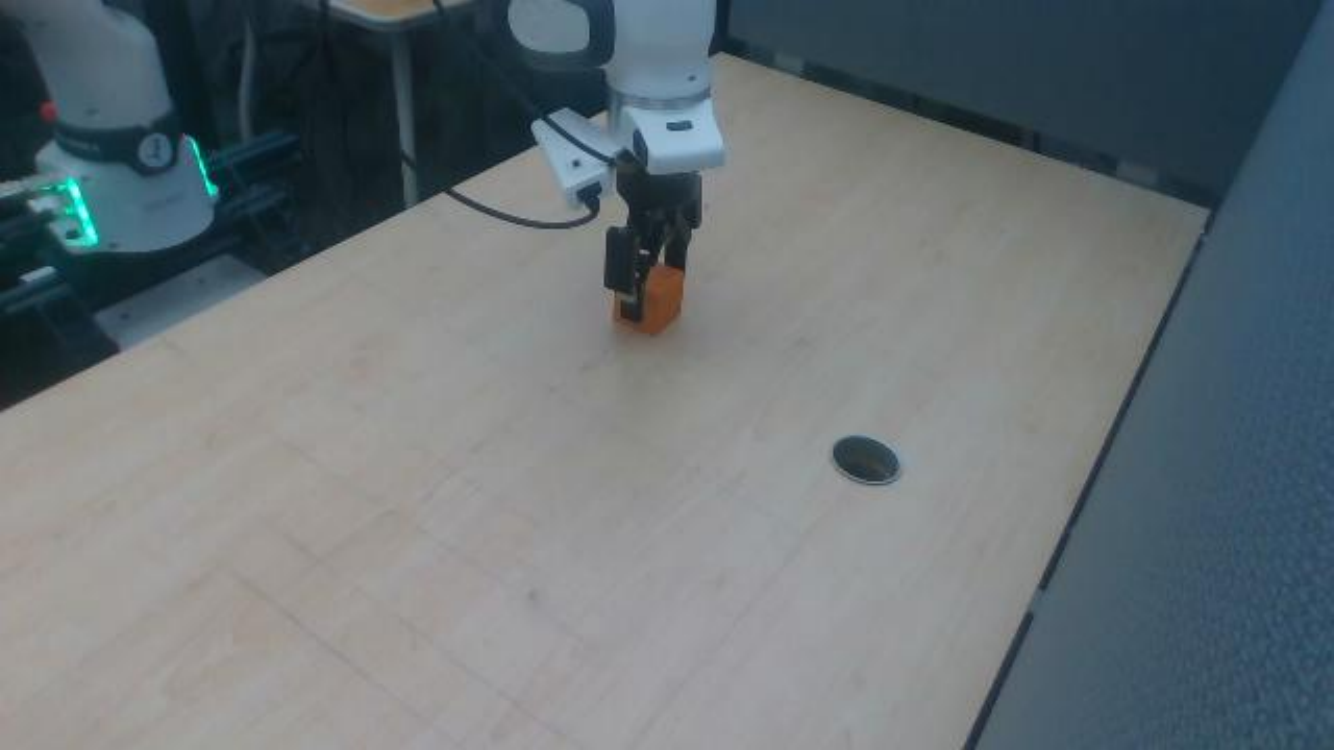} & 
                \includegraphics[width=0.9\linewidth]{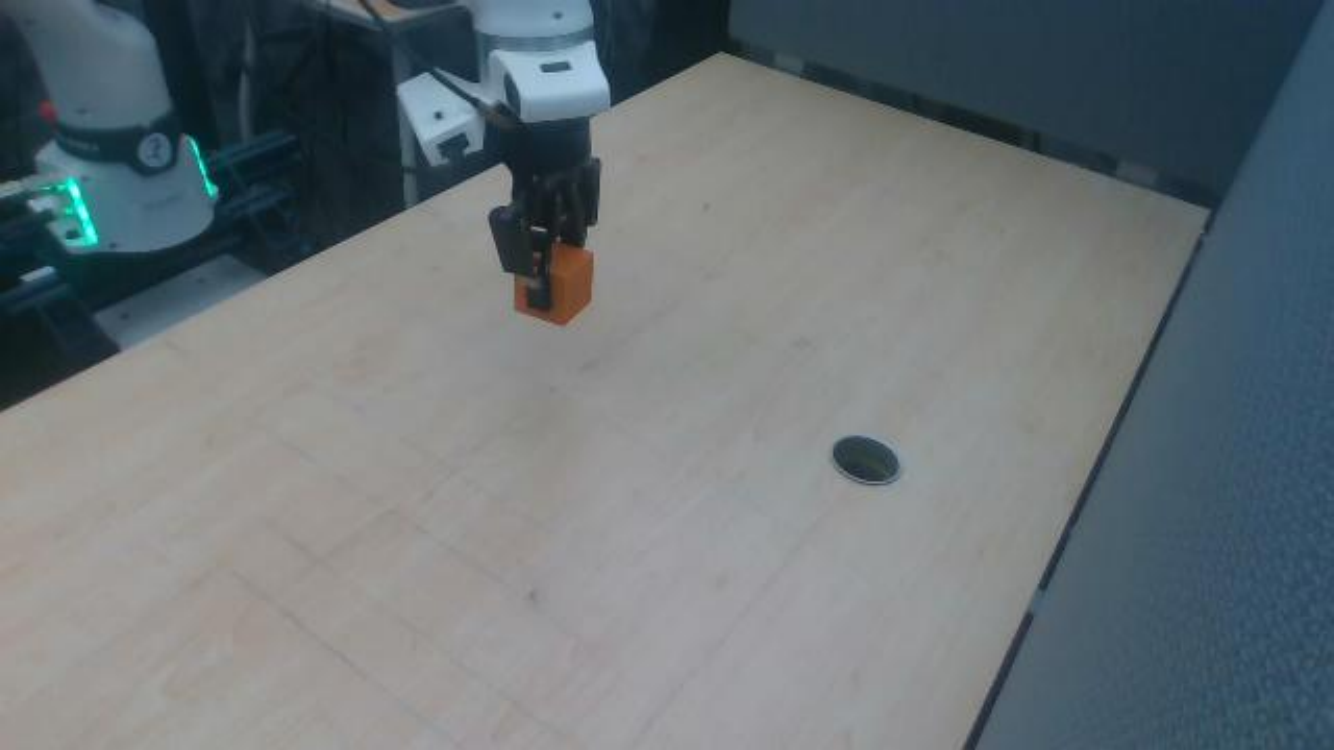} & 
                \includegraphics[width=0.9\linewidth]{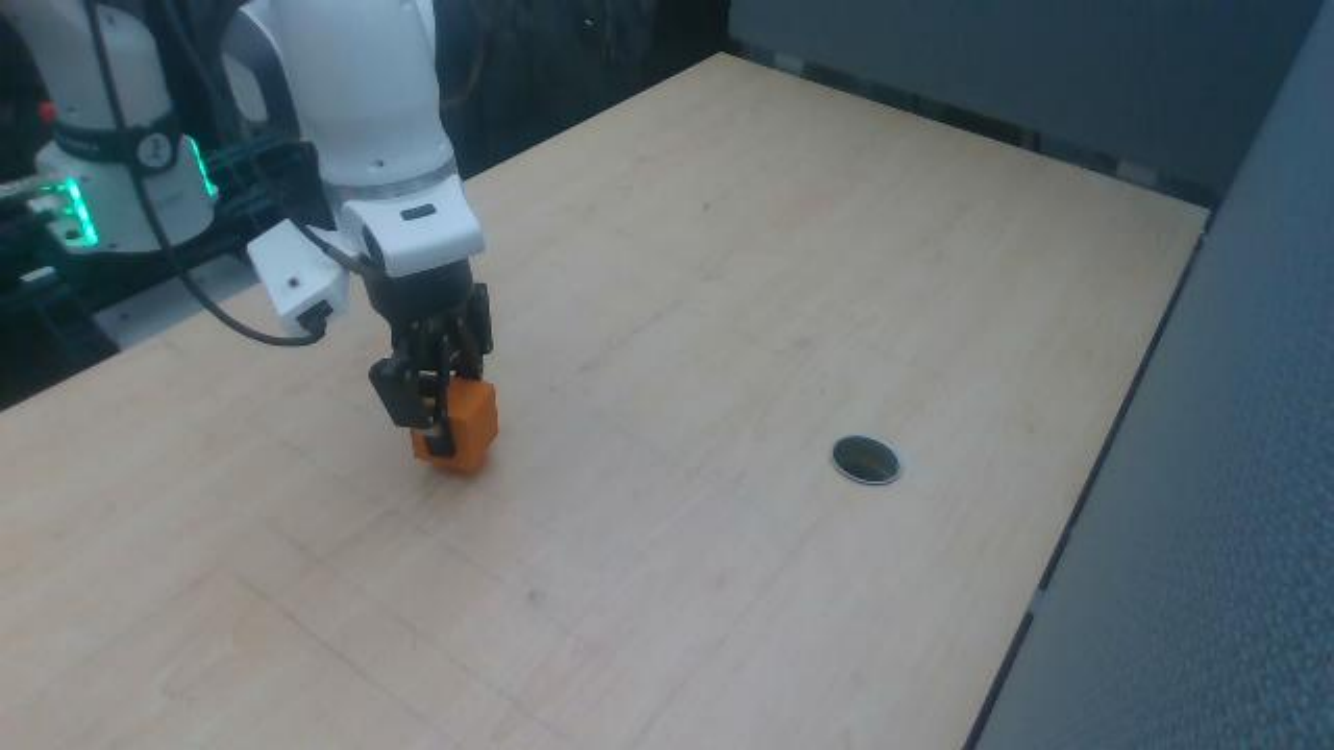} & 
                \includegraphics[width=0.9\linewidth]{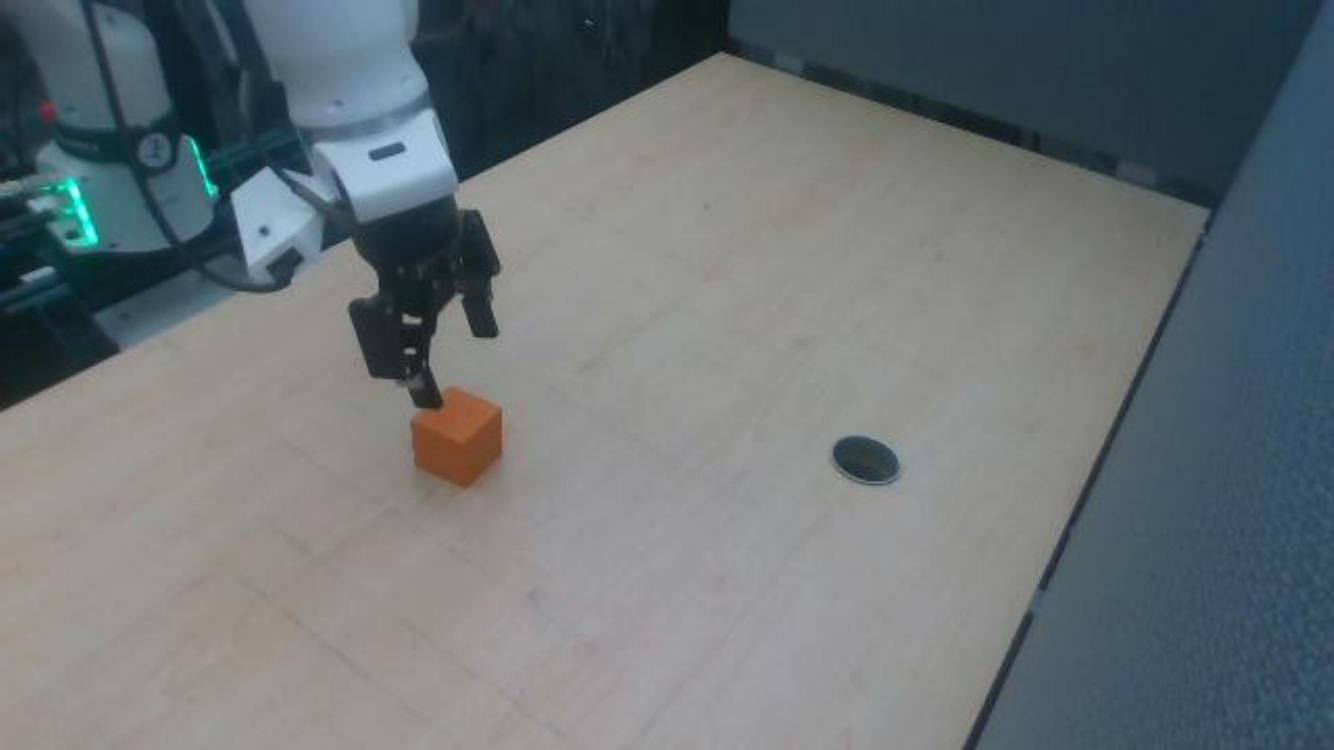} \\
            & Allegro & 
                \includegraphics[width=0.9\linewidth]{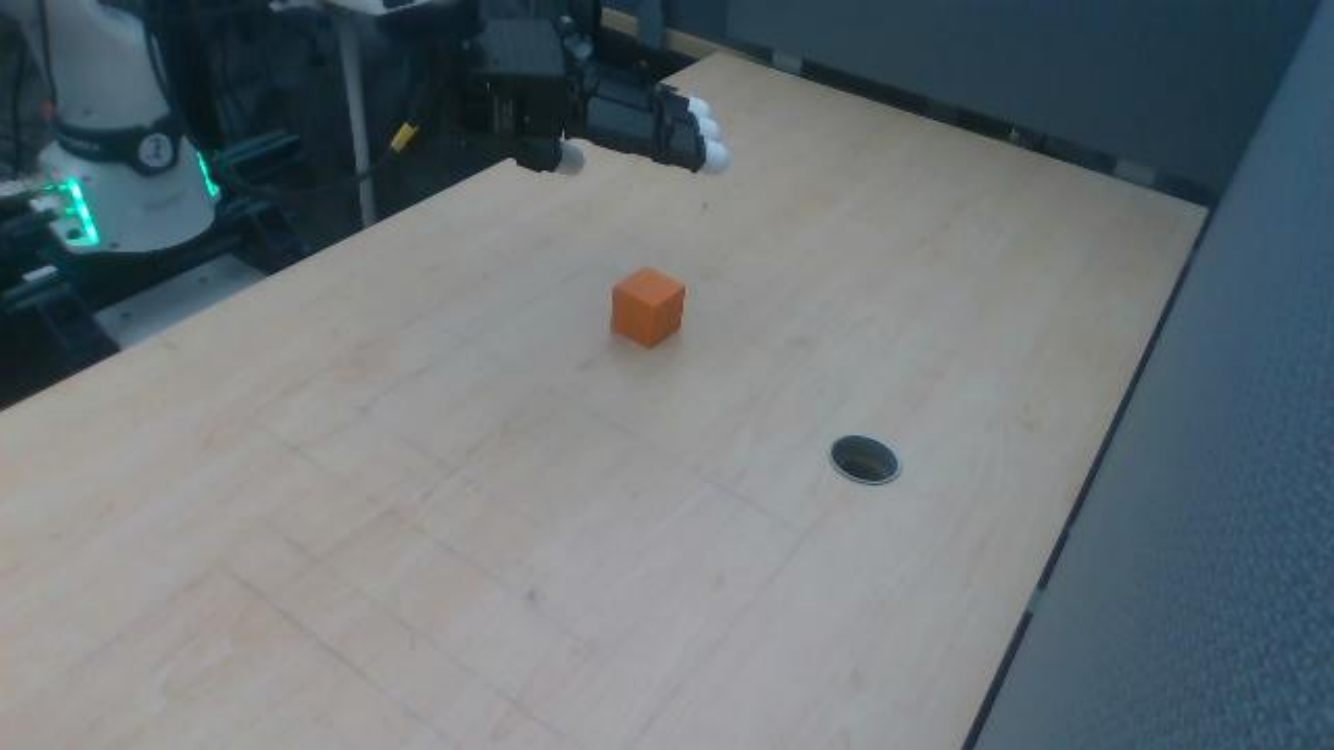} & 
                \includegraphics[width=0.9\linewidth]{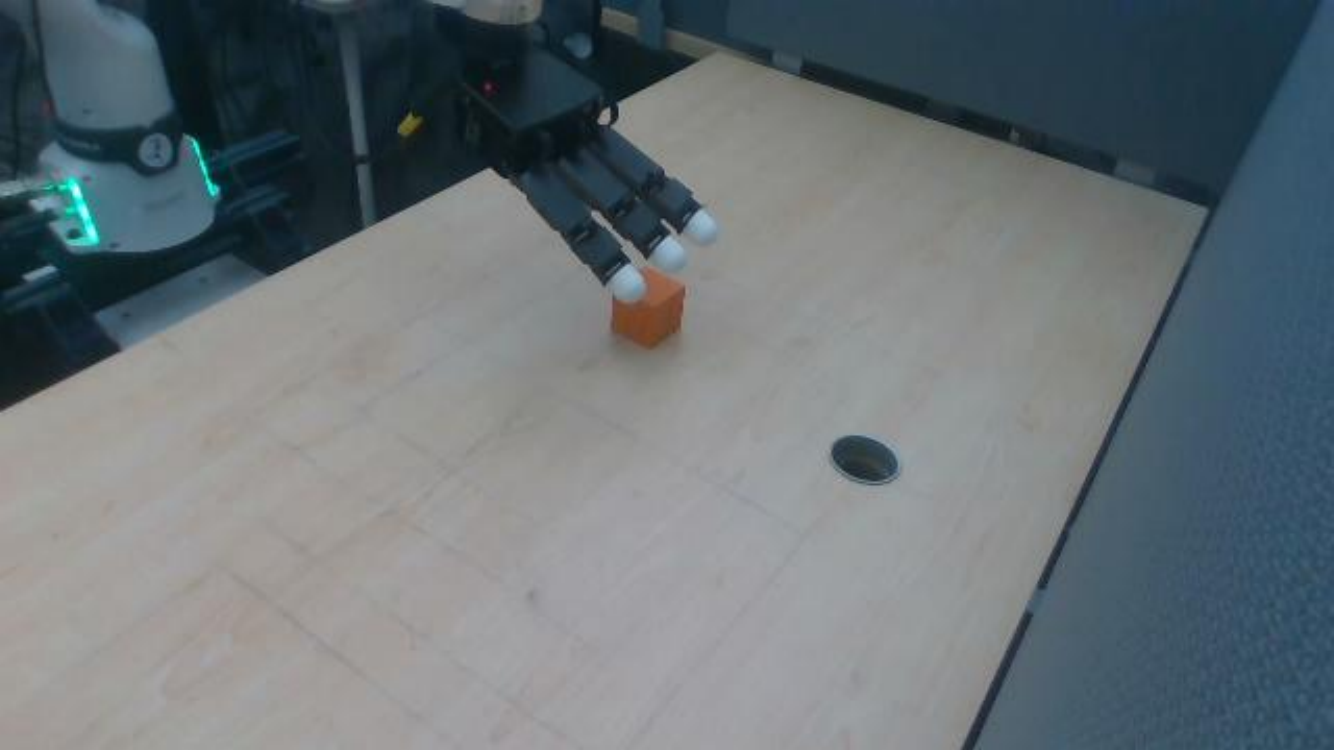} & 
                \includegraphics[width=0.9\linewidth]{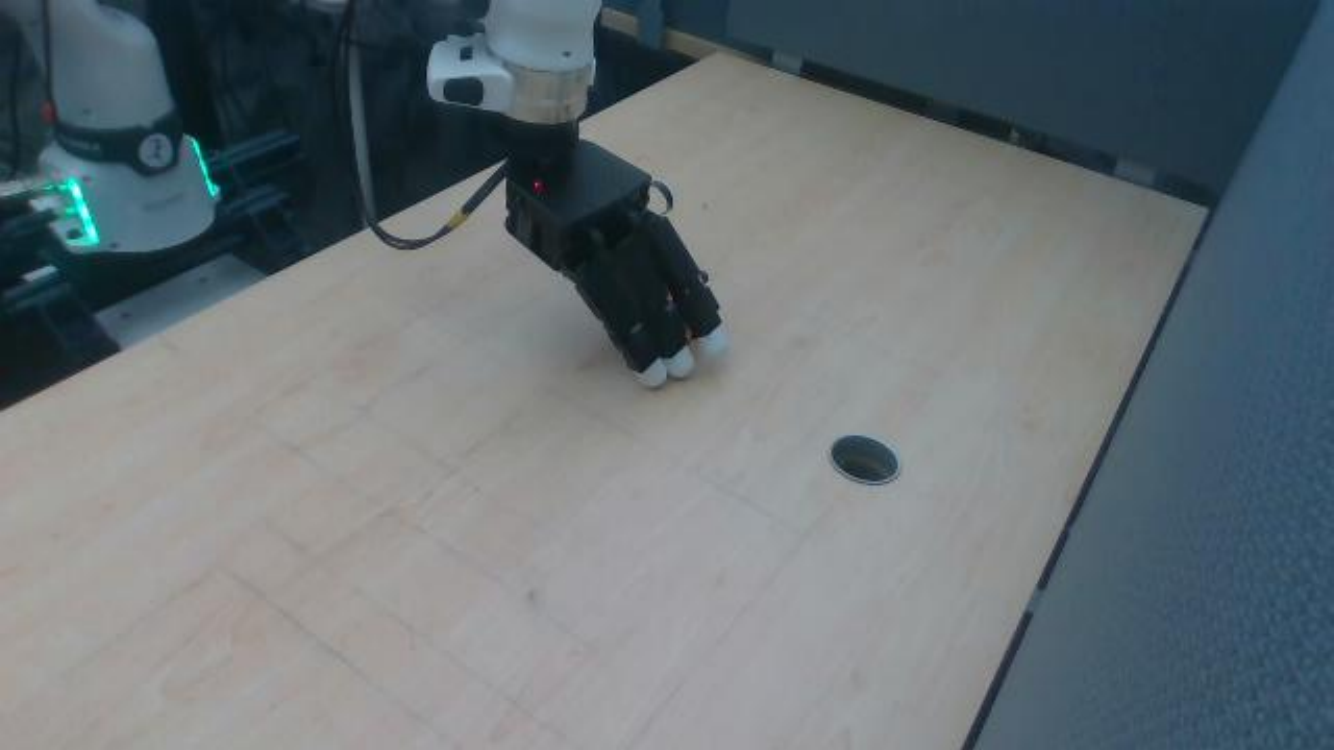} & 
                \includegraphics[width=0.9\linewidth]{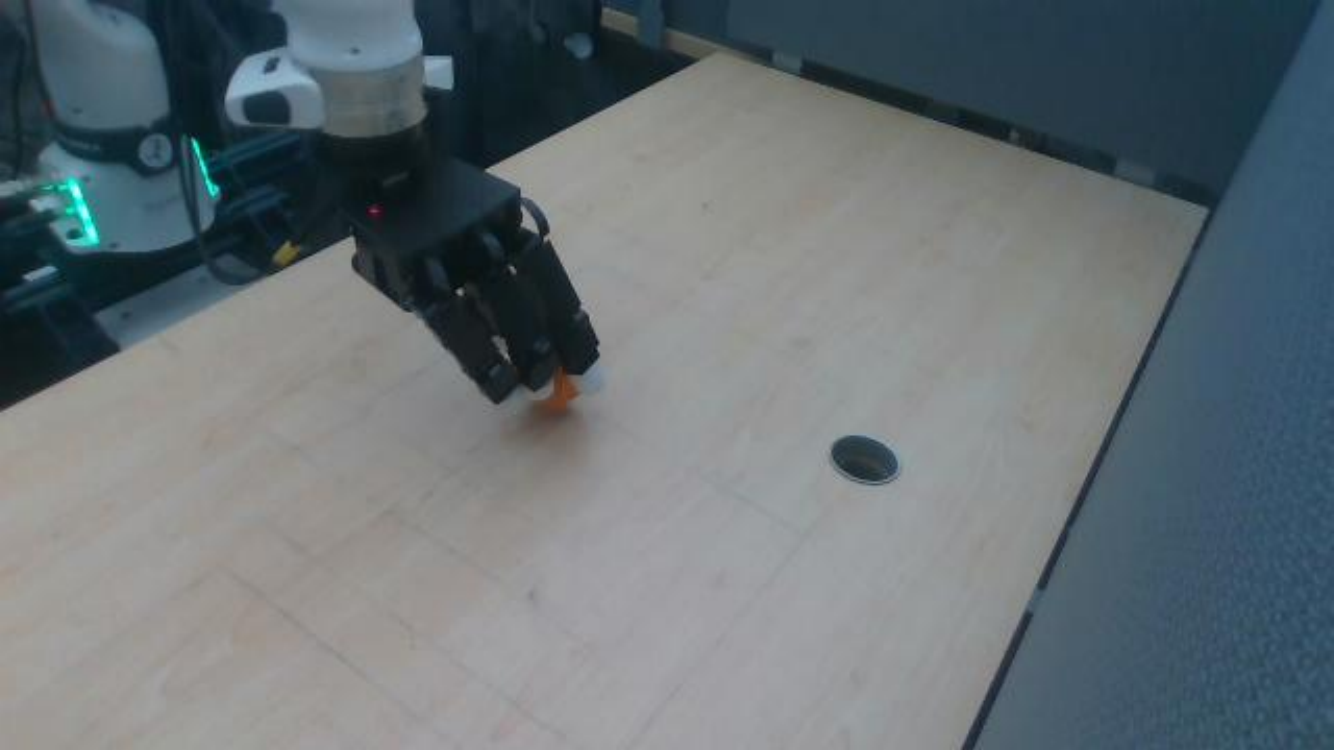} & 
                \includegraphics[width=0.9\linewidth]{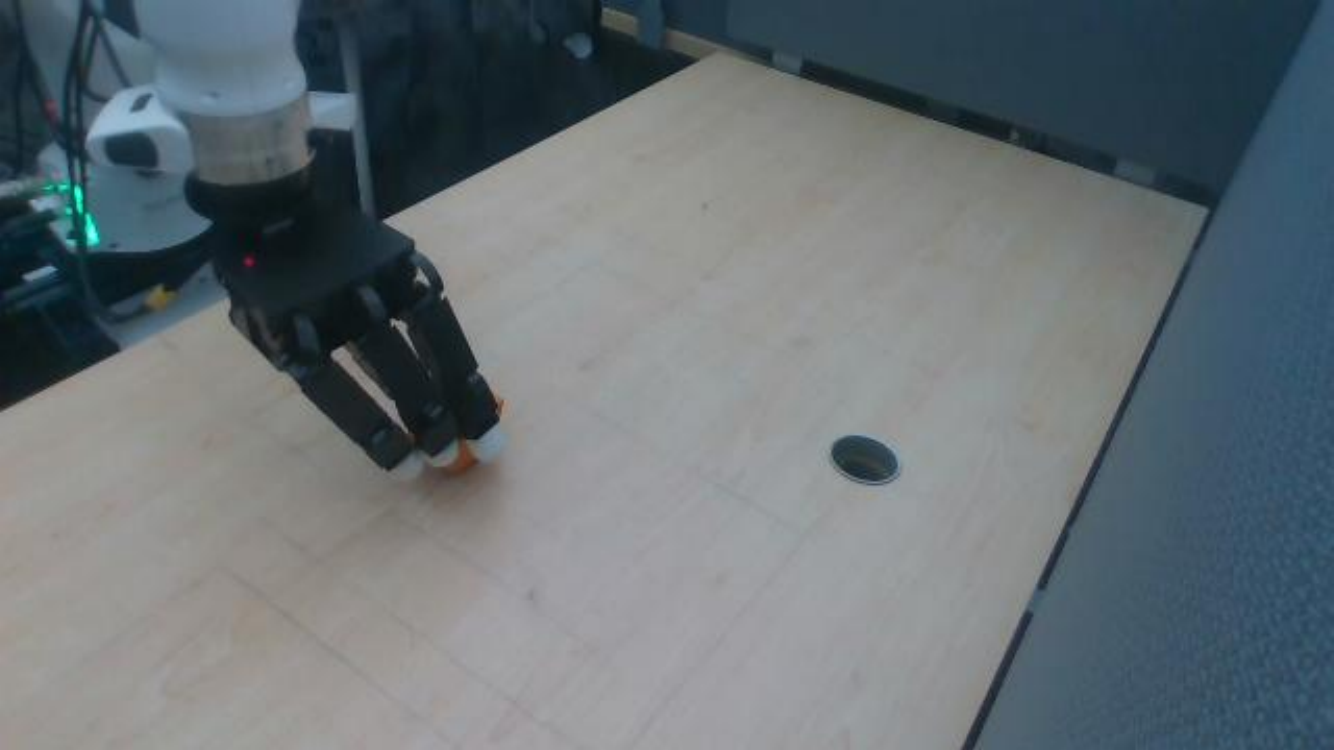} & 
                \includegraphics[width=0.9\linewidth]{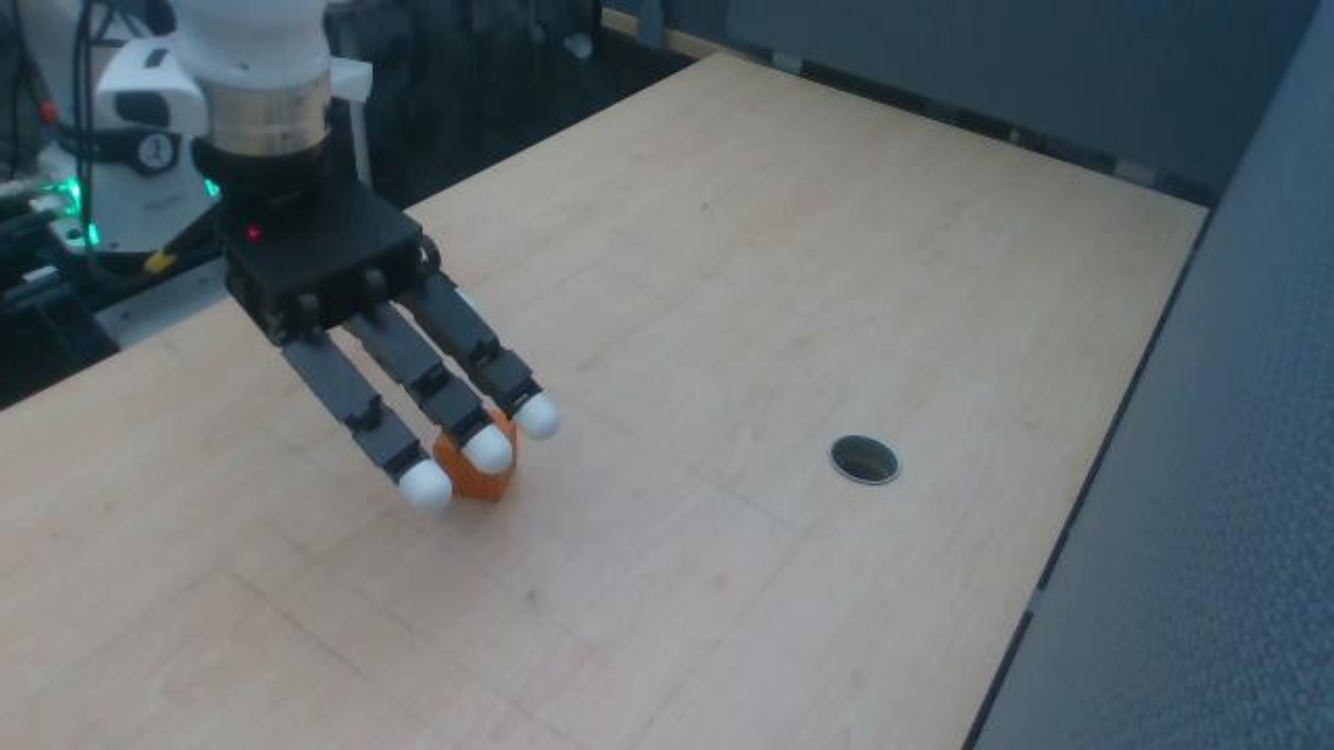} \\
            & Ability & 
                \includegraphics[width=0.9\linewidth]{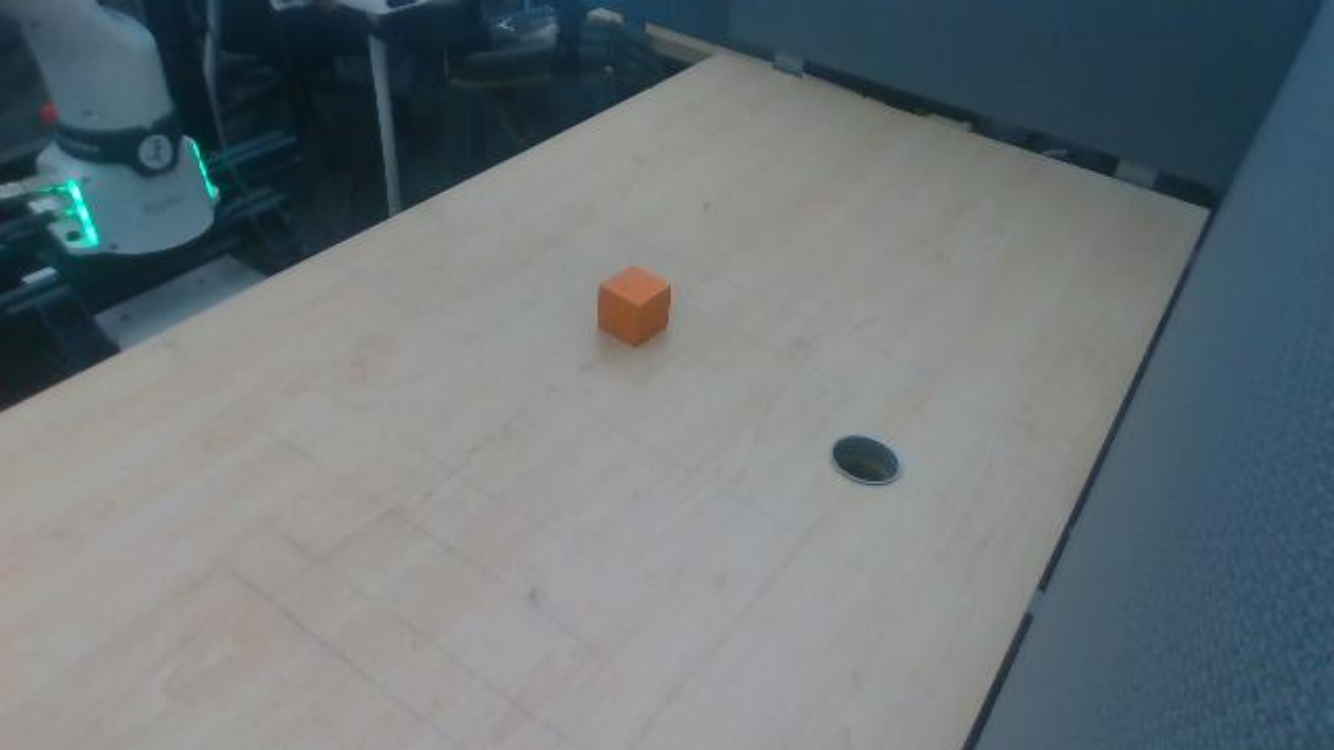} & 
                \includegraphics[width=0.9\linewidth]{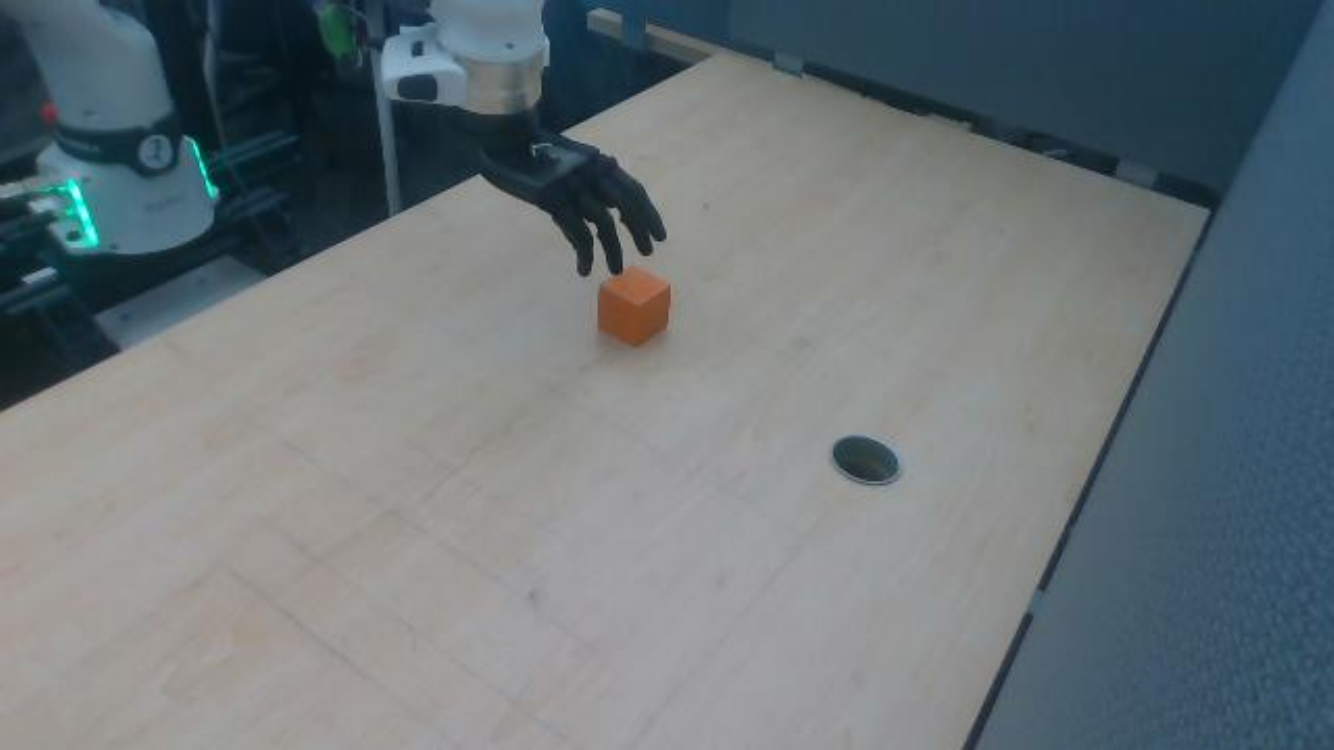} & 
                \includegraphics[width=0.9\linewidth]{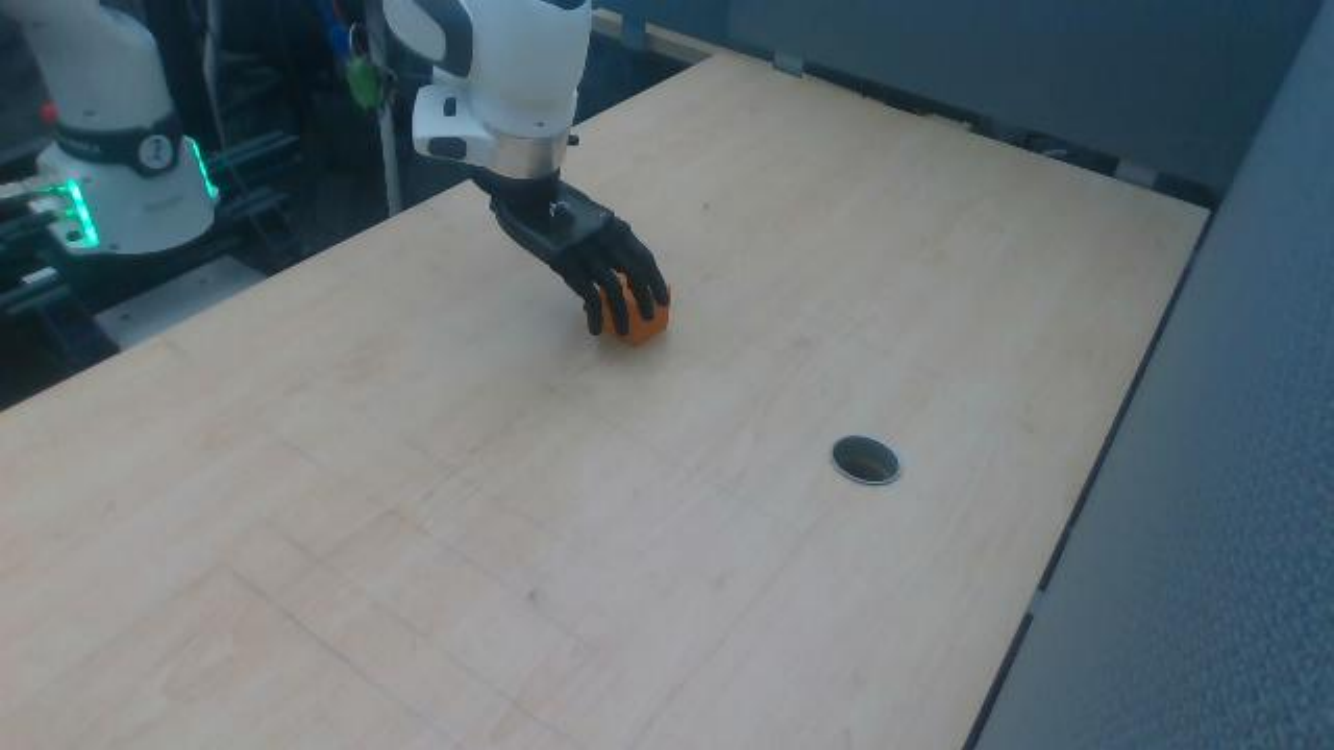} & 
                \includegraphics[width=0.9\linewidth]{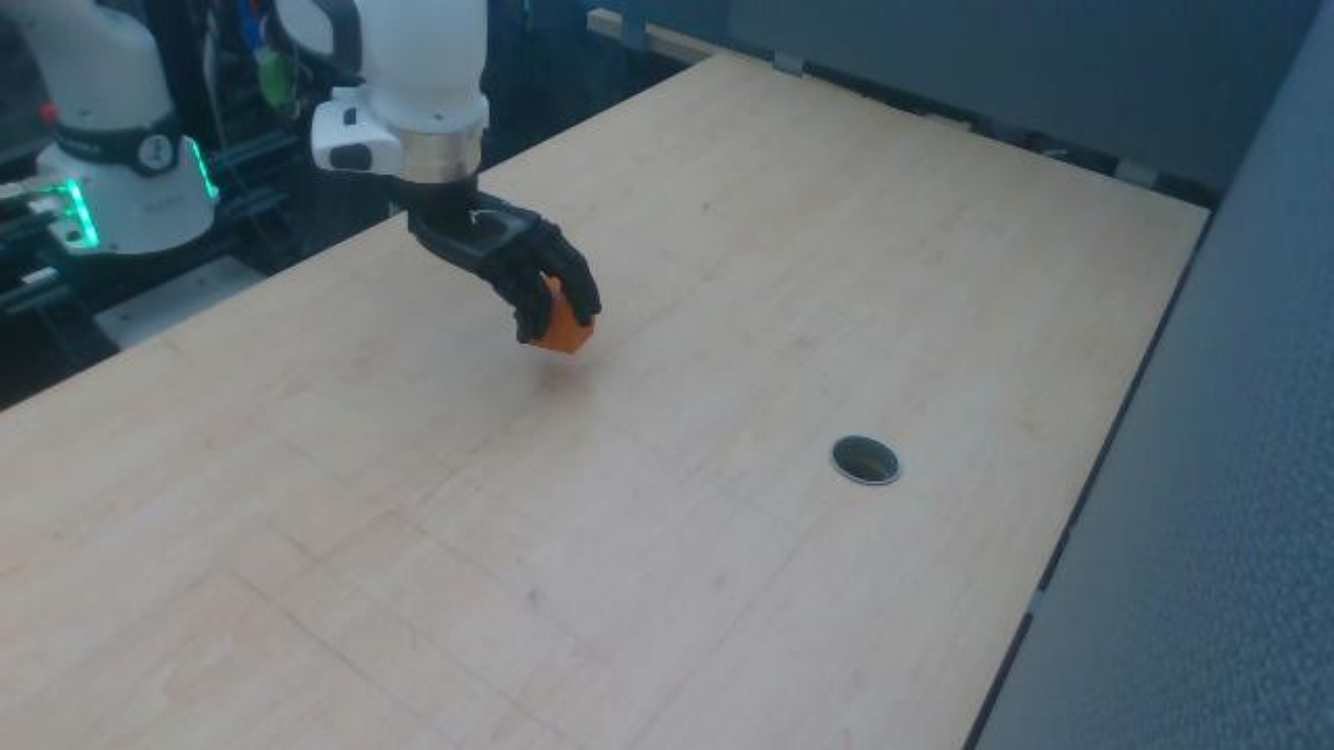} & 
                \includegraphics[width=0.9\linewidth]{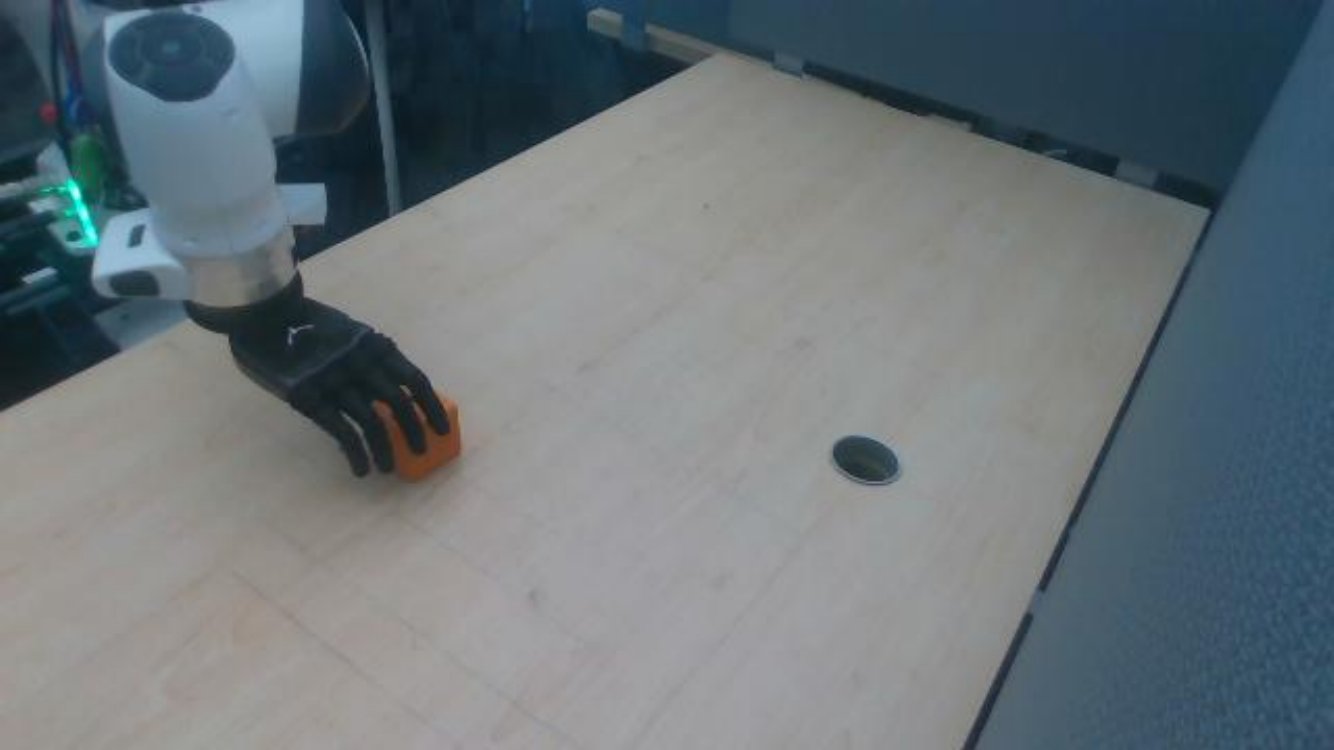} & 
                \includegraphics[width=0.9\linewidth]{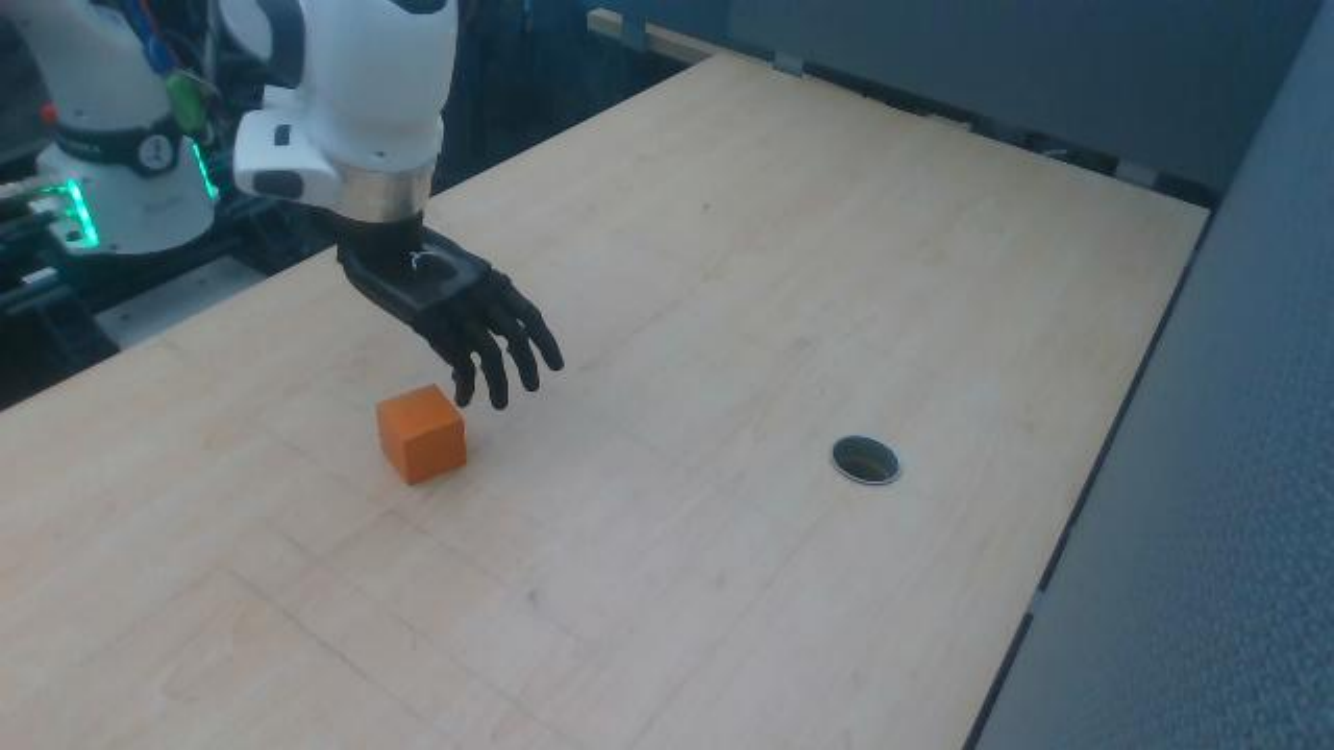} \\
        \midrule
        \multirow{5}{*}{\vspace*{-10ex}Push}
            & Human & 
                \includegraphics[width=0.9\linewidth]{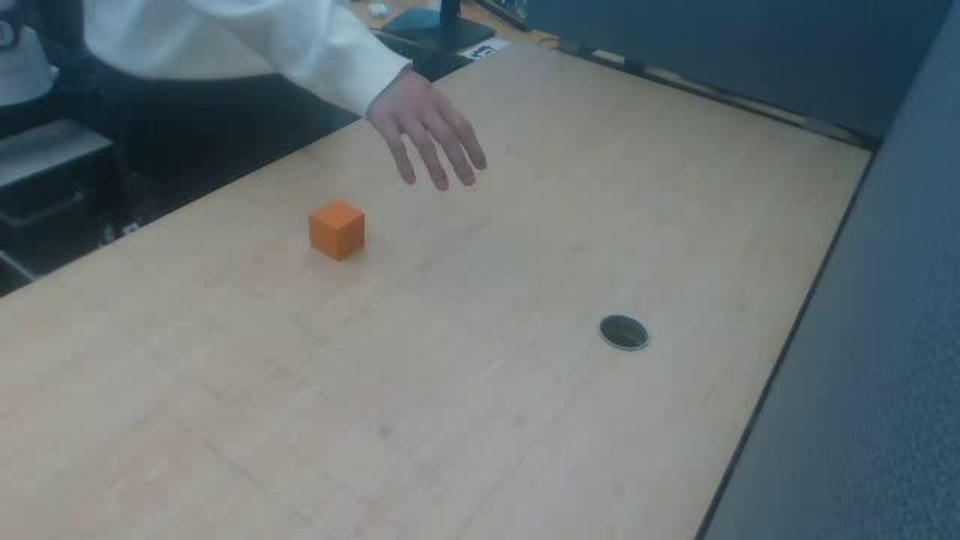} & 
                \includegraphics[width=0.9\linewidth]{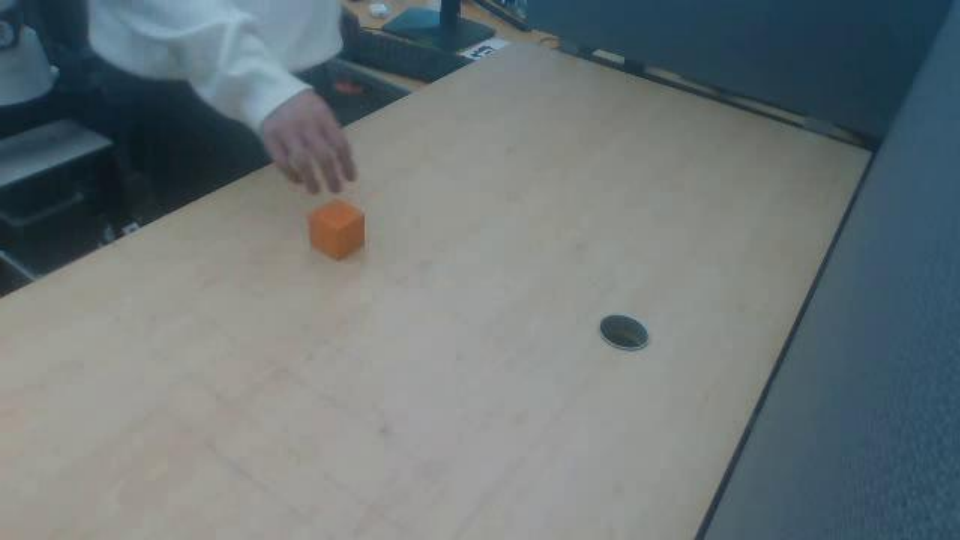} & 
                \includegraphics[width=0.9\linewidth]{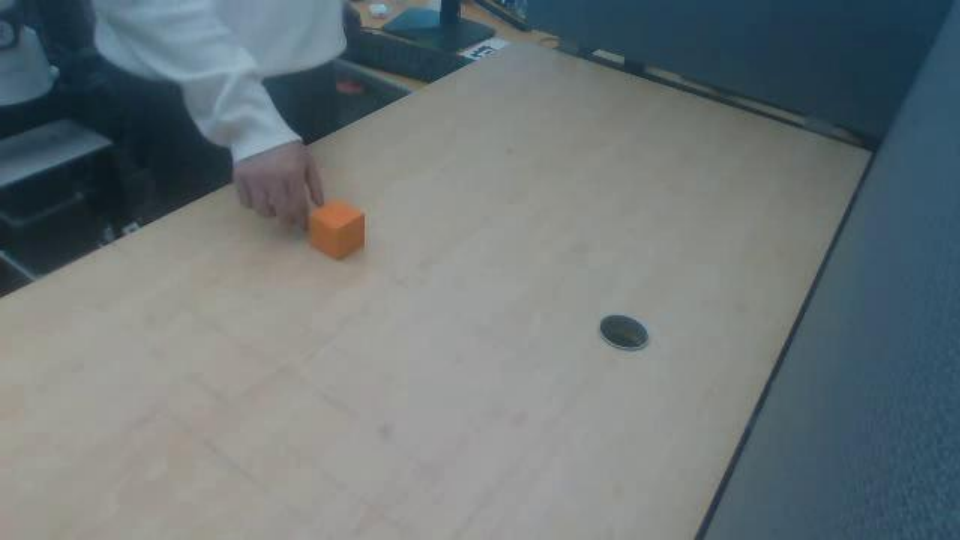} & 
                \includegraphics[width=0.9\linewidth]{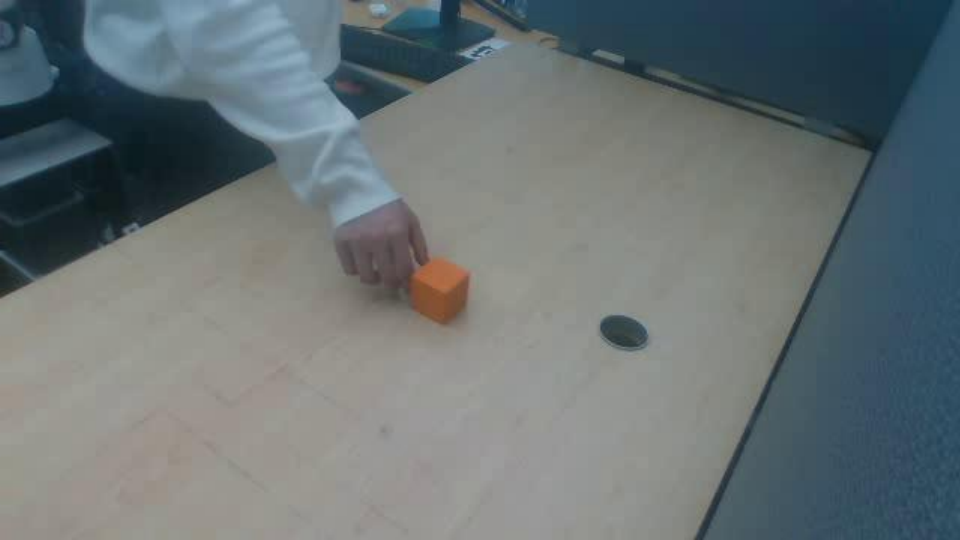} & 
                \includegraphics[width=0.9\linewidth]{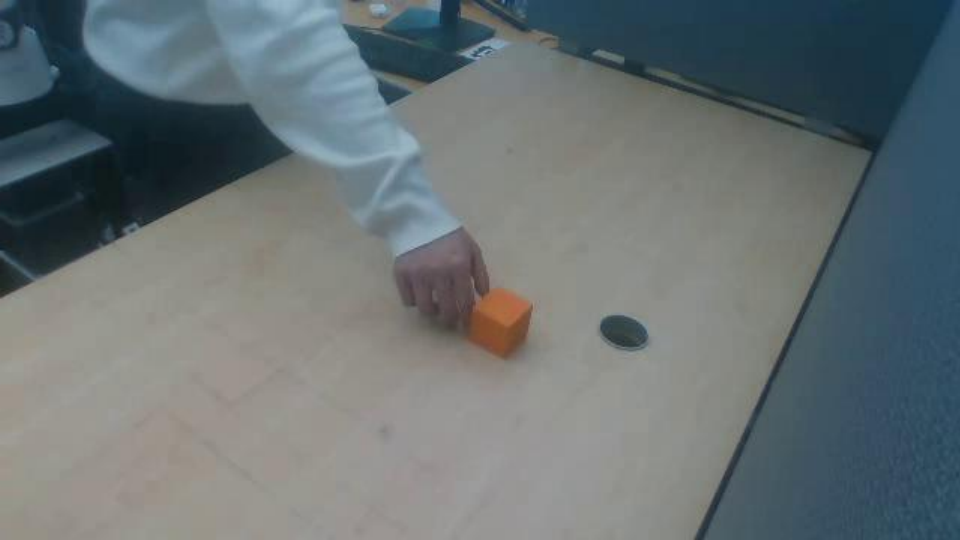} & 
                \includegraphics[width=0.9\linewidth]{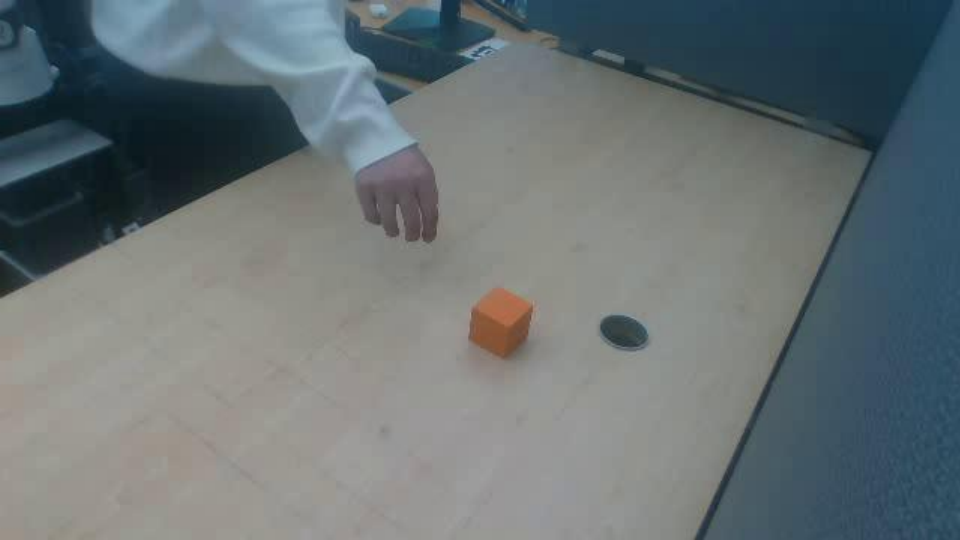} \\
            & FR & 
                \includegraphics[width=0.9\linewidth]{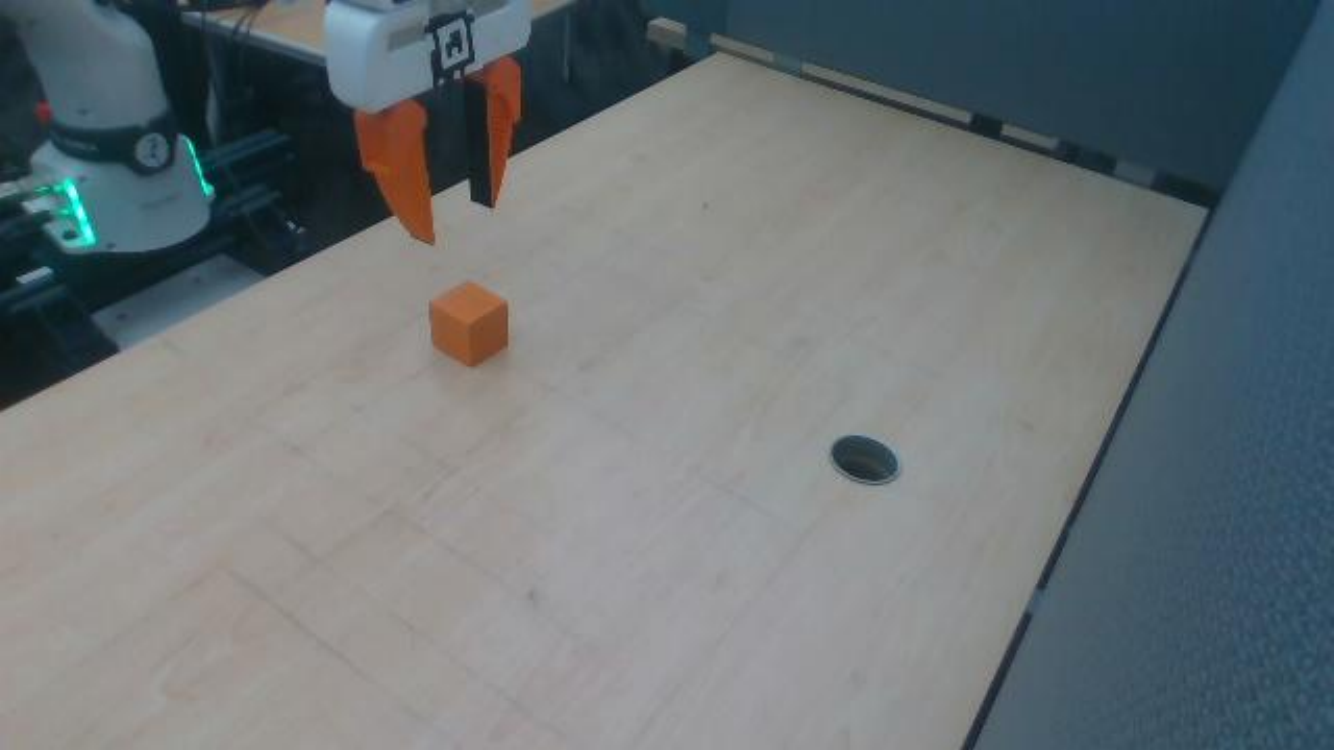} & 
                \includegraphics[width=0.9\linewidth]{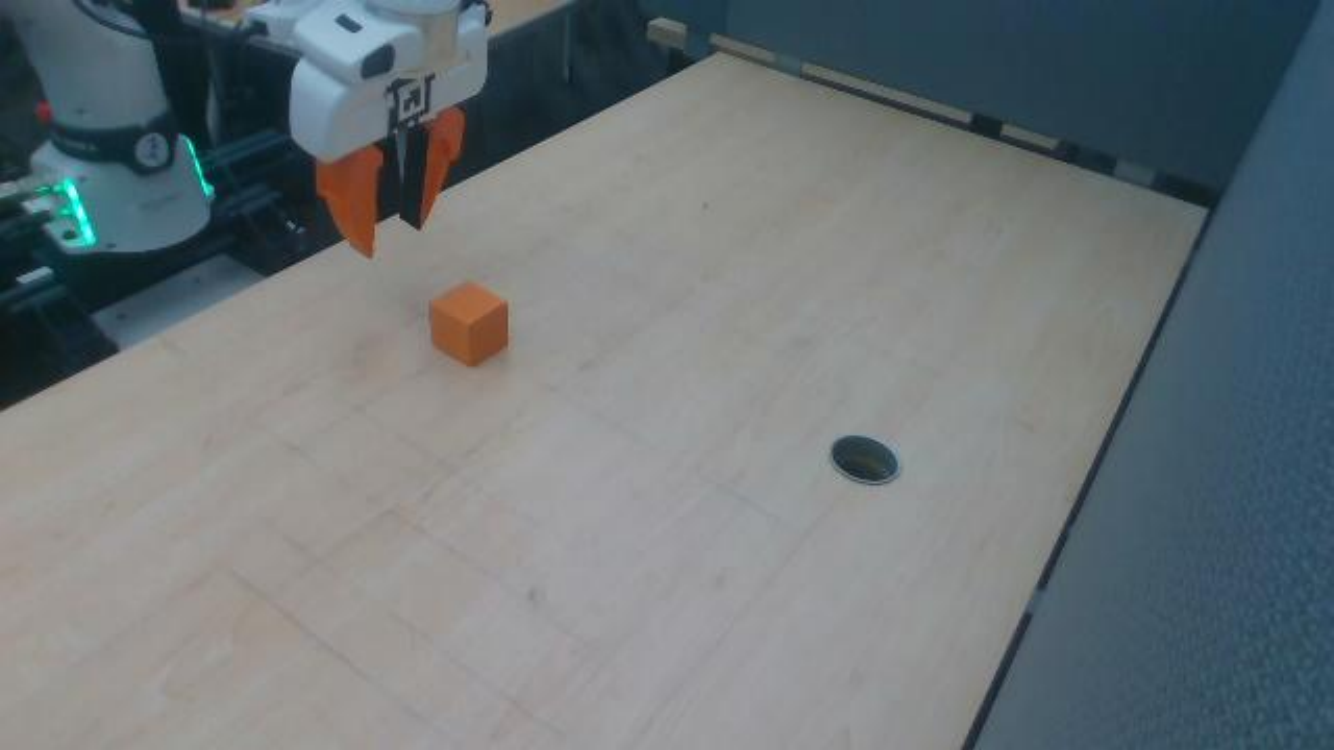} & 
                \includegraphics[width=0.9\linewidth]{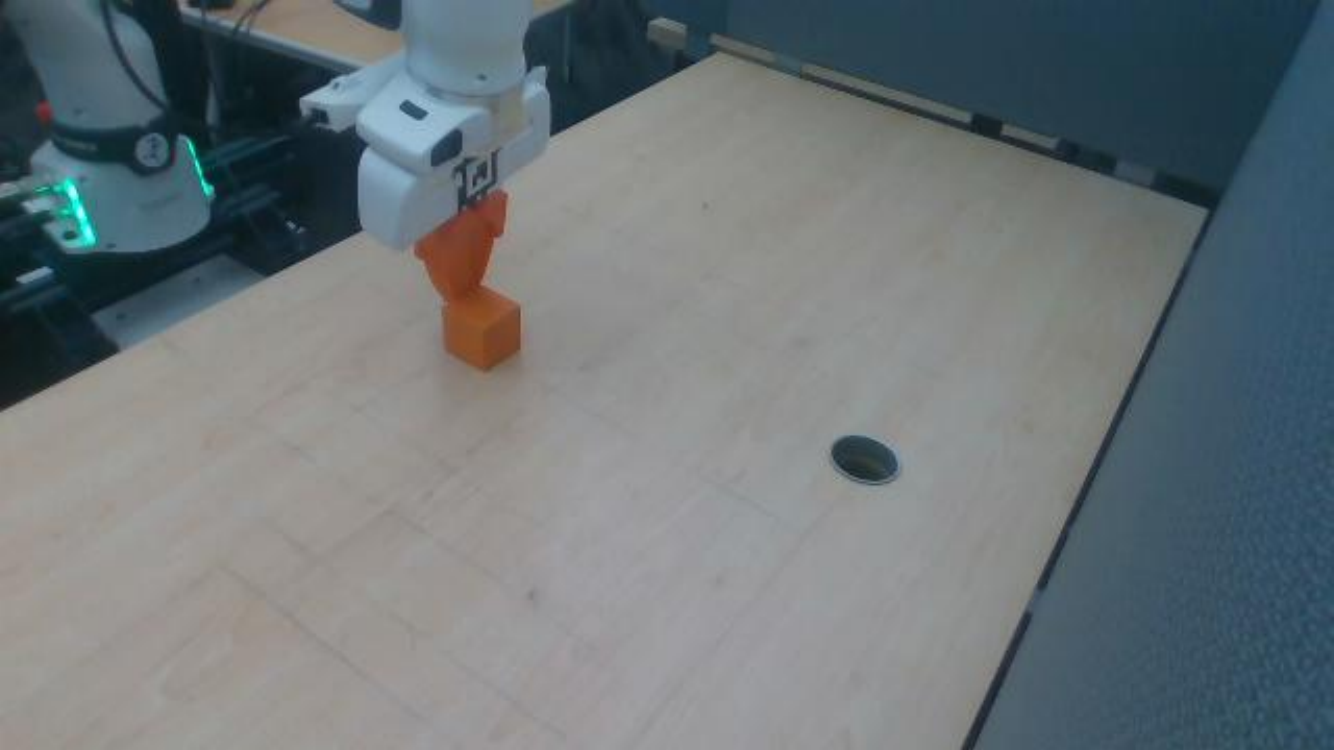} & 
                \includegraphics[width=0.9\linewidth]{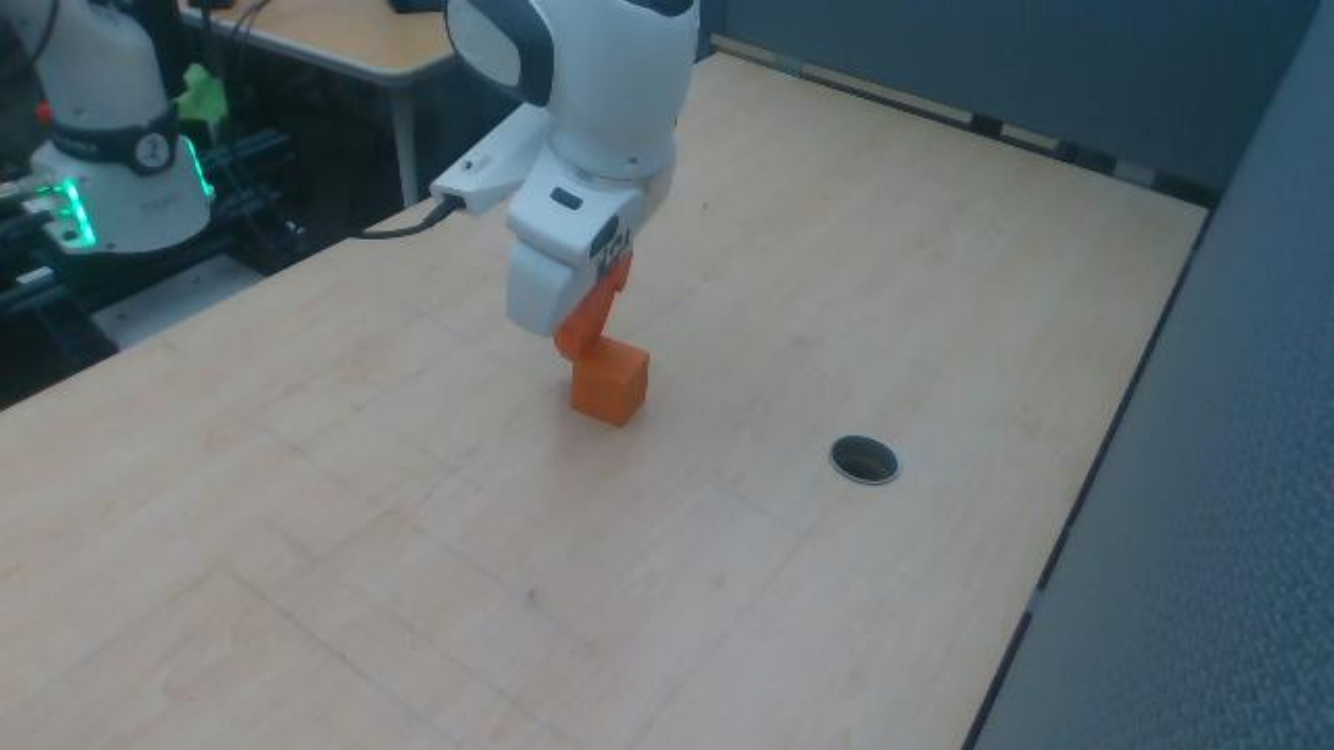} & 
                \includegraphics[width=0.9\linewidth]{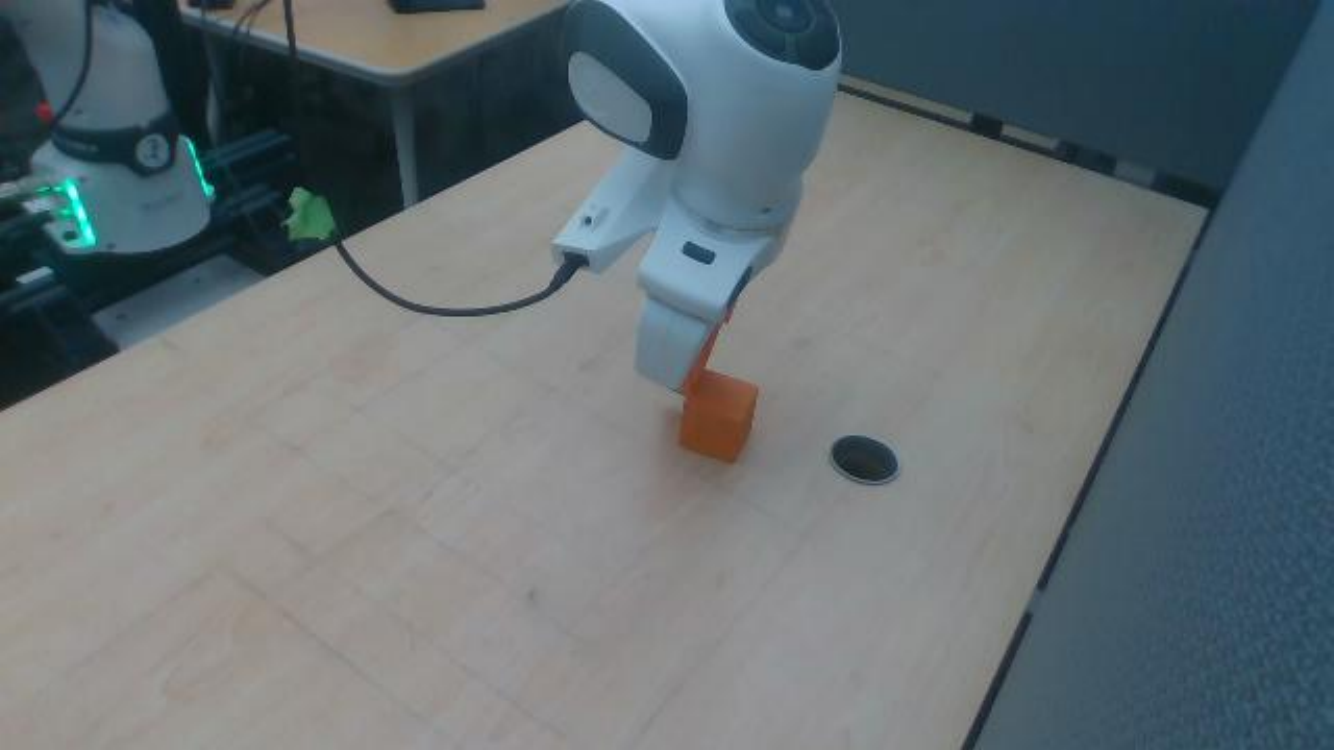} & 
                \includegraphics[width=0.9\linewidth]{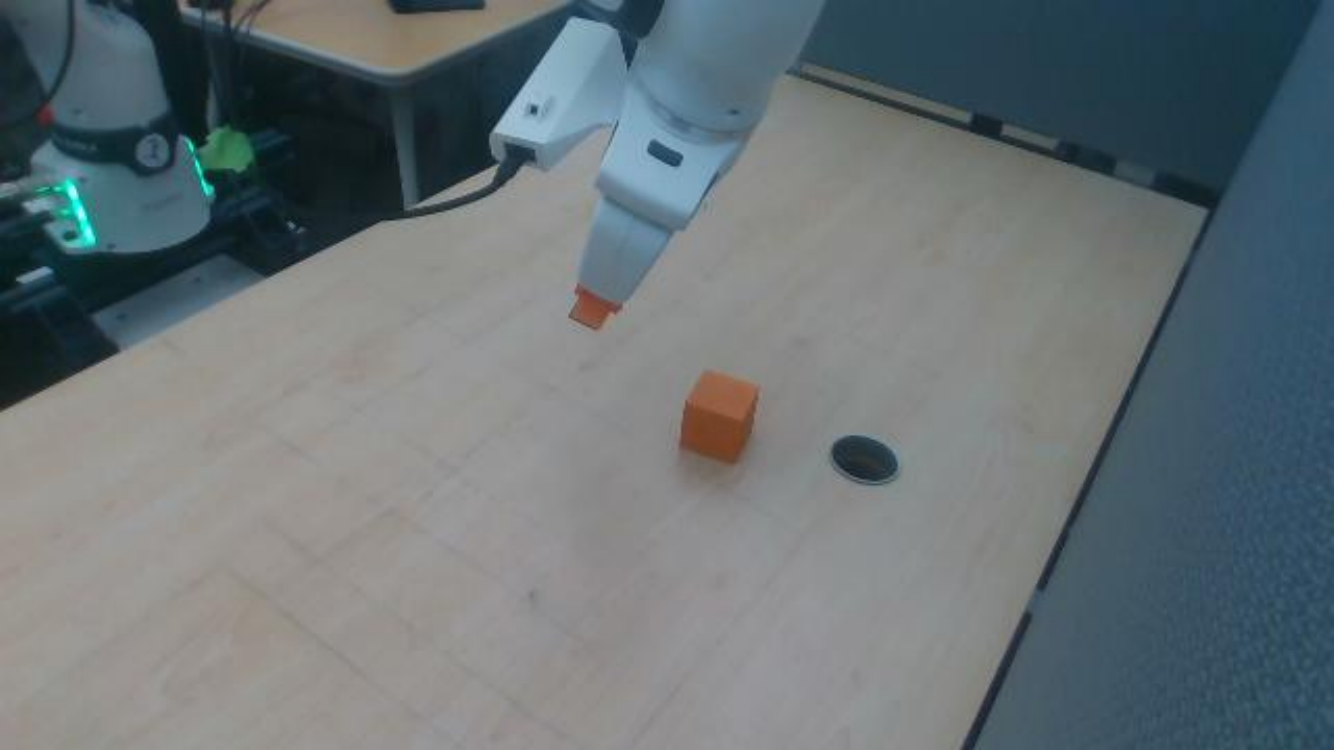} \\
            & Robotiq & 
                \includegraphics[width=0.9\linewidth]{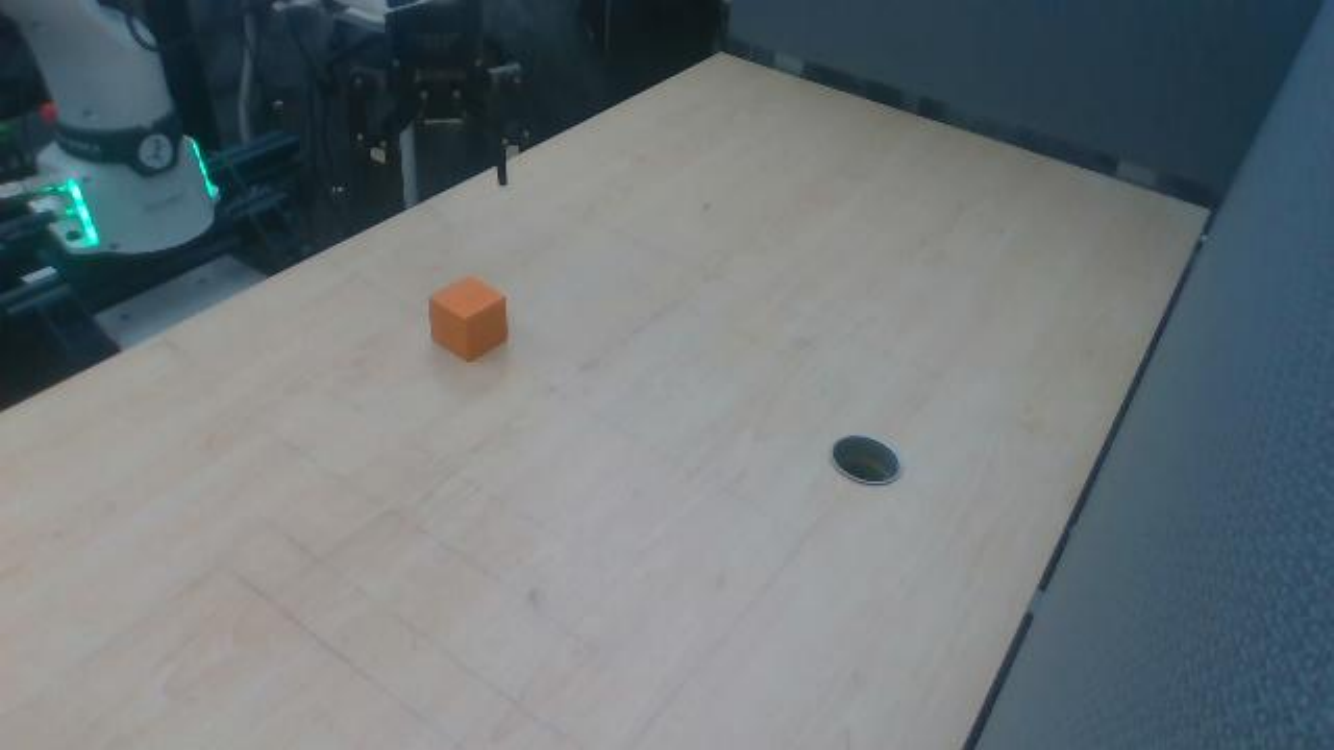} & 
                \includegraphics[width=0.9\linewidth]{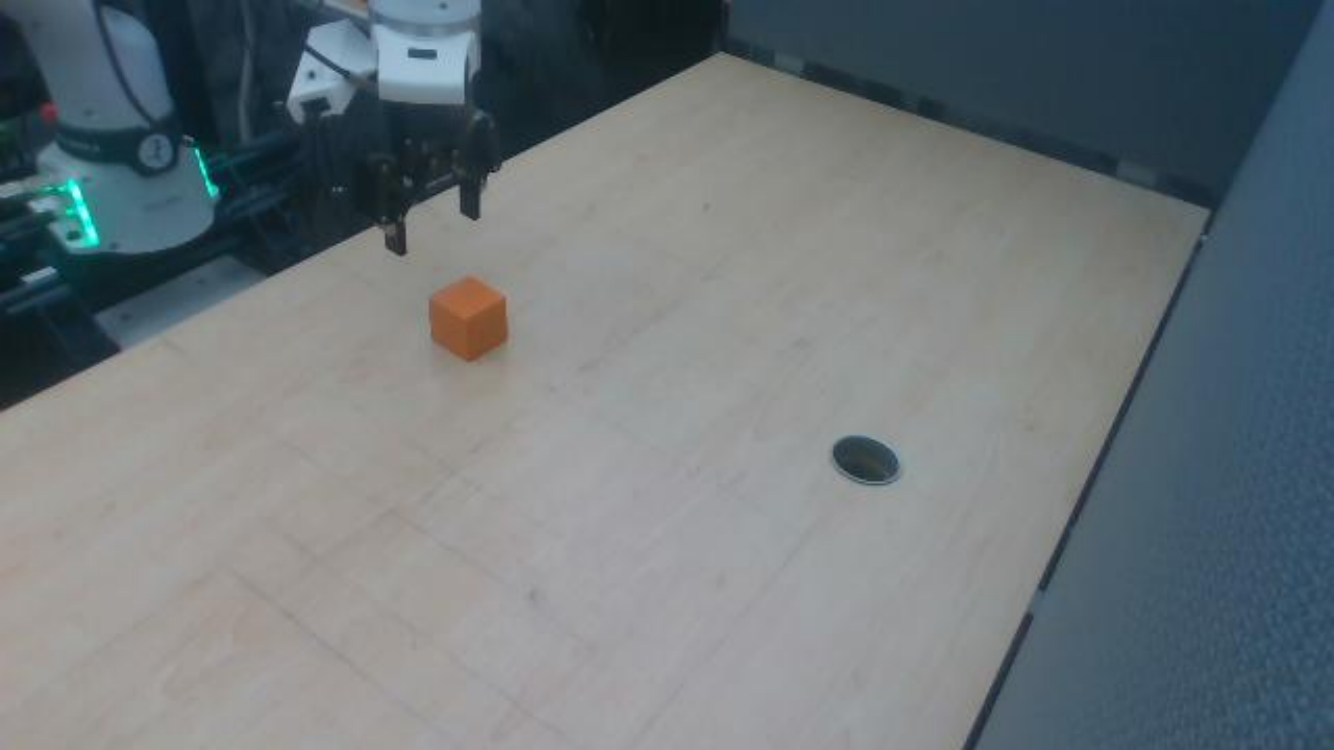} & 
                \includegraphics[width=0.9\linewidth]{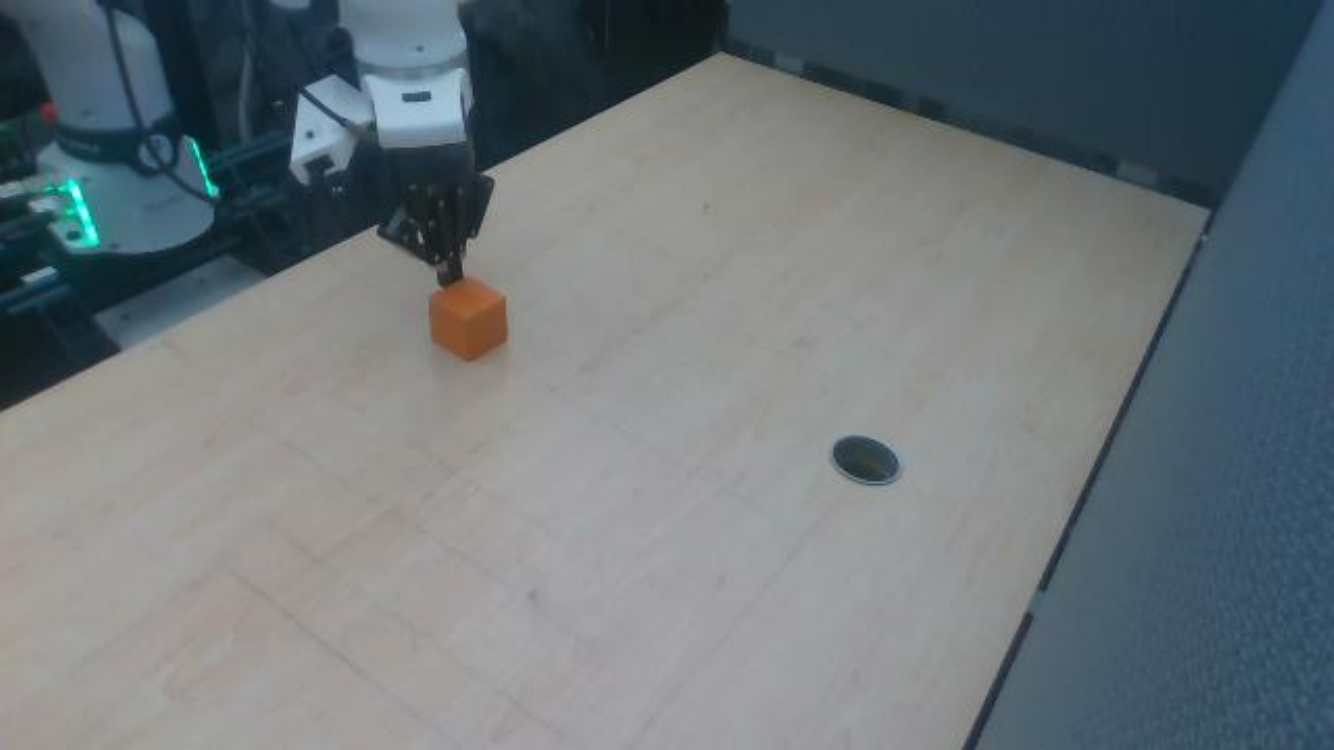} & 
                \includegraphics[width=0.9\linewidth]{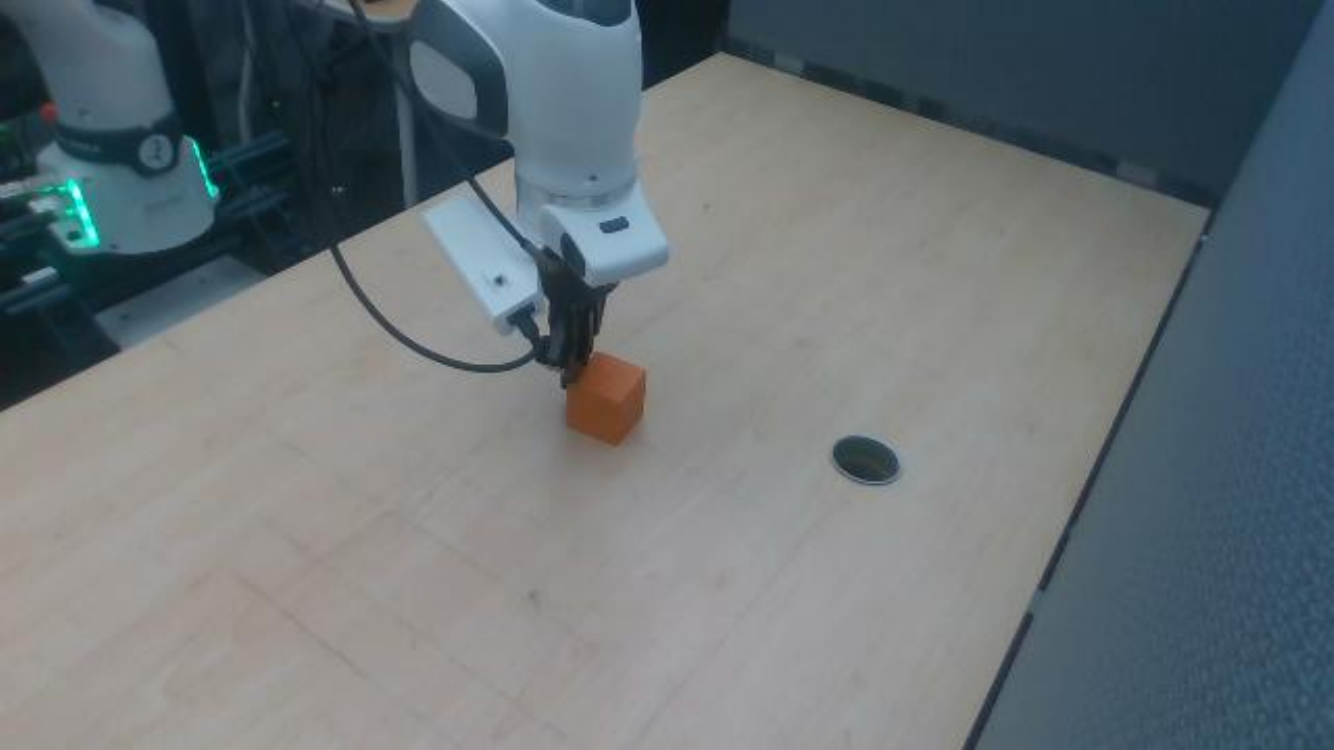} & 
                \includegraphics[width=0.9\linewidth]{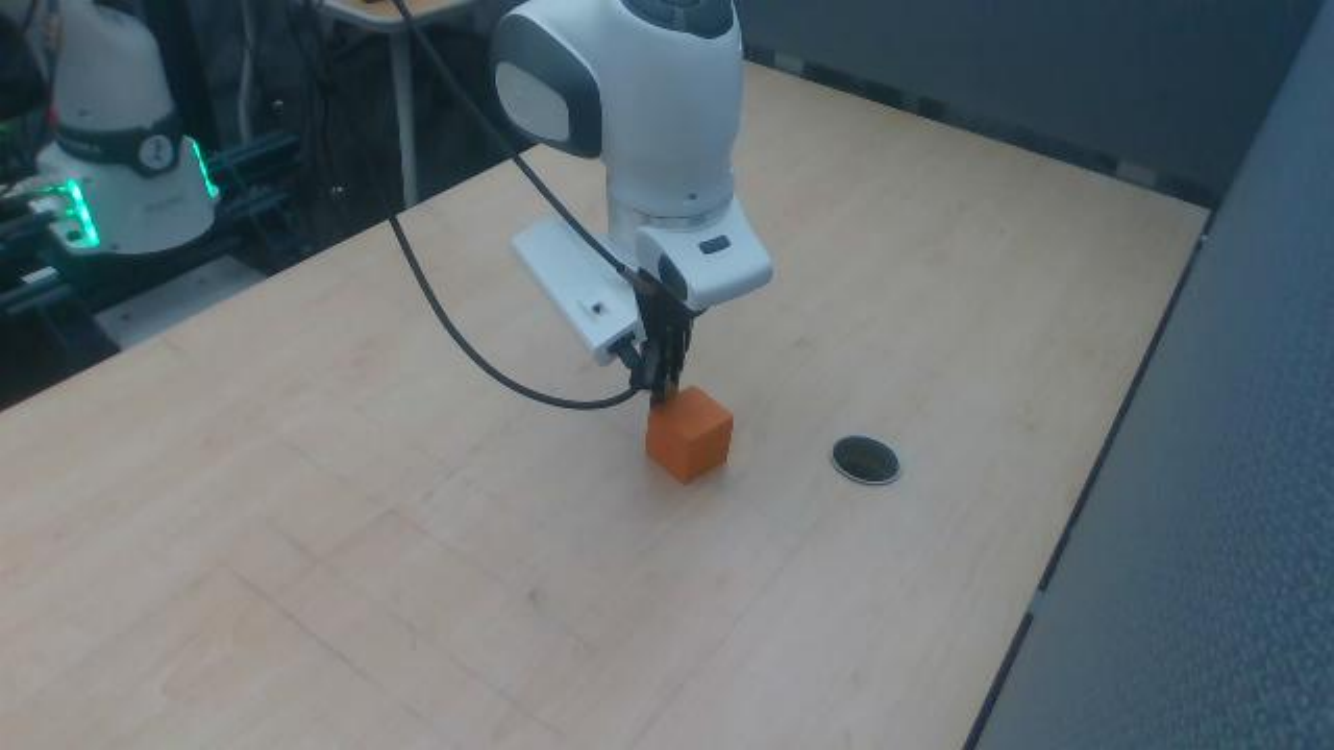} & 
                \includegraphics[width=0.9\linewidth]{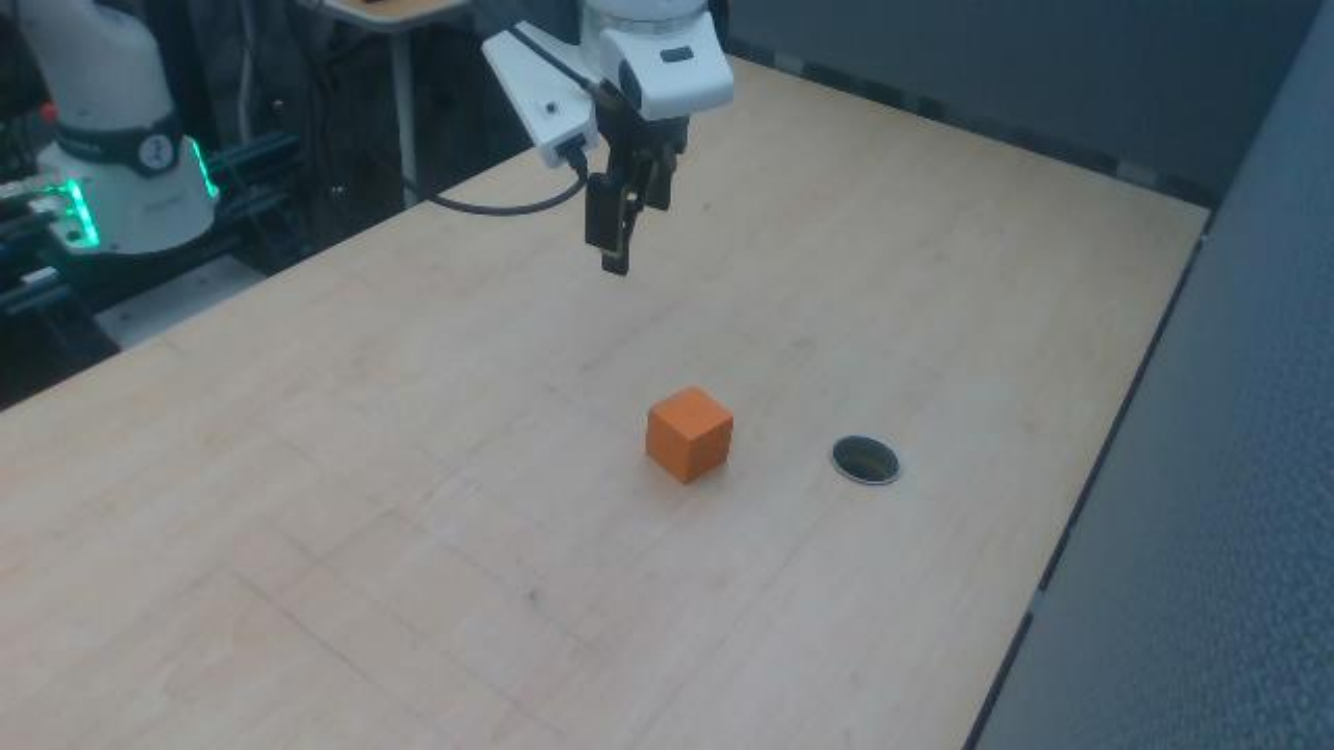} \\
            & Allegro & 
                \includegraphics[width=0.9\linewidth]{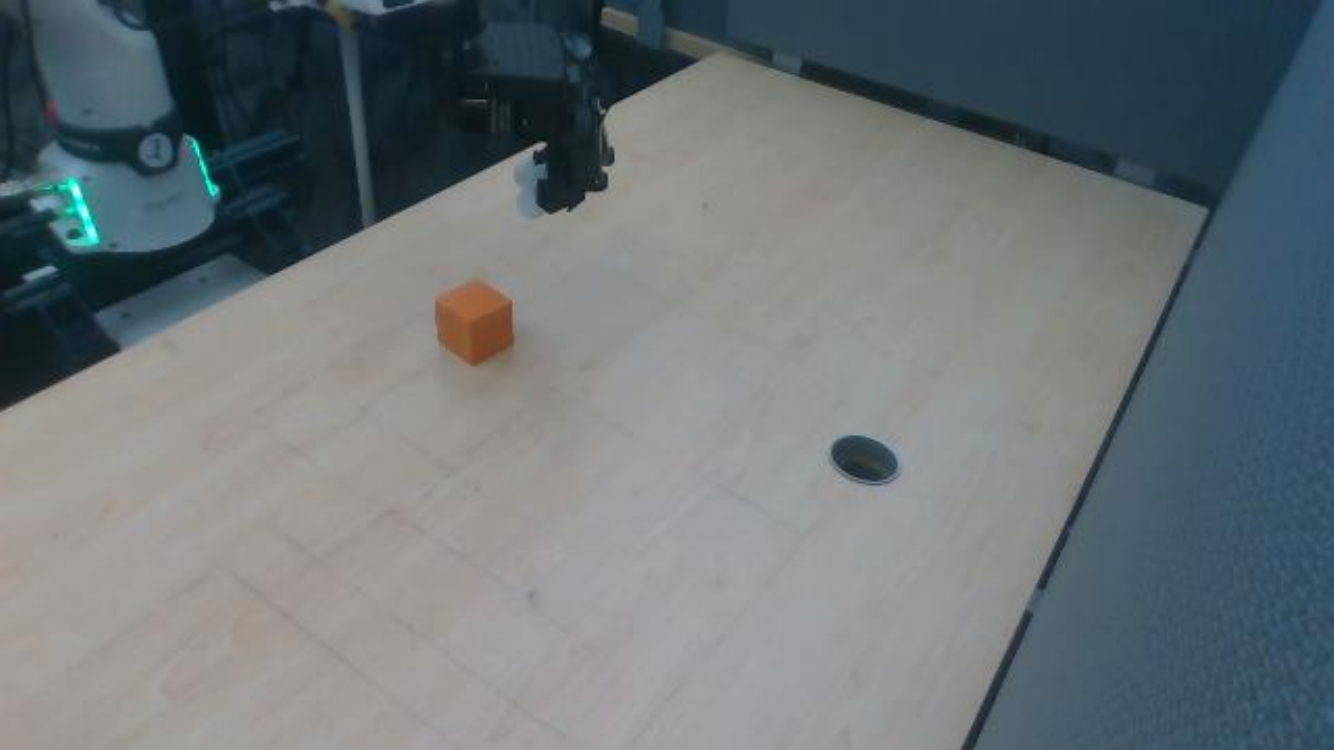} & 
                \includegraphics[width=0.9\linewidth]{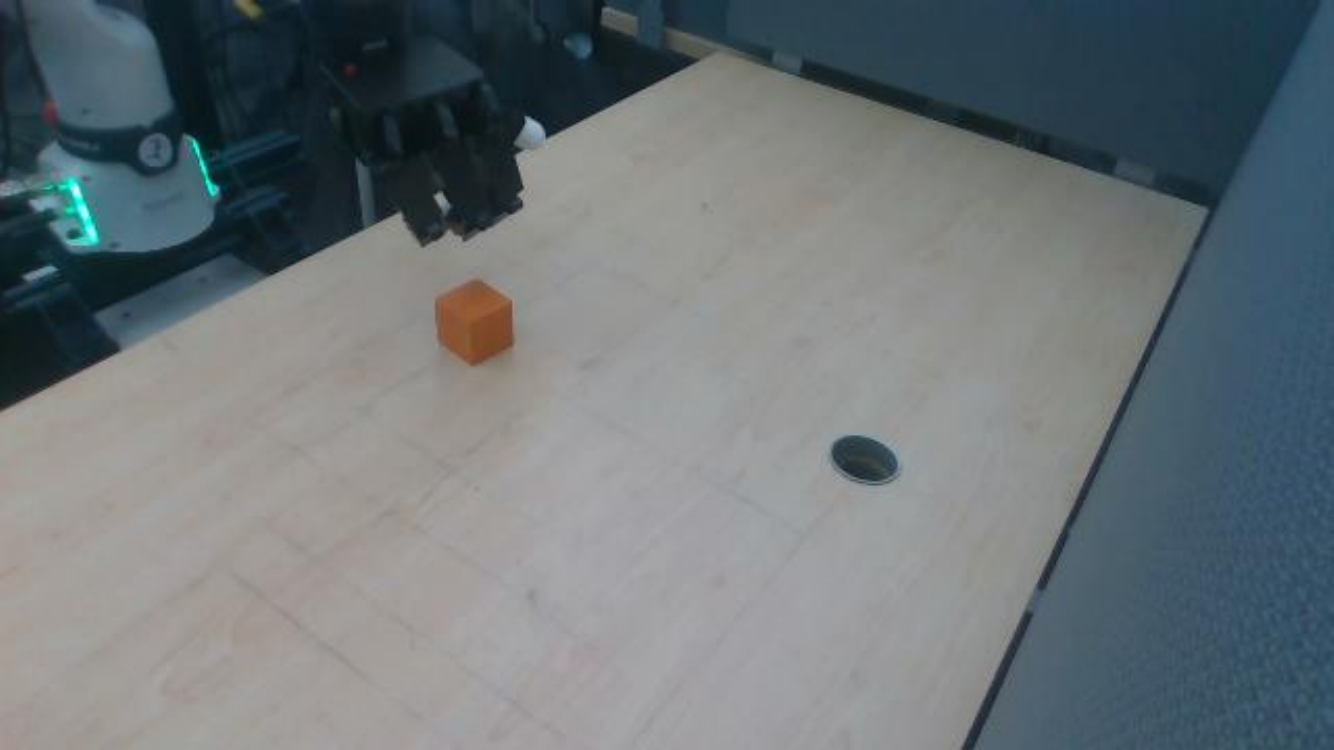} & 
                \includegraphics[width=0.9\linewidth]{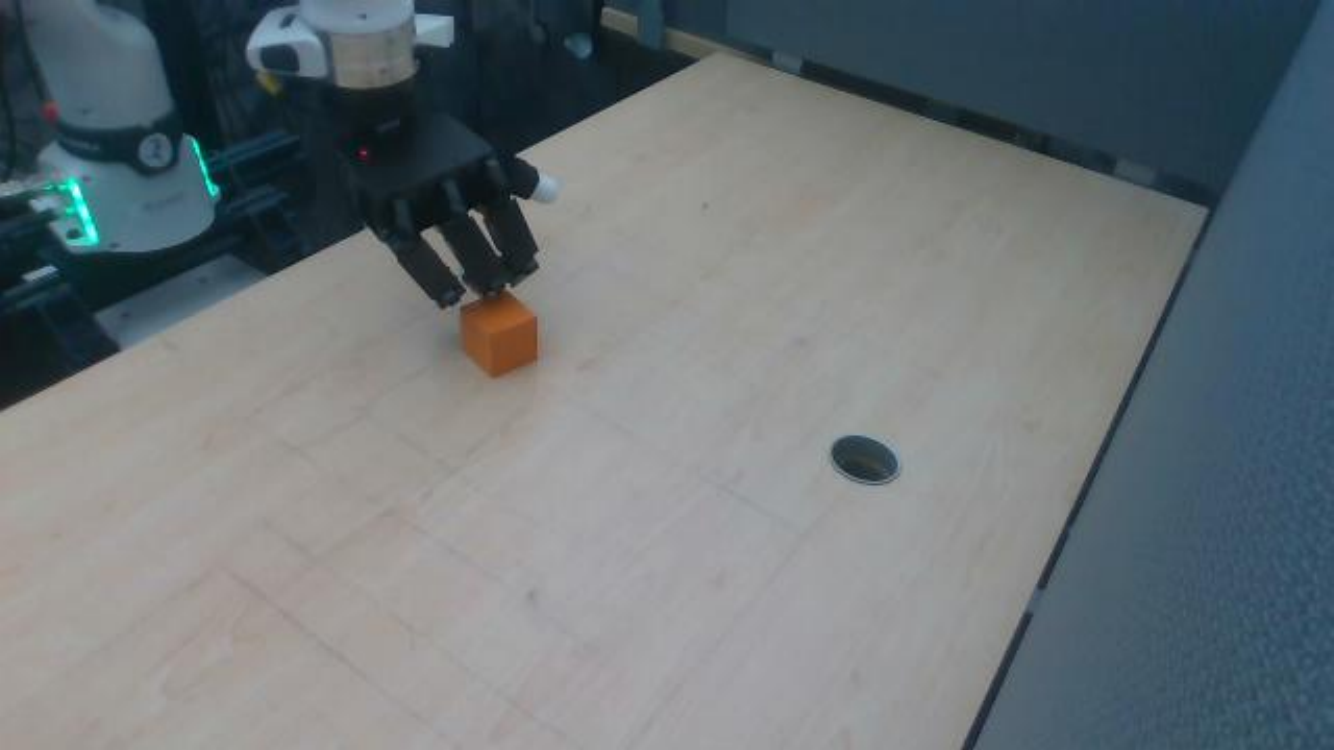} & 
                \includegraphics[width=0.9\linewidth]{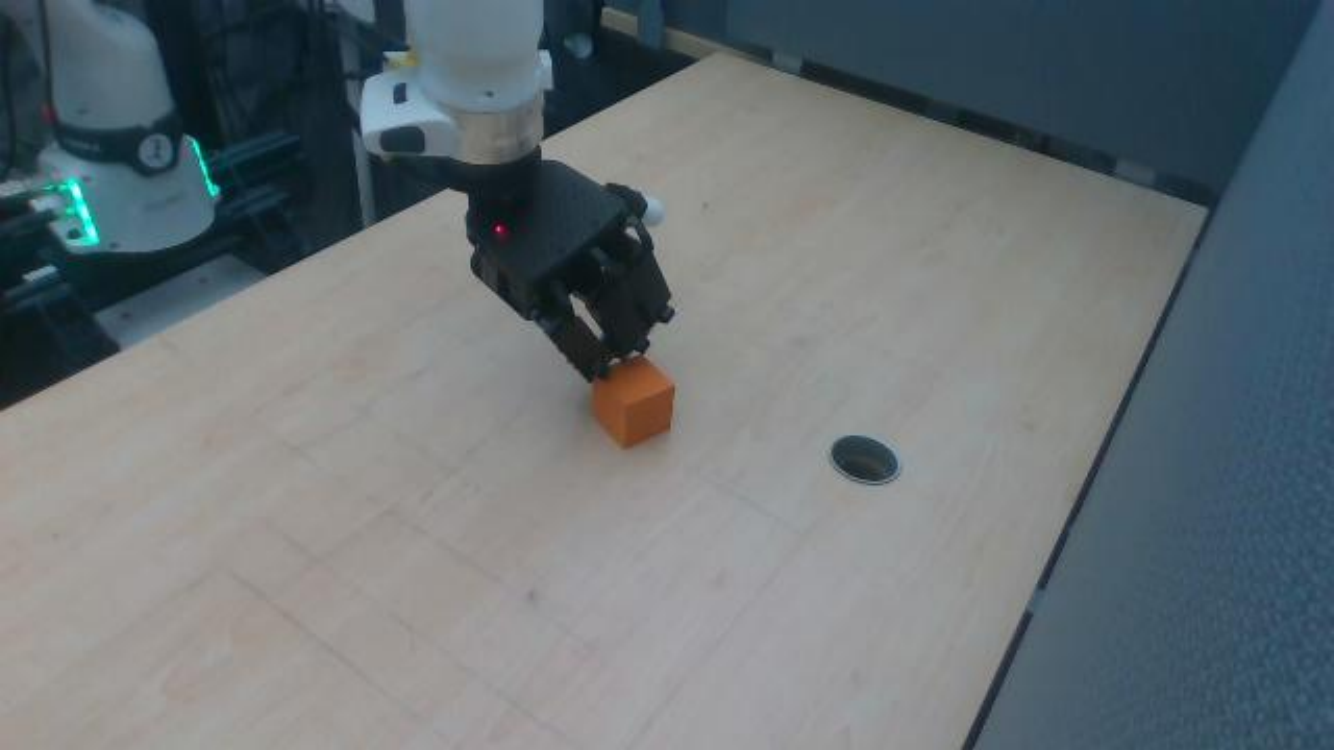} & 
                \includegraphics[width=0.9\linewidth]{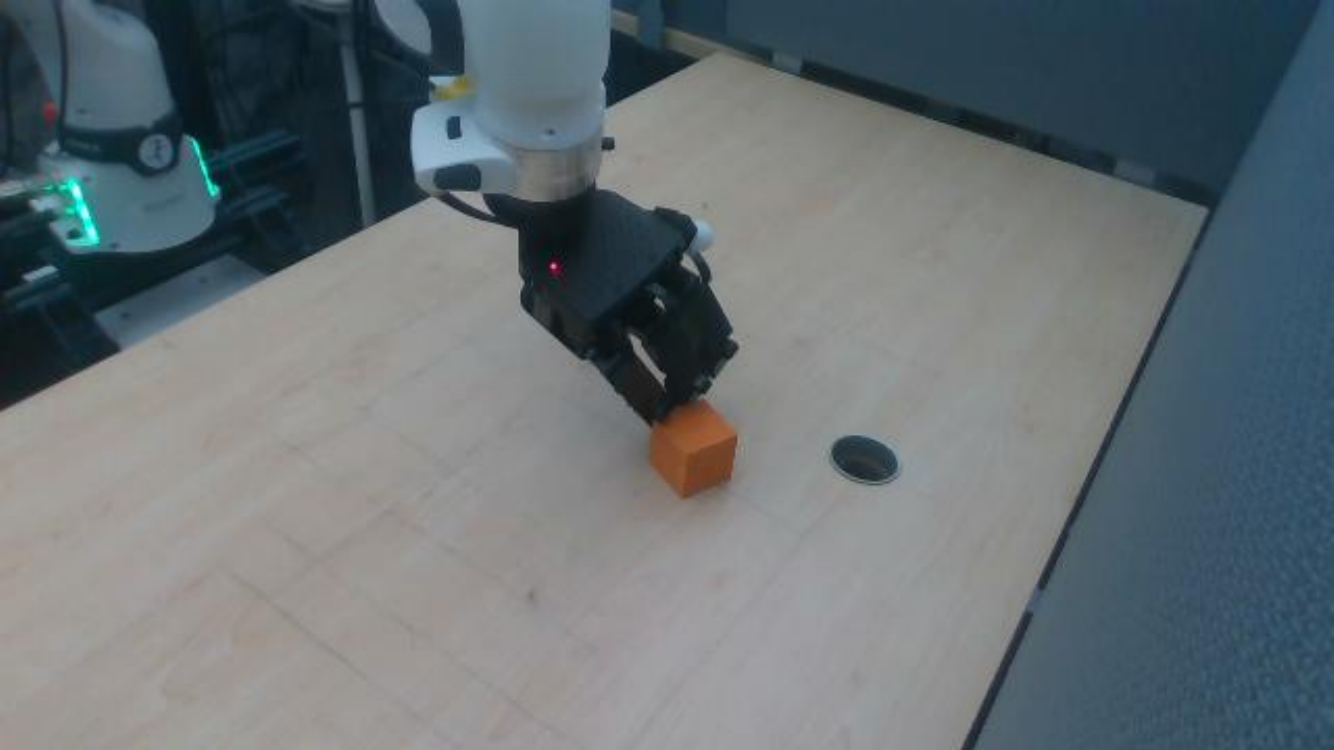} & 
                \includegraphics[width=0.9\linewidth]{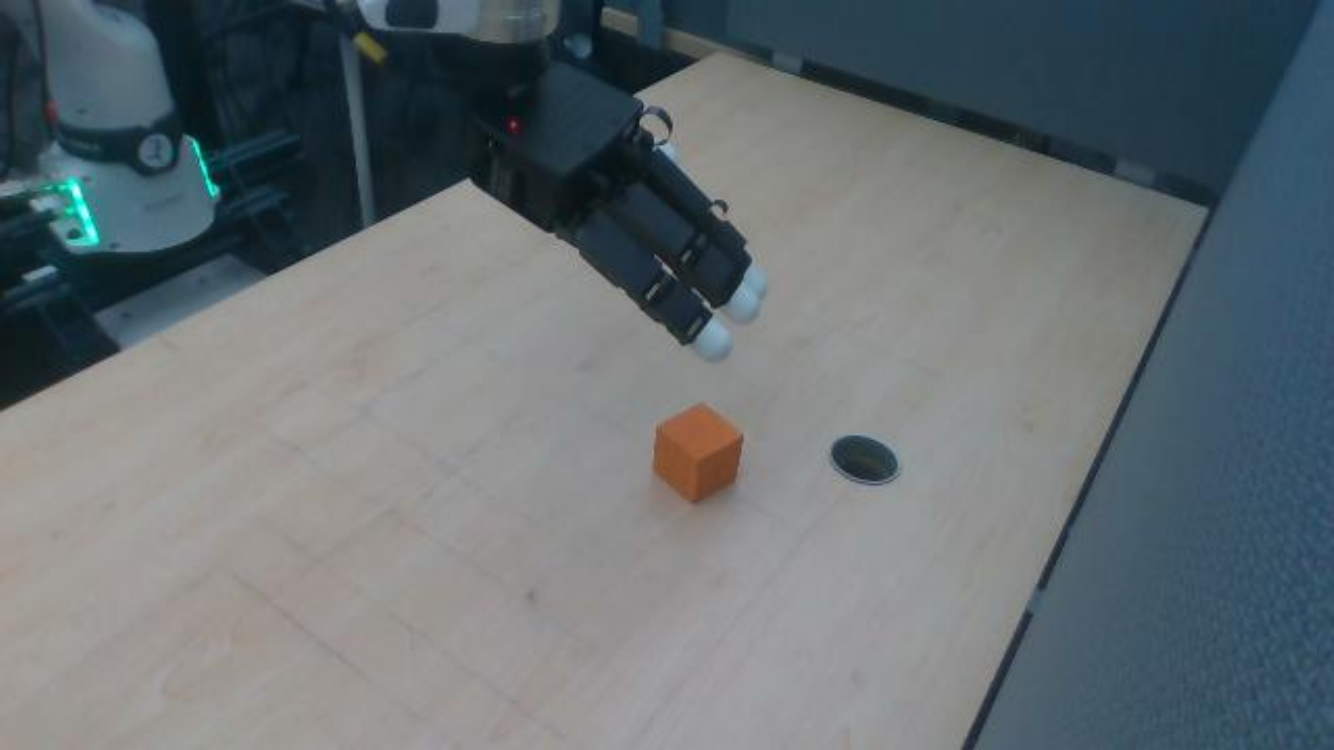} \\
            & Ability & 
                \includegraphics[width=0.9\linewidth]{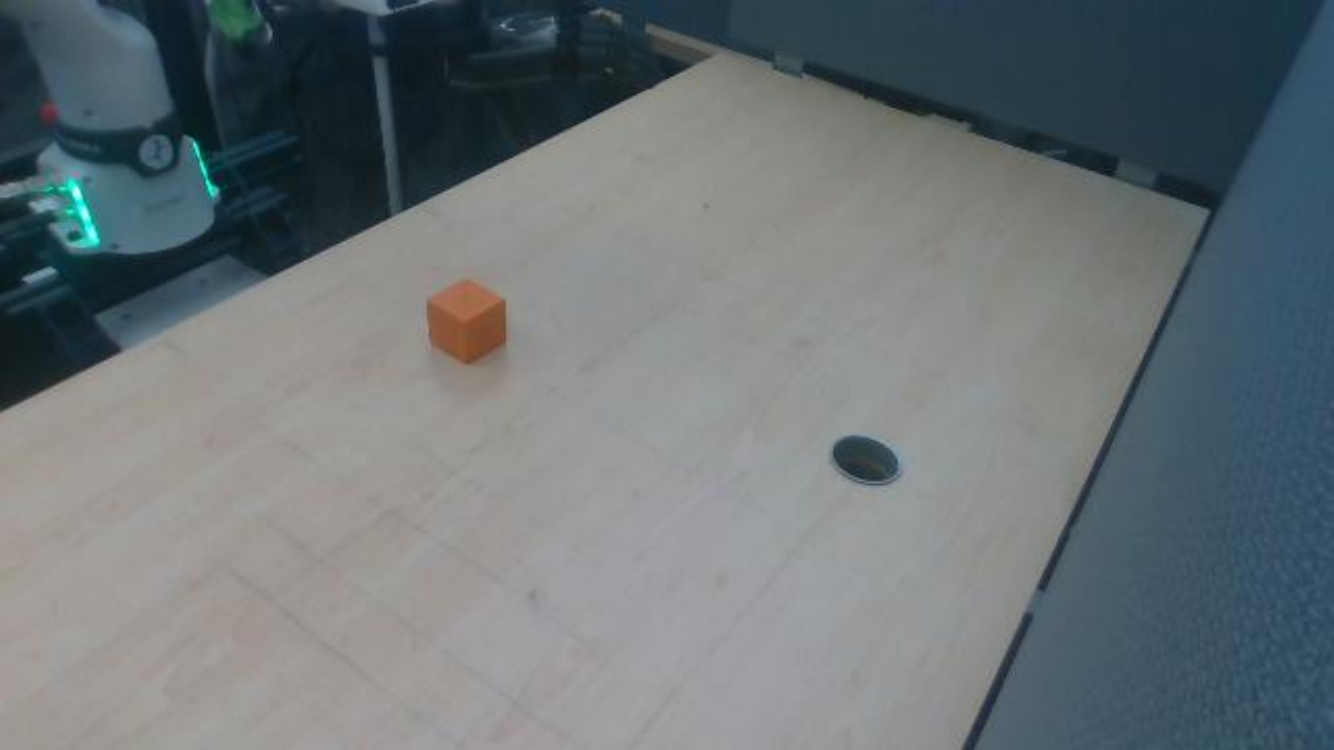} & 
                \includegraphics[width=0.9\linewidth]{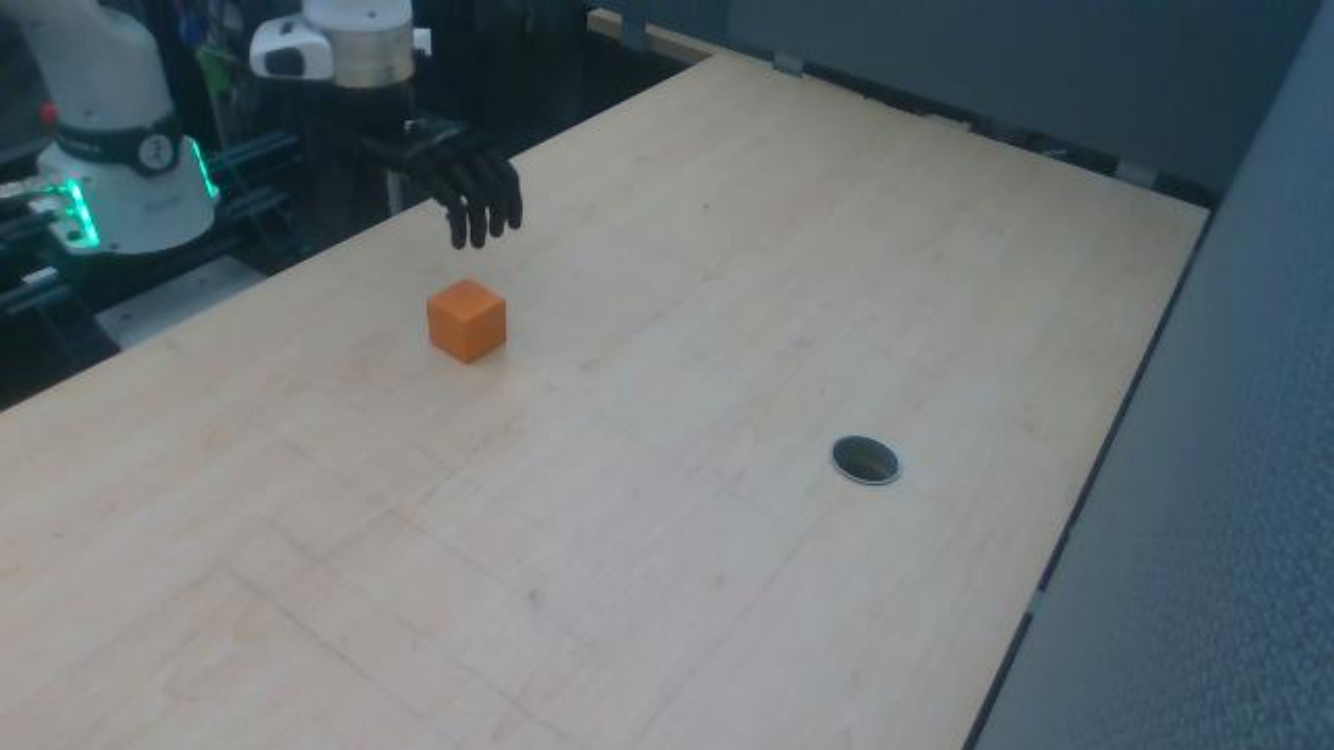} & 
                \includegraphics[width=0.9\linewidth]{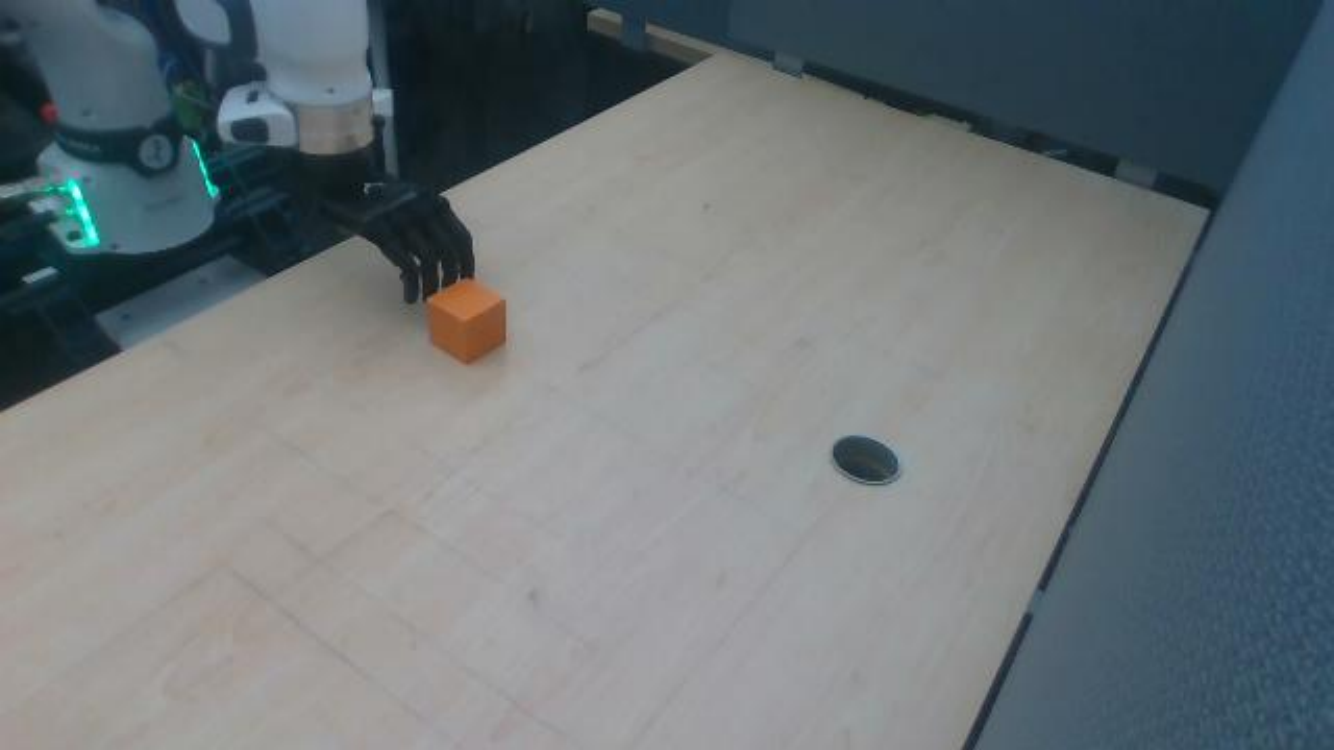} & 
                \includegraphics[width=0.9\linewidth]{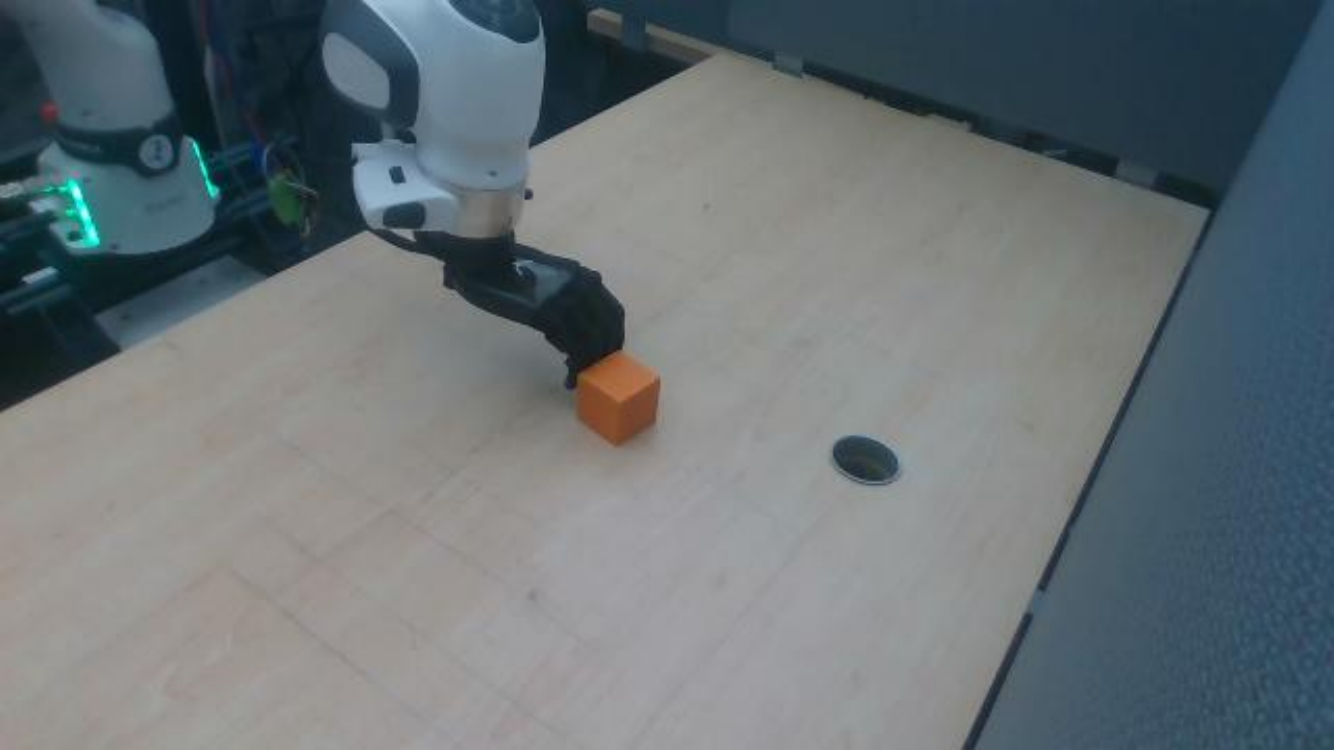} & 
                \includegraphics[width=0.9\linewidth]{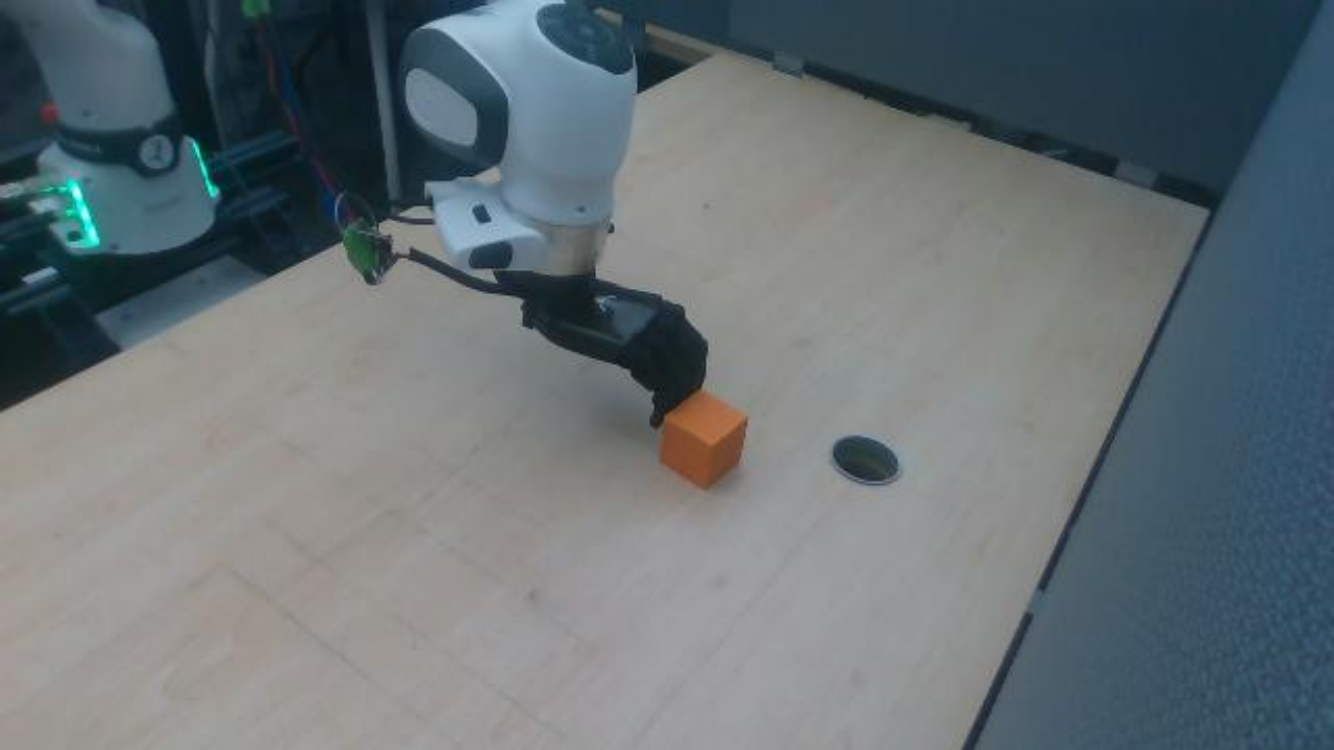} & 
                \includegraphics[width=0.9\linewidth]{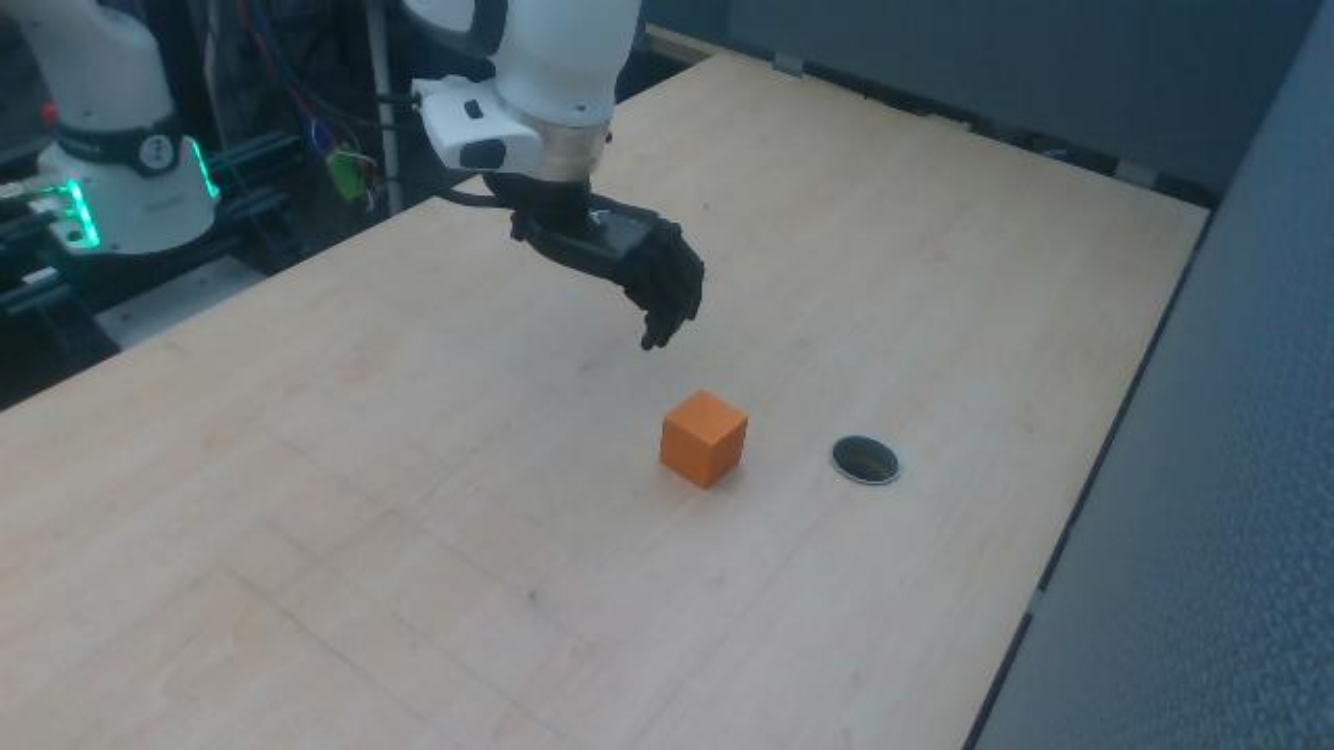} \\
        \midrule
        \multirow{5}{*}{\vspace*{-10ex}Hammer}
            & Human & 
                \includegraphics[width=0.9\linewidth]{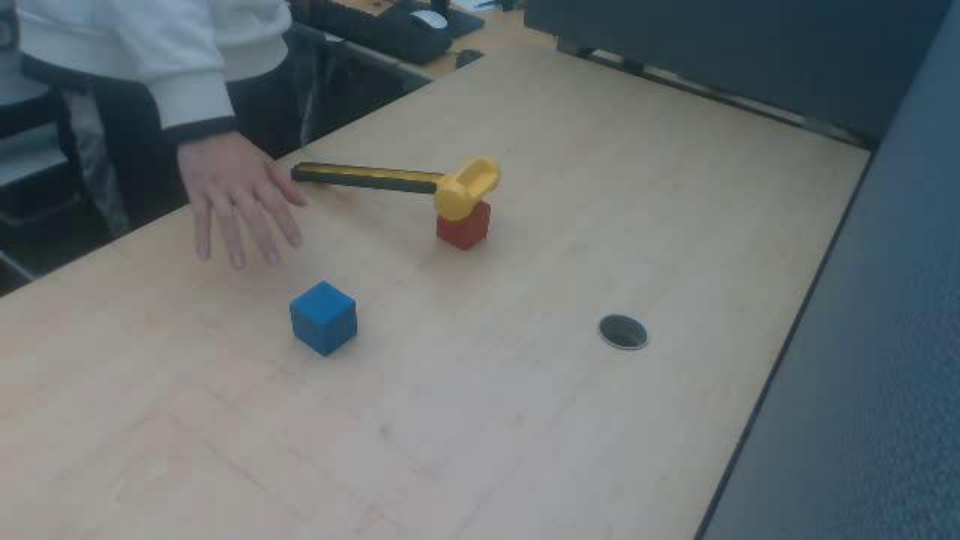} & 
                \includegraphics[width=0.9\linewidth]{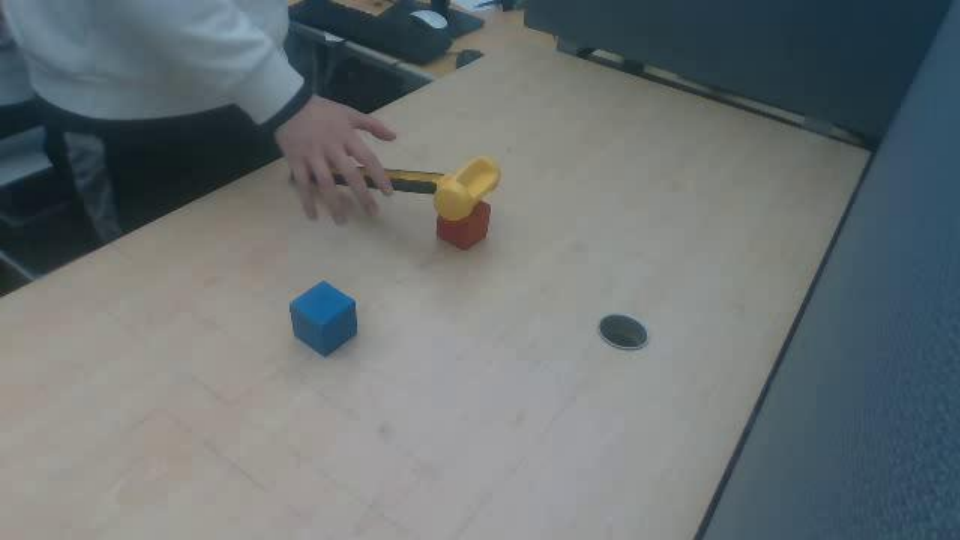} & 
                \includegraphics[width=0.9\linewidth]{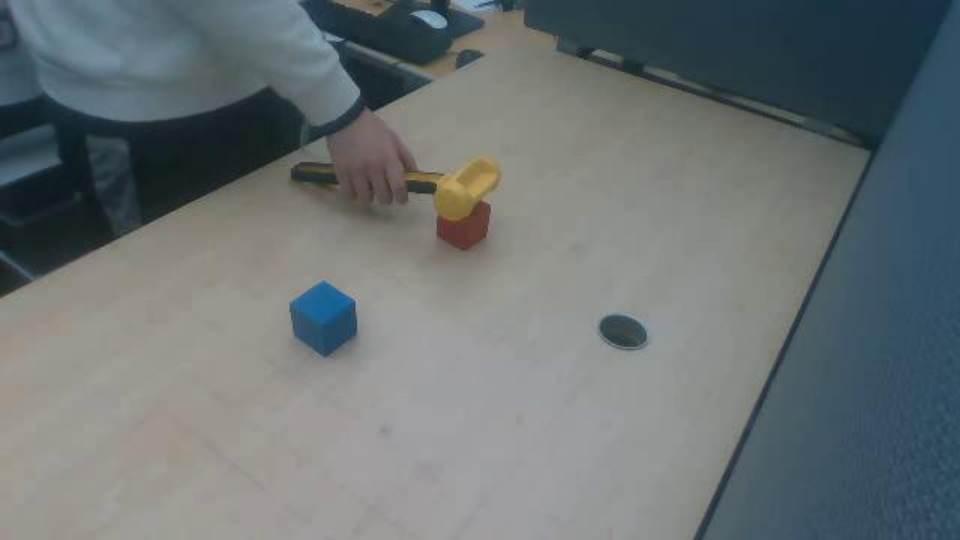} & 
                \includegraphics[width=0.9\linewidth]{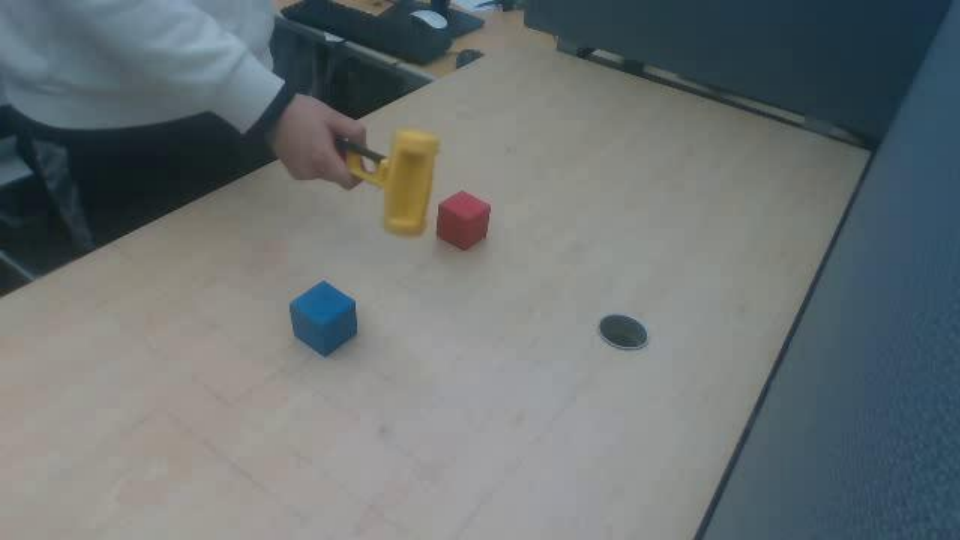} & 
                \includegraphics[width=0.9\linewidth]{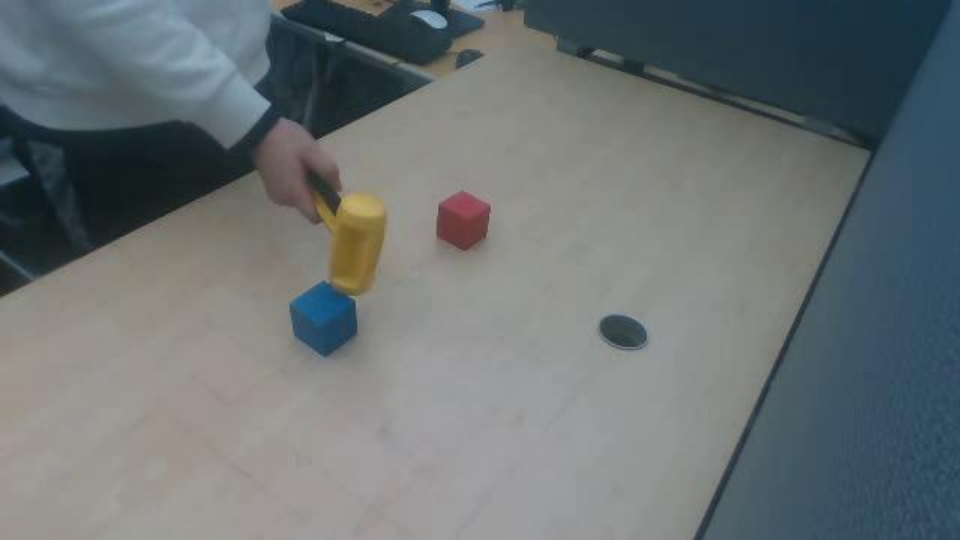} & 
                \includegraphics[width=0.9\linewidth]{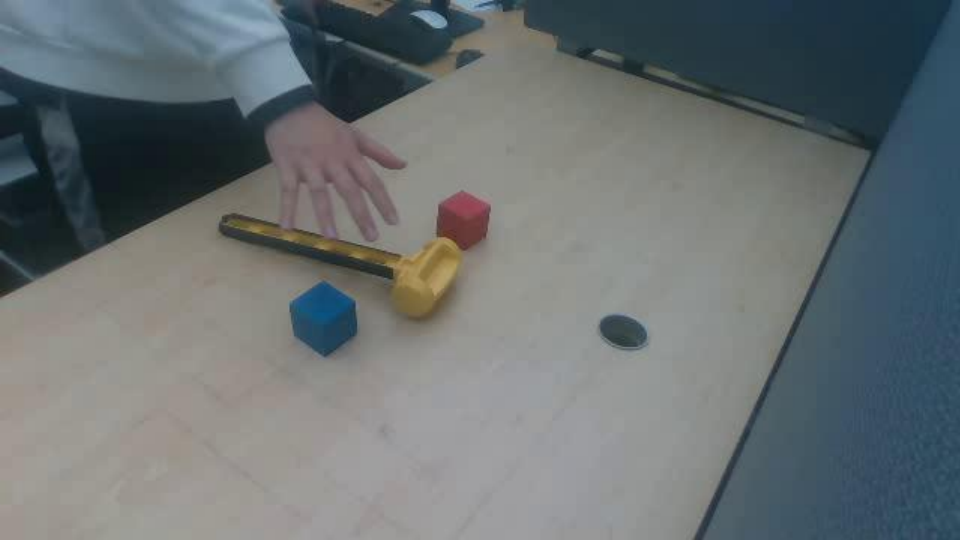} \\
            & FR & 
                \includegraphics[width=0.9\linewidth]{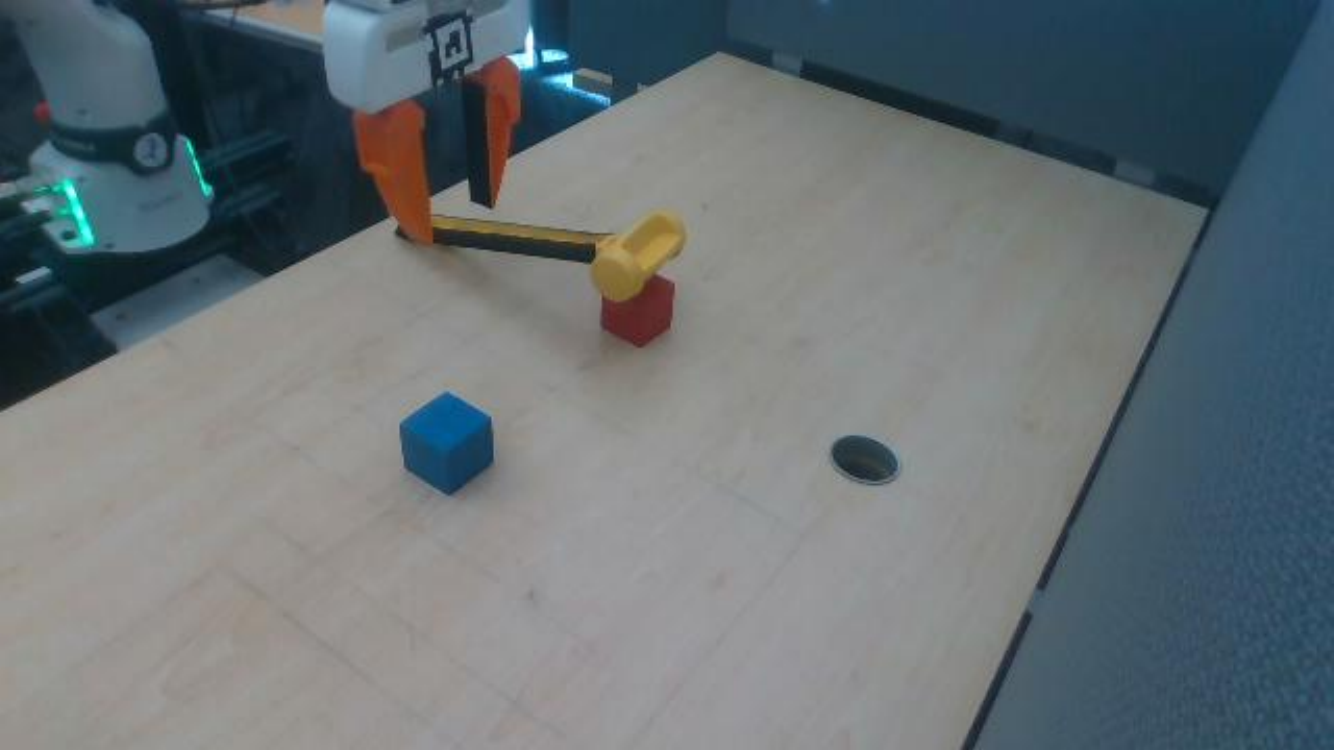} & 
                \includegraphics[width=0.9\linewidth]{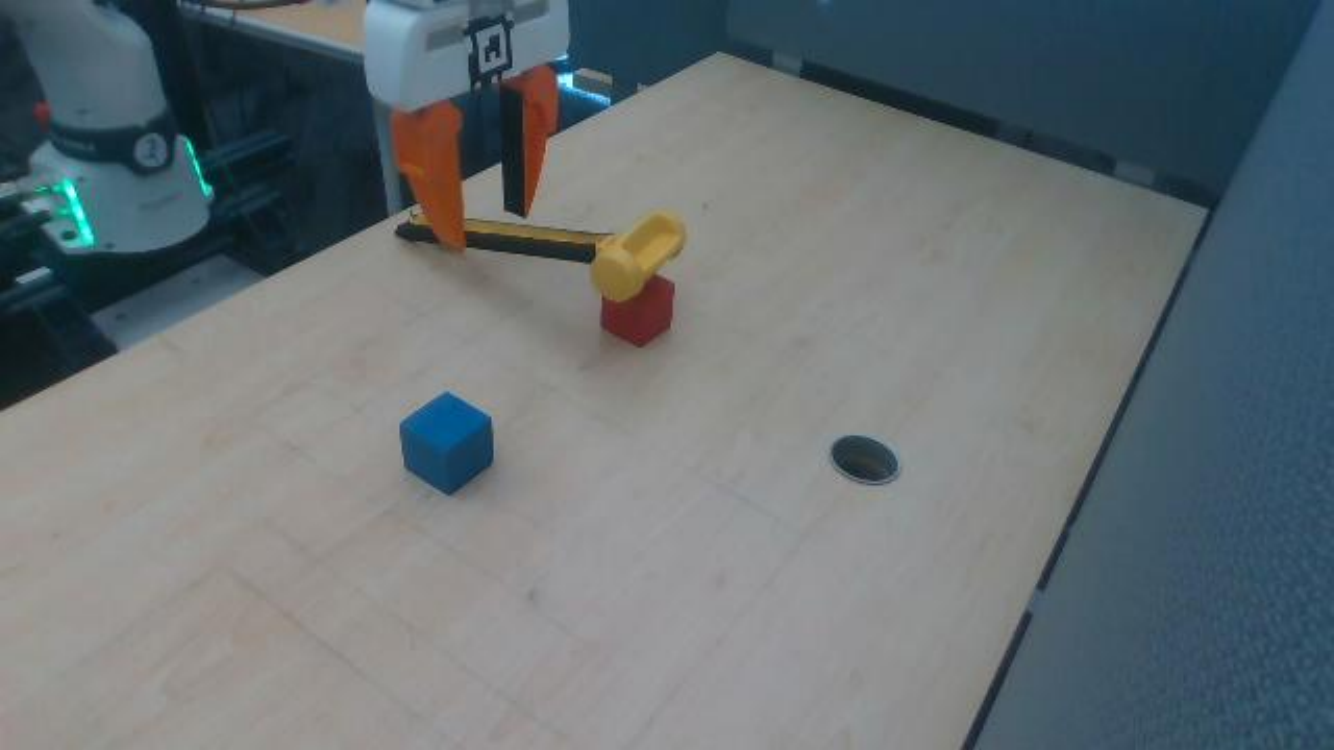} & 
                \includegraphics[width=0.9\linewidth]{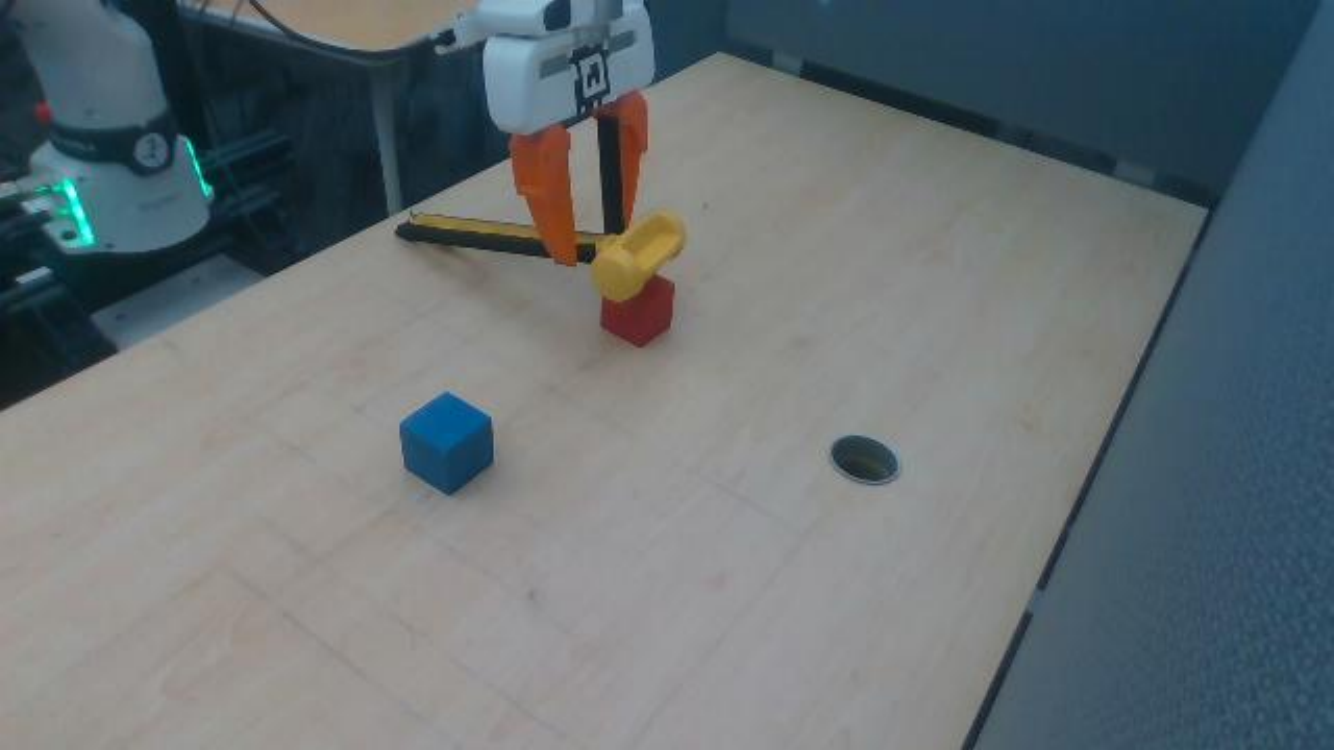} & 
                \includegraphics[width=0.9\linewidth]{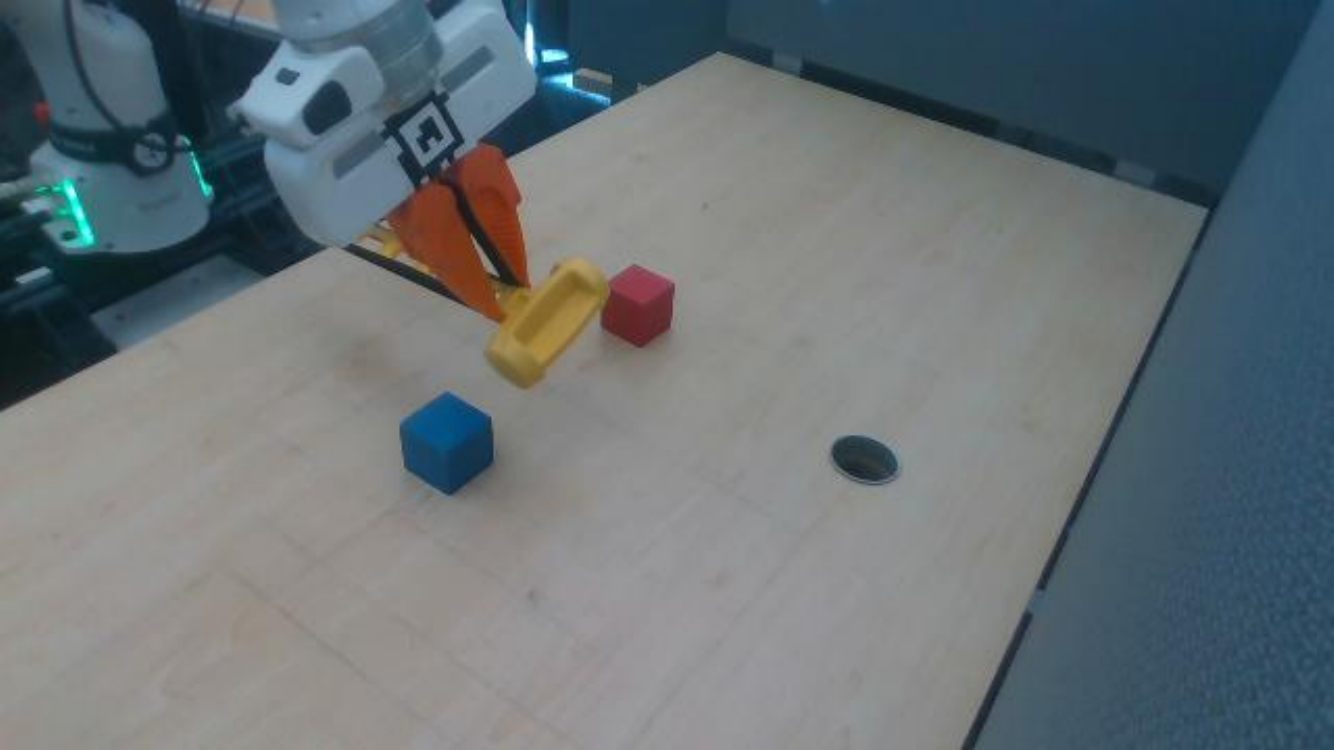} & 
                \includegraphics[width=0.9\linewidth]{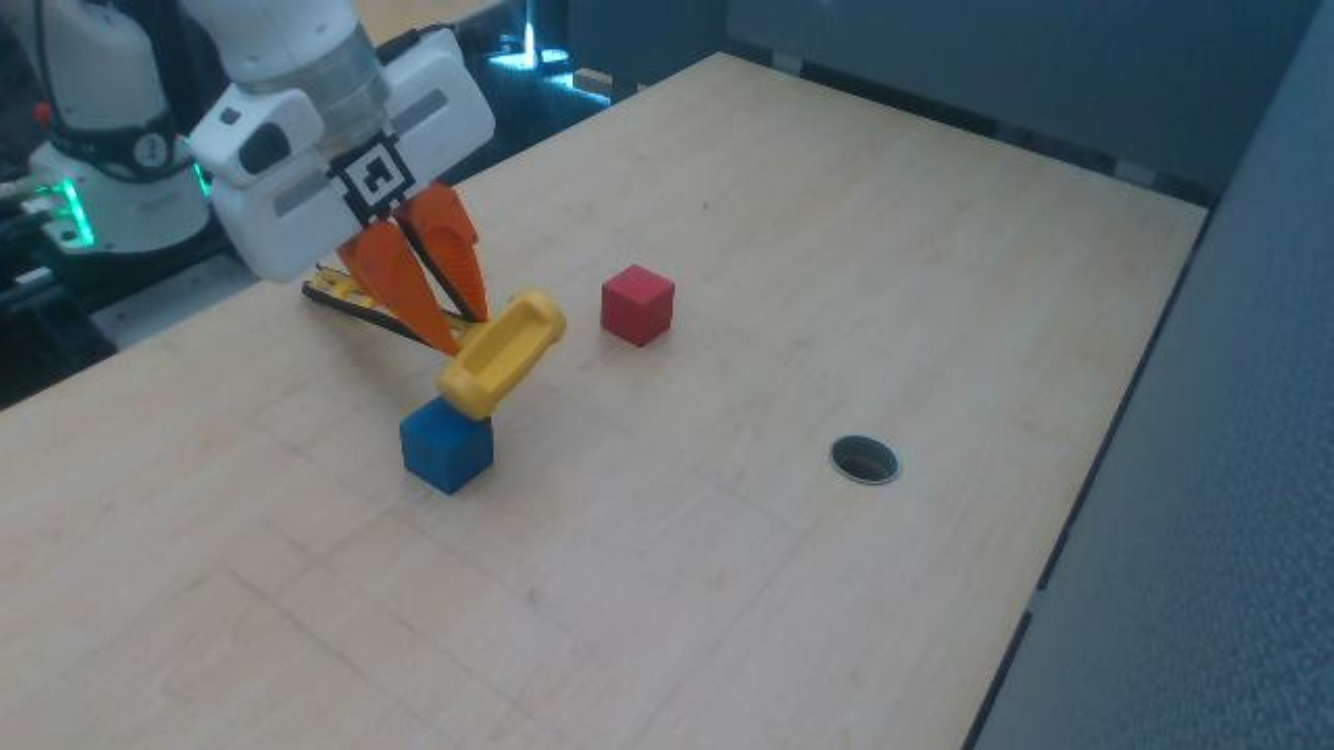} & 
                \includegraphics[width=0.9\linewidth]{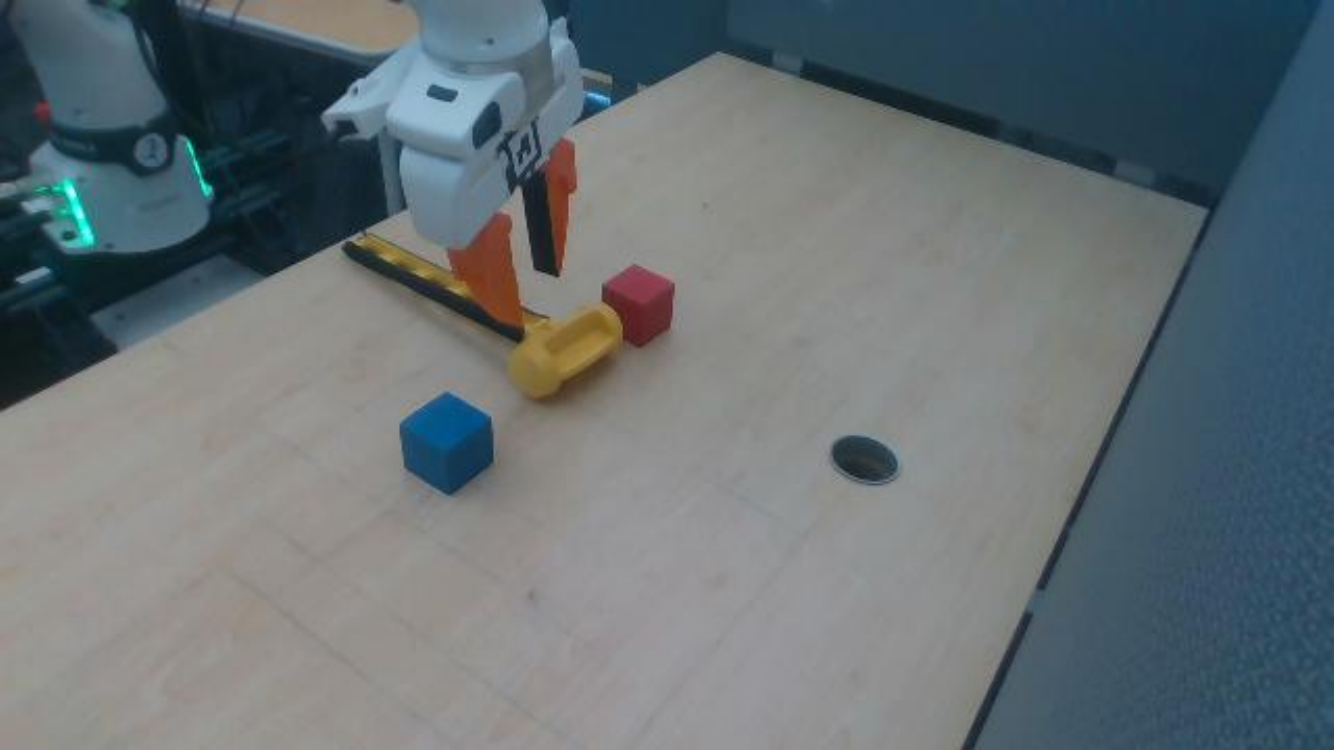} \\
            & Robotiq & 
                \includegraphics[width=0.9\linewidth]{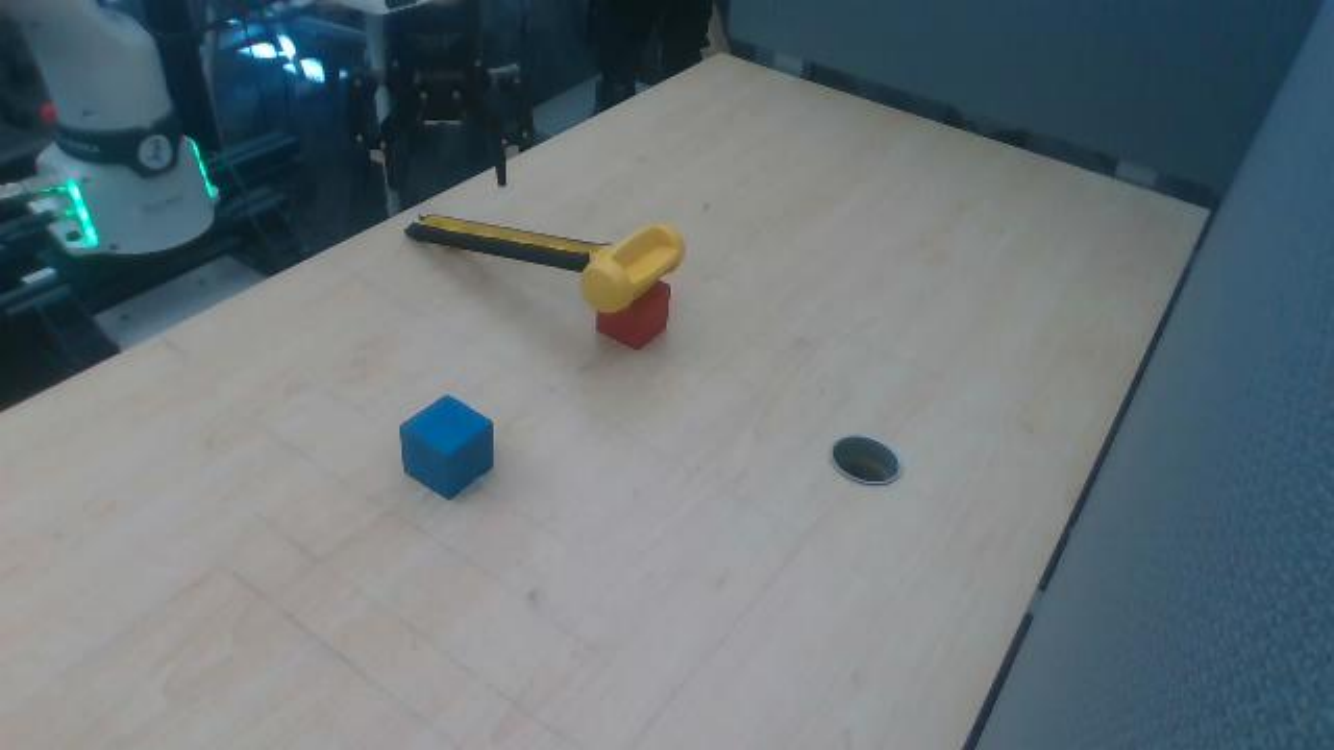} & 
                \includegraphics[width=0.9\linewidth]{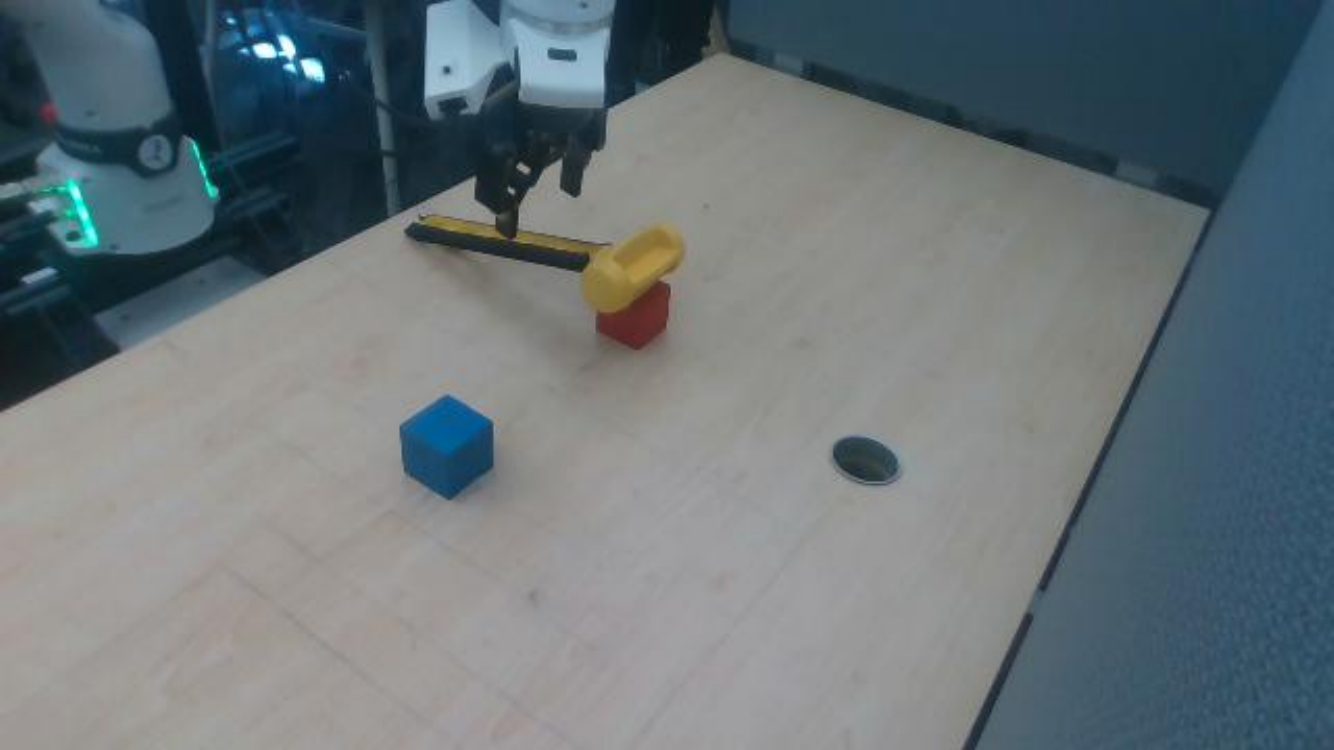} & 
                \includegraphics[width=0.9\linewidth]{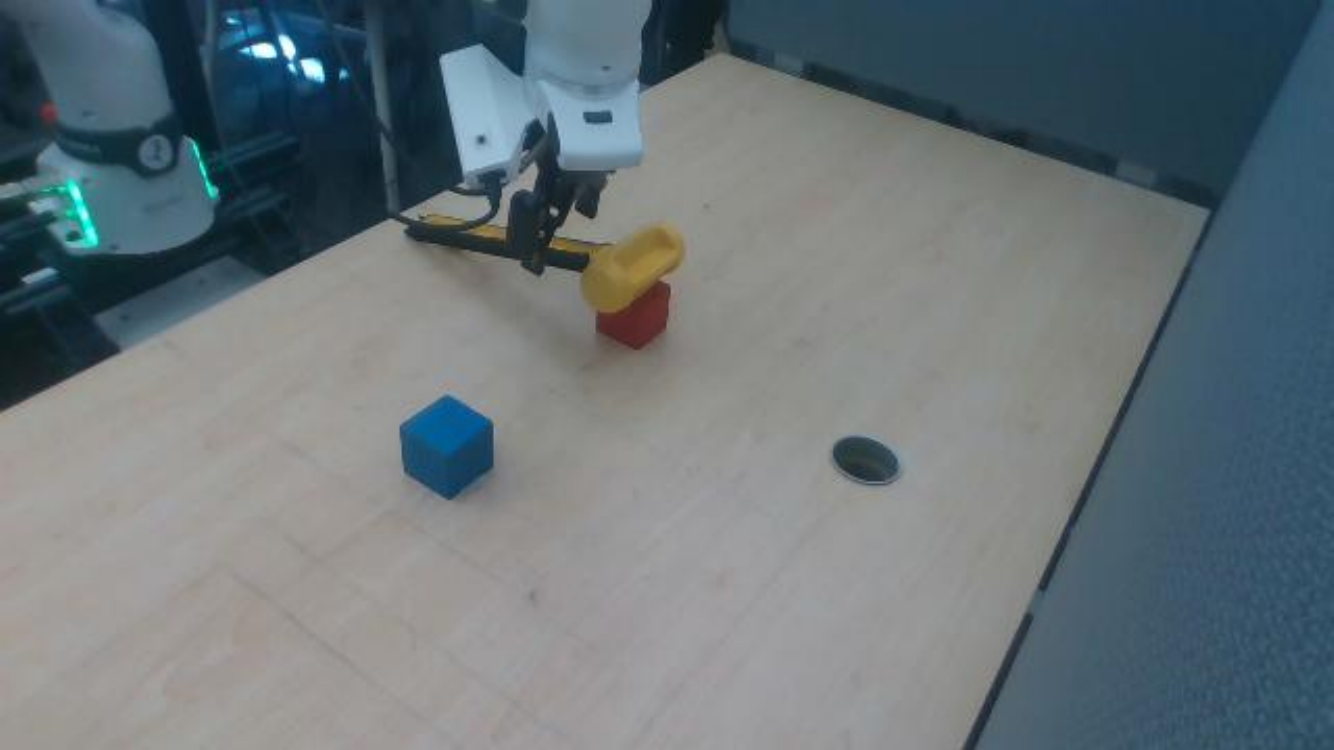} & 
                \includegraphics[width=0.9\linewidth]{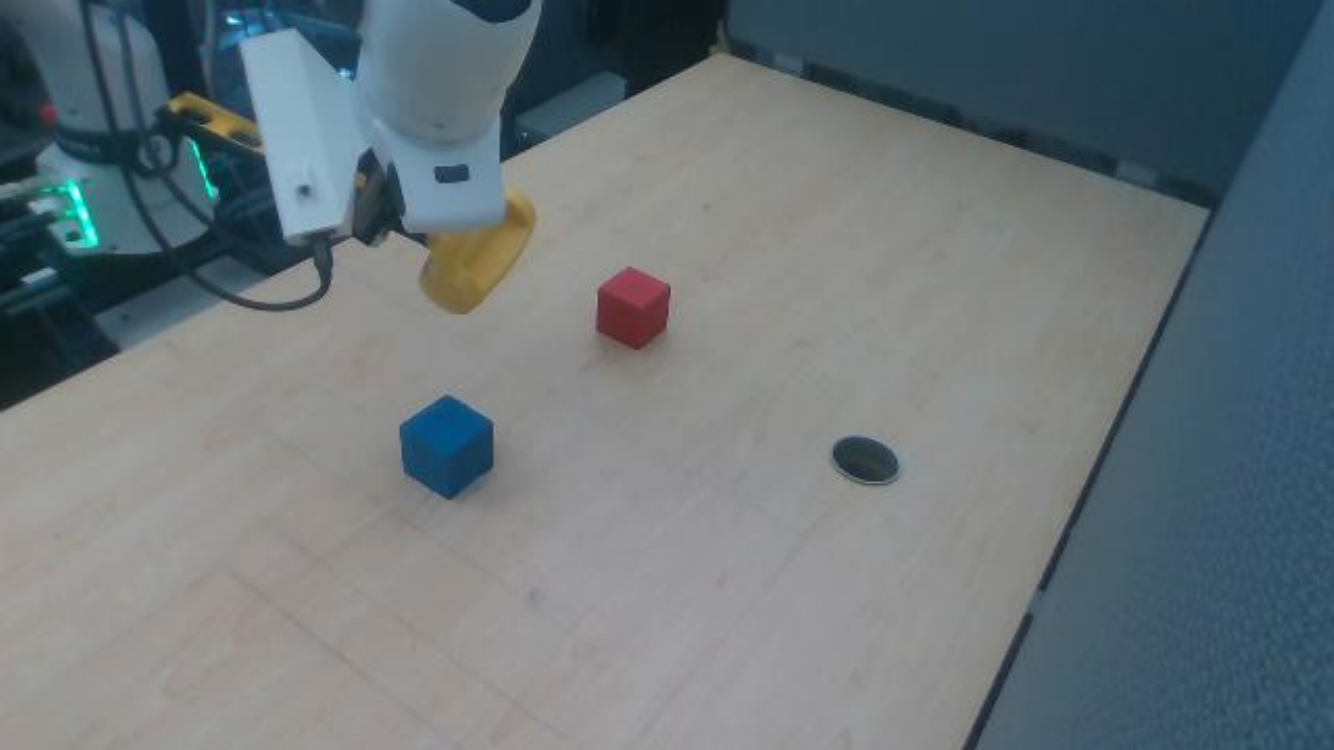} & 
                \includegraphics[width=0.9\linewidth]{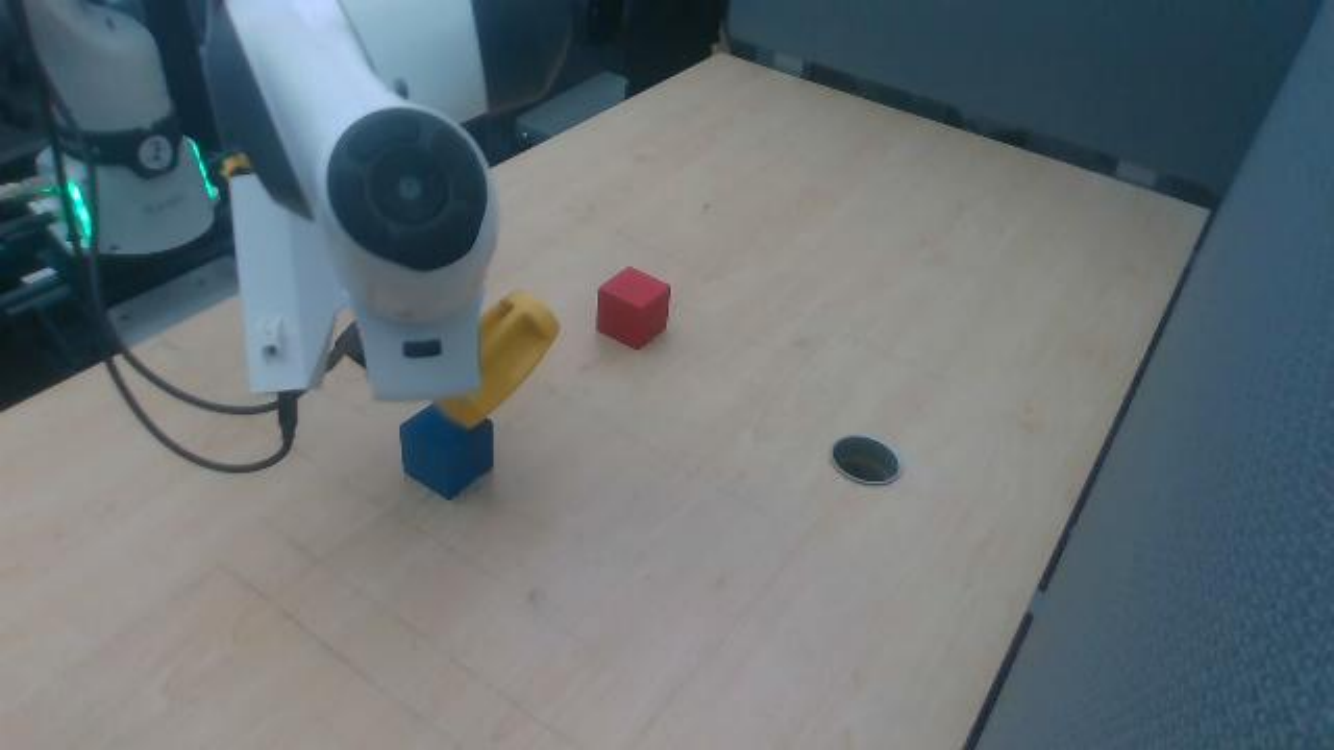} & 
                \includegraphics[width=0.9\linewidth]{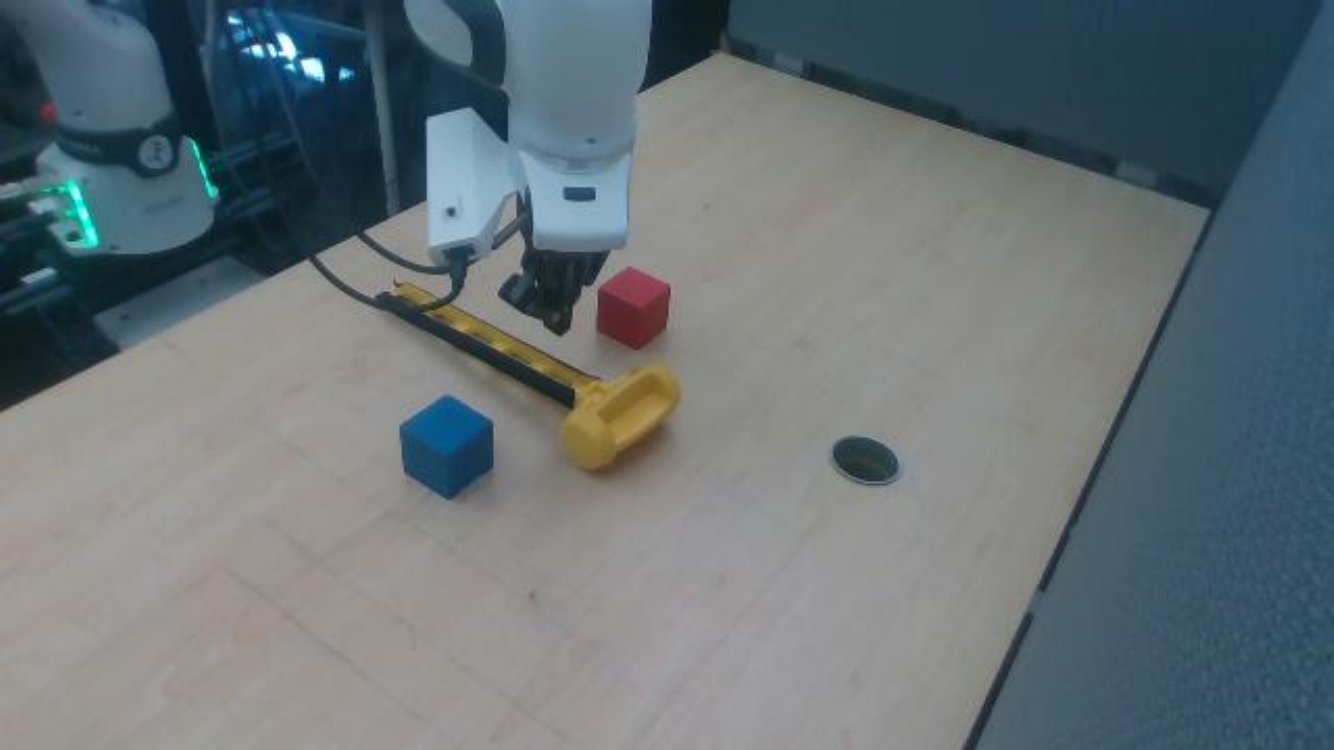} \\
            & Allegro & 
                \includegraphics[width=0.9\linewidth]{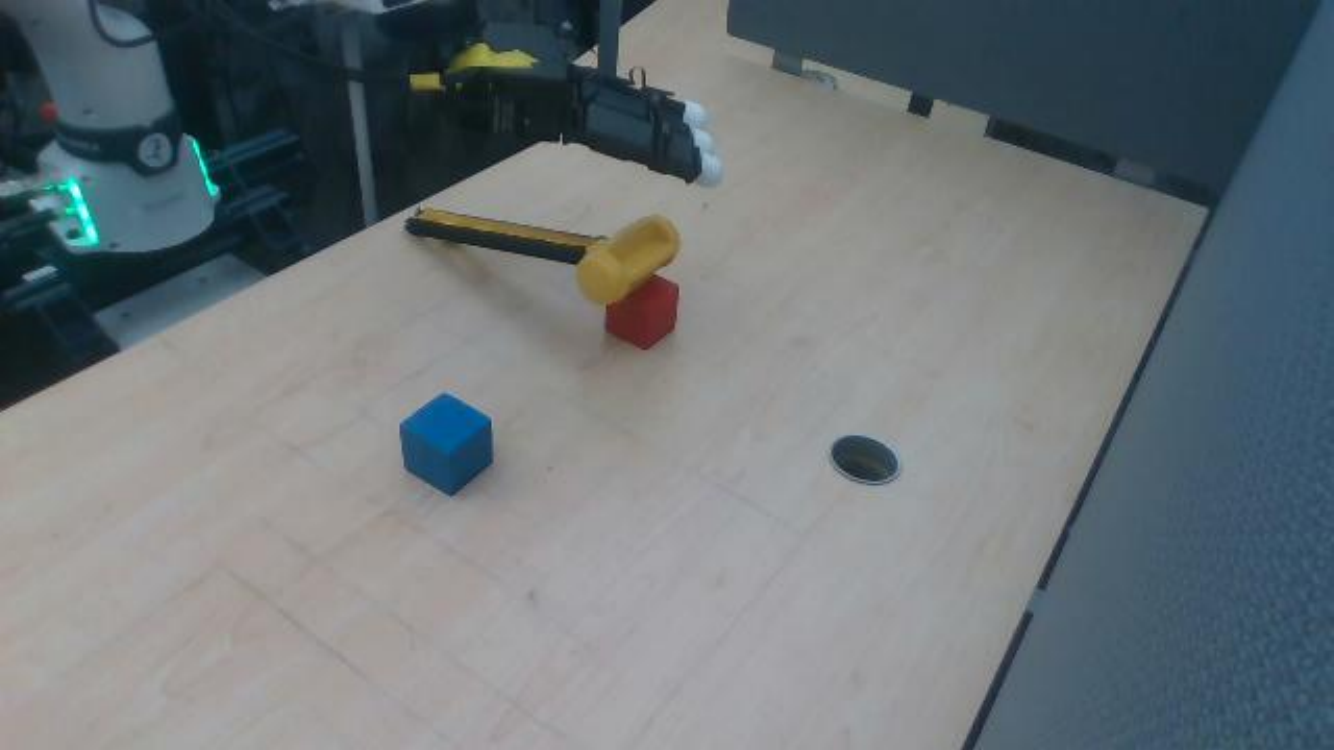} & 
                \includegraphics[width=0.9\linewidth]{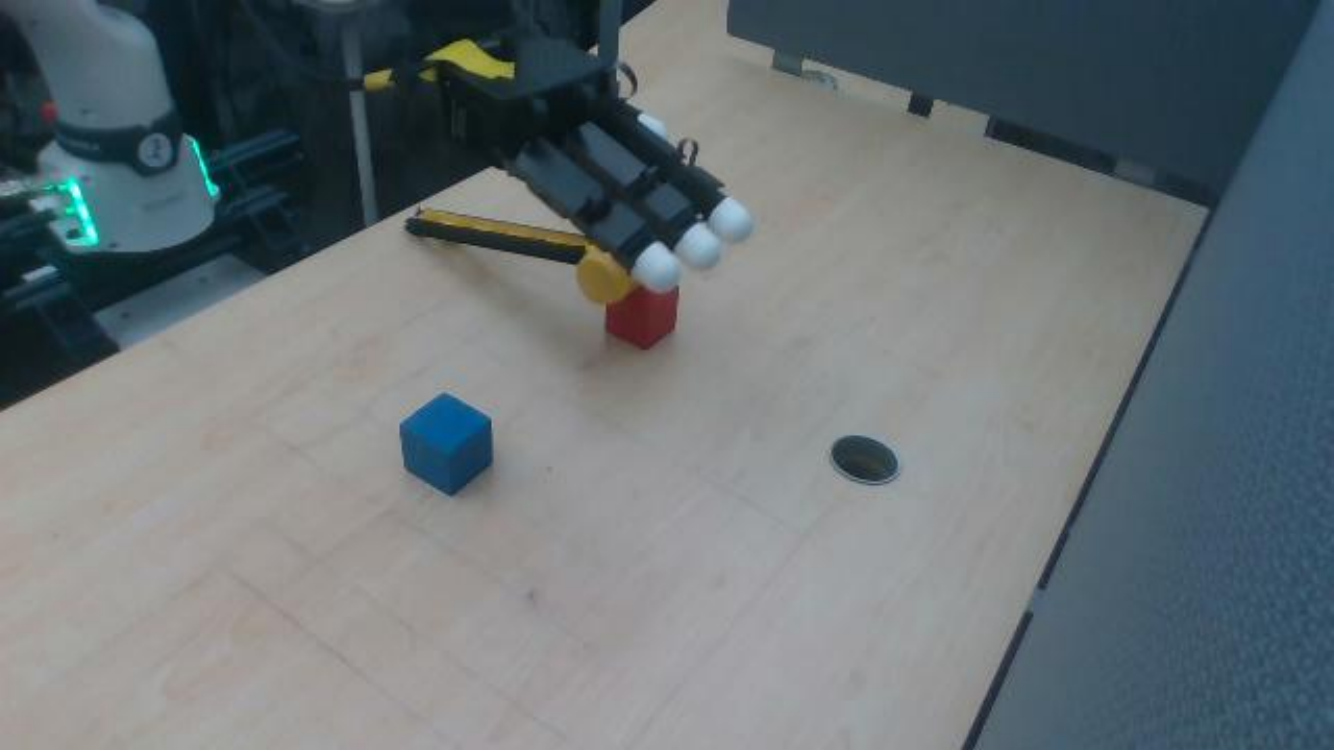} & 
                \includegraphics[width=0.9\linewidth]{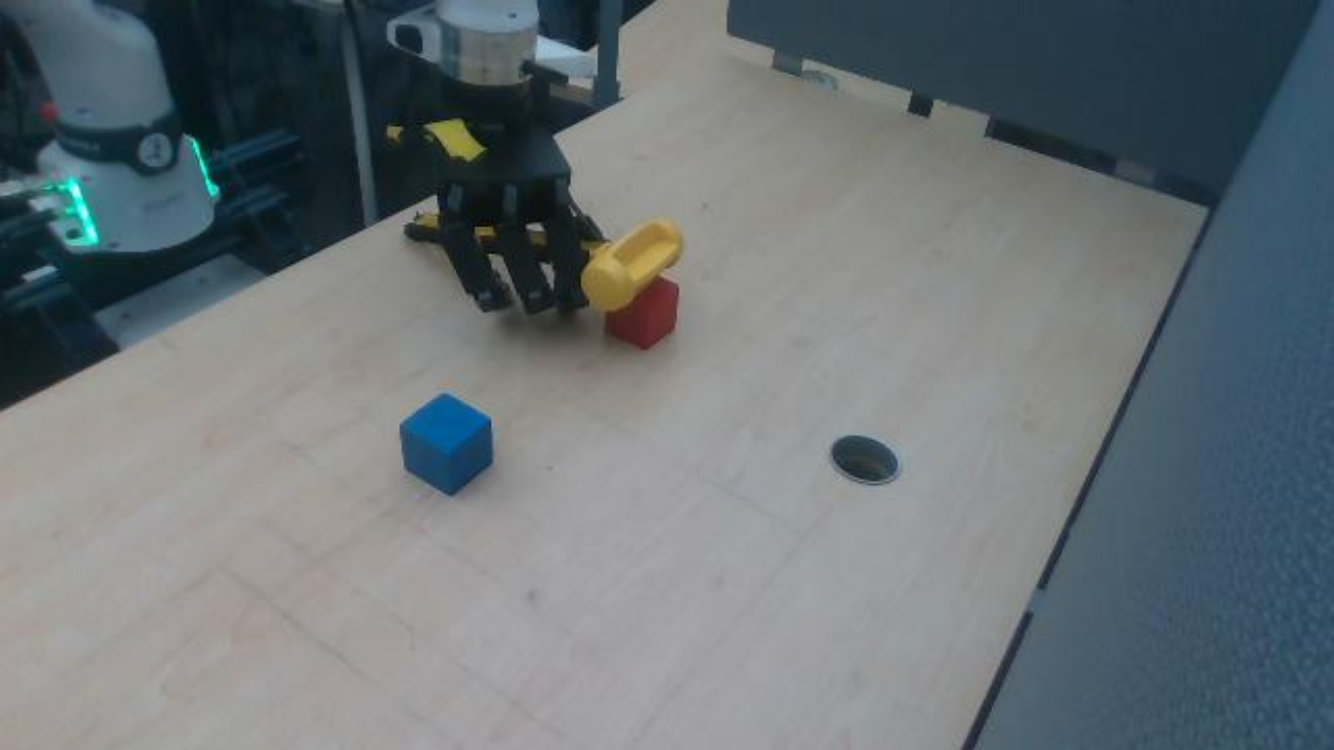} & 
                \includegraphics[width=0.9\linewidth]{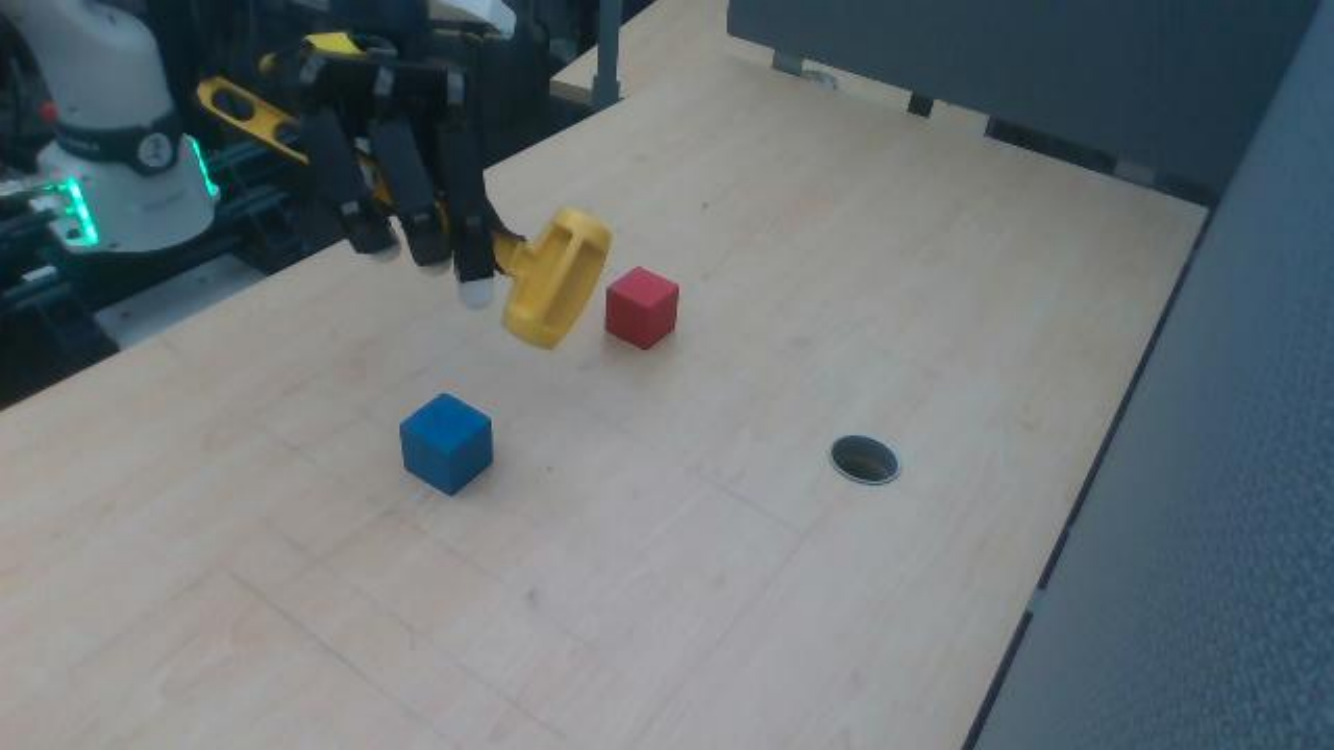} & 
                \includegraphics[width=0.9\linewidth]{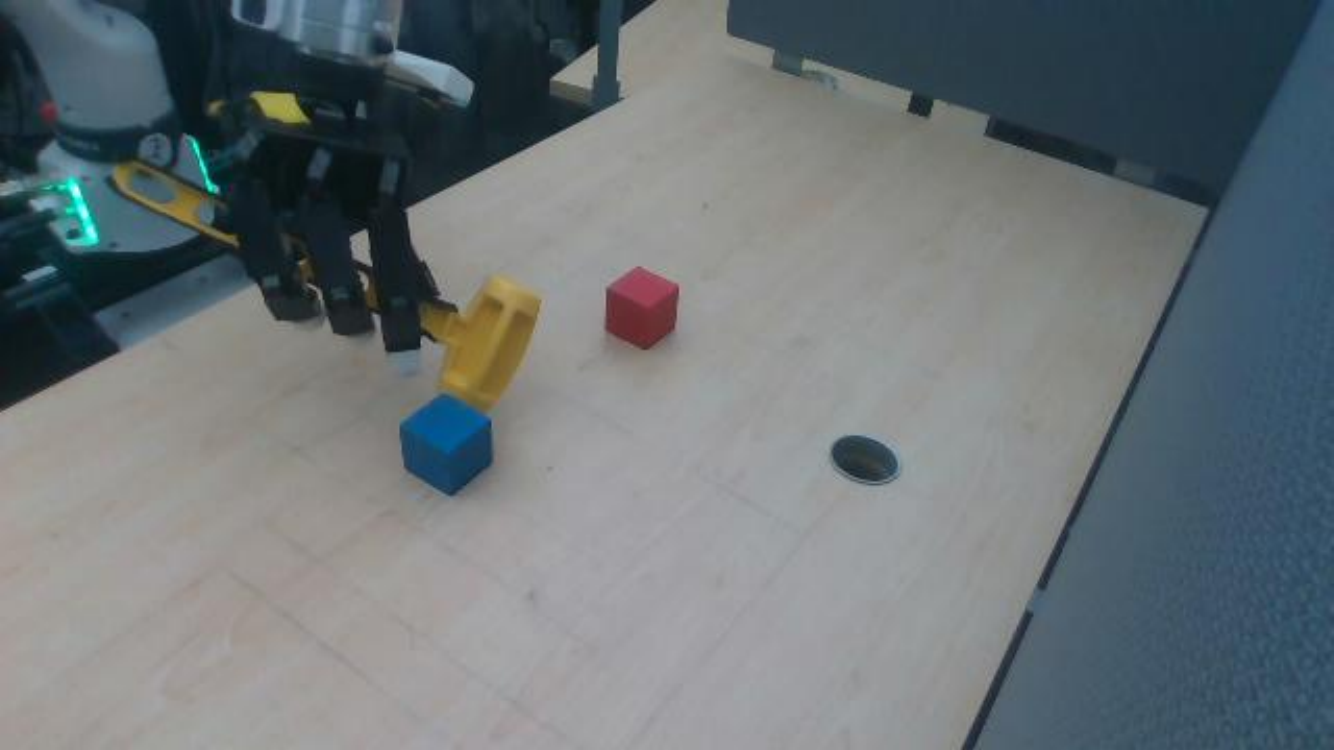} & 
                \includegraphics[width=0.9\linewidth]{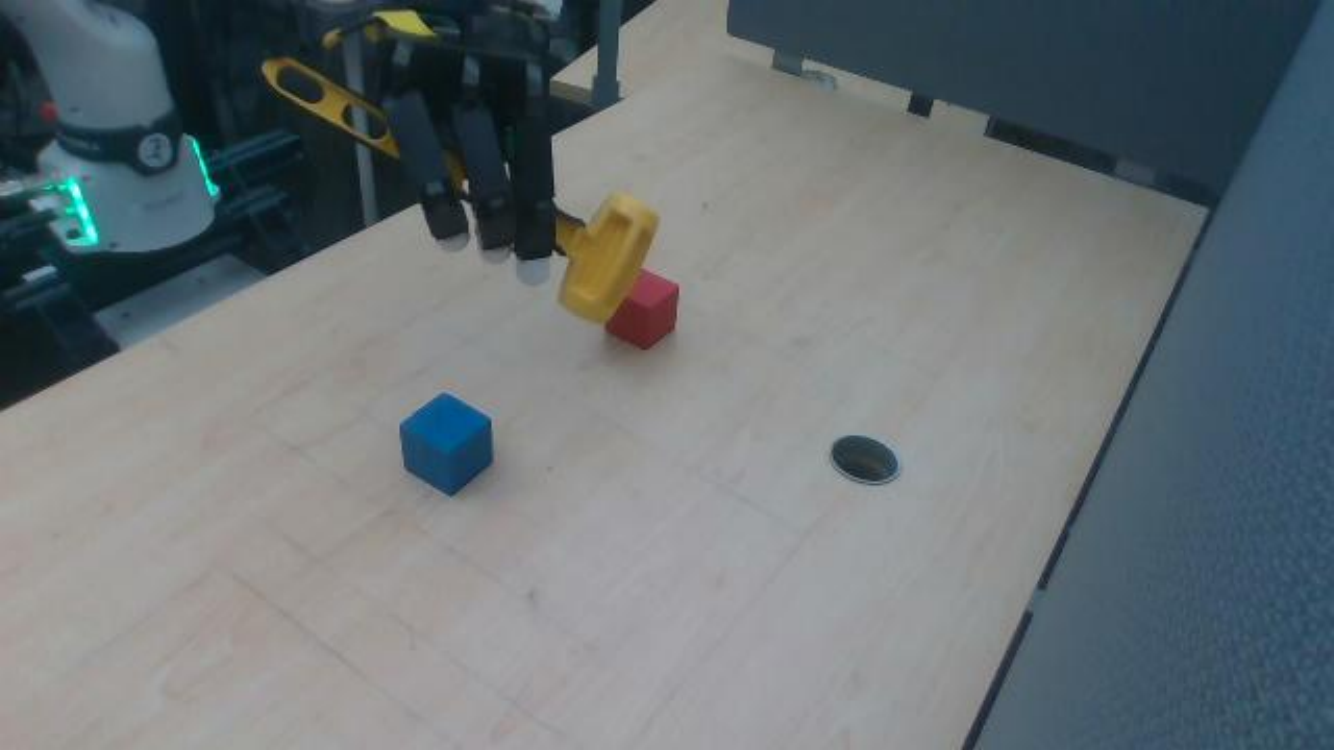} \\
            & Ability & 
                \includegraphics[width=0.9\linewidth]{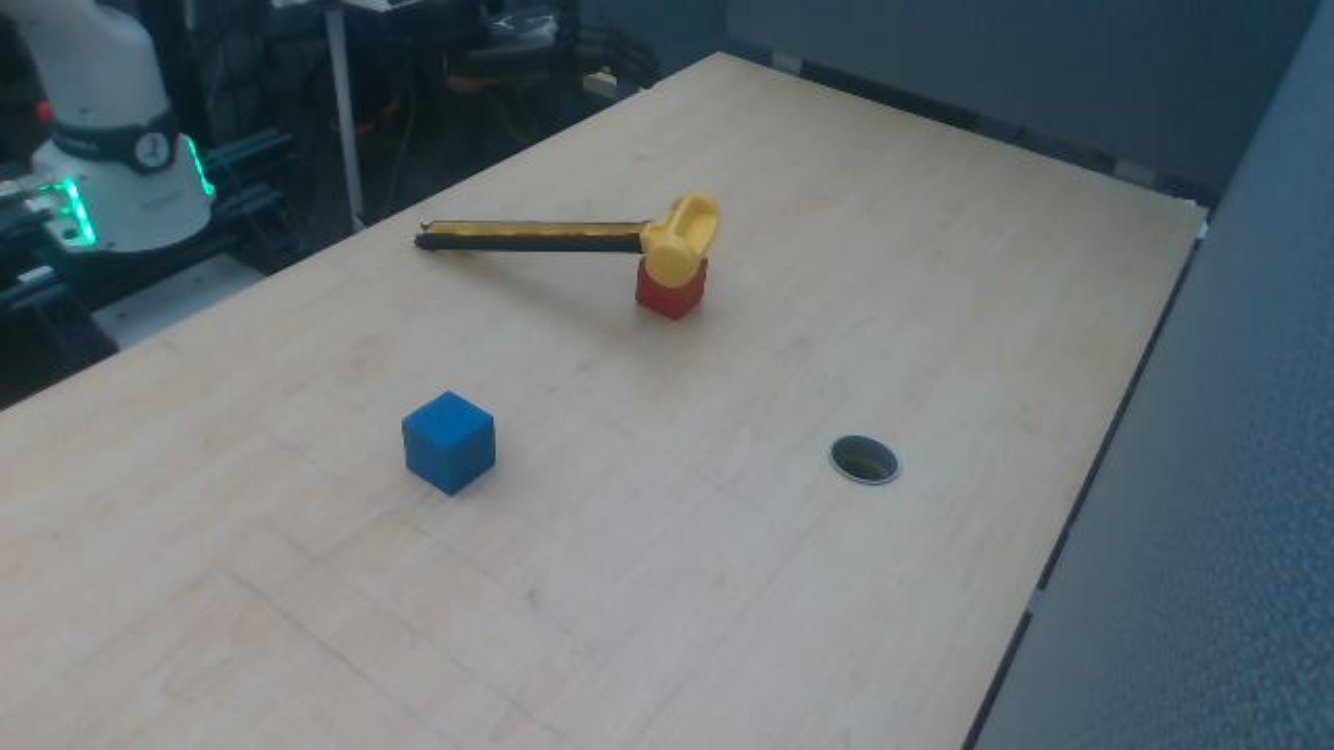} & 
                \includegraphics[width=0.9\linewidth]{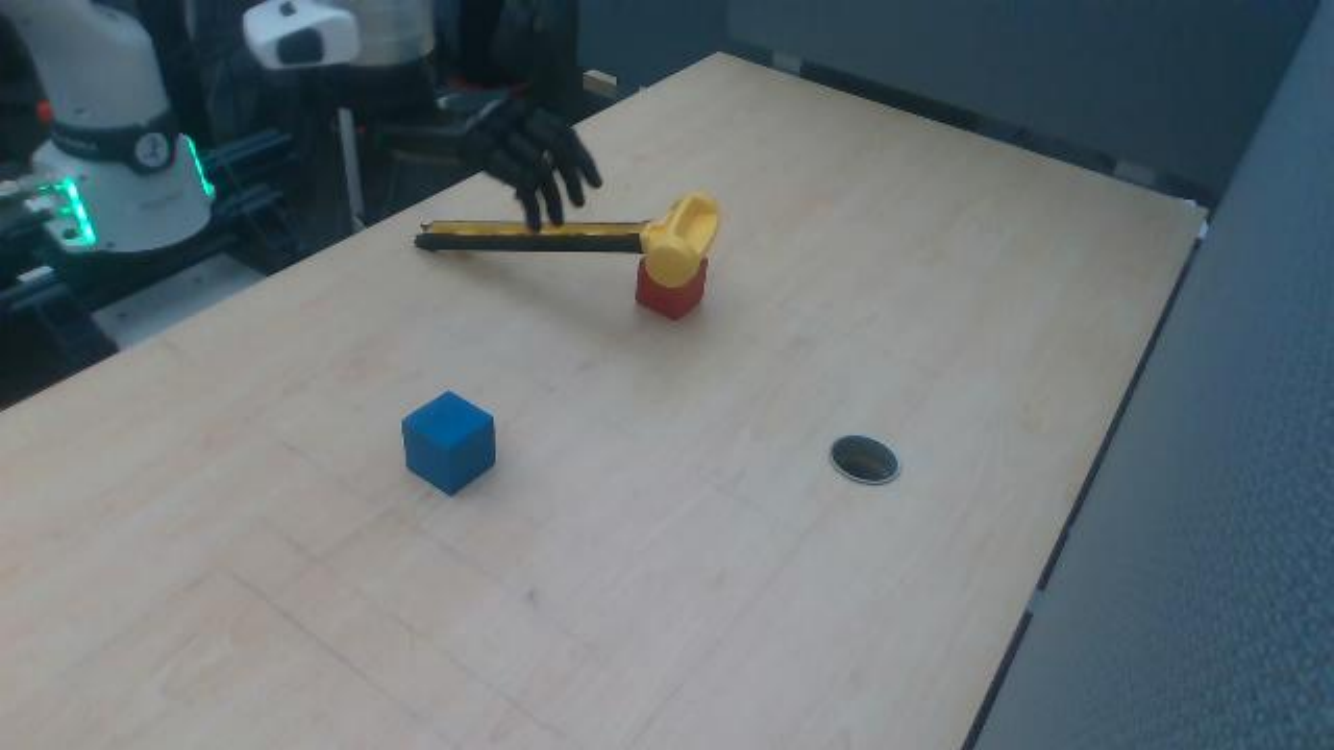} & 
                \includegraphics[width=0.9\linewidth]{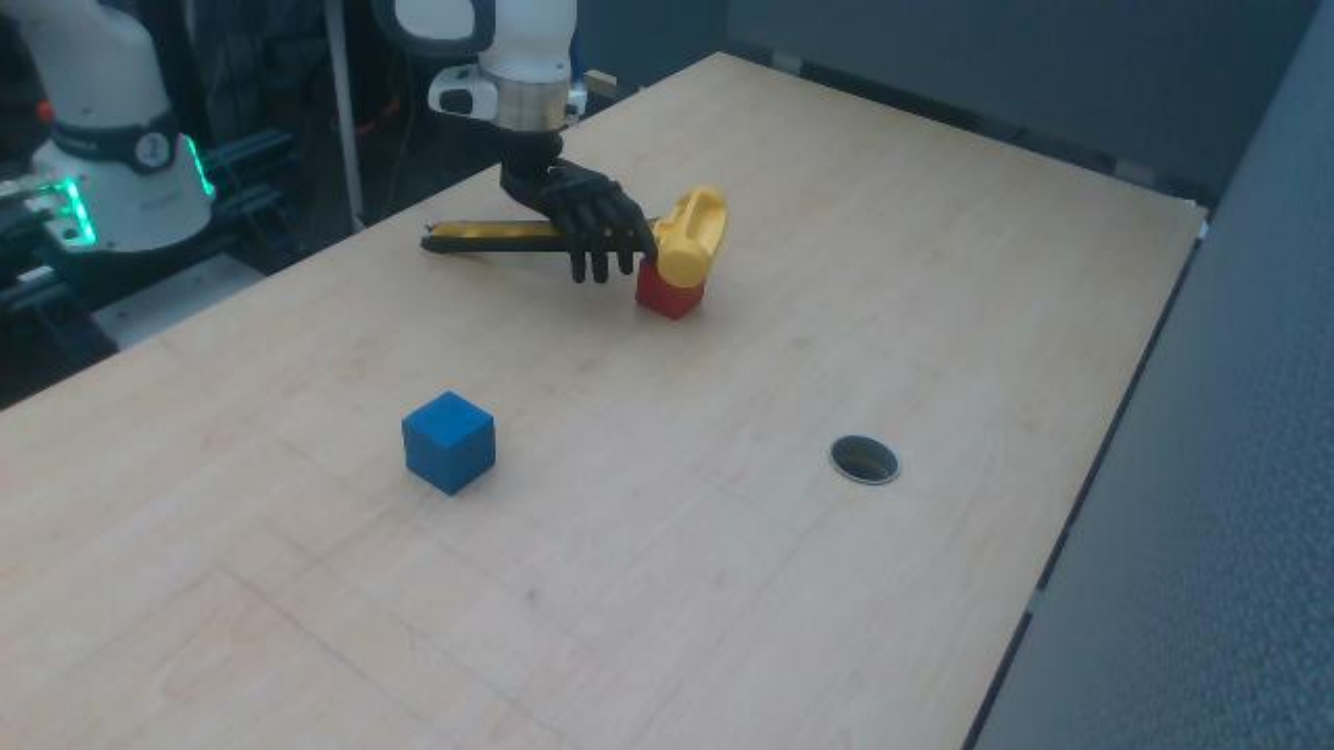} & 
                \includegraphics[width=0.9\linewidth]{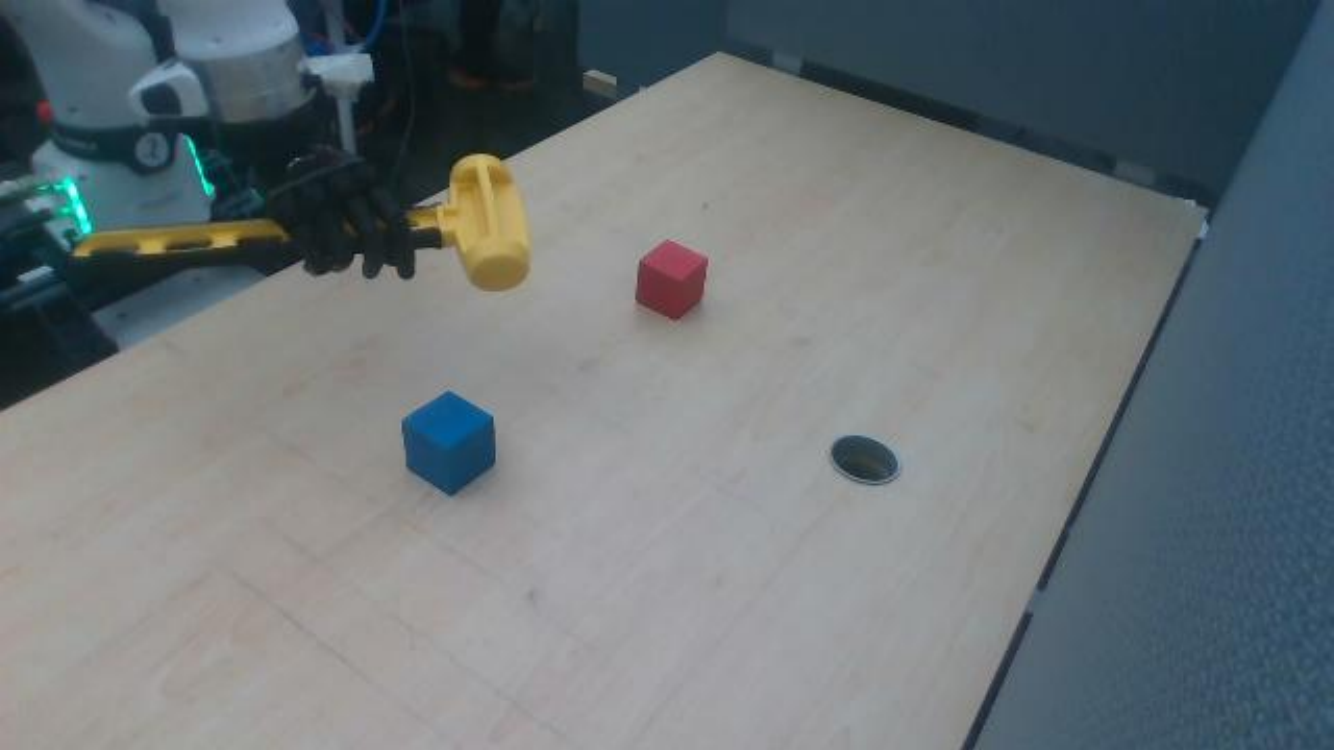} & 
                \includegraphics[width=0.9\linewidth]{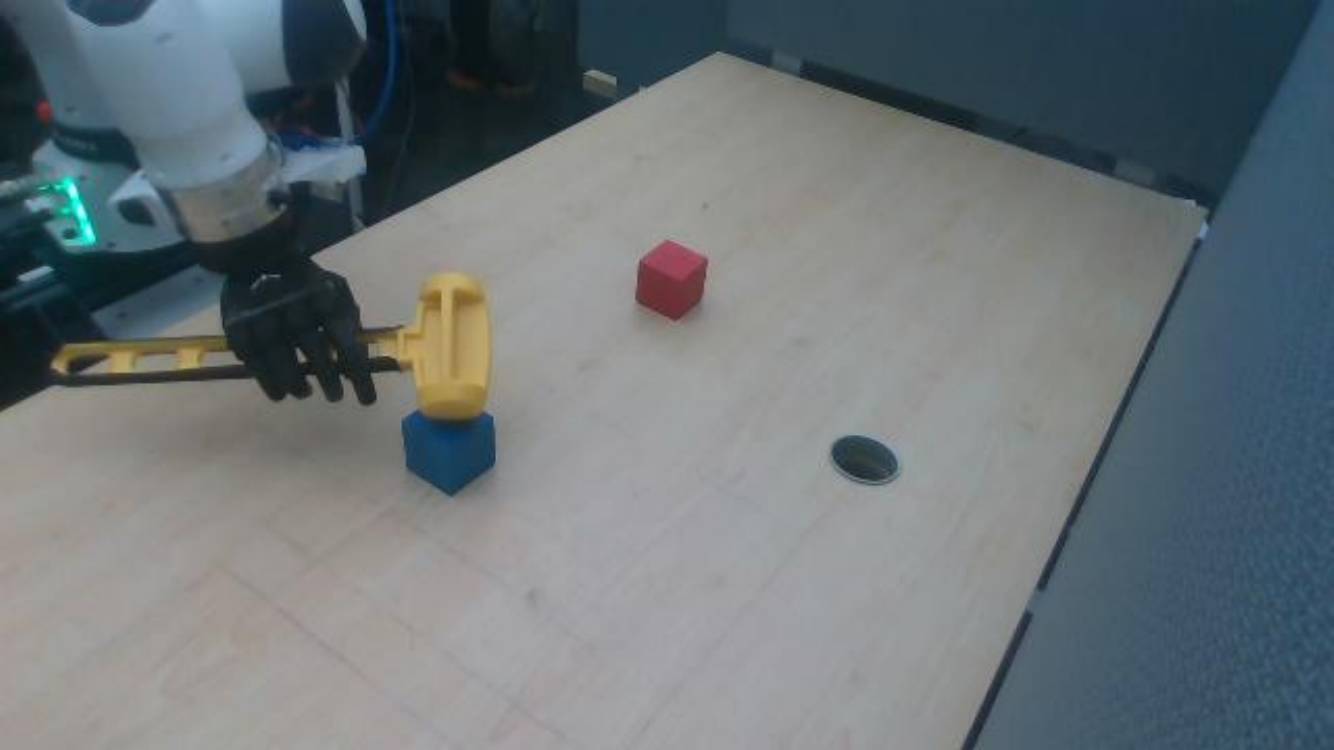} & 
                \includegraphics[width=0.9\linewidth]{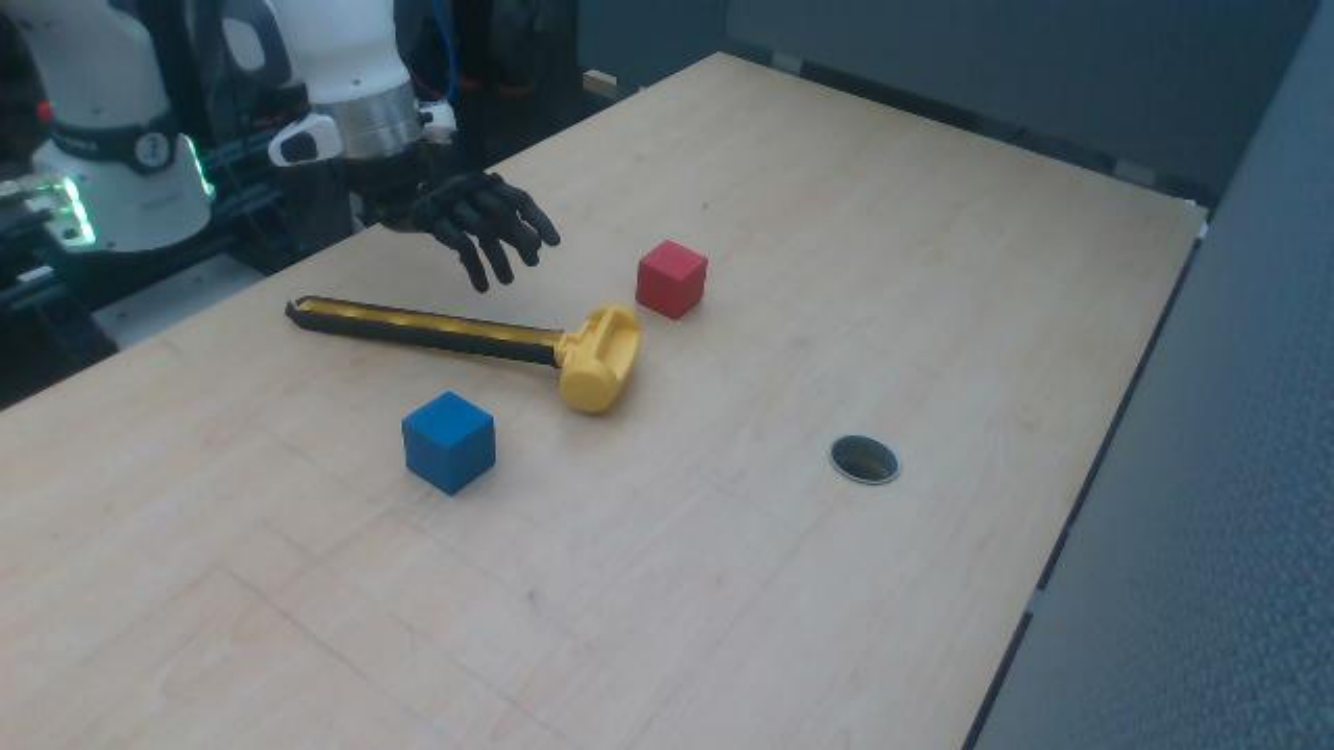} \\
        \midrule
        \multirow{5}{*}{\vspace*{-10ex}Flip}
            & Human & 
                \includegraphics[width=0.9\linewidth]{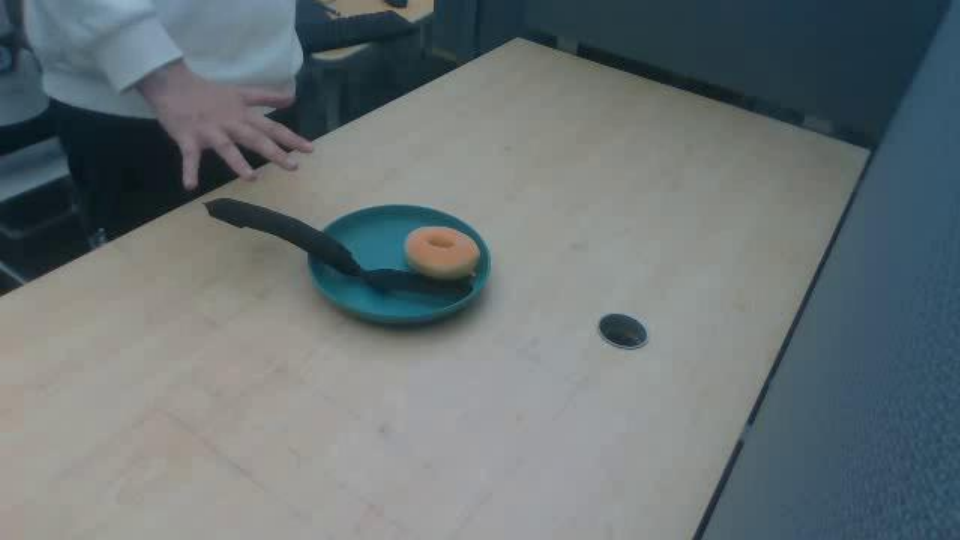} & 
                \includegraphics[width=0.9\linewidth]{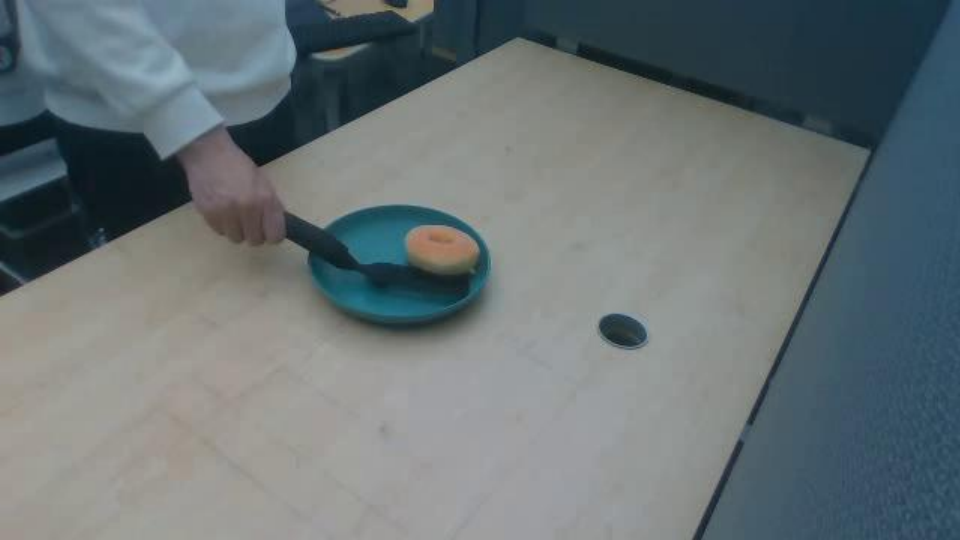} & 
                \includegraphics[width=0.9\linewidth]{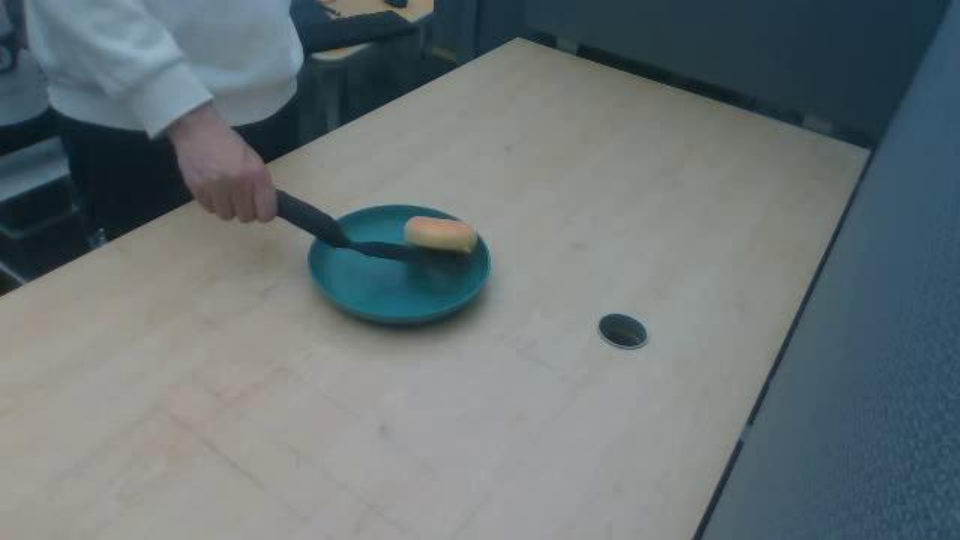} & 
                \includegraphics[width=0.9\linewidth]{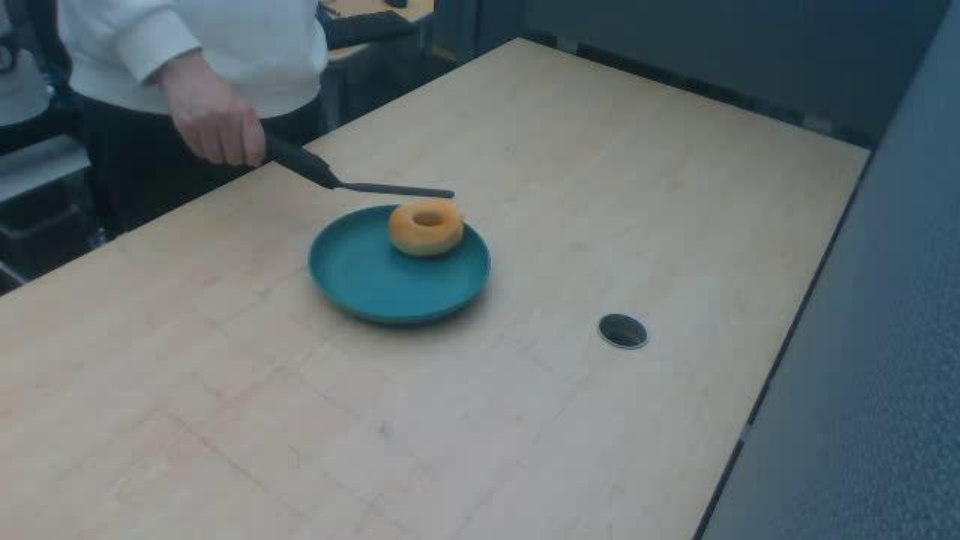} & 
                \includegraphics[width=0.9\linewidth]{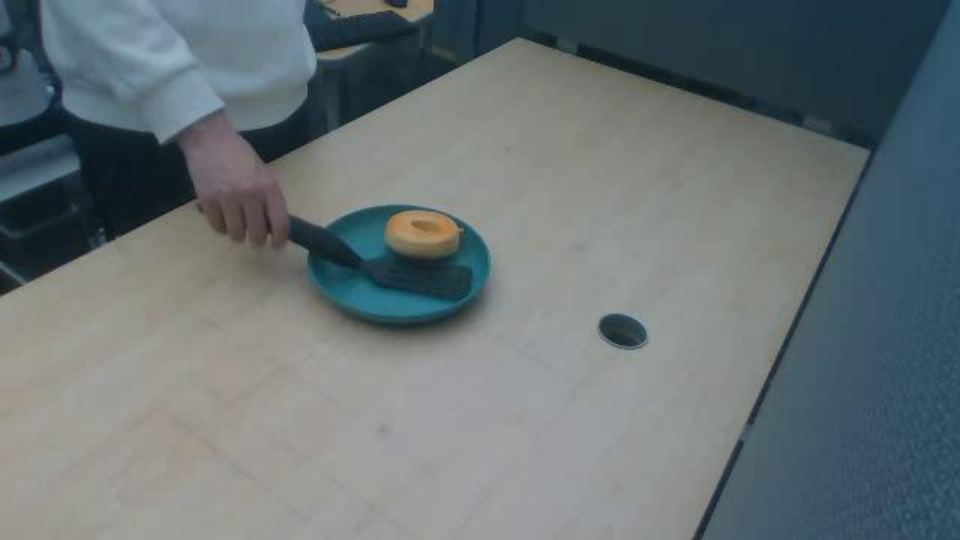} & 
                \includegraphics[width=0.9\linewidth]{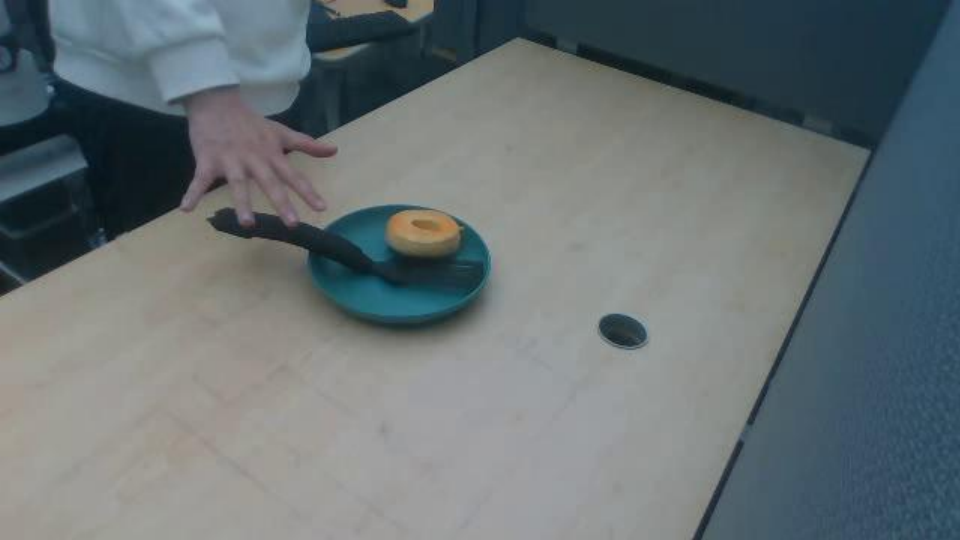} \\
            & FR & 
                \includegraphics[width=0.9\linewidth]{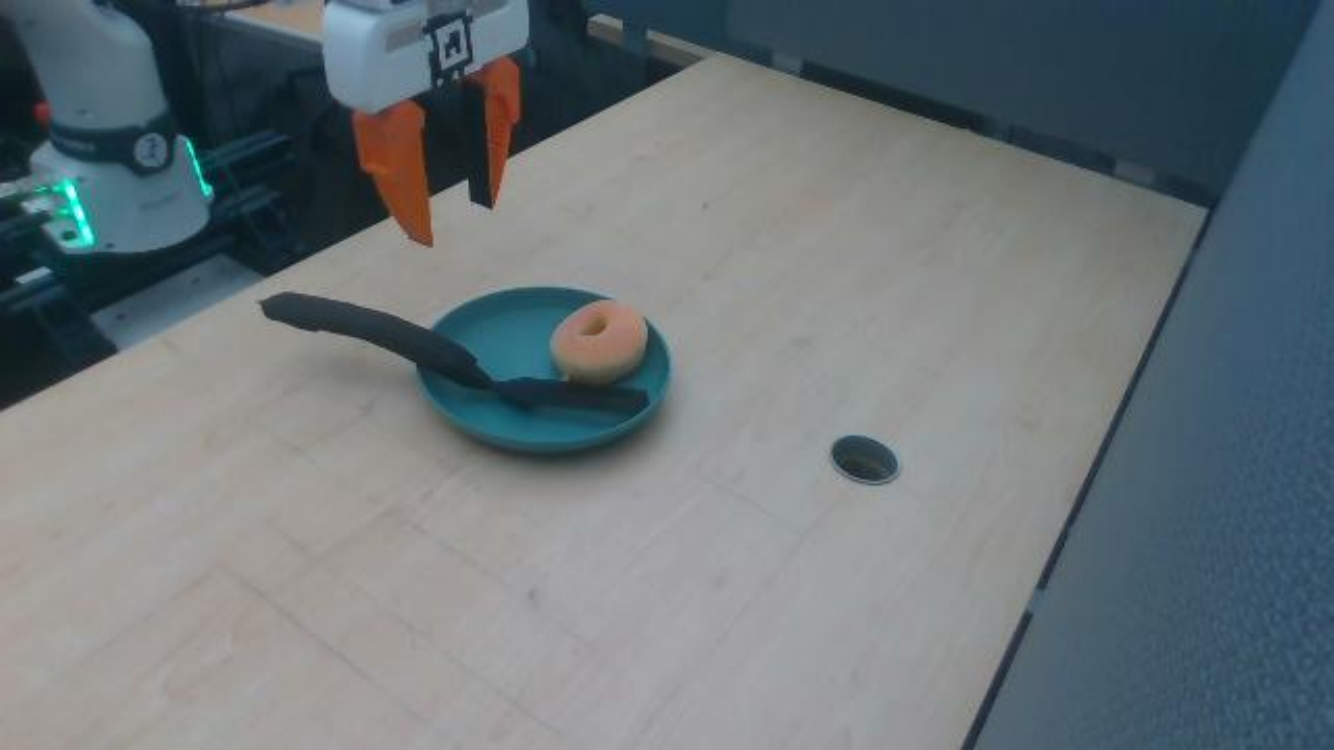} & 
                \includegraphics[width=0.9\linewidth]{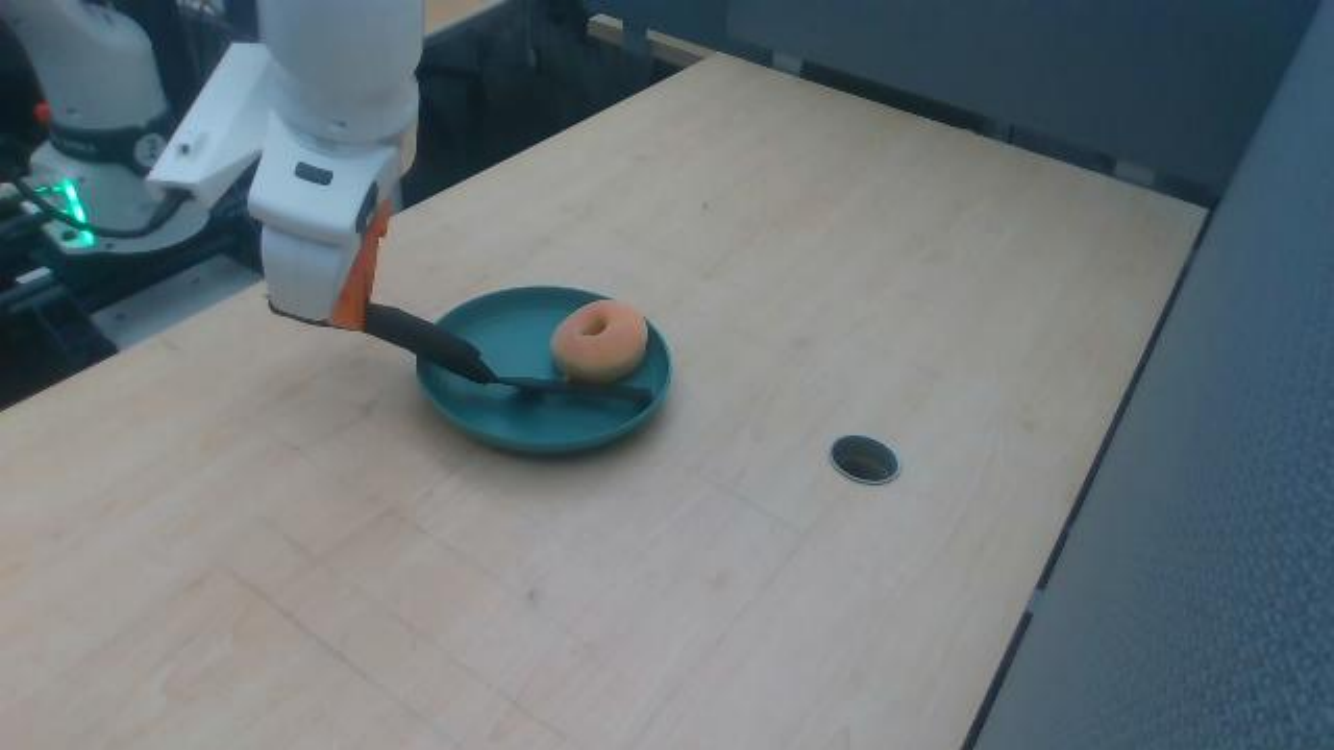} & 
                \includegraphics[width=0.9\linewidth]{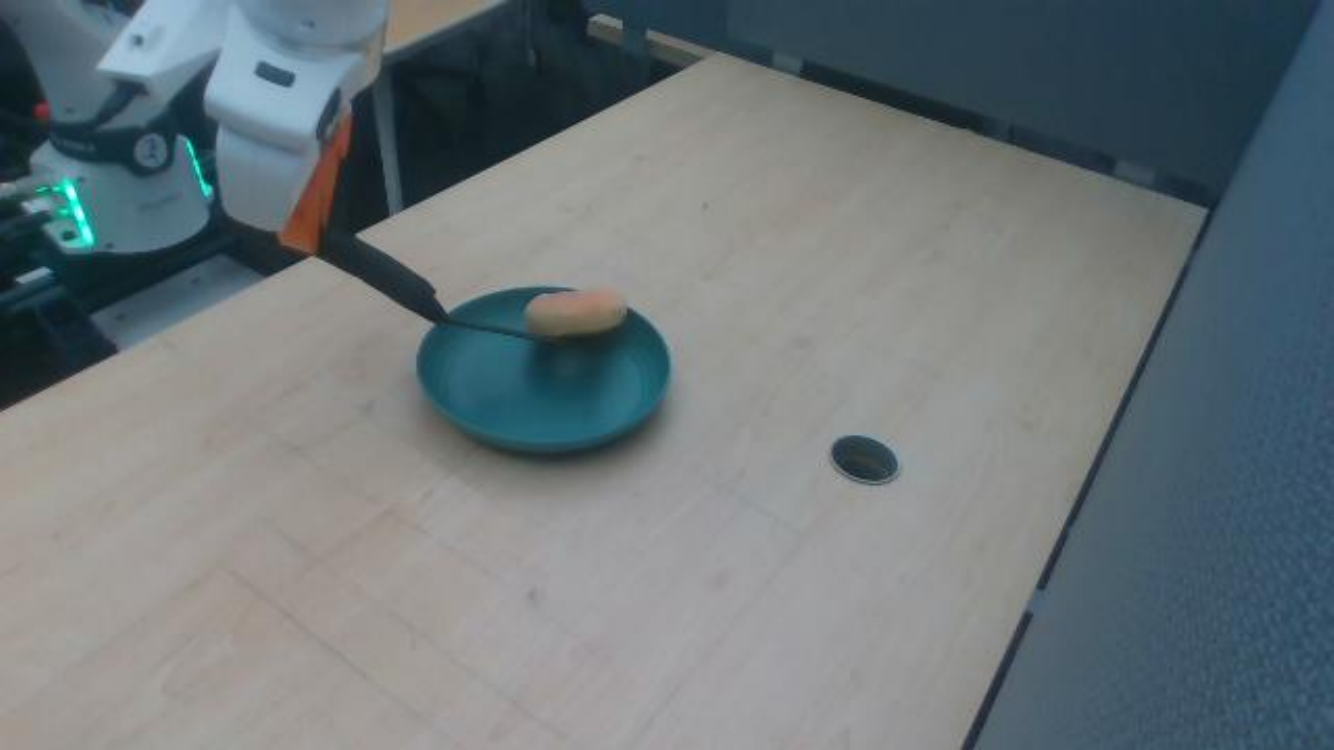} & 
                \includegraphics[width=0.9\linewidth]{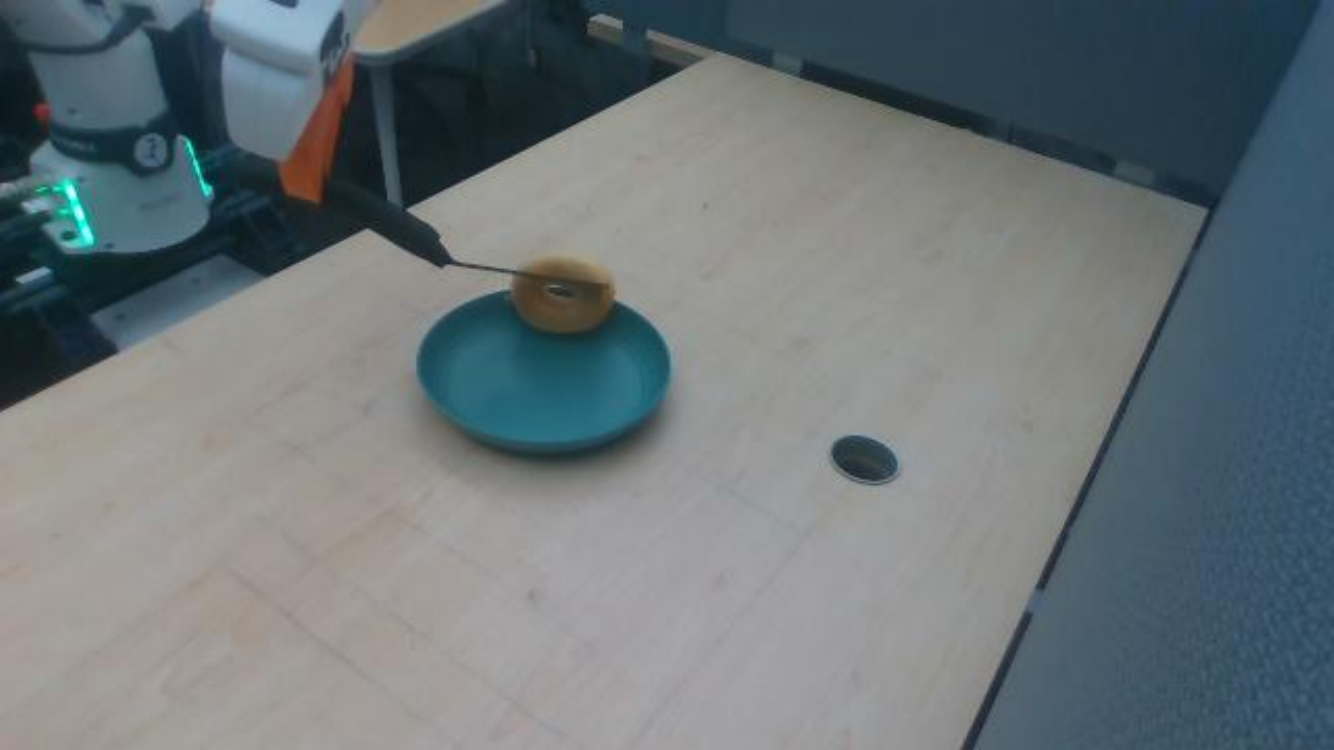} & 
                \includegraphics[width=0.9\linewidth]{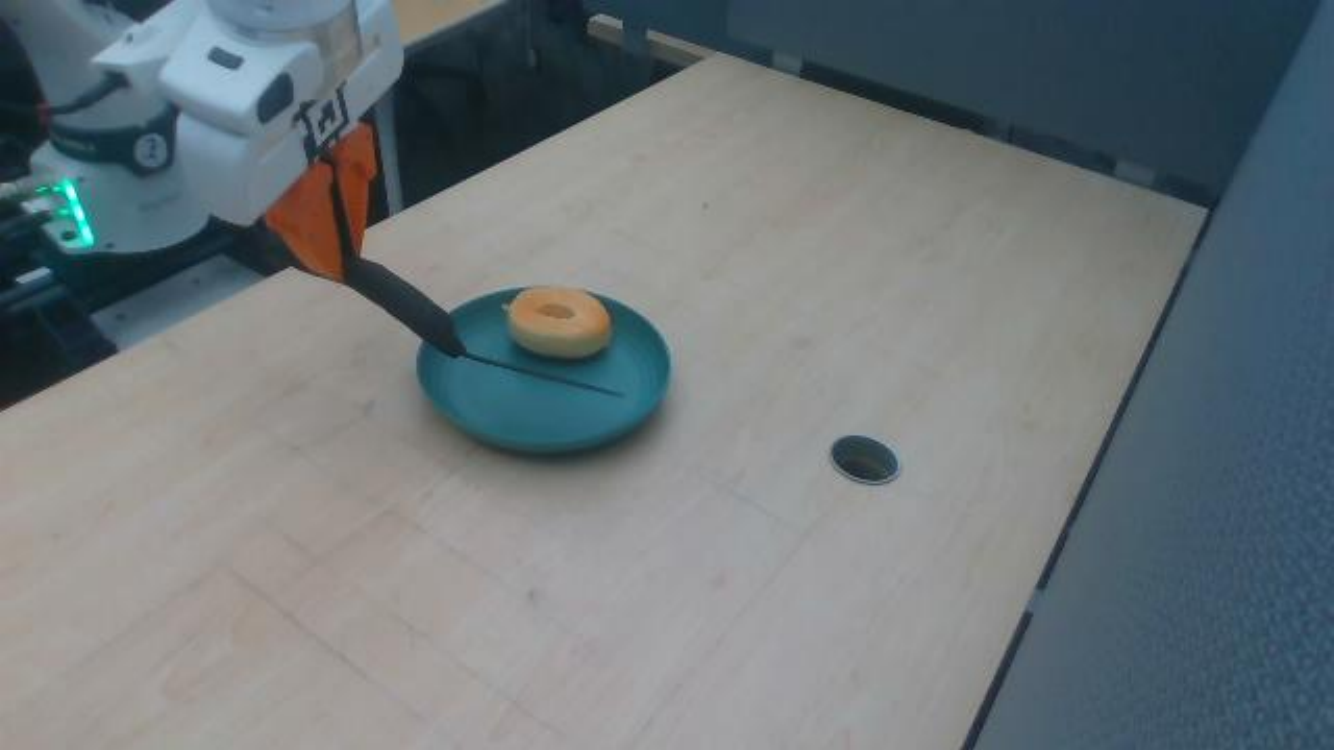} & 
                \includegraphics[width=0.9\linewidth]{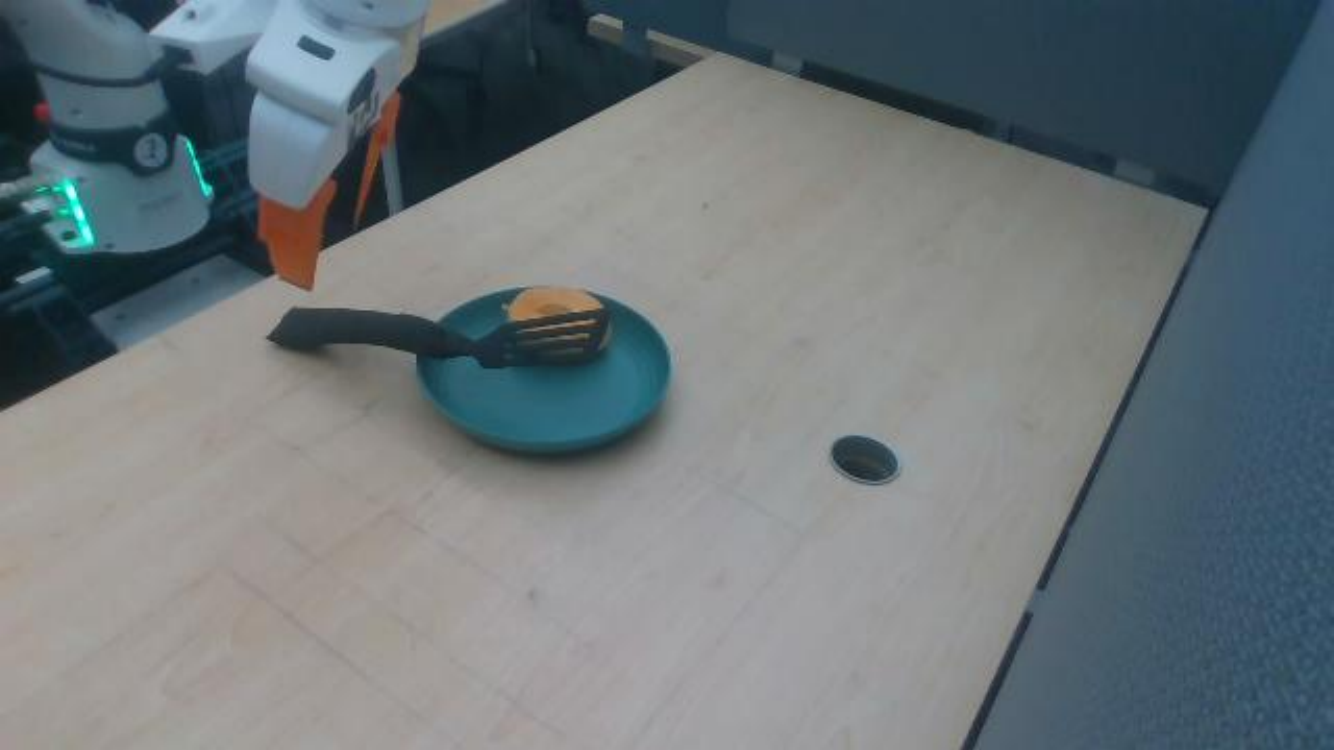} \\
            & Robotiq & 
                \includegraphics[width=0.9\linewidth]{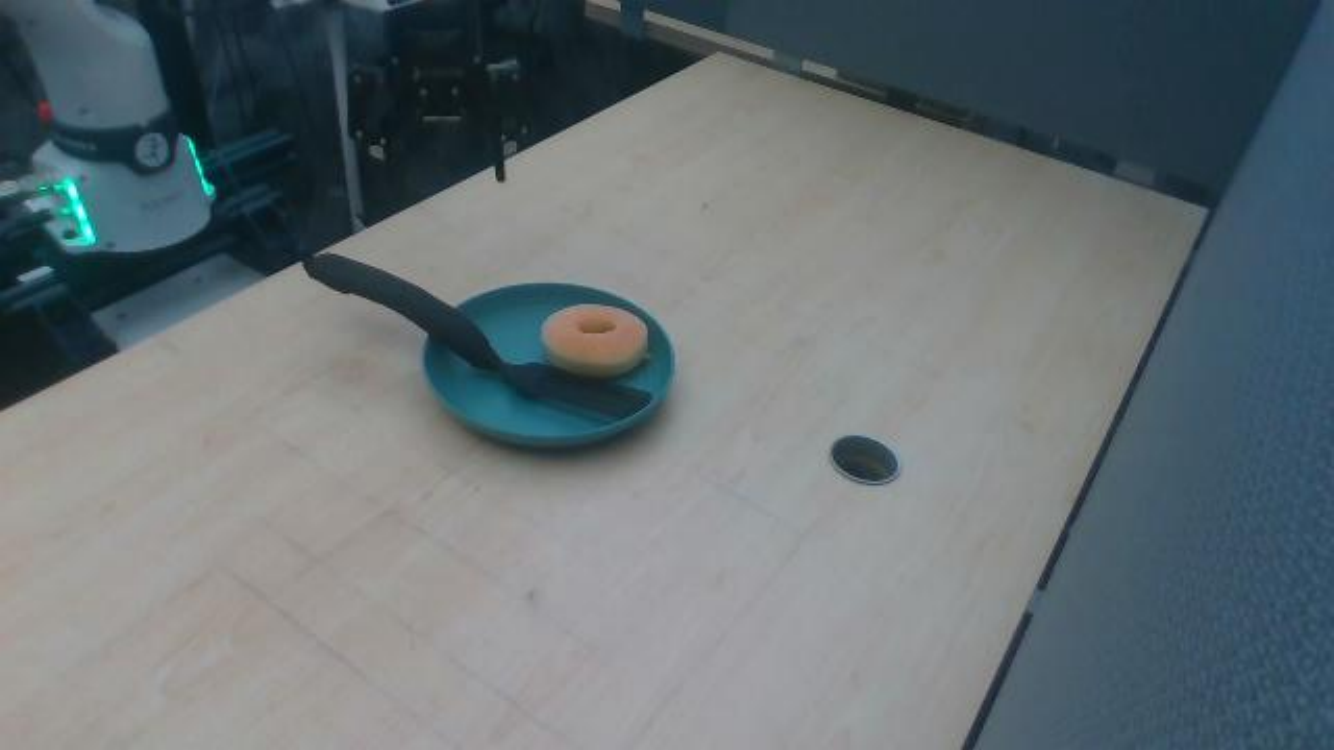} & 
                \includegraphics[width=0.9\linewidth]{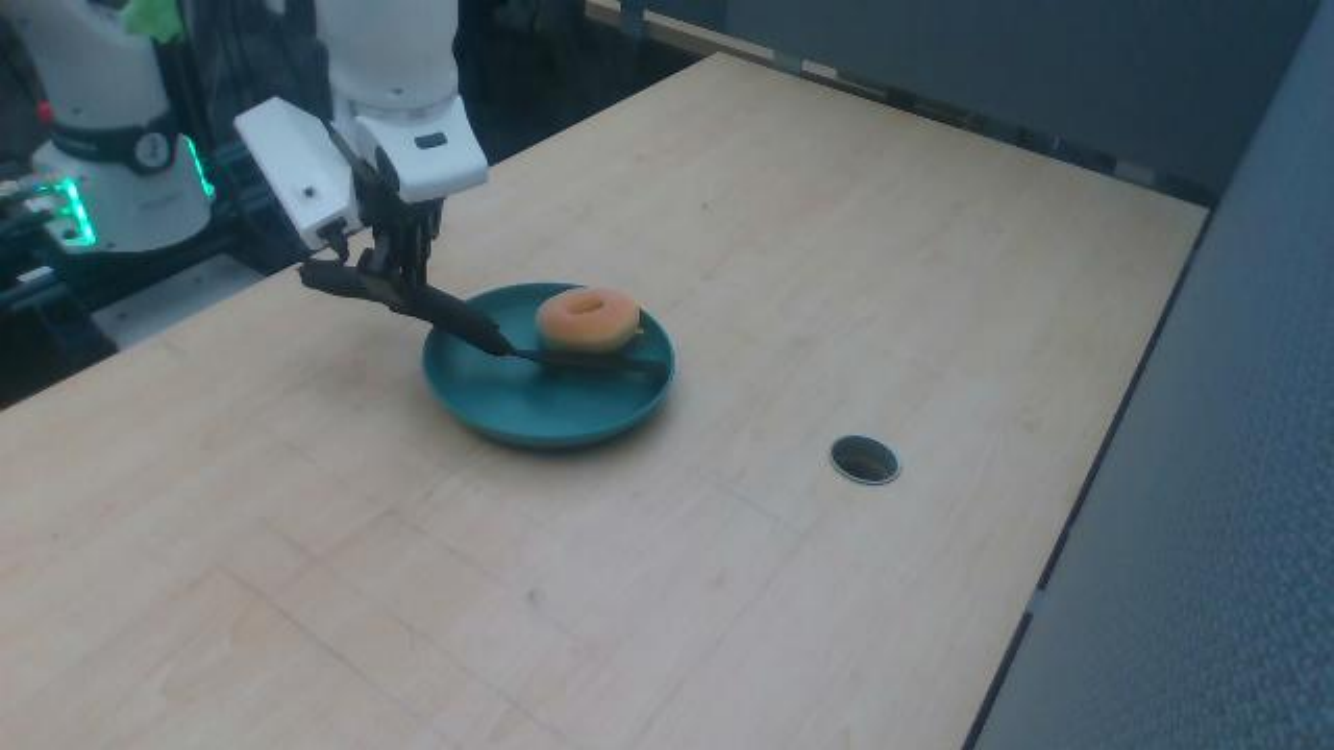} & 
                \includegraphics[width=0.9\linewidth]{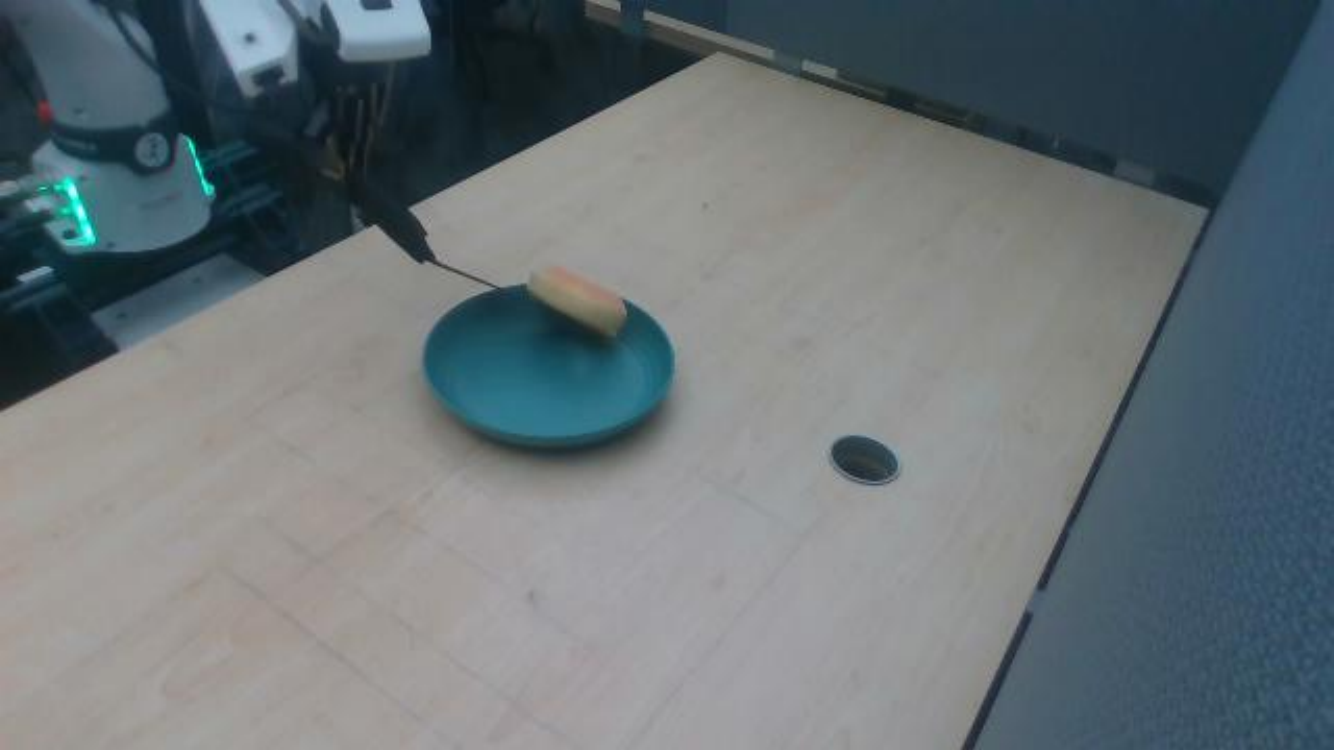} & 
                \includegraphics[width=0.9\linewidth]{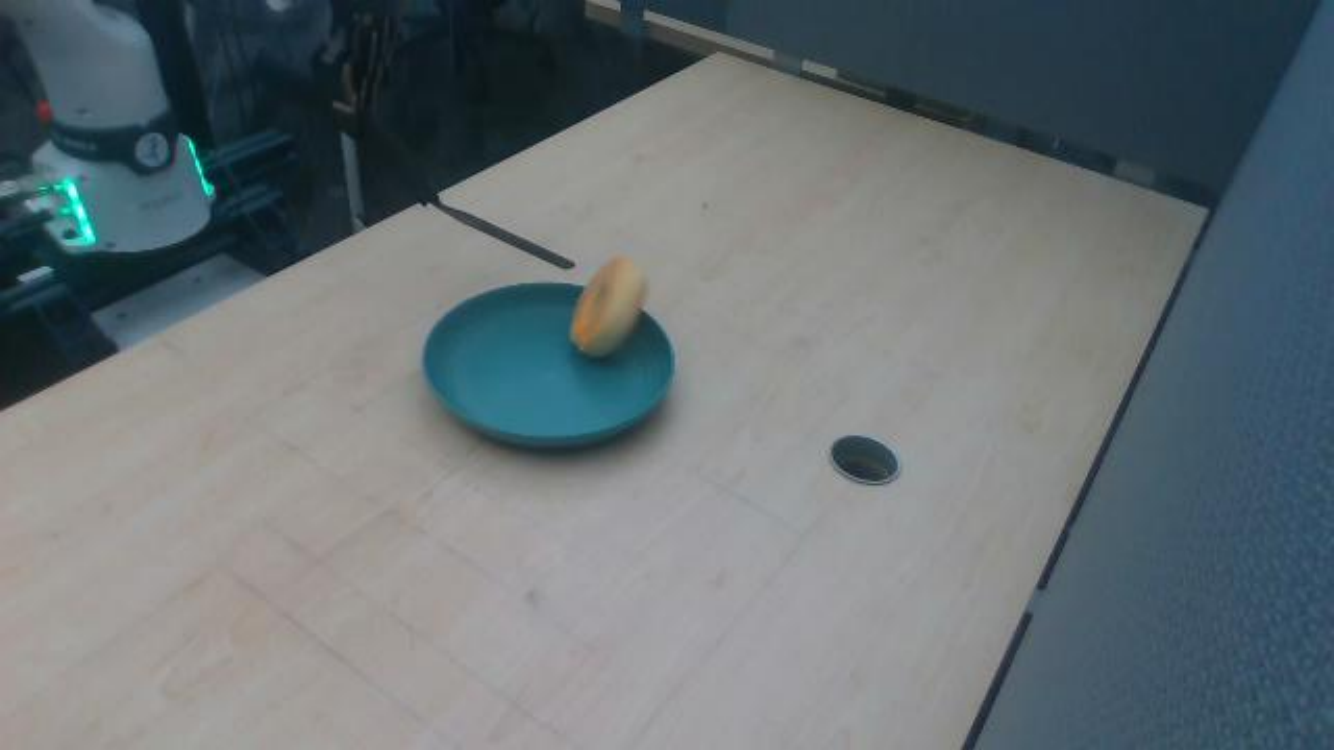} & 
                \includegraphics[width=0.9\linewidth]{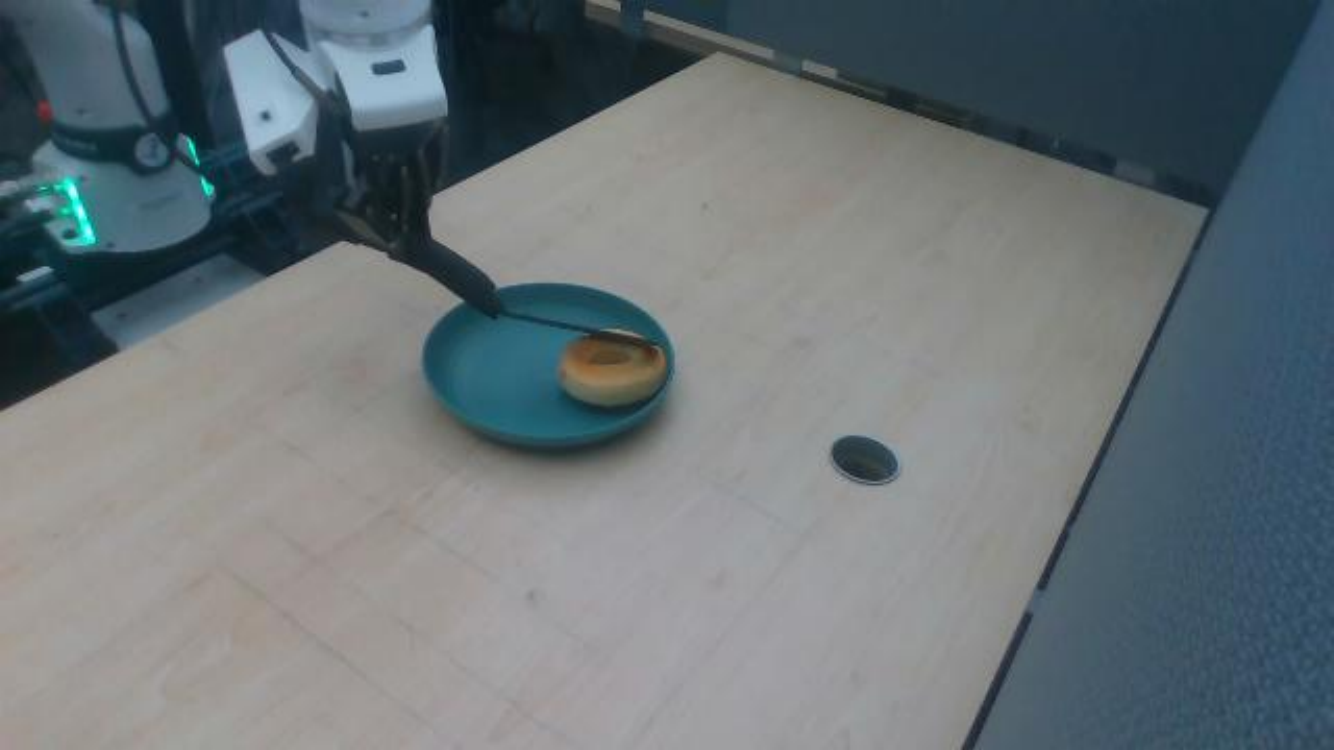} & 
                \includegraphics[width=0.9\linewidth]{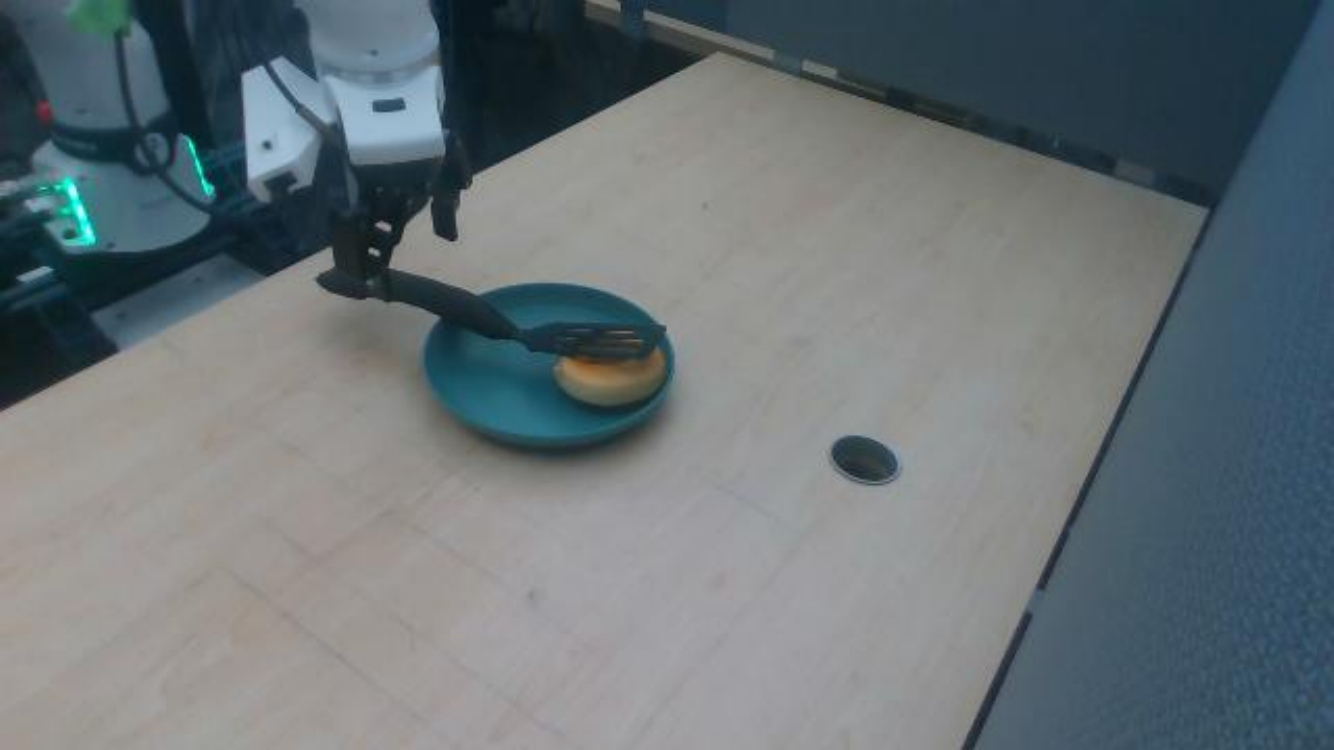} \\
            & Allegro & 
                \includegraphics[width=0.9\linewidth]{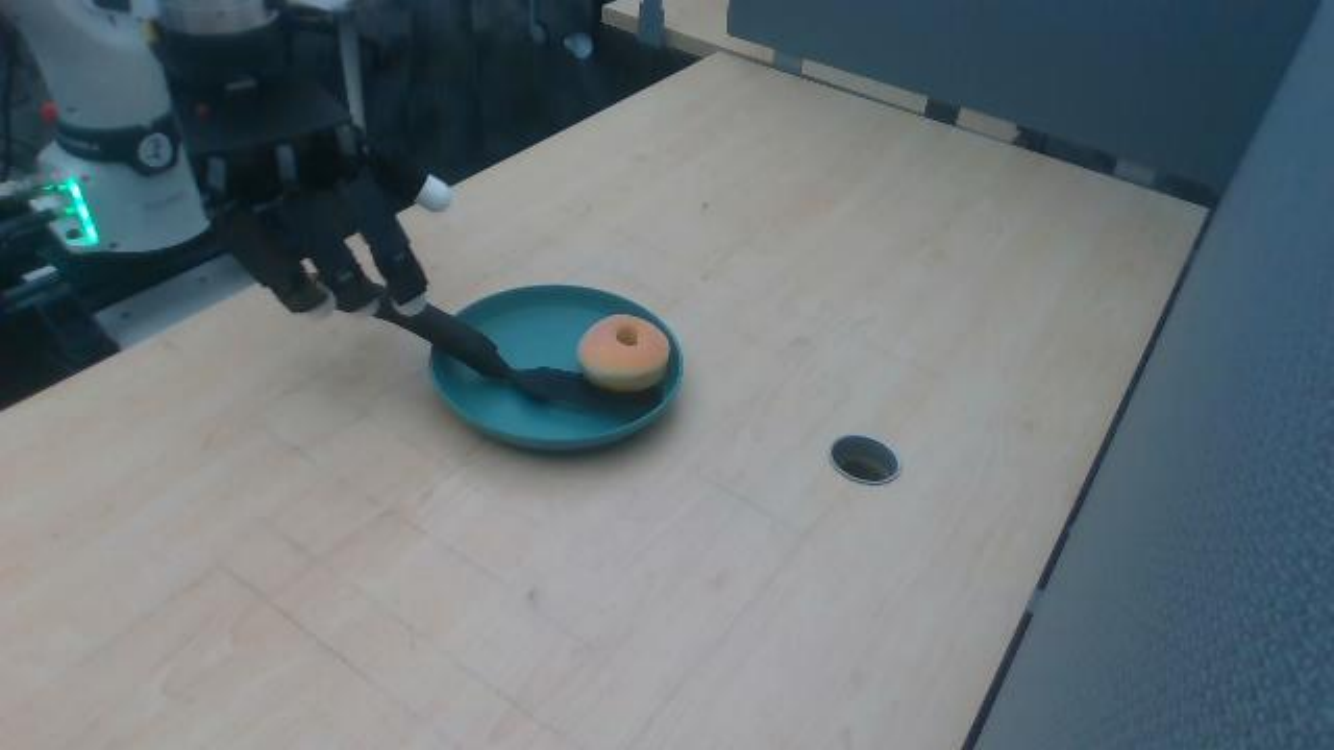} & 
                \includegraphics[width=0.9\linewidth]{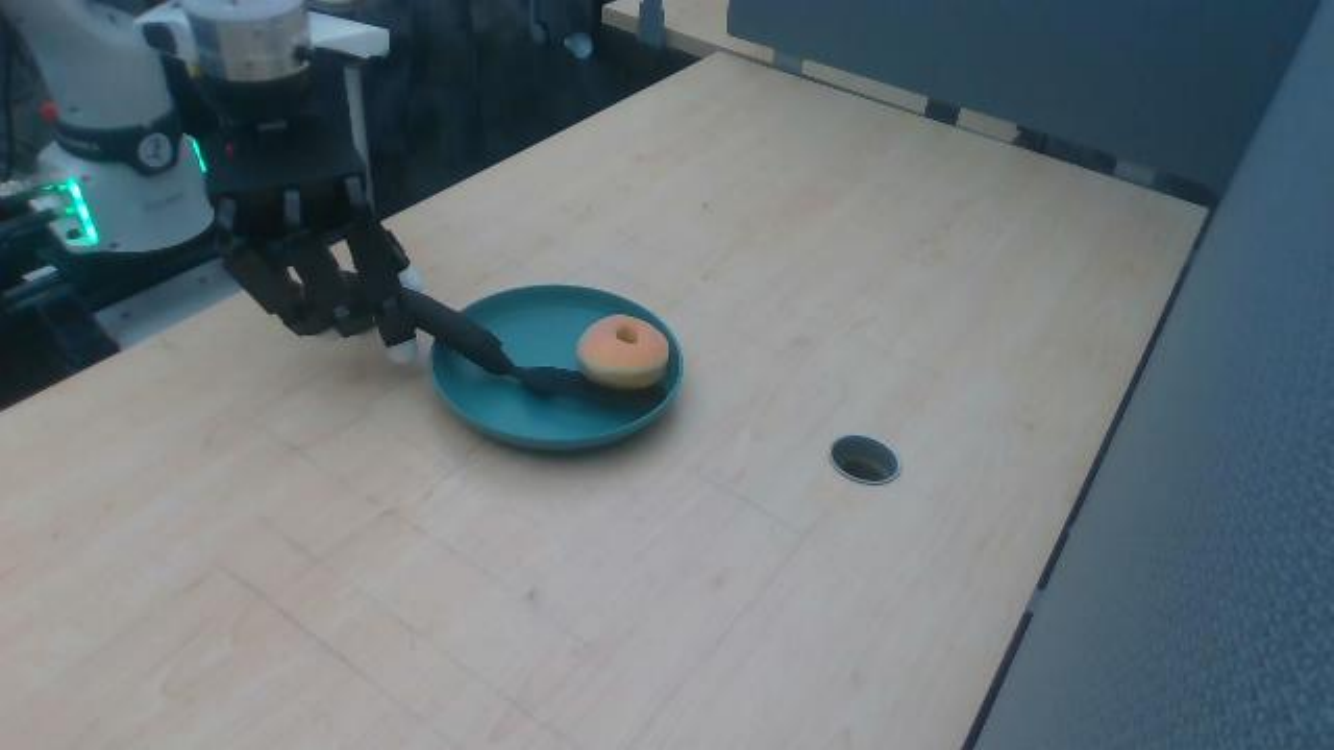} & 
                \includegraphics[width=0.9\linewidth]{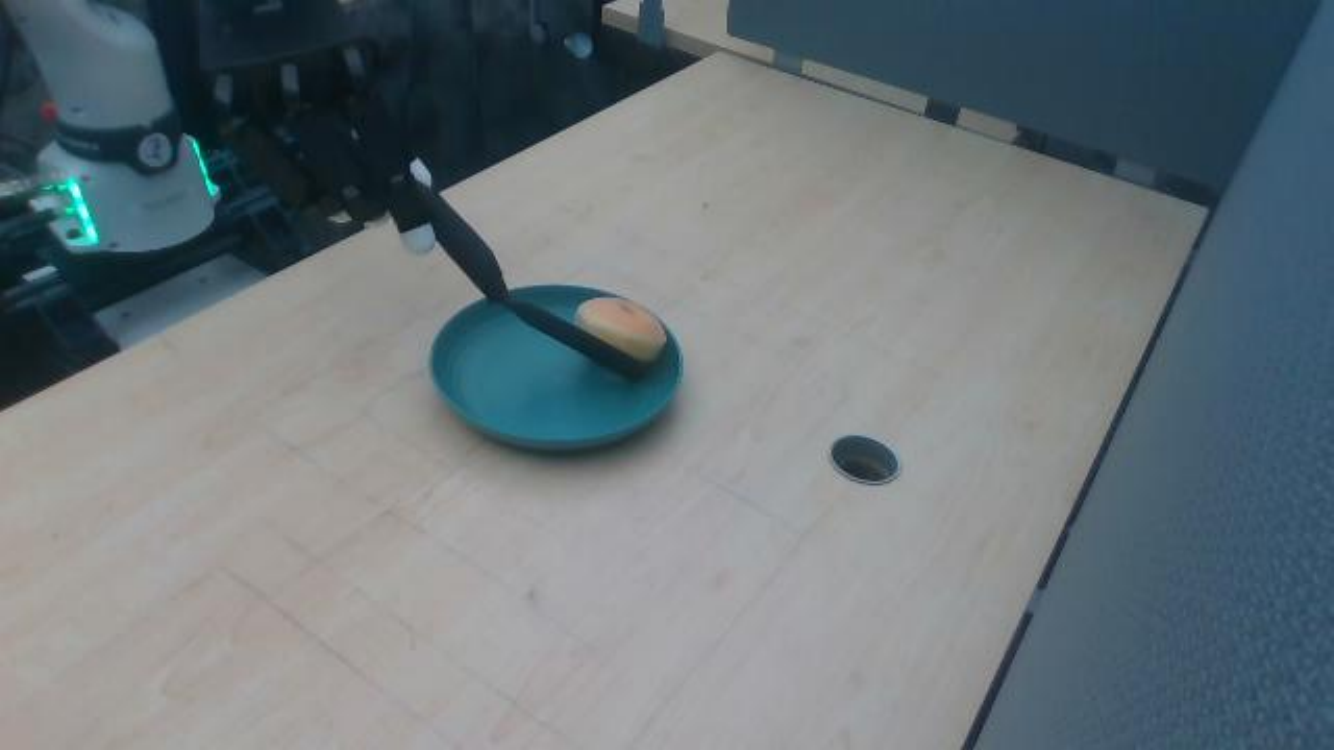} & 
                \includegraphics[width=0.9\linewidth]{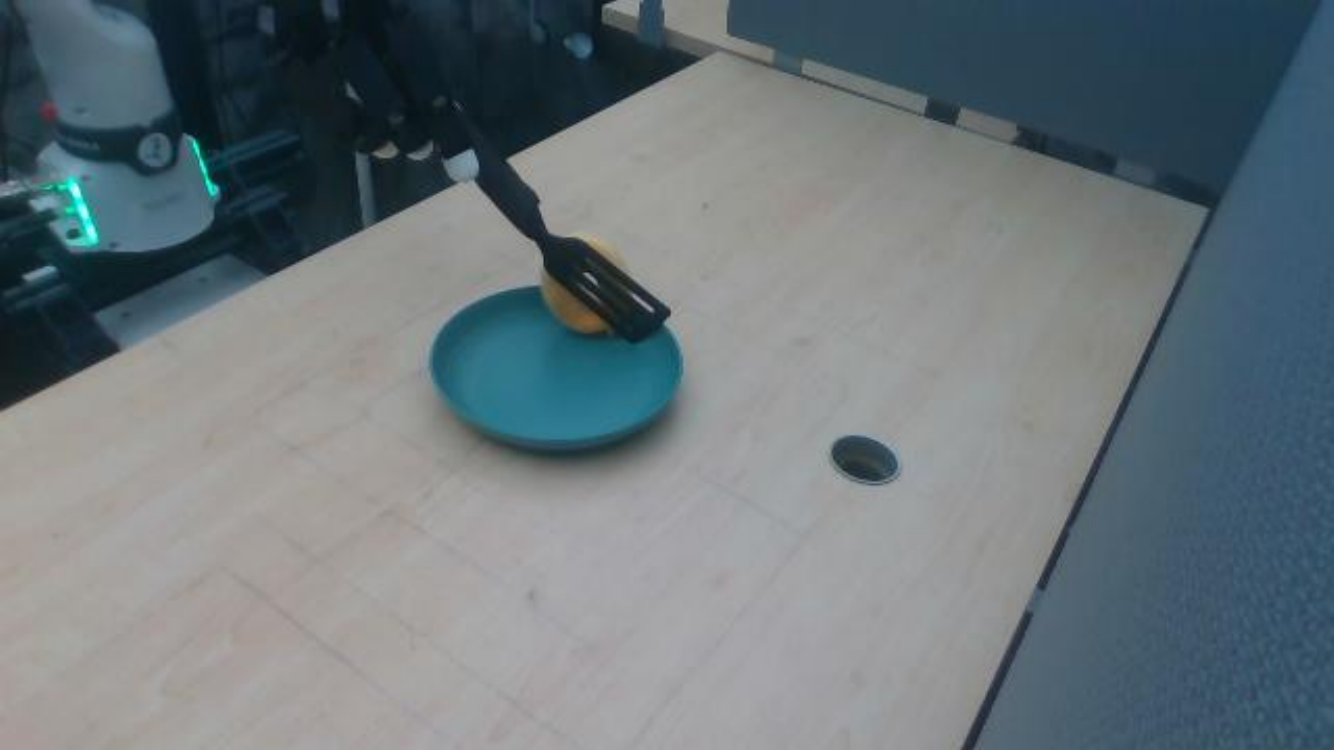} & 
                \includegraphics[width=0.9\linewidth]{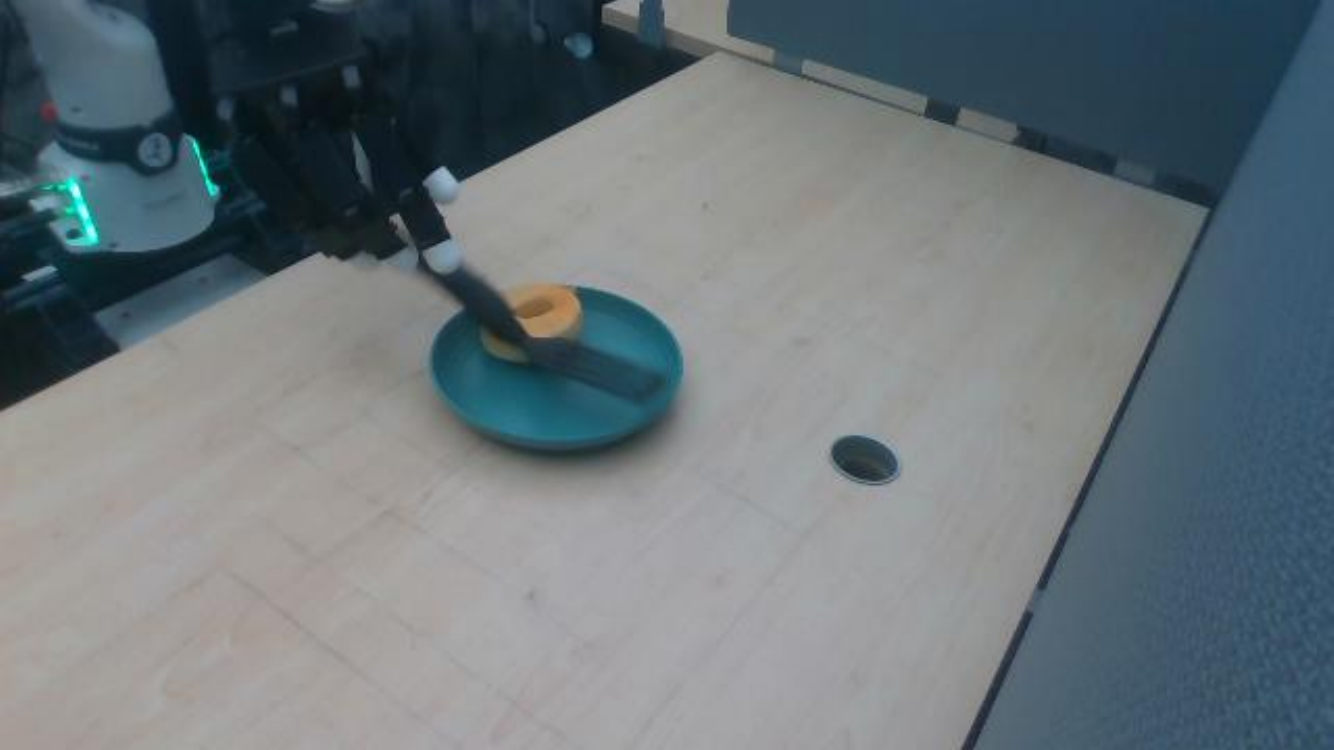} & 
                \includegraphics[width=0.9\linewidth]{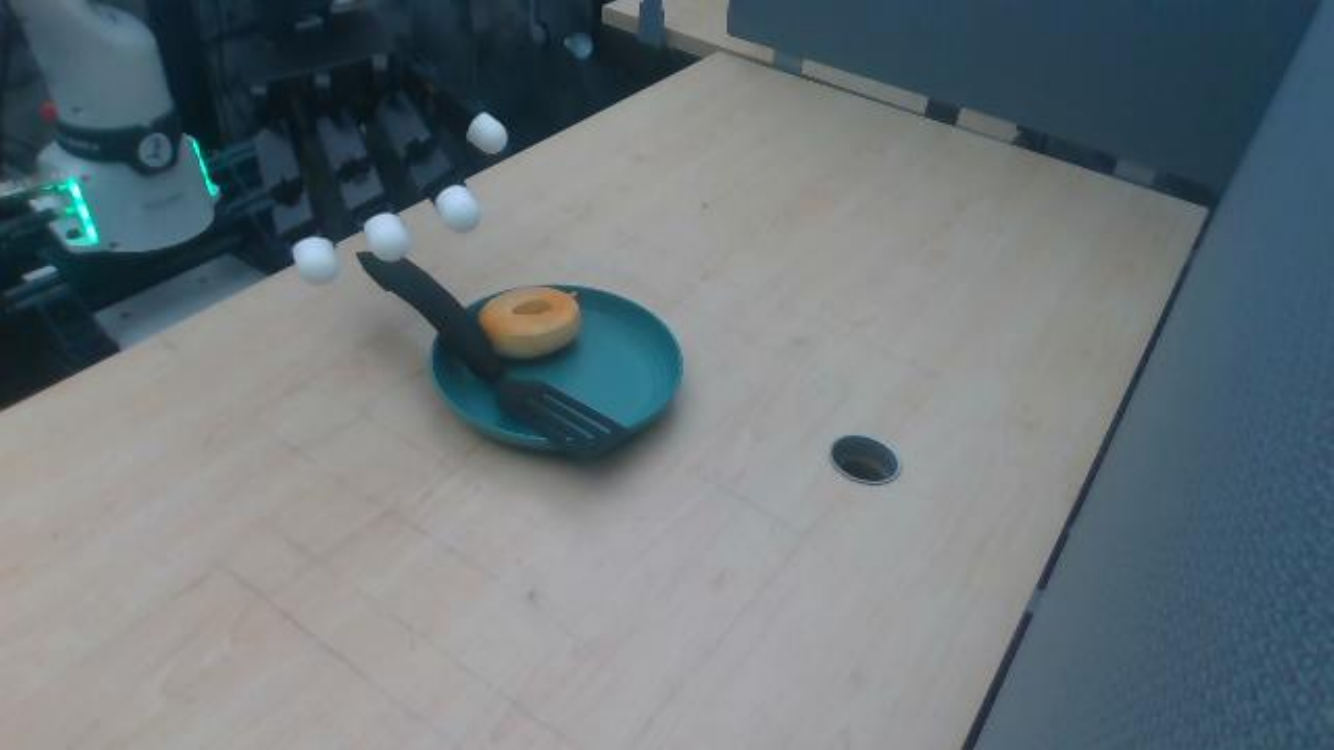} \\
            & Ability & 
                \includegraphics[width=0.9\linewidth]{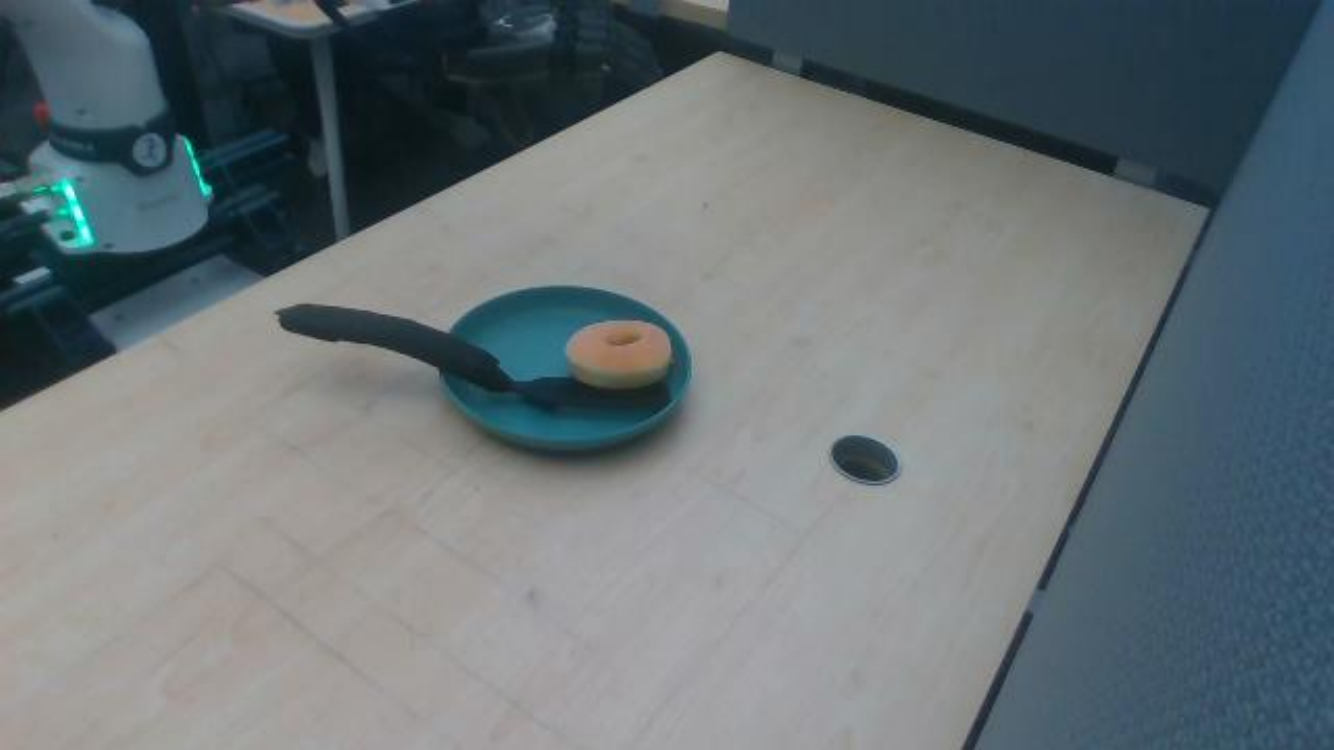} & 
                \includegraphics[width=0.9\linewidth]{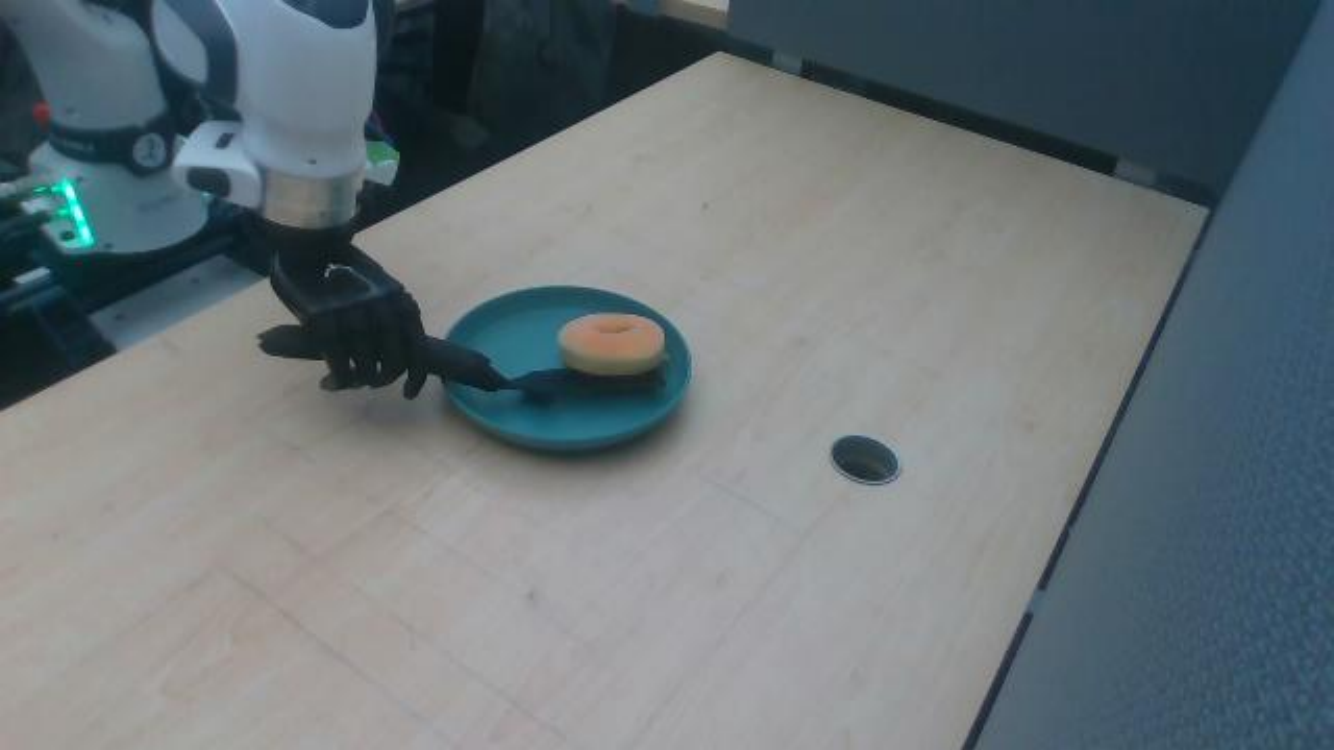} & 
                \includegraphics[width=0.9\linewidth]{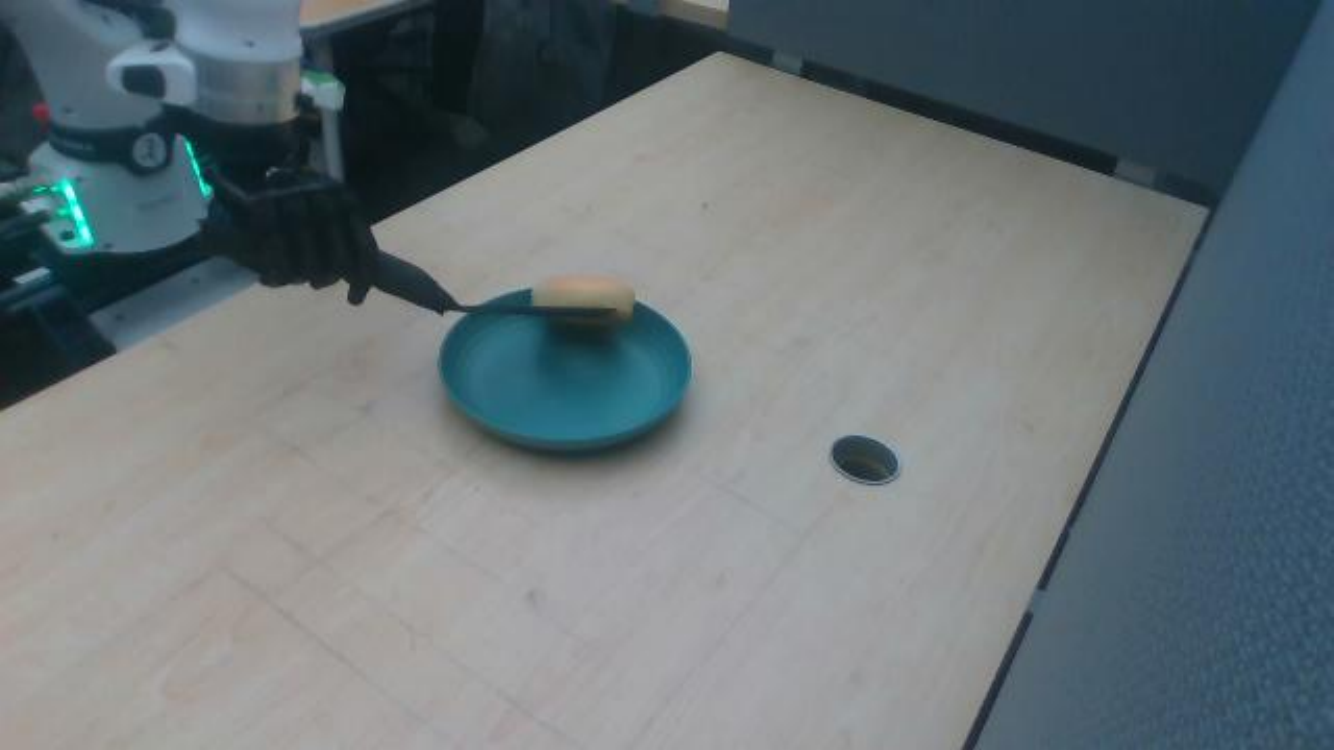} & 
                \includegraphics[width=0.9\linewidth]{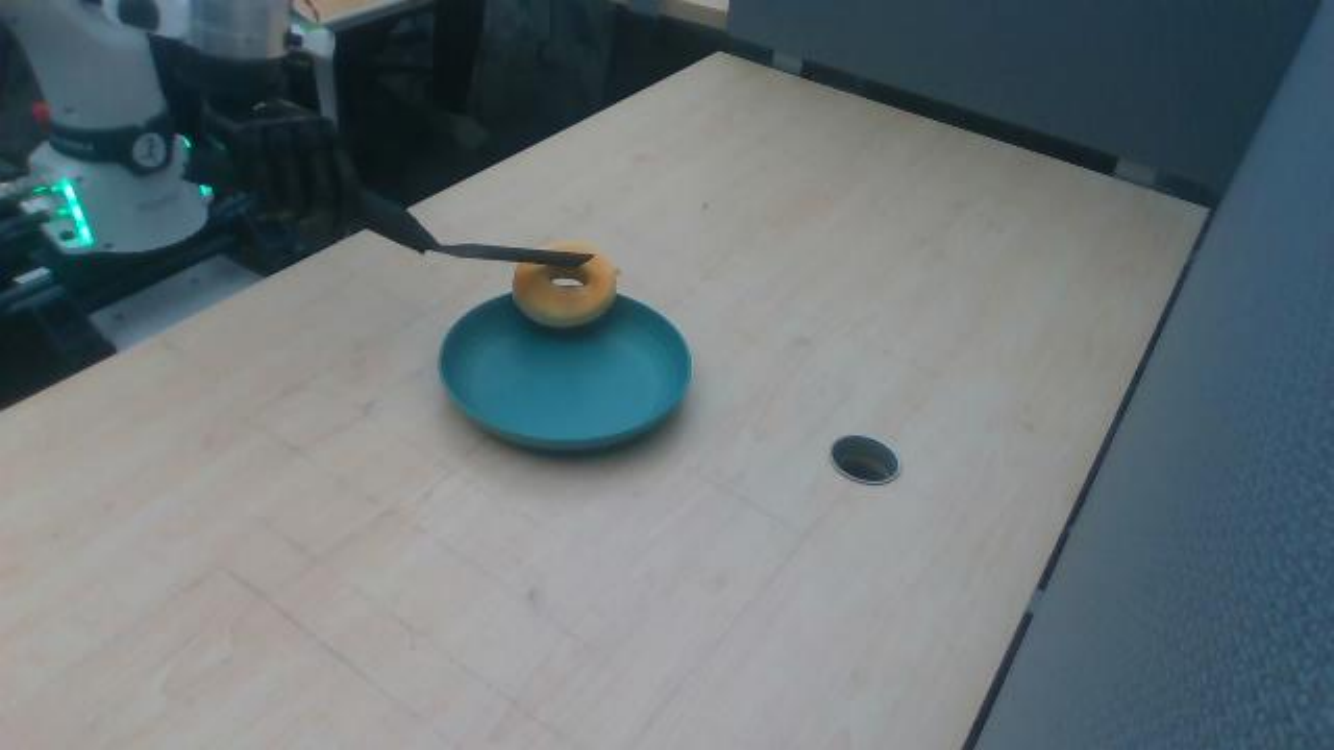} & 
                \includegraphics[width=0.9\linewidth]{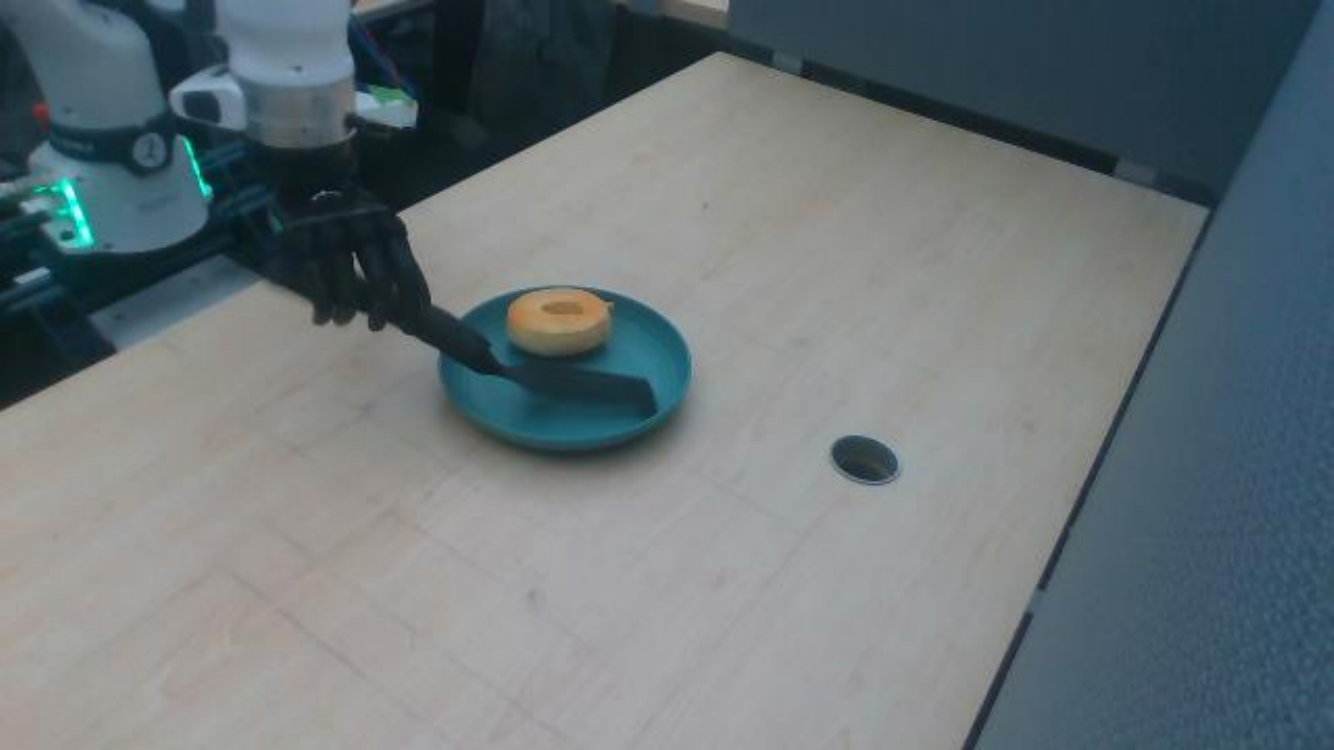} & 
                \includegraphics[width=0.9\linewidth]{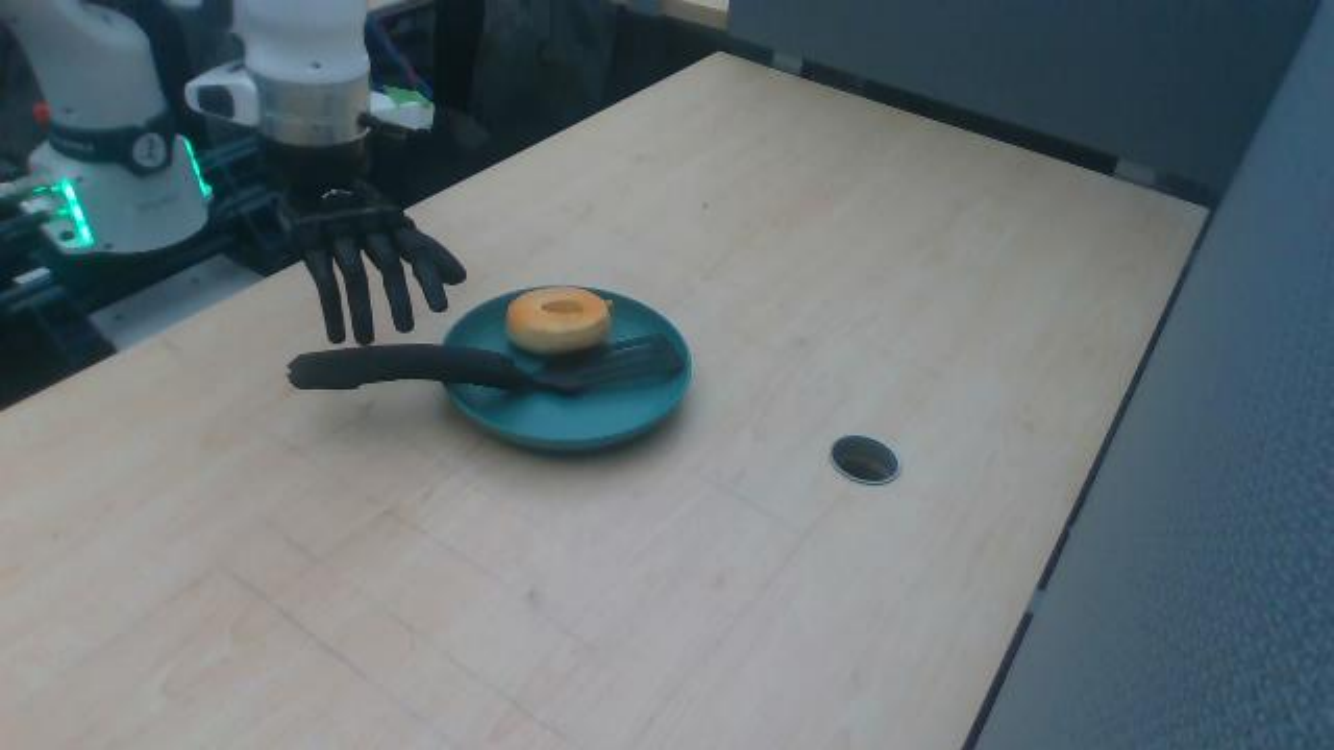} \\
        \bottomrule
    \end{tabularx}
    \caption{Agent-view visualization for human and four different embodiments (FR, Robotiq, Allegro, Ability) performing four tasks (Pick and Place, Push, Hammer, Flip).}
    \label{datavis}
    \vspace{-1em}
\end{table}

\subsection{MixUp with $\beta$-distribution}

In several MixUp-based approaches~\cite{mixup,xu2020domainmixup}, $\alpha$ is sampled from a $\beta$-distribution to augment the data distribution.  
In Tab.~\ref{tab:beta_dist_study} we compare ImMimic-A ($\beta$-dist) to ImMimic-A (linear), where $\alpha$ is sampled directly from a $\beta$-distribution.  
Our results show that ImMimic-A (linear), which uses a linearly decreasing $\alpha$ schedule, still outperforms ImMimic-A ($\beta$-dist).

The results confirm that progressive MixUp scheduling enhances policy robustness across domains. 
Models trained with the linear $\alpha$ scheduler achieve better adaptation between human and robot distributions, leading to smoother trajectories and improved task success compared to the $\beta$-distributed variant. 
This demonstrates that controlled, gradual interpolation not only bridges the domain gap but also yields more stable and effective robot behaviors.

\subsection{Success Rate Metrics}

\textbf{Success Rate.} 
The four tasks are designed to evaluate various aspects of robotic manipulation. Each task includes specific disturbances to test robustness.

\emph{1. Basic Object Manipulation.}
(1) \textbf{Pick and Place}: The robot must pick up a cube from a start position and place it at a designated target location.
The initial position of the cube is roughly fixed but includes a random offset within the start area. This task evaluates the robot’s ability to accurately grasp and relocate objects. The task is considered successful if the cube fully covers the target point.
We consider the attempt successful if the cube fully covers the target point.
(2) \textbf{Push}: The robot must push the object from the start position to the target region. This task primarily evaluates finger-free manipulation capabilities. Similar to the Pick and Place, a random offset is applied to the cube's initial position. The task is considered successful if the object reaches the target region after the push.
\emph{2. Tool-based Manipulation.}
(1) \textbf{Hammer}: The robot must pick up a hammer and strike a target cube with its head. This task requires proper tool grasping and precise targeting. The hammer is initially placed on a cube, with its handle orientation randomly disturbed within a 45-degree range. The task is successful if the hammer’s head touches the top surface of the target cube.
(2) \textbf{Flip}: The robot must flip a bagel using a spatula after lifting it. This task emphasizes precise wrist control and rotational dexterity. The spatula is placed at an angle within 45 degrees, and the bagel is positioned randomly on different parts of its head. Success is defined as the bagel being flipped over.

\textbf{Failure Cases.}
We summarize common failure modes observed across the four robotic embodiments.

\emph{Robotiq Gripper.}  
In Push, Robotiq Gripper struggles to maintain a straight trajectory due to its thin fingertips, leading to unstable contact and frequent path corrections. In Flip, limited wrist articulation and low contact area make it difficult to control the spatula through the full rotation, resulting in intermittent slippage. Additionally, a structural gap above the fingertips can cause the gripper to grasp the spatula within this space, leading to an unstable grip. These issues are visually highlighted in Fig. 7(a,b), where Robotiq’s fingertip geometry and palm gap contribute to contact instability and slippage.

\emph{Fin Ray Gripper.}  
In Push, FR Gripper improves on Robotiq Gripper’s stability but still lacks the fine precision of multi-fingered hands. In Flip, its limited wrist articulation leads to occasional loss of control during dynamic movements.

\emph{Allegro Hand.}  
In Hammer, Allegro’s relatively large hand size reduces its ability to generate sufficient lift force, making it difficult to wield heavier tools effectively. In Flip, the same size limitation, combined with weak grip force, often results in the spatula slipping before the motion completes. These failures are illustrated in Fig. 7(f,g), where the hand struggles to maintain stable tool contact during high-torque actions.

\emph{Ability Hand.}  
In Pick and Place, the short thumb and limited wrist flexibility of the Panda arm often result in unstable grasps and frequent object drops. In Hammer, the same constraints hinder stable tool grasping and force transmission. As shown in Fig. 7(c,d), the shorter thumb may also contribute to misaligned grasps, especially when positional offsets are present.

\textbf{Mechanical Design Insights.} 
Analysis of failure cases reveals that no single hand design is universally optimal across all tasks. However, several general insights can inform more effective mechanical design of end-effector:

(1) Increase thumb length relative to other fingers to expand the acceptable grasping margin and reduce off-center spinning (supported by biological evidence~\cite{evolution}). A longer thumb increases the moment arm and provides greater contact redundancy, improving robustness when objects shift under load.

(2) Account for mounting and arm constraints. Most current end-effector mounts lack an additional wrist degree of freedom, limiting the ability to perform human-like reorientation. Introducing a swivel or universal joint at the mounting interface can restore this degree of freedom, enabling more favorable tool approaches without compromising the robot’s kinematic reach.

(3) Enable firm, adaptive grasps by incorporating an adjustable thumb–finger aperture mechanism and compliant interface materials. A variable-spacing mechanism allows the hand to conform to different tool cross-sections, while soft, high-friction coatings compensate for local misalignments and absorb minor impacts, preventing slippage throughout the workspace.

\subsection{Smoothness Metrics}
\label{ssec:appendix:smoothness}

Spectral Arc Length (SPARC) quantifies smoothness by measuring the arc length of the normalized magnitude spectrum of a trajectory’s speed profile in the frequency domain \cite{sparc}, building on the original Spectral Arc Length (SAL) \cite{sal}. Given a speed profile $s_t$, the normalized spectrum is defined as:
\begin{equation}
    \hat{S}(\omega) = \frac{S(\omega)}{S(0)}
\end{equation}
The SAL metric is then computed as:
\begin{equation}
    \text{SAL} \triangleq -\int_{0}^{\omega_c} \sqrt{\left(\frac{1}{\omega_c}\right)^2 + \left(\frac{d\,\hat{S}(\omega)}{d\omega}\right)^2}\, d\omega
\end{equation}
SPARC improves upon SAL by adaptively selecting the cutoff frequency $\omega_c$ based on an amplitude threshold $\overline{S}$ and an upper frequency limit $\omega_c^{\max}$:
\begin{equation}
    \omega_c \triangleq \min\Biggl\{\omega_c^{\max},\, \min\Bigl\{\omega \,\Big|\, \hat{S}(\gamma) < \overline{S},\ \forall\, \gamma > \omega\Bigr\}\Biggr\}
\end{equation}
In our implementation, we apply zero-padding to the speed trajectory with a factor of $K=4$, and set the parameters $\omega_c^{\max}=15$, $\overline{S}=0.05$. 
A higher SPARC score corresponds to a smoother trajectory.
With the metric, we are able to show that our ImMimic improves the smoothness for the rollout policy to both Robot-Only and Co-Training.

\subsection{Training Setup and Deployment Details}
All models are trained for 300 epochs using an NVIDIA A40 GPU, with a batch size of 128.
For deployment, we perform policy rollout with both inference and control running at 30 Hz on a desktop equipped with an NVIDIA RTX 4090 GPU.
All robot sensors operate at 30 Hz, while the Zed and RealSense cameras stream at 30 FPS.

\end{document}